\documentclass[twoside,11pt]{article}
\usepackage[preprint]{jmlr2e}
\usepackage{amsmath}
\usepackage{subfigure}
\usepackage{float}
\usepackage{booktabs}
\usepackage{wrapfig}
\usepackage{multirow}
\usepackage{arydshln}
\usepackage{pifont}
\usepackage{comment}
\usepackage{enumitem}
\usepackage{lastpage}
\usepackage{xcolor}
\usepackage{graphicx}
\graphicspath{{Shared/}}
\usepackage{adjustbox}

\newcommand{\bm}[1]{\boldsymbol{#1}}
\newcommand{\e}{\textnormal{e}}
\allowdisplaybreaks[2]
\hypersetup{hidelinks,hypertexnames=false}
\jmlrheading{}{2026}{1-\pageref{LastPage}}{}{}{}{Qiufu Li, Huibin Xiao, Linlin Shen, and Yongdong Huang}
\ShortHeadings{Neural Collapse on Imbalanced Data}{Li, Xiao, Shen, and Huang}
\firstpageno{1}
\begin{document}
\title{Neural Collapse on Imbalanced Datasets \\ and Open-Set Scenario}
\author{\name Qiufu Li \email liqiufu@szu.edu.cn\\
  \name Huibin Xiao \email hb\_xiao@outlook.com \\
  \name Linlin Shen \email llshen@szu.edu.cn \\
  \addr School of Artificial Intelligence; National Engineering Laboratory for Big Data System Computing Technology;\\
  Guangdong Provincial Key Laboratory of Intelligent Information Processing; Intelligent Computing Center,\\
  Shenzhen University, Shenzhen 518060, China
  \AND
  \name Yongdong Huang \email huang\_yongdong@163.com\\
  \addr Institute of Image Understanding Research, North Minzu University, Yinchuan 750021, China
}


\maketitle
\begin{abstract}
Neural collapse (NC) describes the elegant geometric structure that sample features and classifier vectors exhibit when a classification model's loss reaches its minimum. Class imbalance, however, generally distorts standard NC. In this paper, we theoretically reformulate NC on imbalanced datasets and in open-set settings. Our analysis is based on an optimization problem over class-wise averaged losses, rather than the global averaged losses used in prior NC studies. We prove that when class-wise averaged losses attain their minima, standard NC is achieved on balanced or imbalanced datasets, for both cross-entropy (CE) and binary cross-entropy (BCE) losses. We further extend the analysis to an open-set-inspired scenario, where some classes contain no training samples yet are still assigned classifier vectors in the losses. We find that in this scenario, when class-wise averaged losses reach their minima, standard NC still emerges on the known classes with training samples, while the classifier vectors of unknown classes without training samples overlap with each other and become orthogonal to those of the known classes. Finally, we analyze Softmax- and Sigmoid-based contrastive losses re-balanced by inverse sample frequency, proving that at their minima on balanced or imbalanced datasets, sample features of different classes likewise form standard NC.
\end{abstract}
\section{Introduction}
Neural collapse (NC) \citep{Papyan2020_prevalence,hong2026beyond} refers to the elegant geometric structure formed by the classifier vectors and sample feature vectors at the terminal phase of training a classification model.
Under NC, feature vectors from the same class converge to the class center and align with the corresponding classifier vector, while the class centers are maximally separated and arranged as an equiangular tight frame (ETF).
The NC geometry theoretically corresponds to maximizing within-class compactness and between-class separability of the learned features.
On balanced datasets, extensive empirical and in-depth theoretical researches have shown that NC occurs when cross-entropy (CE) loss \citep{lu2022neural} or binary cross-entropy (BCE) loss \citep{li2025bce} reach their minimum values in training classification models.

In real-world scenarios, sample distributions are typically class-imbalanced.
When training classification models on imbalanced datasets, the training process and the final models performance are easily dominated by majority classes containing large numbers of training samples, making it difficult for the final models to learn well discriminative features and classifier vectors for minority classes containing few training samples \citep{2022_maxnorm,Kang_2020_Decoupling}.
NC provides guidance for empirical classification studies on imbalanced datasets.
To enhance the feature properties and classification accuracy for minority classes of imbalanced datasets, in studies by \citet{yang_2022inducing_nc,kasarla_2022_maximum_NC,liu_2023_inducing}, the classifier vectors were fixed before the model training so that they form an ETF with balanced between-class separability,
and a learnable rotation matrix was employed in the work of \citet{Peifeng_2023_nc_LTR} during the model training to rotate the ETF classifier and align it with the final sample features.
However, the theoretical studies shows that the sample features and classifier vectors of unconstrained feature models (UFM) \citep{fang2021exploring}, trained on imbalanced datasets, cannot attain the standard NC and therefore cannot reach the ETF with maximal and balanced within-class compactness and between-class separability.

On imbalanced datasets that contain only two distinct class scales, when the imbalance factor exceeding a finite threshold, the CE losses' minimizers lead to minority collapse, which means that the classifier vectors of minority classes will collapse into a single one \citep{fang2021exploring}; the sharp value of the imbalance ratio threshold was given by \citet{hong2024neural}.
For contrastive learning on imbalanced datasets, it also results in similar minority collapse \citep{yi2025geometry}.
On general imbalanced datasets, when CE or MSE loss of classification models is minimized, the features from the same class still collapse to their class center, while the norms and directions of the class-center vectors and classifier vectors for different classes depend on the number of samples in each class, exhibiting uneven between-class separability and resulting in asymmetric geometric structures, such as simplex-encoded-labels interpolation (SELI) geometry \citep{Christos2022Imbalance}, $(\delta,R)$-SELI geometry \citep{Tina2023implicit}, and general orthogonal frame (GOF) \citep{Hien2023Neural}, et al.

Some useful techniques can balance the between-class separability of features learned on imbalanced datasets.
If a ReLU activation was integrated into the final layer of the model, then when the supervised contrastive loss \citep{kini2024symmetric} or CE loss \citep{dang2024neural} attain its minima, feature vectors from different classes become orthogonal to each other.
Two NC regularization terms were proposed by \citet{liu_2023_inducing} to equalize feature properties, and empirical results show that integrating them into the CE or MSE loss enables obtaining sample features whose distribution exhibits a symmetric structure on imbalanced datasets.
A generalized classification rule was proposed for classification on imbalanced datasets, which assigns multiple classifier vectors to different classes, and with this rule, the sample features and classifier vectors form NC to multiple centers (NCMC) when the MSE loss reaches its minimum \citep{Hongren2024Neural}.

Most existing NC studies are based on the globally averaged losses, which have shown that the class imbalance distorts the standard NC at the extrema of the losses. More recently, in the work of \citet{wang2026inverse}, the authors solved for the optimal reweighted loss with Tikhonov regularizer on imbalanced datasets and provided a closed-form solution for the optimal weighting factors that form the ideal NC. Compatible with their conclusion, in this paper we directly demonstrate that by re-balancing the losses using the number of samples in each class and optimizing the resulted class-wise averaged losses, the standard NC can be achieved on imbalanced datasets, and even in open-set scenarios.


\section{Preliminaries}\label{sec:preliminaries}
Suppose $\mathcal D = \bigcup_{k=1}^K \mathcal D_k$ be a dataset captured from $K$ classes, where its subset $\mathcal D_k$ contains $n_k = |\mathcal D_k|$ samples from class $k$, for $\forall k\in[K] = \{1,2,\cdots,K\}$.
For any sample $\bm X \in \mathcal D$, a model $\mathcal M$ extracts its feature vector $\bm x = \mathcal M (\bm X) \in \mathbb R^d$;
then $\mathcal F = \bigcup_{k=1}^K\mathcal F_k = \bigcup_{k=1}^K \{\bm x = \mathcal M(\bm X): \bm X \in \mathcal D_k\}$ denotes the feature set.
For convenience, the feature vector is normalized into unit ball with $\|\bm x\|=1$.

\subsection{Classification}
To classify a sample $\bm X$, a linear classifier with a weight matrix $\bm W = [\bm w_1, \cdots, \bm w_K]^\top \in \mathbb R^{K\times d}$ and a bias vector $\bm b = [b_1, \cdots, b_K]^\top \in \mathbb R^K$, is first applied on its feature vector $\bm x = \mathcal M(\bm X)$ and transforms it into a decision vector $\bm z = \gamma\bm W \bm x - \bm b = [\gamma \langle\bm w_k, \bm x\rangle - b_k]_{k=1}^K$, where $\gamma$ is a scale factor, $\|\bm w_k\| =1$ for $\forall k\in[K]$, and $\langle\cdot,\cdot\rangle$ is operation of inner product for any two vectors.
Then, $\bm X$ is classified into class $\hat{k} = \arg\max_{j}\{\gamma \langle\bm w_j, \bm x\rangle - b_j\}_{j=1}^K$,
where $\bm w_k \in\mathbb R^d$ is classifier vector for class $k$.
To correctly classify a sample $\bm X^{(k)}\in \mathcal D_k$ from class $k$, it requires
\begin{align}
    \gamma \bm w_k^\top \bm x^{(k)} - b_k \geq \max\big\{\gamma \bm w_j^\top \bm x^{(k)} - b_j:j\in[K], j\neq k\big\},
    \label{eq_classification_condition_feature}
\end{align}
where $\bm x^{(k)} = \mathcal M(\bm X^{(k)})$ is feature vector of the sample $\bm X^{(k)}$.

In training the classification model, using Softmax, the decision vector $\bm z^{(k)} = [\gamma \langle\bm w_k, \bm x^{(k)}\rangle - b_k]_{k=1}^K$ of $\bm X^{(k)} \in \mathcal D_k$ is transformed into probabilities that the sample belong to each class, and then the probabilities are applied to calculate a loss using cross-entropy (CE),
\begin{align}
    L_{\text{ce}} (\bm X^{(k)}) = -\log\bigg(\frac{\exp\big(\gamma \langle\bm w_k, \bm x^{(k)}\rangle - b_k\big)}{\sum_{j=1}^K\exp\big(\gamma \langle\bm w_j, \bm x^{(k)}\rangle - b_j\big)}\bigg), \label{equ-ce-single-sample}
\end{align}
which has been widely used in deep learning.

\subsection{Feature learning}
In deep learning, Softmax-based contrastive losses, such as SupCon \cite{khosla2020supervised} and InfoNCE \cite{he2020momentum}, are commonly used to train feature extraction models.
However, for feature learning, it expects good within-class compactness and between-class separability among the sample feature vectors.
Intuitively, for feature vectors of class $k$, taking its classifier vector $\bm w_k$ as an anchor, the high compactness requires
\begin{align}
    \gamma \langle \bm w_k, \bm x^{(k)}\rangle > b_k, \quad \forall \bm x^{(k)} \in \mathcal F_k,
    \label{eq_compactness condition_feature}
\end{align}
for the arbitrary reasonable threshold $b_k$.
Using Sigmoid, one can calculate the probability $p_k(\bm x^{(k)}) = \sigma(\gamma\langle\bm w_k, \bm x^{(k)}\rangle - b_k)$ that the compactness condition (\ref{eq_compactness condition_feature}) holds, and binary cross-entropy (BCE) could be used to calculate a loss, $-\log\big(p_k(\bm x^{(k)})\big)$, for the model training.

Similarly, for high separability between class $k$ and class $j$, taking $\bm w_j$ as the anchor, it wishes
\begin{align}
    \gamma \langle \bm w_j, \bm x^{(k)}\rangle < b_j,\quad \forall \bm x^{(k)}\in\mathcal F_k,
    \label{eq_separability_condition_feature}
\end{align}
for the arbitrary reasonable threshold $b_j$.
Then, one can calculate the probability $p_j(\bm x^{(k)}) = 1-\sigma(\gamma\langle\bm w_j, \bm x^{(k)}\rangle - b_j)$ that the separability condition (\ref{eq_separability_condition_feature}) holds, which could be used to calculate a BCE loss, $-\log\big(p_j(\bm x^{(k)})\big)$, for model training.

Totally, for the feature learning, a multi-class BCE loss for the sample $\bm X^{(k)}$ is
\begin{align}
    L_{\text{bce}}(\bm X^{(k)}) = &~\log\big(1+\e^{-\gamma\langle\bm w_k, \bm x^{(k)}\rangle + b_k}\big) + \sum_{j=1, j\neq k}^K \log\big(1 + \e^{\gamma\langle\bm w_j, \bm x^{(k)}\rangle - b_j}\big), \label{equ-bce-single-sample}
\end{align}
which has been widely used in multi-label classification, and we here reveal its relationship to the feature learning.

\subsection{NC on balanced datasets}
On the whole dataset $\mathcal D$, using unconstrained feature model or layer-peeled model \cite{fang2021exploring}, the existing works analyze the minimization of loss functions,
\begin{align}
    \hat{f}_{\mu}(\bm W, \bm b, \mathcal F) = \frac{1}{N} \sum_{k=1}^K \sum_{\bm X^{(k)} \in \mathcal D_k} L_{\mu}(\bm X^{(k)}),
    \label{eq_optimization_global_losses}
\end{align}
where $\mu \in \{\text{ce}, \text{bce}\}$ and $N = \sum_{k = 1}^K n_k$.
On the balanced datasets with $n = n_k = |\mathcal D_k|$ for $\forall k\in[K]$, the current works have proved that the CE \cite{lu2022neural} and BCE \cite{li2025bce} losses lead to standard neural collapse (NC) when reaching their minima in training models.
The standard NC means that,
\begin{itemize}
  \item within the same class $k$ for $\forall k\in[K]$, the sample feature vectors $\mathcal F_k = \{\bm x = \mathcal M(\bm X): \bm X\in\mathcal D_k\}$ collapse to their class center $\bar {\bm x}_k = \frac{1}{n_k}\sum\limits_{\bm x\in \mathcal F_k} \bm x$, i.e., $\bm x = \bar{\bm x}_k$ for $\forall \bm x\in\mathcal F_k$,
  \item for any class $k$, its class center align with the classifier vector, i.e., $\bar{\bm x}_k = \bm w_k$ for $\forall k\in [K]$,
  \item between the classes, the angles of different class centers $\{\bar{\bm x}_k\}_{k=1}^K$ are equal and maximized, forming an ETF, i.e., $\bar{\bm x}_k^\top \bar{\bm x}_j = -\frac{1}{K-1}$ for $\forall k\neq j\in[K]$.
\end{itemize}
When the losses reach their minima, the classifier biases become equal, i.e., $b_j=b_k=b^\star\in\mathbb R$.
Then, the standard NC means that, for any sample, it satisfies classification condition (\ref{eq_classification_condition_feature}) with maximum possible likelihood; meanwhile, across the entire dataset, the within-class compactness of sample features and the between-class separability are both uniformly maximized.
In other words, the classification and feature extraction models trained with CE or BCE possess the optimal theoretical expectations.

\section{Main Results}
\subsection{Neural Collapse on Imbalanced Datasets}
On imbalanced datasets, when the globally averaged losses defined in Eq. (\ref{eq_optimization_global_losses}) reach their minimum values, it is found that the standard NC cannot be achieved.
In this case, although the feature vectors of samples within the same class can still collapse to the class center, the class center is not necessarily aligned with its classifier vector, and the angle between classifier vectors $\{\bm w_k\}_{k=1}^K$ and that between class centers $\{\bar{\bm x}_k\}_{k=1}^K$ vary with the number of samples contained in the corresponding classes.
In practice, the training of models and their final performance are often dominated by the majority classes containing large numbers of training samples.
Although various strategies have been proposed to rebalance their training and performance across classes, none of them can theoretically reformulate the standard NC on imbalanced datasets.

It can be observed that the losses in Eq. (\ref{eq_optimization_global_losses}) are global mean of the CE and BCE losses on the entire dataset $\mathcal D$.
In this paper, we define class-wise averaged losses,
\begin{align}
    f_{\mu}(\bm W, \bm b, \mathcal F) = \frac{1}{K} \sum_{k=1}^K \frac{1}{n_k}\sum_{\bm X^{(k)} \in \mathcal D_k} L_{\mu}(\bm X^{(k)}),
    \label{eq_optimization_local_losses}
\end{align}
which first computes the averaged loss for each class and then calculates the total averaged loss over the $K$ classes.
We prove that, on an imbalanced dataset, the class-wise averaged CE or BCE loss lead to standard NC when reaching their minimum values.

\begin{theorem}\label{theorem_NC_imbalanced_SC}
On a dataset $\mathcal D$ containing $K$ classes, each class contains at least one sample, i.e., $n_k = |\mathcal D_k| > 0$ for $\forall k \in [K]$.
Suppose that the feature vectors extracted by the model $\mathcal M$ have dimensionality $d\geq K-1$.
Then, when the losses $f_{\mu}(\bm W, \bm b, \mathcal F)$ defined in Eq. (\ref{eq_optimization_local_losses}) achieving their minima, they satisfy
    \begin{align}
        &\bm x^{(k)} = \bm w_k,\quad\forall\bm x^{(k)} \in \mathcal F_k,~\forall k\in[K], \label{eq_x=w_nc_imb_sc_mb}\\
        &\langle\bm w_k, \bm w_j\rangle = - \frac1{K-1},\quad \forall k\neq j\in[K], \label{eq_w=w_etf_imb_sc_mb}\\
        &b_k = b_j = b^\star,\quad\forall k,j\in[K], \label{eq_b=b*_imb_sc_mb}
    \end{align}
while, for $\mu = \rm{bce}$, $b^\star$ is the unique real solution of
    \begin{align}
		{\color{black} \e^{\frac{\gamma}{K-1}}\e^{2b} - (K-2)\e^b - (K-1)\e^{\gamma} =0. \label{eq_b*_bce_function}}
    \end{align}
and for $\mu = \rm{ce}$, $b^\star\in\mathbb R$ could be arbitrary number.\hfill$\square$
\end{theorem}
\begin{proof}
	The proofs are provided in Appendix, i.e., Theorems \ref{them:BCE collapse_supp} (BCE) and \ref{them:lemmaB.1} (CE) .
\end{proof}

Eq. (\ref{eq_x=w_nc_imb_sc_mb}) implies that the feature vectors of samples of the $k$-th class not only collapse to their class center but also align with their classifier vector, $\bm w_k$;
Eq. (\ref{eq_w=w_etf_imb_sc_mb}) signifies that both the class centers and the classifier vectors form ETF.
In total, according to Theorem \ref{theorem_NC_imbalanced_SC}, training classification models on dataset $\mathcal D$, balanced or imbalanced, minimizing the class-wise averaged losses induces standard NC.

For the CE and BCE, the difference between their minimizations lies in the classifier biases.
At a minimum point of CE, the classifier biases can take any value as long as the $K$ biases are equal; therefore, setting the classifier biases to 0 in the CE still yields the standard NC.
In contrast, at the minimum point of BCE, the classifier biases have only a unique value, which means that the classifier biases play a substantive role in training models using BCE.
Furthermore, when removing the classifier biases in BCE, the final classifier vector $\bm w_k$ and class center $\bar{\bm x}_k$ cannot be aligned, failing to form a standard NC, except for the case $K=2$ where its theoretical optimal bias is zero.

\subsubsection{The NC-optimal Re-balancing Strategy}
When training models on imbalanced datasets, practitioners commonly use re-balancing strategies such as re-sampling and re-weighting to achieve good performance in practice.
Theorem \ref{theorem_NC_imbalanced_SC} implies that re-weighting the loss using the inverse frequency, $\frac{1}{n_k}$, is theoretically the NC-optimal re-balancing strategy.

In practice, logit adjustment (LA) has been a commonly used re-balancing strategy in model training on imbalanced datasets.
For each sample $\bm X$, LA typically modulates the difference between its positive and negative logits (i.e., the decision scores), or adjust its logits using the sample number corresponding to the class to which the classifier vectors belong, such as
\begin{equation}
    L_{\text{ce-la}} (\bm X^{(k)})= -\log\bigg(\frac{\e^{(\gamma \bm w_k^\top \bm x^{(k)} - b_k)/{\delta_{kk}}}}{\sum_{j=1}^K\e^{(\gamma \bm w_j^\top \bm x^{(k)} - b_j)/{\delta_{kj}}}}\bigg), \nonumber
\end{equation}
\begin{equation}
    L_{\text{bce-la}}(\bm X^{(k)})=\log\Big(1+\e^{(-\gamma\bm w_k^\top \bm x^{(k)} + b_k)/{\delta_{kk}}}\Big)+\sum_{j=1, j\neq k}^K \log\Big(1+\e^{(\gamma\bm w_j^\top \bm x^{(k)} - b_j)/{\delta_{kj}}}\Big), \nonumber
\end{equation}
where $\delta_{kj}$ is typically a quantity related to $n_k$ and $n_j$.

The LA strategy operates at the individual sample level, adjusting the similarities between the feature vector and the imbalanced classifier vectors within the loss function. When the LA loss reaches its minimum, the feature and classifier vectors will form a certain geometric structure, but there is a failure to prove that they would form a standard NC.

\subsection{Neural Collapse in Open-set Scenarios}
In \cite{fang2021exploring}, the authors proved that minority collapse occurs when the imbalance factor (IF) tends to infinity.
In practice, since the number of samples in majority classes cannot be infinite, an IF approaching infinity implies that the minority classes contain zero samples, shifting the classification task into open-set recognition (OSR).
OSR refers to a classification task in which the test or validation set contains samples from classes that do not appear in the training set, meaning that some classes have zero training samples when optimizing the classification loss.
In the loss function for training OSR model, assigning classifier placeholders \citep{ZhouYZ21Learning} or dummy classifiers \citep{liu2022psdc} for unknown classes can improve OSR performance. NC structure was also used to develop OSR method \cite{li2025unlocking}.
In an open-set setting, this paper investigates the geometry of sample features and classifier vectors at the loss minimum.

In our open-set setting, we assume that the first $K_1$ classes in the dataset $\mathcal D$ are \textbf{known classes} containing \emph{non-zero} training samples, with $1<K_1< K$, while the remaining $K-K_1$ classes are \textbf{unknown classes} containing \emph{zero} training samples.
For each class $\ell \in [K]$, regardless of that it is known or unknown class, we assign a classifier vector $\bm w_\ell$ and bias $b_\ell$ to it in training models.
Then, for any sample $\bm X^{(k)}$ from a known class $k\in[K_1]$, its CE and BCE losses are still expressed by Eqs. (\ref{equ-ce-single-sample}) and (\ref{equ-bce-single-sample}).

\begin{theorem}\label{theorem_NC_open_set_BCE_main}
For dataset $\mathcal D=\bigcup_{k=1}^K\mathcal D_k$ with $1<K_1<K$, assume that its last $K-K_1$ classes are empty, i.e., $n_k = |\mathcal D_k| > 0$ for $\forall k \in [K_1]$ and $n_k = |\mathcal D_k| = 0$ for $ K_1 < k \leq K$.
Suppose $d\geq K_1$ and denote $\tilde{\bm x} = \frac1{K_1}\sum_{k=1}^{K_1}\bar{\bm x}_k$ with $\bar{\bm x}_k = \frac1{n_k}\sum_{\bm x\in\mathcal F_k}\bm x$.
Then, for BCE loss
    \begin{align}
        f_{\mathrm{bce}\text{-}\mathrm{os}}(\bm W,\bm{b},\mathcal F)\triangleq \frac{1}{K_1}\sum_{k=1}^{K_1}\frac1{n_k}\sum_{\bm X^{(k)}\in \mathcal D_k} L_{\mathrm{bce}}(\bm X^{(k)}),\label{equ-minf_object_os_bce_main}
    \end{align}
    it satisfies
    \begin{align}
        f_{\mathrm{bce}\text{-}\mathrm{os}}(\bm W,\bm{b},\mathcal F) \geq &~\log\big(1 + \e^{b^\star - \gamma\sqrt{1 - \|\tilde{\bm x}\|^2}}\big) + (K - K_1)\log\big(1 + \e^{- \gamma \|\tilde{\bm x}\| - b^\dag}\big) \nonumber \\
        &~+ (K_1 - 1)\log\big(1 + \e^{- \frac{\gamma}{K_1-1}\sqrt{1 - \|\tilde{\bm x}\|^2} - b^\star}\big), 
        \label{eq_bce_lower_bound_os_main}
    \end{align}
    where $b^\dag\in\mathbb R$ and $b^\star$ is the unique real solution of
    \begin{align}
         \e^{\frac{\gamma}{K_1-1}\sqrt{1-\|\tilde{\bm x}\|^2}}\e^{2b} - (K_1-2)\e^b - (K_1-1)\e^{\gamma \sqrt{1-\|\tilde{\bm x}\|^2}} =0.\nonumber
    \end{align}
    When $f_{\mathrm{bce}\text{-}\mathrm{os}}(\bm W,\bm{b},\mathcal F)$ reaching the lower bound in Eq. (\ref{eq_bce_lower_bound_os_main}),
    \begin{itemize}
      \item (collapse) for $\forall k\in[K_1]$,
          \begin{align}
            \bm x^{(k)} = \bar{\bm x}_k = \frac1{n_k}\sum_{\bm x\in\mathcal F_k} \bm x,\quad\forall \bm x^{(k)} \in\mathcal F_k, \nonumber
          \end{align}
      \item (equiangular) for $k,j\in[K_1]$ with $j\neq k$,
        \begin{align}
            &\bm w_k^\top \bm w_j = -\frac{1}{K_1-1}, \quad \bar{\bm x}_k^\top \bar{\bm x}_j = -\frac{1}{K_1-1} + \frac{K_1}{K_1-1}\|\tilde{\bm x}\|^2, \nonumber
        \end{align}
      \item (isogonal) for $\forall k,j\in[K_1]$ with $j\neq k$,
        \begin{align}
            &\bm w_k^\top \bar{\bm x}_k = \sqrt{1-\|\tilde{\bm x}\|^2}, \quad \bm w_k^\top \bar{\bm x}_j = -\frac1{K_1-1}\sqrt{1-\|\tilde{\bm x}\|^2}, \quad \bm w_k^\top \tilde{\bm x} = 0, \nonumber
        \end{align}
      \item (minority collapse) for $K_1<j\leq K$, if $\|\tilde{\bm x}\|\neq0$,
        \begin{align}
            &\bm w_j = -\frac{\tilde{\bm x}}{\|\tilde{\bm x}\|}, \nonumber
        \end{align}
        and for $K_1<j\leq K$ with $\forall k\in[K_1]$
        \begin{align}
            &\bm w_j^\top \bar{\bm x}_k = -\|\tilde{\bm x}\|, \quad \bm w_j^\top \bm w_k = 0, \nonumber
        \end{align}
    \end{itemize}
    and for $\forall k\in[K_1]$, $K_1<j\leq K$,
    \begin{align}
        b_k = b^\star, \quad
        b_j = b^\dag.\nonumber
    \end{align}

    Furthermore, when $f_{\mathrm{bce}\text{-}\mathrm{os}}(\bm W,\bm{b},\mathcal F)$ approaching its minimum value, $b^\dag\rightarrow+\infty$ and $\|\tilde{\bm x}\|\rightarrow0$.\hfill$\square$
\end{theorem}
\begin{proof}
	The proof is given in Appendix, i.e., Theorem \ref{them:BCE collapse_supp_open_set}.
\end{proof}

\begin{theorem}\label{theorem_NC_open_set_CE_main}
For dataset $\mathcal D=\bigcup_{k=1}^K\mathcal D_k$ with $1<K_1<K$, assume that its last $K-K_1$ classes are empty, i.e., $n_k = |\mathcal D_k| > 0$ for $\forall k \in [K_1]$ and $n_k = |\mathcal D_k| = 0$ for $ K_1 < k \leq K$.
Suppose $d\geq K_1$ and denote $\tilde{\bm x} = \frac1{K_1}\sum_{k=1}^{K_1}\bar{\bm x}_k$ with $\bar{\bm x}_k = \frac1{n_k}\sum_{\bm x\in\mathcal F_k}\bm x$.
Then, for CE loss
    \begin{align}
        f_{\mathrm{ce}\text{-}\mathrm{os}}(\bm W, \bm b, \mathcal F) \triangleq
        \frac1{K_1}\sum_{k=1}^{K_1}\frac1{n_k}\sum_{\bm X^{(k)}}L_{\mathrm{ce}}(\bm X^{(k)}), \label{equ-minf_object-ce_os_main}
    \end{align}
    it satisfies
    \begin{align}
        f_{\mathrm{ce}\text{-}\mathrm{os}}(\bm W, \bm b, \mathcal F) \geq \log(1+c_1+c_2), \label{eq_ce_lower_bound_os_main}
    \end{align}
    where
    \begin{align}
        c_1 &~= (K_1-1)\exp\bigg(\frac{\gamma}{\sqrt{g(q^\star)}}\frac{K_1}{K_1-1}\big(\|\tilde{\bm x}\|^2 - 1\big)\bigg)\nonumber\\
        c_2 &~= K_1\frac{1-q^\star}{q^\star}\exp\bigg(\frac{\gamma}{\sqrt{g(q^\star)}}\frac{K_1}{K_1-1}\big(\|\tilde{\bm x}\|^2 - 1\big)\bigg)\nonumber\\
        c_3 &~= \frac{1}{q^\star}\sqrt{g(q^\star)}\exp\bigg(\frac{\gamma}{\sqrt{g(q^\star)}}\frac{K_1}{K_1-1}\big(\|\tilde{\bm x}\|^2 - 1\big)\bigg), \nonumber
    \end{align}
    and $g(q) = 1-2 q \|\tilde{\bm x}\|^2 + q^2 \|\tilde{\bm x}\|^2$ with a certain $q^\star\in(0,1)$.
    When $f_{\mathrm{ce}\text{-}\mathrm{os}}(\bm W, \bm b, \mathcal F)$ reaching the lower bound in Eq. (\ref{eq_ce_lower_bound_os_main}),
    \begin{itemize}
      \item (collapse) for $\forall k\in [K_1],$
        \begin{align}
            &\bm x^{(k)} = \bar{\bm x}_k = \frac1{n_k}\sum_{\bm x\in\mathcal F_k} \bm x,\quad\forall \bm x^{(k)}\in\mathcal F_k,\nonumber
        \end{align}
      \item (equiangular) for $\forall k,j\in[K_1]$ with $j\neq k$,
        \begin{align}
            &\bm w_k^\top \bm w_j = - \frac{1}{K_1-1} +
        \frac{1}{K_1}\frac{1}{K_1-1}\big(\frac{c_2}{c_3}\big)^2\|\tilde{\bm x}\|^2,\nonumber\\
            &\bar{\bm x}_k^\top\bar{\bm x}_j = -\frac1{K_1-1}+\frac{K_1}{K_1-1}\|\tilde{\bm x}\|^2,\nonumber
        \end{align}
      \item (isogonal) for $\forall k\in[K_1]$ and $\forall j\in [K_1]$ with $j\neq k$,
        \begin{align}
            \bm w_k^\top \bar{\bm x}_k =&~  \bigg(\frac{c_1}{K_1-1}+\frac{c_2}{K_1}\bigg)\frac{1}{c_3} - \frac{c_1}{K_1-1}\frac1{c_3}\|\tilde{\bm x}\|^2, \quad \bm w_k^\top \tilde{\bm x} =  \frac{c_2}{c_3}\frac{\|\tilde{\bm x}\|^2}{K_1}, \nonumber\\
            \bm w_k^\top \bar{\bm x}_j =&~ \frac{1}{c_3}\bigg(\frac{1}{K_1-1}\frac{c_1}{K_1-1} + \frac{K_1}{K_1-1}\frac{c_2}{K_1}\bigg)\|\tilde{\bm x}\|^2 - \frac{1}{c_3}\frac{1}{K_1-1}  \bigg(\frac{c_1}{K_1-1} + \frac{c_2}{K_1}\bigg),  \nonumber \\
        \end{align}
      \item (minority collapse) for $K_1 < j \leq K$, if $\|\tilde{\bm x}\|\neq0$,
        \begin{align}
            &\bm w_j = -\frac{\tilde{\bm x}}{\|\tilde{\bm x}\|},\nonumber
        \end{align}
        and for $K_1 < j \leq K$ with $\forall k\in[K_1]$
        \begin{align}
            &\bm w_j^\top \bar{\bm x}_k = -\|\tilde{\bm x}\|, \quad \bm w_j^\top \bm w_k = -\frac{c_2}{c_3}\frac{\|\tilde{\bm x}\|}{K_1}, \nonumber
        \end{align}
    \end{itemize}
    and for $\forall k\in[K_1]$, $K_1<j\leq K$, $\exists b^\star, b^\dag\in\mathbb R$, such that
    \begin{align}
            b_k = b^\star,\quad
            b_j = b^\dag.\nonumber
    \end{align}

    Furthermore, when $f_{\mathrm{ce}\text{-}\mathrm{os}}(\bm W, \bm b, \mathcal F)$ approaching its minimum value, $b^\dag - b^\star\rightarrow+\infty$, $q^\star\rightarrow1$, and $\|\tilde{\bm x}\|\rightarrow0$.\hfill$\square$
\end{theorem}
\begin{proof}
	The complete proof is given in Appendix, i.e., Theorem \ref{them:lemmaB.1_OS}.
\end{proof}

According to Theorems \ref{theorem_NC_open_set_BCE_main} and \ref{theorem_NC_open_set_CE_main}, in the open-set scenarios, before the BCE and CE losses approach their global minima, they respectively reach the lower bound in Eqs. (\ref{eq_bce_lower_bound_os_main}) and (\ref{eq_ce_lower_bound_os_main}), at which point their classifier vectors and feature vectors have already formed relatively concise geometric structures.
In the geometric structures, similar to standard NC, sample features $\{\bm x = \mathcal M(\bm X): \bm X\in\mathcal D_k\}$ of a certain known class $k$ collapse to the class center $\bar{\bm x}_k$; the difference is that although the class means $\{\bar{\bm x}_k\}_{k=1}^{K_1}$ and classifier vectors $\{\bm w_k\}_{k=1}^{K_1}$ of $K_1$ known classes form equiangular frames respectively, they do not necessarily have the maximum pairwise angles, $\arccos[-1/(K_1-1)]$.
In the geometric structures, the classifier vectors $\{\bm w_j: K_1< j\leq K\}$ of the unknown class coincide with each other and are opposite to the direction of the global feature mean $\tilde{\bm x}$.
Simultaneously, the classifier biases of known and unknown classes converge to specific values, $b^\star$ and $b^\dag$.

In the structures, the classifier vectors $\{\bm w_k\}_{k=1}^{K_1}$ of different known classes under the BCE have a uniform and largest angular separation, whereas those under the CE are uniform but with a smaller angular separation.
Between the known and unknown classes, their classifier vectors, $\{\bm w_k\}_{k=1}^{K_1}$ and $\{\bm w_j: K_1<j\leq K\}$, under the BCE are orthogonal, while they under the CE have a larger angular separation

As BCE and CE respectively approach their minima in the open-set scenario, the global feature mean $\tilde{\bm x}$ tend toward the zero vector. At this point, standard NC still forms over the $K_1$ known classes.
Theoretically, the classifier vectors $\{\bm w_j: K_1< j\leq K\}$ of unknown classes always coincide with each other and remain perpendicular to the classifier vectors $\{\bm w_k\}_{k=1}^{K_1}$ of known classes under the BCE, while under the CE they gradually approach perpendicularity to $\{\bm w_k\}_{k=1}^{K_1}$.
Meanwhile, as the losses approach their minima, $\{\bm w_j: K_1< j\leq K\}$ consistently remains in the direction opposite to the global feature mean $\tilde{\bm x}$ in theoretical, i.e., $\bm w_j = -\frac{\tilde{\bm x}}{\|\tilde{\bm x}\|}$; however, in practice, as $\|\tilde{\bm x}\|$ approaches $0$, any random perturbation can dramatically change the direction of $\tilde{\bm x}$, thereby altering its angle with $\bm w_j$.
\subsection{Extensions to Contrastive Losses}
In contrastive learning, the common losses used for training feature extraction models include Softmax-based losses such as SupCon \cite{khosla2020supervised} and N-pair loss \cite{sohn2016improved}, as well as Sigmoid-based losses such as MS loss \cite{wang2019multi} and SimPLE \cite{wen2023pairwise}.
The Softmax-based contrastive loss originates from noise contrastive estimation \citep{oord2018representation}.
In this paper, for $\forall~\bm X^{(k)}$ from class $k$, $\forall k\in[K]$, we define its Softmax-based contrastive loss as
    \begin{align}
        &~L_{\text{soft}}(\bm X^{(k)}) = -\frac{1}{n_k}\sum_{\bm x_\text{p} \in \mathcal F_k}\log\bigg(\frac{\e^{\gamma\bm x_\text{p}^\top \bm x^{(k)}}}{\sum_{j=1}^K\frac{1}{n_j}\sum_{\bm x\in\mathcal F_j}\e^{\gamma\bm x^\top \bm x^{(k)}}}\bigg), \label{eq_softmax_CL_single_sample}
    \end{align}
where $\bm x^{(k)} = \mathcal M(\bm X^{(k)})$ and we apply the numbers of positive and negative sample pairs to re-balance the model training.
After we completed the proof of Theorem \ref{theorem_NC_imbalanced_CL} and began drafting this paper, we found that Zhu et al. \cite{zhu2022balanced} established on each batch of samples a re-balanced Softmax-based contrastive loss in the batch algorithm, similar to Eq. (\ref{eq_softmax_CL_single_sample}), and they proved that it can induce standard NC.

Similar to the BCE loss in Eq. (\ref{equ-bce-single-sample}), one can define Sigmoid-based contrastive loss from the first principle of feature learning.
For any sample $\bm X^{(k)}$ from class $k$, with $\bm x^{(k)} = \mathcal M(\bm X^{(k)})$, it is expected that the feature similarity $\bm x_{\text{p}}^\top\bm x^{(k)}$ with feature vector $\bm x_{\text{p}}$ of any positive sample from the same class exceeds any reasonable threshold $t_k = b_k/\gamma$, while the feature similarity $\bm x_{\text{n}}^\top\bm x^{(k)}$ between it and $\bm x_{\text{n}}$ of any negative sample from any other class, such as $j$ with $j\neq k$, is below any reasonable threshold $t_j = b_j / \gamma$.
Then, we apply Sigmoid to transform $\gamma \bm x_{\text{p}}^\top \bm x^{(k)} - b_k$ and $\gamma \bm x_{\text{n}}^\top \bm x^{(k)}-b_j$ into probabilities that meet the above expectations, and take BCE to construct the Sigmoid-based contrastive loss for model training.
In total, for any sample $\bm X^{(k)}$ from class $k$,
\begin{align}
    &L_{\text{sigm}}(\bm X^{(k)})
   =\frac{1}{n_k}\sum_{\bm x_\text{p} \in \mathcal F_k}\log\big(1 + \e^{- \gamma \bm x_\text{p}^\top \bm x^{(k)} + b_k}\big) + \lambda\sum_{j=1, j\neq k}^K\frac{1}{n_j}\sum\limits_{\bm x_\text{n}\in\mathcal F_j}\log\big(1+\e^{\gamma \bm x_\text{n}^\top \bm x^{(k)} - b_j}\big), \label{eq_sigmoid_CL_single_sample}
\end{align}
where $1/n_k$ and $1/n_j$ are used to eliminate the impact of the different numbers of positive sample pairs and negative ones of sample $\bm X^{(k)}$ on the balanced training, and $\lambda > 0$ is used to adjust the learning rate of the feature similarities of negative sample pairs that reflecting between class separability.

\begin{theorem}\label{theorem_NC_imbalanced_CL}
Suppose that each class in dataset $\mathcal D$ contains at least two samples, and that the feature dimension is greater than the number of classes, i.e., $n_k = |\mathcal D_k| > 2$ for $\forall k \in [K]$ and $d \geq K-1$.
On the dataset $\mathcal D$, define losses
\begin{align}
    f_{\mu}(\mathcal F) = \frac{1}{K}\sum_{k=1}^K\frac{1}{n_k}\sum_{\bm X^{(k)}\in\mathcal D_k} L_{\mu}(\bm X^{(k)}),
    \label{eq_losses_CL_dataset}
\end{align}
where $\mu \in \{\mathrm{soft}, \mathrm{sigm}\}$.
Then, at its minimum point,
    \begin{align}
        &\bm x^{(k)} = \frac{1}{n_k}\sum_{\bm x\in\mathcal F_k} \bm x = \bar{\bm x}_k ,\quad\forall\bm x^{(k)} \in \mathcal F_k,~\forall k\in[K],\\
        &\langle\bar{\bm x}_k,\bar{\bm x}_j\rangle = -\frac1{K-1},\quad\forall k\neq j\in[K],
    \end{align}
and for the Sigmoid-based loss, it also satisfies
    \begin{align}
        b_k = b^\star,\quad\forall k\in[K],
    \end{align}
where $b^\star$ is the unique solution of
    \begin{align}
        \e^{\frac{\gamma}{K-1}}\e^{2b} - \big[\lambda(K-1)-1\big]\e^{b} - \lambda(K-1)\e^{\gamma} = 0.
    \end{align}
\end{theorem}
\begin{proof}
	The proofs are provided in Appendix, i.e., Theorems \ref{theorem_NC_imbalanced_CL_sigmoid_supp} (Sigmoid) and \ref{theorem_NC_imbalanced_CL_softmax_supp}
	(Softmax).
\end{proof}

Similar to Eq. (\ref{eq_optimization_local_losses}) in Theorem \ref{theorem_NC_imbalanced_SC}, which re-balances the model training using the inverse frequency of each class, Eq. (\ref{eq_losses_CL_dataset}) in Theorem \ref{theorem_NC_imbalanced_CL} re-balances the model training by using the reciprocal of the number of sample pairs $(n_k,n_j)$ that can be formed by any two classes, such as classes $k$ and $j$.


As shown in Eq. (\ref{eq_sigmoid_CL_single_sample}), in the Sigmoid-based contrastive loss, a weighting parameter $\lambda$ can be flexibly employed to balance the learning rates of feature similarities of positive and negative sample pairs that reflecting within-class compactness and between-class separability. In contrast, the different feature similarities in Eq. (\ref{eq_softmax_CL_single_sample}) are coupled together through Softmax, making it difficult to flexibly adjust their learning intensity in the Softmax-based contrastive loss.


Furthermore, although the biases in the Sigmoid-based contrastive loss still play a substantive role during the feature learning, it can be proved that if the they are removed, the main conclusion of Theorem \ref{theorem_NC_imbalanced_CL} remains valid.


\section{Experiments}\label{sec:experiments}

\begin{table*}[!htb]
\centering
\begin{tabular}{c|cc|cccc}
\hline
\multirow{2}{*}{Loss} & \multirow{2}{*}{Dataset} & \multirow{2}{*}{IF} & \multicolumn{4}{c}{ResNet18} \\
\cline{4-7}
 &  &  & $s^{\text{xx}}_{\text{pos}}$ & $s^{\text{wx}}_{\text{pos}}$ & $s^{\text{wx}}_{\text{neg}}$ & $b$ \\
\hline
\multirow{10}{*}{CE} & \multirow{5}{*}{CIFAR10} & 1 & $.9799\pm.0029$ & $.9895\pm.0015$ & $-.1099\pm.0079$ & $.0000\pm.0221$ \\
 &  & 10 & $.9778\pm.0056$ & $.9886\pm.0028$ & $-.1098\pm.0081$ & $.0000\pm.0131$ \\
 &  & 50 & $.9709\pm.0113$ & $.9850\pm.0056$ & $-.1093\pm.0105$ & $.0000\pm.0246$ \\
 &  & 100 & $.9697\pm.0134$ & $.9843\pm.0067$ & $-.1092\pm.0110$ & $.0000\pm.0347$ \\
\cdashline{2-7}
 & \multicolumn{2}{c|}{NC theor. value} & $1.0$ & $1.0$ & $-1/9\approx-.1111$ & — \\
\cline{2-7}
 & \multirow{4}{*}{CIFAR100} & 1 & $.9941\pm.0018$ & $.9967\pm.0010$ & $-.0101\pm.0306$ & $.0000\pm.0041$ \\
 &  & 10 & $.9914\pm.0033$ & $.9953\pm.0017$ & $-.0100\pm.0787$ & $.0000\pm.0148$ \\
 &  & 50 & $.9898\pm.0063$ & $.9945\pm.0032$ & $-.0100\pm.0728$ & $.0000\pm.0240$ \\
 &  & 100 & $.9888\pm.0081$ & $.9940\pm.0041$ & $-.0100\pm.0737$ & $.0000\pm.0259$ \\
\cdashline{2-7}
 & \multicolumn{2}{c|}{NC theor. value} & $1.0$ & $1.0$ & $-1/99\approx-.0101$ & — \\
\hline
\multirow{10}{*}{BCE} & \multirow{5}{*}{CIFAR10} & 1 & $.9993\pm.0001$ & $.9996\pm.0001$ & $-.1111\pm.0031$ & $4.669\pm.0009$ \\
 &  & 10 & $.9989\pm.0005$ & $.9994\pm.0002$ & $-.1110\pm.0028$ & $4.668\pm.0040$ \\
 &  & 50 & $.9960\pm.0020$ & $.9979\pm.0010$ & $-.1109\pm.0036$ & $4.663\pm.0101$ \\
 &  & 100 & $.9967\pm.0018$ & $.9983\pm.0009$ & $-.1109\pm.0034$ & $4.664\pm.0073$ \\
\cdashline{2-7}
 & \multicolumn{2}{c|}{NC theor. value} & $1.0$ & $1.0$ & $-1/9\approx-.1111$ & $4.6698$ \\
\cline{2-7}
 & \multirow{5}{*}{CIFAR100} & 1 & $.9993\pm.0007$ & $.9994\pm.0003$ & $-.0101\pm.0025$ & $6.330\pm.0009$ \\
 &  & 10 & $.9989\pm.0004$ & $.9993\pm.0002$ & $-.0101\pm.0021$ & $6.328\pm.0042$ \\
 &  & 50 & $.9978\pm.0014$ & $.9987\pm.0007$ & $-.0101\pm.0154$ & $6.329\pm.0066$ \\
 &  & 100 & $.9971\pm.0021$ & $.9983\pm.0011$ & $-.0101\pm.0246$ & $6.332\pm.0075$ \\
\cdashline{2-7}
 & \multicolumn{2}{c|}{NC theor. value} & $1.0$ & $1.0$ & $-1/99\approx-.0101$ & $6.3437$ \\
\hline
\end{tabular}
\caption{Statistical results (mean $\pm$ std) for positive/negative similarities ($s_{\text{pos}}$/$s_{\text{neg}}$) and biases ($b$) of different models trained by CE and BCE on CIFAR10-LT and CIFAR100-LT with $\text{IF} = 1, 10, 50$, and $100$.
	The CE and BCE are class-wise averaged using Eq. (\ref{eq_optimization_local_losses}). In the table, when the absolute value of a number if less than $1$, we omit the `0' in its integer part to save space.
}
\label{table_results_on_datasets_wo_empty}
\end{table*}

\begin{table*}[!htb]
	\centering
	\begin{tabular}{c|cc|cccc}
		\hline
		\multirow{2}{*}{Loss} & \multirow{2}{*}{Dataset} & \multirow{2}{*}{IF} & \multicolumn{4}{c}{ViT-Tiny} \\
		\cline{4-7}
		& & & $s^{\text{xx}}_{\text{pos}}$ & $s^{\text{wx}}_{\text{pos}}$ & $s^{\text{wx}}_{\text{neg}}$ & \multicolumn{1}{c}{$b$} \\
		\hline
		\multirow{10}{*}{CE} & \multirow{5}{*}{CIFAR10} & 1 & $.9997\pm.0000$ & $.9999\pm.0000$ & $-.1111\pm.0004$ & $.0011\pm.0012$ \\
		&  & 10 & $.9993\pm.0001$ & $.9997\pm.0001$ & $-.1111\pm.0003$ & $-.0004\pm.0013$ \\
		&  & 50 & $.9992\pm.0001$ & $.9996\pm.0000$ & $-.1111\pm.0004$ & $-.0002\pm.0012$ \\
		&  & 100 & $.9985\pm.0002$ & $.9993\pm.0001$ & $-.1110\pm.0003$ & $-.0008\pm.0008$ \\
		\cdashline{2-7}
		& \multicolumn{2}{c|}{NC theor. value} & $1.0$ & $1.0$ & $-1/9\approx-.1111$ & \multicolumn{1}{c}{—} \\
		\cline{2-7}
		& \multirow{4}{*}{CIFAR100} & 1 & $.9893\pm.0052$ & $.9942\pm.0026$ & $-.0101\pm.0070$ & $.0208\pm.0041$ \\
		&  & 10 & $.9999\pm.0000$ & $1.000\pm.0000$ & $-.0101\pm.0005$ & $.0145\pm.0013$ \\
		&  & 50 & $.9999\pm.0000$ & $.9999\pm.0000$ & $-.0101\pm.0007$ & $.0145\pm.0014$ \\
		&  & 100 & $.9968\pm.0007$ & $.9984\pm.0004$ & $-.0101\pm.0067$ & $.0142\pm.0072$ \\
		\cdashline{2-7}
		& \multicolumn{2}{c|}{NC theor. value} & $1.0$ & $1.0$ & $-1/99\approx-.0101$ & \multicolumn{1}{c}{—} \\
		\hline
		\multirow{10}{*}{BCE} & \multirow{5}{*}{CIFAR10} & 1 & $.9999\pm.0000$ & $.9999\pm.0000$ & $-.1111\pm.0016$ & $4.670\pm.0003$ \\
		&  & 10 & $1.000\pm.0000$ & $1.000\pm.0000$ & $-.1111\pm.0012$ & $4.670\pm.0002$ \\
		&  & 50 & $.9996\pm.0002$ & $.9998\pm.0001$ & $-.1111\pm.0007$ & $4.675\pm.0092$ \\
		&  & 100 & $.9945\pm.0046$ & $.9954\pm.0038$ & $-.1109\pm.0162$ & $4.658\pm.0507$ \\
		\cdashline{2-7}
		& \multicolumn{2}{c|}{NC theor. value} & $1.0$ & $1.0$ & $-1/9\approx-.1111$ & \multicolumn{1}{c}{$4.6698$} \\
		\cline{2-7}
		& \multirow{5}{*}{CIFAR100} & 1 & $.9963\pm.0036$ & $.9980\pm.0019$ & $-.0101\pm.0037$ & $6.337\pm.0017$ \\
		&  & 10 & $.9999\pm.0000$ & $.9999\pm.0000$ & $-.0101\pm.0010$ & $6.344\pm.0011$ \\
		&  & 50 & $.9996\pm.0001$ & $.9998\pm.0000$ & $-.0101\pm.0008$ & $6.342\pm.0018$ \\
		&  & 100 & $.9983\pm.0010$ & $.9992\pm.0005$ & $-.0101\pm.0088$ & $6.325\pm.0079$ \\
		\cdashline{2-7}
		& \multicolumn{2}{c|}{NC theor. value} & $1.0$ & $1.0$ & $-1/99\approx-.0101$ & \multicolumn{1}{c}{$6.3437$} \\
		\hline
	\end{tabular}
	\noindent\textbf{Table~\ref{table_results_on_datasets_wo_empty} (continued)} \par
\end{table*}

To verify the theorems, we applied the CE and BCE losses to train models such as ResNet and ViT on CIFAR10 and CIFAR100 \citep{krizhevsky2009learning}, respectively.
The detailed experimental settings can be found in the supplementary.

In each experiment, after the model training, we collect the following sets of different types of values,
\begin{align}
    &s_{\mathrm{pos}}^{\text{xx}}=\big\{\bm{x}^{\top}\bm{x}_*: \forall \bm x, \bm x_* \in\mathcal F_k, \forall k\in[K]\big\}, \quad s_{\mathrm{neg}}^{\text{xx}}=\big\{\bar{\bm{x}}^{\top}_k\bar{\bm{x}}_j: \forall k\neq j \in[K]\big\}, \nonumber \\
    &s_{\mathrm{pos}}^{\text{wx}}=\big\{\bm{w}_k^{\top}\bar{\bm{x}}_k: \forall k\in[K]\big\}, \quad s_{\mathrm{neg}}^{\text{wx}}=\big\{\bm{w}_j^{\top}\bar{\bm{x}}_k: \forall j\neq k\in[K]\big\}, \nonumber \\
    &s_{\mathrm{neg}}^{\text{ww}}=\big\{\bm{w}_j^{\top}\bm{w}_k: \forall j\neq k\in[K]\big\}. \nonumber
\end{align}
According to Theorem \ref{theorem_NC_imbalanced_SC}, when the losses reaching their minima, the numbers in $s_{\mathrm{pos}}^{\text{xx}}$ and $s_{\mathrm{pos}}^{\text{wx}}$ will converge to $1$, while that in $s_{\mathrm{neg}}^{\text{xx}}, s_{\mathrm{neg}}^{\text{wx}}$, and $s_{\mathrm{neg}}^{\text{ww}}$ will converge to $-\frac1{K-1}$.

In open-set scenario, we collect the above sets on the first $K_1$ non-empty classes, and the numbers will converge to specific values related to $K_1$ and $\|\tilde{\bm x}\|$, given by Theorems \ref{theorem_NC_open_set_BCE_main} and \ref{theorem_NC_open_set_CE_main}.
Besides that, for the last $K-K_1$ empty unknown classes, we collect the sets as
\begin{align}
    &s_{\mathrm{emp}}^{\text{ww}}=\big\{{\bm{w}_j^{\top}{\bm{w}}_{j'}}: K_1 < j\neq j'\leq K \big\},\nonumber\\
    &s_{\mathrm{emp2}}^{\text{ww}}=\big\{{\bm{w}_k^{\top}{\bm{w}}_{j}}: k\in[K_1], K_1 < j\leq K \big\},\nonumber\\
    &s_{\mathrm{emp}}^{\text{wx}}=\big\{{\bm{w}_j^{\top}\tilde{\bm{x}}}/{\|\tilde{\bm x}\|}: K_1 < j\leq K \big\},\nonumber
\end{align}
and in these three sets, the numbers will respectively converge to $1, 0$, and $-1$ when minority collapse occurs.

\subsection{On Imbalanced Datasets}
We first verify Theorem \ref{theorem_NC_imbalanced_SC}.
Table \ref{table_results_on_datasets_wo_empty} shows the statistical results of $s_{\mathrm{pos}}^{\text{xx}}$, $s_{\mathrm{pos}}^{\text{wx}}$, $s_{\mathrm{neg}}^{\text{wx}}$, and that of biases $\{b_j\}_{j=1}^K$ for ResNet18 and ViT-Tiny trained on training data of CIFAR10-LT and CIFAR100-LT \citep{cao2019learning} with different imbalanced factors, IF $\in\{1, 10, 50, 100\}$, where the model training is driven using the class-wise (CW) averaged CE and BCE defined in Eq. (\ref{eq_optimization_local_losses}).
In the experiments, {\color{black} we trained the models for $800$ epochs with a batch size of $128$. The optimizer and learning rate setting are detailed in Table \ref{tab_supp_supervised_settings}.}
No weight decay was applied. The classifier biases were initialized to $0.0$, and the scale factor $\gamma$ was set to $8.0$.

According to Theorem \ref{theorem_NC_imbalanced_SC}, when the CW losses reaching their global minima, the features of samples of the same class align with their corresponding classifier vectors, while different classes form an ETF.
From Table \ref{table_results_on_datasets_wo_empty}, one can find that the results are highly consistent with the theoretical analysis, regardless of whether the CE or BCE is used to train the ResNet or ViT.
On CIFAR10-LT, except for the data of ResNet18 trained with CE, which deviates slightly more from their theoretical values, the means of $s_{\mathrm{pos}}^{\text{xx}}$ and $s_{\mathrm{pos}}^{\text{wx}}$ for the other models are basically all around $0.99$, very close to the theoretical value of $1$, while the means of the $s_{\mathrm{neg}}^{\text{wx}}$ are all around $-0.11$, close to the theoretical value of $-1/9$; moreover, these all data have relatively small standard deviations, even IF $=100$.
Similar observations can also be made on CIFAR100-LT.
These results indicate that, regardless of the imbalance ratio, class number, or the model architecture, training a model with the CW averaged loss ultimately causes the sample features and classifier vectors of different classes to form an ETF, reformulating standard NC on imbalanced datasets.

Furthermore, the results of final classifier biases $\{b_k\}_{k=1}^K$ in Table \ref{table_results_on_datasets_wo_empty} illustrate the distinction between the CE and BCE losses.
For the CE loss, Theorem \ref{theorem_NC_imbalanced_SC} only requires the biases of different classes to converge to the same value, which can be any real number, with no unique theoretical value. In the experiments, since the initial values were set to zero mean, after training the models using CE, their all classifier biases still concentrated around $0.0$.
In contrast, when BCE loss reaches a minimum, there is a unique theoretical classifier bias. As shown in Table \ref{table_results_on_datasets_wo_empty}, the theoretical bias values on CIFAR10 and CIFAR100 are $4.6698$ and $6.3437$, respectively. The final biases obtained in practical training concentrate near these two values and exhibit small standard deviations across different models and datasets with different IFs.


\begin{figure*}[!htb]
	\centering
	\includegraphics[width=0.49\linewidth]{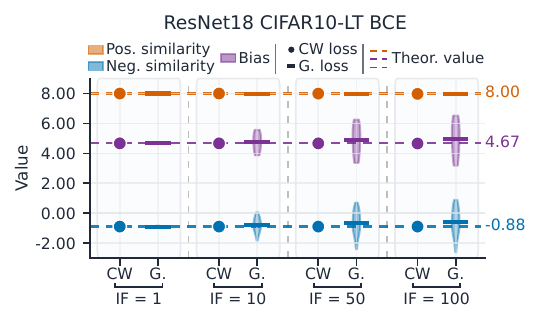}%
	\hfill
	\includegraphics[width=0.49\linewidth]{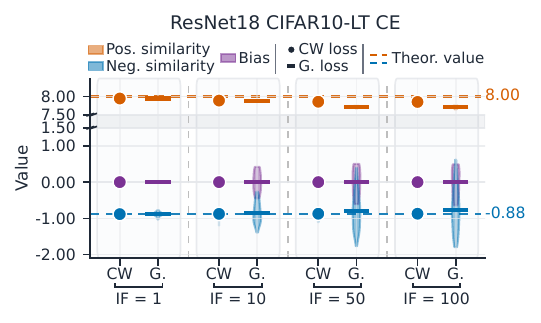}
	\caption{Distribution of positive and negative scaled similarities ($\gamma\boldsymbol{w}_k^{\top}\boldsymbol{\bar x}_j$) and biases of ResNet18 trained by class-wise (CW) and globally (G.) averaged CE and BCE.}
	\label{fig_dis_resnet18_cifar10}
\end{figure*}
To demonstrate the theoretical advantages of the class-wise (CW) averaged losses defined in Eq. (\ref{eq_optimization_local_losses}), we also train the models using the globally (G.) averaged losses defined in Eq. (\ref{eq_optimization_global_losses}), with $\gamma=8.0$.
Using violin plots, Fig. \ref{fig_dis_resnet18_cifar10} shows the distributions of positive/negative scaled similarities ($\gamma \bm w^\top \bm x$) and biases of ResNet18 trained by the different losses on CIFAR10-LT with IF $= 1, 10, 50$, and $100$.
When IF$=1$, the CW loss and the G. loss are equivalent, and their final results are both close to NC solution.
However, as IF increases, there is a significant difference between the results of these two types of losses.
With different IFs, the three metrics obtained by the CW averaged loss are all concentrated around the NC theoretical values.
In contrast, for the G. averaged losses, as IF increases, although the positive similarities are still mainly concentrated around the theoretical value, the negative similarities and classifier biases become increasingly dispersed and increasingly difficult to converge.
These results demonstrate the theoretical advantages of the CW averaged losses over the G. averaged loss, i.e., on imbalanced datasets, the former can achieve standard NC, theoretically maximizing the between-class compactness of sample features and, in a balanced manner, maximizing the between-class separability between different sample features and classifier vectors of different classes, whereas the latter cannot.

\begin{figure*}[tbp]
	\centering
	\subfigure[Similarity matrices learned by class-wise (CW) averaged BCE loss defined in Eq. \ref{eq_optimization_local_losses}.]{
		\includegraphics*[scale=0.36, viewport=5 0 355 275]{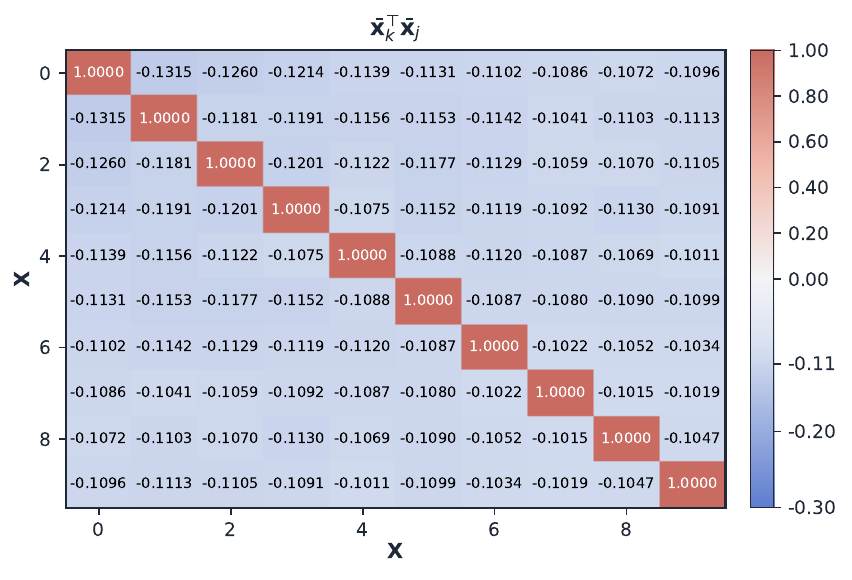}
		\hspace{-1.5mm}
		\includegraphics*[scale=0.36, viewport=5 0 355 275]{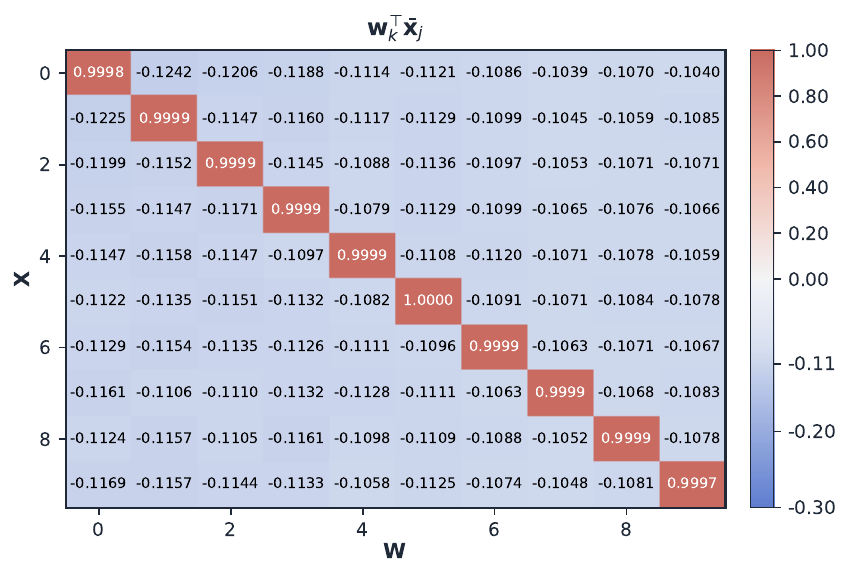}
		\hspace{-1.5mm}
		\includegraphics*[scale=0.36, viewport=5 0 440 275]{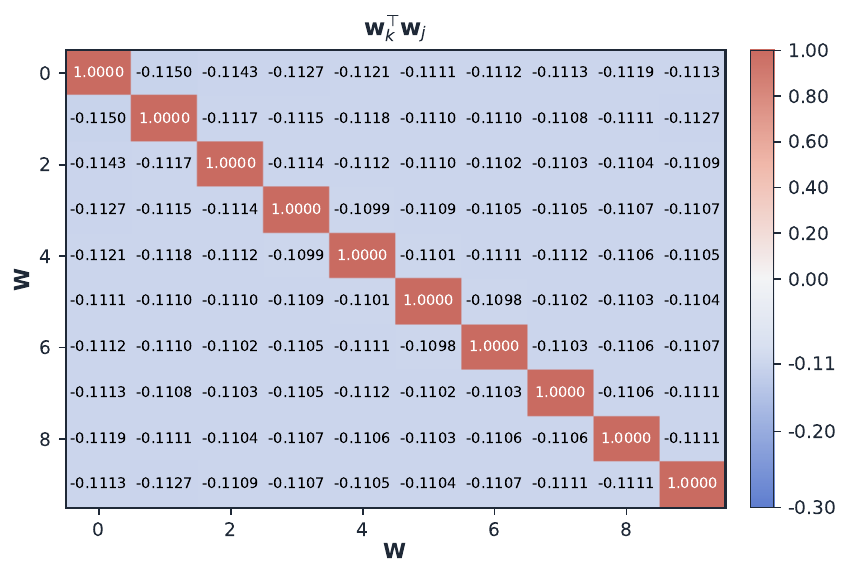}
	}
	\subfigure[Similarity matrices learned by global (G.) averaged BCE loss defined in Eq. \ref{eq_optimization_global_losses}.]{
		\includegraphics*[scale=0.36, viewport=5 0 355 275]{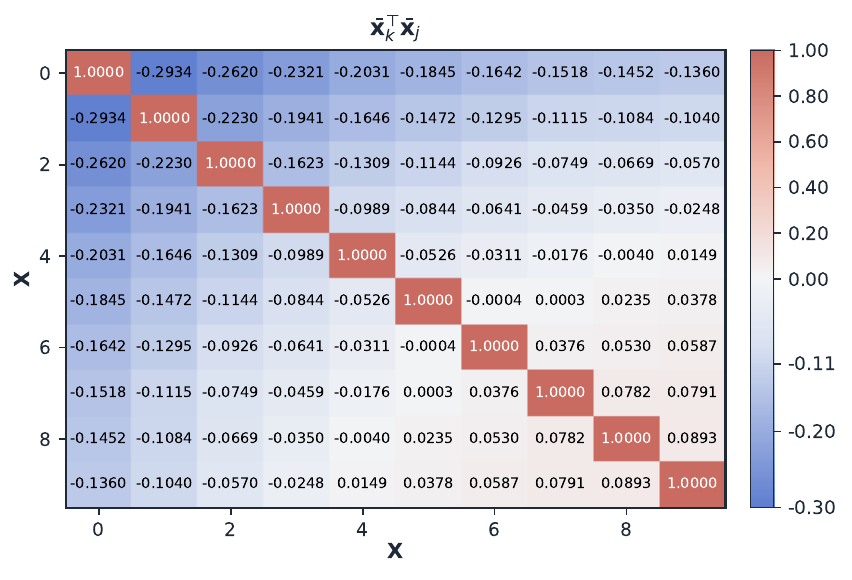}
		\hspace{-1.5mm}
		\includegraphics*[scale=0.36, viewport=5 0 355 275]{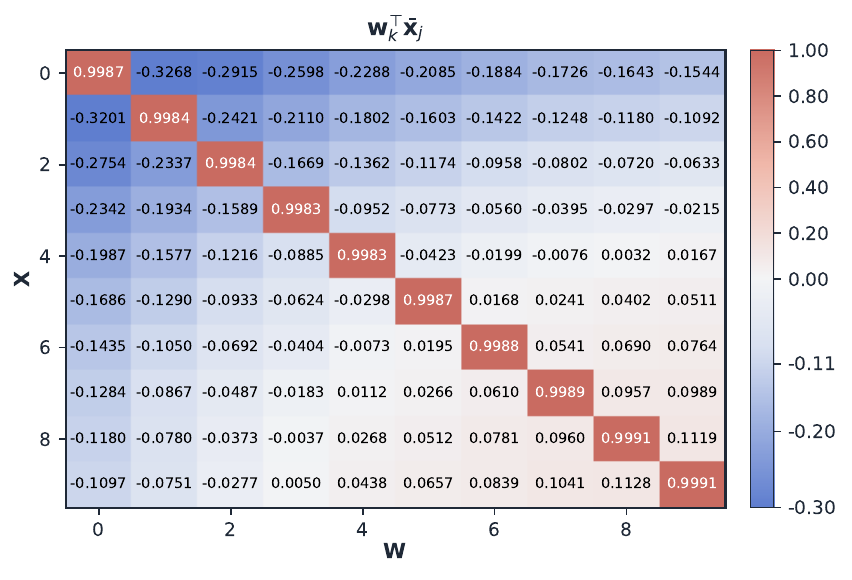}
		\hspace{-1.5mm}
		\includegraphics*[scale=0.36, viewport=5 0 440 275]{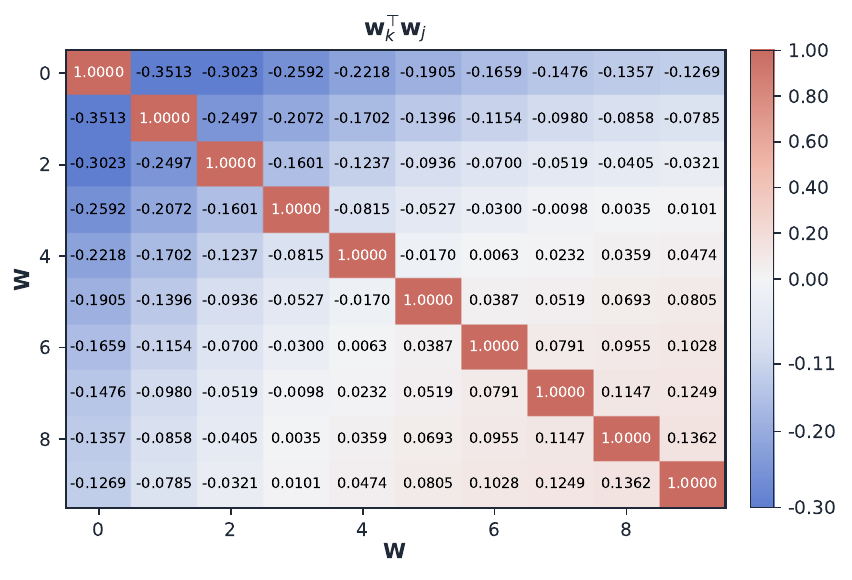}
	}
	\caption{Similarity matrices for class centers $\{\bar{\bm x}_k\}_{k=1}^K$ and classifier vectors $\{\bm w_k\}_{k=1}^K$ of ResNet18 trained by different BCE losses on CIFAR10-LT with IF = 100.}
	\label{fig_matrices_ce_bce_two_cifar10}
\end{figure*}

Fig. \ref{fig_matrices_ce_bce_two_cifar10} shows the similarity matrices for class centers $\{\bar{\bm x}_k\}_{k=1}^K$ and classifier vectors $\{\bm w_k\}_{k=1}^K$ of ResNet18 trained by CW and G. averaged BCE on CIFAR10-LT with IF $=100$.
For the CW loss, in its different similarity matrices, the off-diagonal elements that reflect the between-class separabilities are uniformly smaller than $-0.1$, close to the NC theoretical value of $-1/9$, and lie within a similarity interval with a relative length of less than $0.024$. These matrices indicate that the CW loss achieves balanced between-class separabilities.
In contrast, for the G. loss, the between-class geometric structures among sample features and classifier vectors are significantly distorted. The off-diagonal elements of all three matrices exhibits obvious differences, lying within a similarity interval of about $0.4$ in length; meanwhile, these off-diagonal values gradually increase from about $-0.3$ in the upper-left corner to about $0.1$ in the lower-right corner, which is clearly correlated with the number of samples contained in the classes from head to tail.
These results illustrate that, when training model on imbalanced datasets by using the globally (G.) averaged loss, the training is significantly affected by the sample number, and the between-class separabilities exhibit obvious imbalanced effect across the different classes, whereas the class-wise (CW) averaged loss could effectively suppresses this imbalance.

\begin{table*}[!tb]
\centering
\begin{tabular}{cc|ccccc}
\hline
\multirow{2}{*}{$\mathcal D$} & \multirow{2}{*}{$K_1$} & \multicolumn{5}{c}{ResNet18 trained using CE} \\
\cline{3-7}
 &  & $s^{\text{xx}}_{\text{pos}}$ & $s^{\text{xx}}_{\text{neg}}$ & $s^{\text{wx}}_{\text{pos}}$ & $s^{\text{wx}}_{\text{neg}}$ & $s^{\text{ww}}_{\text{neg}}$ \\
\hline
\multirow{4}{*}{\rotatebox{90}{CIFAR10}} & 9 & $.9938\pm.0005$ & $-.1241\pm.0051$ & $.9968\pm.0002$ & $-.1246\pm.0034$ & $-.1250\pm.0004$ \\
 & 7 & $.9556\pm.0067$ & $-.1221\pm.0175$ & $.9649\pm.0022$ & $-.1558\pm.0137$ & $-.1659\pm.0102$ \\
 & 5 & $.9532\pm.0060$ & $-.1155\pm.0220$ & $.9429\pm.0063$ & $-.2095\pm.0189$ & $-.2442\pm.0111$ \\
 & 3 & $.9385\pm.0073$ & $-.1476\pm.0145$ & $.9271\pm.0050$ & $-.3155\pm.0188$ & $-.4307\pm.0162$ \\
\cline{1-7}
\multirow{4}{*}{\rotatebox{90}{CIFAR100}} & 90 & $.9948\pm.0010$ & $-.0108\pm.0541$ & $.9971\pm.0006$ & $-.0112\pm.0541$ & $-.0112\pm.0541$ \\
 & 70 & $.9922\pm.0011$ & $-.0131\pm.0485$ & $.9953\pm.0006$ & $-.0144\pm.0486$ & $-.0145\pm.0487$ \\
 & 50 & $.9944\pm.0006$ & $-.0164\pm.0210$ & $.9954\pm.0003$ & $-.0200\pm.0208$ & $-.0204\pm.0207$ \\
 & 30 & $.9878\pm.0017$ & $-.0211\pm.0085$ & $.9891\pm.0009$ & $-.0322\pm.0062$ & $-.0342\pm.0033$ \\
\hline\hline
\multirow{2}{*}{$\mathcal D$} & \multirow{2}{*}{$K_1$} & \multicolumn{5}{c}{ResNet18 trained using BCE} \\
\cline{3-7}
 &  & $s^{\text{xx}}_{\text{pos}}$ & $s^{\text{xx}}_{\text{neg}}$ & $s^{\text{wx}}_{\text{pos}}$ & $s^{\text{wx}}_{\text{neg}}$ & $s^{\text{ww}}_{\text{neg}}$ \\
\hline
\multirow{4}{*}{\rotatebox{90}{CIFAR10}} & 9 & $.9935\pm.0032$ & $-.1230\pm.0152$ & $.9958\pm.0015$ & $-.1245\pm.0113$ & $-.1250\pm.0028$ \\
 & 7 & $.9954\pm.0003$ & $-.1653\pm.0039$ & $.9974\pm.0002$ & $-.1662\pm.0026$ & $-.1667\pm.0002$ \\
 & 5 & $.9936\pm.0009$ & $-.2323\pm.0137$ & $.9903\pm.0003$ & $-.2476\pm.0087$ & $-.2500\pm.0002$ \\
 & 3 & $.9859\pm.0012$ & $-.2713\pm.0125$ & $.9124\pm.0019$ & $-.4107\pm.0091$ & $-.4999\pm.0002$ \\
\cline{1-7}
\multirow{4}{*}{\rotatebox{90}{CIFAR100}} & 90 & $.9990\pm.0006$ & $-.0109\pm.0031$ & $.9993\pm.0003$ & $-.0112\pm.0023$ & $-.0112\pm.0012$ \\
 & 70 & $.9990\pm.0003$ & $-.0137\pm.0032$ & $.9991\pm.0001$ & $-.0145\pm.0023$ & $-.0145\pm.0007$ \\
 & 50 & $.9994\pm.0003$ & $-.0184\pm.0037$ & $.9987\pm.0001$ & $-.0204\pm.0026$ & $-.0204\pm.0005$ \\
 & 30 & $.9985\pm.0002$ & $-.0098\pm.0038$ & $.9871\pm.0002$ & $-.0342\pm.0027$ & $-.0345\pm.0006$ \\
\hline
\end{tabular}
\caption{Statistical results (mean $\pm$ std) for different metrics of ResNet18 trained by CE and BCE on CIFAR10 and CIFAR100 with $K-K_1$ empty classes.
	The CE and BCE are class-wise averaged using Eqs. (\ref{equ-minf_object_os_bce_main}) and (\ref{equ-minf_object-ce_os_main}).} \label{table_results_on_datasets_w_empty}
\end{table*}

\begin{table*}[!tb]
\centering
\begin{tabular}{cc|cccc}
\hline
\multirow{2}{*}{$\mathcal D$} & \multirow{2}{*}{$K_1$} & \multicolumn{4}{c}{ResNet18 trained using CE} \\
\cline{3-6}
 &  & $s^{\text{wx}}_{\text{emp}}$ & $s^{\text{ww}}_{\text{emp}}$ & $s^{\text{ww}}_{\text{emp2}}$ & $\|\tilde{\bm x}\|$ \\
\hline
\multirow{4}{*}{\rotatebox{90}{CIFAR10}} & 9 & $-.1668\pm.0000$ & — & $.0000\pm.0028$ & $0.0114$ \\
 & 7 & $-.9261\pm.0083$ & $.8431\pm.0171$ & $-.0220\pm.0170$ & $0.1784$ \\
 & 5 & $-.8663\pm.0092$ & $.7383\pm.0214$ & $-.0583\pm.0103$ & $0.3135$ \\
 & 3 & $-.7569\pm.0068$ & $.5581\pm.0234$ & $-.1603\pm.0281$ & $0.4631$ \\
\cline{1-6}
\multirow{4}{*}{\rotatebox{90}{CIFAR100}} & 90 & $-.2541\pm.0156$ & $.4836\pm.0309$ & $-.0001\pm.0158$ & $0.0204$ \\
 & 70 & $-.7047\pm.0156$ & $.6822\pm.0247$ & $-.0013\pm.0115$ & $0.0359$ \\
 & 50 & $-.8269\pm.0133$ & $.7630\pm.0202$ & $-.0039\pm.0082$ & $0.0620$ \\
 & 30 & $-.7977\pm.0170$ & $.6489\pm.0251$ & $-.0136\pm.0089$ & $0.1121$ \\
\hline\hline
\multirow{2}{*}{$\mathcal D$} & \multirow{2}{*}{$K_1$} & \multicolumn{4}{c}{ResNet18 trained using BCE} \\
\cline{3-6}
 &  & $s^{\text{wx}}_{\text{emp}}$ & $s^{\text{ww}}_{\text{emp}}$ & $s^{\text{ww}}_{\text{emp2}}$ & $\|\tilde{\bm x}\|$ \\
\hline
\multirow{4}{*}{\rotatebox{90}{CIFAR10}} & 9 & $-.5390\pm.0000$ & — & $.0004\pm.0032$ & $0.0328$ \\
 & 7 & $-.9545\pm.0006$ & $.9991\pm.0003$ & $.0000\pm.0016$ & $0.0216$ \\
 & 5 & $-.9905\pm.0003$ & $.9994\pm.0002$ & $.0002\pm.0012$ & $0.1133$ \\
 & 3 & $-.9990\pm.0003$ & $.9990\pm.0004$ & $.0078\pm.0026$ & $0.3845$ \\
\cline{1-6}
\multirow{4}{*}{\rotatebox{90}{CIFAR100}} & 90 & $-.2166\pm.0000$ & $1.000\pm.0000$ & $.0000\pm.0013$ & $0.0186$ \\
 & 70 & $-.7830\pm.0000$ & $1.000\pm.0000$ & $.0000\pm.0003$ & $0.0279$ \\
 & 50 & $-.9253\pm.0000$ & $1.000\pm.0000$ & $.0000\pm.0003$ & $0.0444$ \\
 & 30 & $-.9963\pm.0000$ & $1.000\pm.0000$ & $.0009\pm.0007$ & $0.1544$ \\
\hline
\end{tabular}

\noindent \textbf{Table~\ref{table_results_on_datasets_w_empty} (continued)} \par
\end{table*}

\begin{table*}[!tb]
\centering
\begin{tabular}{c|ccccc}
\hline
\multirow{2}{*}{epoch} & \multicolumn{5}{c}{ViT-Tiny trained by BCE} \\
\cline{2-6}
 & $s^{\text{xx}}_{\text{pos}}$ & $s^{\text{xx}}_{\text{neg}}$ & $s^{\text{wx}}_{\text{pos}}$ & $s^{\text{wx}}_{\text{neg}}$ & $s^{\text{ww}}_{\text{neg}}$ \\
\hline
50 & $.9726\pm.0256$ & $.4947\pm.0079$ & $.0783\pm.0199$ & $-.4608\pm.0076$ & $.3899\pm.0066$ \\
100 & $.9902\pm.0045$ & $.4096\pm.0041$ & $.2305\pm.0053$ & $-.3929\pm.0044$ & $.3302\pm.0044$ \\
200 & $.9949\pm.0052$ & $.2601\pm.0026$ & $.5001\pm.0047$ & $-.2629\pm.0032$ & $.2074\pm.0034$ \\
300 & $.9937\pm.0038$ & $.1203\pm.0037$ & $.7573\pm.0040$ & $-.1368\pm.0041$ & $.0844\pm.0036$ \\
400 & $.9954\pm.0059$ & $.0052\pm.0035$ & $.9688\pm.0037$ & $-.0329\pm.0034$ & $-.0136\pm.0026$ \\
600 & $.9939\pm.0107$ & $-.0200\pm.0045$ & $.9967\pm.0056$ & $-.0204\pm.0043$ & $-.0204\pm.0034$ \\
\hline\hline
\multirow{2}{*}{epoch} & \multicolumn{5}{c}{ViT-Tiny trained by CE} \\
\cline{2-6}
 & $s^{\text{xx}}_{\text{pos}}$ & $s^{\text{xx}}_{\text{neg}}$ & $s^{\text{wx}}_{\text{pos}}$ & $s^{\text{wx}}_{\text{neg}}$ & $s^{\text{ww}}_{\text{neg}}$ \\
\hline
50 & $.9092\pm.0216$ & $.0057\pm.0355$ & $.9422\pm.0125$ & $-.0147\pm.0372$ & $-.0195\pm.0382$ \\
100 & $.9880\pm.0014$ & $-.0101\pm.0069$ & $.9900\pm.0008$ & $-.0191\pm.0069$ & $-.0203\pm.0068$ \\
200 & $.9882\pm.0008$ & $-.0193\pm.0060$ & $.9937\pm.0004$ & $-.0202\pm.0063$ & $-.0204\pm.0065$ \\
300 & $.9604\pm.0186$ & $-.0185\pm.0127$ & $.9783\pm.0100$ & $-.0200\pm.0129$ & $-.0204\pm.0104$ \\
400 & $.9743\pm.0103$ & $-.0184\pm.0132$ & $.9855\pm.0056$ & $-.0201\pm.0125$ & $-.0204\pm.0096$ \\
600 & $.9997\pm.0003$ & $-.0204\pm.0010$ & $.9998\pm.0001$ & $-.0204\pm.0012$ & $-.0204\pm.0014$ \\
\hline
\end{tabular}
\par\medskip
\begin{tabular}{c|cccc}
\hline
\multirow{2}{*}{epoch} & \multicolumn{4}{c}{ViT-Tiny trained by BCE} \\
\cline{2-5}
 & $s^{\text{wx}}_{\text{emp}}$ & $s^{\text{ww}}_{\text{emp}}$ & $s^{\text{ww}}_{\text{emp2}}$ & $\|\tilde{\bm x}\|$ \\
\hline
50 & $-.9995\pm.0001$ & $.9997\pm.0002$ & $.6340\pm.0030$ & $0.7101$ \\
100 & $-.9999\pm.0000$ & $1.000\pm.0000$ & $.5862\pm.0040$ & $0.6490$ \\
200 & $-.9999\pm.0000$ & $1.000\pm.0000$ & $.4725\pm.0046$ & $0.5243$ \\
300 & $-.9998\pm.0000$ & $1.000\pm.0000$ & $.3205\pm.0055$ & $0.3712$ \\
400 & $-.9992\pm.0000$ & $1.000\pm.0000$ & $.0815\pm.0062$ & $0.1581$ \\
600 & $-.6181\pm.0000$ & $1.000\pm.0000$ & $.0002\pm.0010$ & $0.0160$ \\
\hline\hline
\multirow{2}{*}{epoch} & \multicolumn{4}{c}{ViT-Tiny trained by CE} \\
\cline{2-5}
 & $s^{\text{wx}}_{\text{emp}}$ & $s^{\text{ww}}_{\text{emp}}$ & $s^{\text{ww}}_{\text{emp2}}$ & $\|\tilde{\bm x}\|$ \\
\hline
50 & $-.9513\pm.0012$ & $.9983\pm.0016$ & $-.0302\pm.0121$ & $0.1542$ \\
100 & $-.9996\pm.0000$ & $1.000\pm.0000$ & $-.0105\pm.0026$ & $0.0991$ \\
200 & $-.9977\pm.0000$ & $1.000\pm.0000$ & $-.0011\pm.0021$ & $0.0289$ \\
300 & $-.2504\pm.0004$ & $1.000\pm.0000$ & $-.0002\pm.0096$ & $0.0322$ \\
400 & $-.0492\pm.0004$ & $1.000\pm.0000$ & $.0001\pm.0077$ & $0.0378$ \\
600 & $-.0442\pm.0003$ & $.9999\pm.0001$ & $.0000\pm.0007$ & $0.0013$ \\
\hline
\end{tabular}
\caption{Statistical results (mean $\pm$ std) for different metrics of ViT-Tiny trained on CIFAR100 with $K_1=50$.
}
\label{table_results_on_datasets_w_empty_multi_epoch}
\end{table*}

\subsection{On Open-Set Scenario}
We further verify Theorems \ref{theorem_NC_open_set_BCE_main} and \ref{theorem_NC_open_set_CE_main}.
On CIFAR10 ($K=10$) and CIFAR100 ($K=100$), we retain all samples in the first $K_1$ classes while empty the last $K-K_1$ classes.
When $K_1 = 9,7,5,3$ on CIFAR10 and $K_1 = 90,70,50,30$ on CIFAR100, we respectively train ResNet18 using the CE and BCE losses defined in Eqs. (\ref{equ-minf_object_os_bce_main}) and (\ref{equ-minf_object-ce_os_main}).
In the experiments, the losses were class-wise (CW) averaged over the $K_1$ known classes, {\color{black} and the models were trained for 1000 epochs. The other training settings are detailed in Table \ref{tab_supp_supervised_settings}.}

According to the theorems, as the losses decrease, these numbers in $s_{\text{pos}}^{\text{xx}}$, $s_{\text{neg}}^{\text{xx}}$, $s_{\text{pos}}^{\text{wx}}$, $s_{\text{neg}}^{\text{wx}}$, $s_{\text{neg}}^{\text{ww}}$, and $s_{\text{open}}^{\text{wx}}$ will converge to specific values related to $K_1$ and the norm of the global feature center, $\|\tilde{\bm x}\| = \big\|\frac{1}{K_1}\sum_{k=1}^{K_1}\bar{\bm x}_k\big\|$.
Table \ref{table_results_on_datasets_w_empty} shows their mean and standard deviations.


Except for the model trained on CIFAR10 using CE, the features extracted by other models all converge to their class centers, as evidenced by the means of their $s_{\mathrm{pos}}^{\text{xx}}$ being around $0.99$ and having very small standard deviations.
For these models, when $\|\tilde{\bm x}\|$ is less than $0.025$, the final sample features and classifier vectors closely approximate the ETF structure corresponding to a number of classes equal to $K_1$. For example, consider the model trained using BCE on CIFAR100 with $K_1 = 90$.
For all the known classes, the norm of their feature center $\tilde{\bm x}$ is $0.0186$. Correspondingly, the means of $s_{\mathrm{pos}}^{\text{xx}}$ and $s_{\mathrm{pos}}^{\text{wx}}$ are $0.9990$ and $0.9993$, respectively, with very small standard deviations, indicating that the sample features within the same class essentially collapse and align with their classifier vector.
Meanwhile, the means of $s_{\mathrm{neg}}^{\text{xx}}, s_{\mathrm{neg}}^{\text{wx}}$ and $s_{\mathrm{neg}}^{\text{ww}}$ are $-0.0109, -0.0112$ and $-0.0112$, which are relatively close to $-1/89$, indicating that the class centers and classifier vectors basically form an ETF structure that maximizes equiangular separation.

In contrast, when $\|\tilde{\bm x}\|$ is relatively large, although features within each class also collapse, the class centers do not align well with the classifier vectors, and their structure differs substantially form the ETF, while remaining very close to the structure specified by the theorems.
For example, $\|\tilde{\bm x}\| = 0.1544$ for the model trained using BCE on CIFAR100 with $K_1 = 30$.
Correspondingly, the mean of $s_{\mathrm{pos}}^{\text{xx}}$ is $0.9985$, whereas that of $s_{\mathrm{pos}}^{\text{wx}}$ is $0.9871$, indicating that within-class collapse is achieved but the collapsed features do not fully align with their classifier vectors.
Then mean of $s_{\mathrm{neg}}^{\text{ww}}$ is $-0.0345$, which is equal to $-1/29$, indicating that the classifier vectors essentially form an ETF. However, the mean of $s_{\mathrm{neg}}^{\text{xx}}$ is $-0.0098$, which differs substantially from $-1/29$.
Nevertheless, it should be noted that the means of $s_{\mathrm{pos}}^{\text{wx}}$ and $s_{\mathrm{neg}}^{\text{xx}}$ are very close to the theoretical values, $\sqrt{1-\|\tilde{\bm x}\|^2}\approx0.9880$ and $-\frac1{K_1-1} + \frac{K_1}{K_1-1}\|\tilde{\bm x}\|^2\approx -0.0098$, given by Theorem \ref{theorem_NC_open_set_BCE_main}.

In Table \ref{table_results_on_datasets_w_empty}, for models trained with BCE, it is relatively easily to verify that their metrics largely conform to the conclusions of Theorem \ref{theorem_NC_open_set_BCE_main}. In contrast, for models trained with CE, verifying compliance with Theorem \ref{theorem_NC_open_set_CE_main} requires first numerically solving $c_1, c_2$, and $c_3$, which is slightly more difficult.
One can observe that, compared to the models trained by CE, that trained by BCE achieve smaller standard deviations across various metrics, indicating that after sufficient training, the BCE-trained models yield classifiers and sample features with superior geometric structures.
Furthermore, from Table \ref{table_results_on_datasets_w_empty}, it can also be observed that when $K_1$ is closer to $K$, after the same training, the model's $\|\tilde{\bm x}\|$ is smaller, and all metrics are closer to NC values, which is consistent with our intuitive understanding.

In Table \ref{table_results_on_datasets_w_empty}, one can observe that for classifier vectors $\{\bm w_j: K_1<j\leq K\}$ of empty classes, the means of $s_{\text{emp}}^{\text{ww}}$ and $s_{\text{emp2}}^{\text{ww}}$ are respectively close to $1$ and $0$, with very small standard deviations. These results indicate that $\{\bm w_j: K_1<j\leq K\}$ have basically collapsed to a single vector and are orthogonal to $\{\bm w_k: k\in[K_1]\}$, the classifier vectors of non-empty known classes. With Theorems \ref{theorem_NC_open_set_BCE_main} and \ref{theorem_NC_open_set_CE_main}, the theoretical value of the metrics of $s_{\text{emp}}^{\text{wx}}$ is $-1$; however, in Table \ref{table_results_on_datasets_w_empty}, the $s_{\text{emp}}^{\text{wx}}$ of most models deviates markedly from $-1$. We conjecture that it is because when $\tilde{\bm x} = \frac1{K_1}\sum_{k=1}^{K_1}\bar{\bm x}_k$ is very close to $\bm 0$, even minor angular variations between class centers $\{\bar{\bm x}_k: k\in[K_1]\}$ can substantially alter the direction of $\tilde{\bm x}$, and consequently change its similarity with the classifier vectors of the empty unknown classes, i.e., $\bm w_j^\top \tilde{\bm x}/\|\tilde{\bm x}\|, K_1<j\leq K_1$.

In Table \ref{table_results_on_datasets_w_empty_multi_epoch}, we present the statistical results of ViT-Tiny trained on CIFAR100 with $K_1 = 50$ using BCE and CE.
Before the model training, the randomly initialized classifier vectors are almost mutually orthogonal. However, after training begins, the classifier vectors $\{\bm w_j: K_1<j\leq K\}$ of empty unknown classes quickly coincide with one another, as shown by the statistical results for $s_{\text{emp}}^{\text{ww}}$, which are already very close to $1\pm0$ at epoch = 50. For $K_1<j\leq K$, there are no samples from this unknown class attracting its classifier vector $\bm w_j$, while the classifier vectors of different unknown classes are subjected to exactly the same repulsive force; therefore, in a bounded space, they will ultimately coincide. Meanwhile, they become opposite in direction to the global feature averaged vector $\tilde{\bm x}$, as shown by the statistical results for $s_{\text{emp}}^{\text{wx}}$, which rapidly approach $-1\pm0$ within the first 100 epochs. For $\{\bm w_j: K_1<j\leq K\}$, they cannot indirectly receive repulsive forces from the classifier vectors $\{\bm w_k: k\in[K_1]\}$ of non-empty known classes through their own samples, and can only receive repulsive forces from all sample features of other classes, so ultimately they become opposite to the direction of the resultant repulsive force, $\tilde{\bm x}$. However, as $\tilde{\bm x}$ gradually tends toward $\bm 0$, for the model trained with BCE, the mean of $s_{\text{emp}}^{\text{wx}}$ at the $600$th epoch is $-0.6181$, whereas for that trained with CE, at the $300$th, $400$th, and $600$th epochs the means of $s_{\text{emp}}^{\text{wx}}$ change to $-0.2504, -0.0492$, and $-0.0442$, respectively, indicating that the angle between $\tilde{\bm x}$ and $\{\bm w_j: K_1<j\leq K\}$ changes drastically, which is consistent with our earlier conjecture.

In the final stage of training, the structure of classifier vectors $\{\bm w_k: k\in[K_1]\}$ and the sample features of non-empty known classes are very close to an ETF. The numbers in $s_{\text{pos}}^{\text{xx}}$, which reflect the within-class collapse, and $s_{\text{pos}}^{\text{wx}}$, which reflect the alignment between classifier vectors $\{\bm w_k: k\in[K_1]\}$ and class centers $\{\bar{\bm x}_k: k\in[K_1]\}$, are both close to a mean of $1$ and a standard deviation of $0$. For $s_{\text{neg}}^{\text{xx}}$ and $s_{\text{neg}}^{\text{ww}}$, which reflect the between-class separation of sample features and classifier vectors, their means are both close to $-1/49 \approx -0.0204$, with relatively small standard deviations. These results indicate that, regardless of whether the models are trained with CE or BCE, NC is essentially achieved on the non-empty classes.

\subsection{Contrastive Losses on Imbalanced Datasets}
\begin{table*}[!tb]
\centering
\begin{tabular}{c|cc|ccc}
\hline
\multirow{2}{*}{Loss} & \multirow{2}{*}{Dataset} & \multirow{2}{*}{IF} & \multicolumn{3}{c}{ResNet18} \\
\cline{4-6}
 &  &  & $s^{\text{xx}}_{\text{pos}}$ & $s^{\text{xx}}_{\text{neg}}$ & $b$ \\
\hline
\multirow{10}{*}{Softmax-based} & \multirow{5}{*}{CIFAR10} & 1 & $0.9996\pm0.0001$ & $-0.1111\pm0.0036$ & $0.0000\pm0.0002$ \\
 &  & 10 & $0.9996\pm0.0002$ & $-0.1110\pm0.0031$ & $0.0000\pm0.0004$ \\
 &  & 50 & $0.9996\pm0.0002$ & $-0.1111\pm0.0040$ & $0.0000\pm0.0024$ \\
 &  & 100 & $0.9995\pm0.0003$ & $-0.1104\pm0.0054$ & $0.0000\pm0.0044$ \\
\cdashline{2-6}
 & \multicolumn{2}{c|}{NC theo. value} & $1.0$ & $-1/9\approx-0.1111$ & — \\
\cline{2-6}
 & \multirow{5}{*}{CIFAR100} & 1 & $0.9996\pm0.0005$ & $-0.0101\pm0.0029$ & $0.0000\pm0.0003$ \\
 &  & 10 & $0.9995\pm0.0006$ & $-0.0101\pm0.0031$ & $0.0000\pm0.0034$ \\
 &  & 50 & $0.9996\pm0.0003$ & $-0.0100\pm0.0029$ & $0.0000\pm0.0122$ \\
 &  & 100 & $0.9994\pm0.0006$ & $-0.0099\pm0.0033$ & $0.0000\pm0.0232$ \\
\cdashline{2-6}
 & \multicolumn{2}{c|}{NC theo. value} & $1.0$ & $-1/99\approx-0.0101$ & — \\
\hline
\multirow{10}{*}{Sigmoid-based} & \multirow{5}{*}{CIFAR10} & 1 & $0.9998\pm0.0002$ & $-0.1111\pm0.0035$ & $1.7772\pm0.0004$ \\
 &  & 10 & $0.9996\pm0.0002$ & $-0.1110\pm0.0043$ & $1.7769\pm0.0005$ \\
 &  & 50 & $0.9995\pm0.0004$ & $-0.1108\pm0.0061$ & $1.7770\pm0.0009$ \\
 &  & 100 & $0.9995\pm0.0003$ & $-0.1106\pm0.0061$ & $1.7770\pm0.0007$ \\
\cdashline{2-6}
 & \multicolumn{2}{c|}{NC theo. value} & $1.0$ & $-1/9\approx-0.1111$ & $1.7778$ \\
\cline{2-6}
 & \multirow{5}{*}{CIFAR100} & 1 & $0.9999\pm0.0002$ & $-0.0101\pm0.0290$ & $1.9809\pm0.0020$ \\
 &  & 10 & $0.9998\pm0.0004$ & $-0.0101\pm0.0145$ & $1.9788\pm0.0030$ \\
 &  & 50 & $0.9997\pm0.0008$ & $-0.0101\pm0.0512$ & $1.9841\pm0.0026$ \\
 &  & 100 & $0.9995\pm0.0011$ & $-0.0102\pm0.0247$ & $1.9772\pm0.0135$ \\
\cdashline{2-6}
 & \multicolumn{2}{c|}{NC theo. value} & $1.0$ & $-1/99\approx-0.0101$ & $1.9798$ \\
\hline
\end{tabular}
\caption{Statistical results (mean $\pm$ std) of different models trained by Softmax-based and Sigmoid-based losses on CIFAR10-LT and CIFAR100-LT with $\text{IF} = 1, 10, 50$, and $100$.
}
\label{table_results_on_datasets_contrastive_loss}
\end{table*}

\begin{table*}[!tb]
	\centering
	\begin{tabular}{c|cc|ccc}
		\hline
		\multirow{2}{*}{Loss} & \multirow{2}{*}{Dataset} & \multirow{2}{*}{IF} & \multicolumn{3}{c}{ViT-Tiny} \\
		\cline{4-6}
		&  &  & $s^{\text{xx}}_{\text{pos}}$ & $s^{\text{xx}}_{\text{neg}}$ & $b$ \\
		\hline
		\multirow{10}{*}{Softmax-based} & \multirow{5}{*}{CIFAR10} & 1 & $0.9998\pm0.0001$ & $-0.1111\pm0.0036$ & $0.0002\pm0.0002$ \\
		&  & 10 & $0.9997\pm0.0001$ & $-0.1111\pm0.0040$ & $0.0000\pm0.0007$ \\
		&  & 50 & $0.9996\pm0.0002$ & $-0.1110\pm0.0057$ & $0.0000\pm0.0024$ \\
		&  & 100 & $0.9997\pm0.0001$ & $-0.1109\pm0.0053$ & $0.0000\pm0.0046$ \\
		\cdashline{2-6}
		& \multicolumn{2}{c|}{NC theo. value} & $1.0$ & $-1/9\approx-0.1111$ & — \\
		\cline{2-6}
		& \multirow{5}{*}{CIFAR100} & 1 & $0.9998\pm0.0001$ & $-0.0101\pm0.0026$ & $0.0000\pm0.0000$ \\
		&  & 10 & $0.9998\pm0.0001$ & $-0.0101\pm0.0037$ & $0.0000\pm0.0034$ \\
		&  & 50 & $0.9997\pm0.0001$ & $-0.0100\pm0.0300$ & $0.0000\pm0.0132$ \\
		&  & 100 & $0.9997\pm0.0001$ & $-0.0100\pm0.0321$ & $0.0000\pm0.0241$ \\
		\cdashline{2-6}
		& \multicolumn{2}{c|}{NC theo. value} & $1.0$ & $-1/99\approx-0.0101$ & — \\
		\hline
		\multirow{10}{*}{Sigmoid-based} & \multirow{5}{*}{CIFAR10} & 1 & $0.9996\pm0.0001$ & $-0.1111\pm0.0045$ & $1.7790\pm0.0020$ \\
		&  & 10 & $0.9999\pm0.0000$ & $-0.1111\pm0.0020$ & $1.7777\pm0.0003$ \\
		&  & 50 & $0.9999\pm0.0000$ & $-0.1111\pm0.0021$ & $1.7776\pm0.0002$ \\
		&  & 100 & $0.9999\pm0.0000$ & $-0.1110\pm0.0015$ & $1.7776\pm0.0005$ \\
		\cdashline{2-6}
		& \multicolumn{2}{c|}{NC theo. value} & $1.0$ & $-1/9\approx-0.1111$ & $1.7778$ \\
		\cline{2-6}
		& \multirow{5}{*}{CIFAR100} & 1 & $0.9999\pm0.0000$ & $-0.0101\pm0.0245$ & $1.9827\pm0.0028$ \\
		&  & 10 & $0.9996\pm0.0002$ & $-0.0101\pm0.0655$ & $1.9894\pm0.0043$ \\
		&  & 50 & $0.9997\pm0.0001$ & $-0.0101\pm0.0550$ & $1.9835\pm0.0045$ \\
		&  & 100 & $0.9997\pm0.0001$ & $-0.0101\pm0.0616$ & $1.9841\pm0.0052$ \\
		\cdashline{2-6}
		& \multicolumn{2}{c|}{NC theo. value} & $1.0$ & $-1/99\approx-0.0101$ & $1.9798$ \\
		\hline
	\end{tabular}
	
	\noindent\textbf{Table~\ref{table_results_on_datasets_contrastive_loss} (continued)} \par
\end{table*}

Finally, we verify Theorem \ref{theorem_NC_imbalanced_CL} for contrastive learning on balanced and imbalanced datasets. 
In experiments for the Sigmoid-based contrastive loss, we set $\lambda=\frac{1}{K-1}$; then, $b^\star = \frac{\gamma}{2}\Big(1-\frac{1}{K-1}\Big)$.

Table \ref{table_results_on_datasets_contrastive_loss} reports the statistical results of ResNet18 and ViT-Tiny trained using the re-balanced Softmax-based and Sigmoid-based losses on CIFAR10-LT and CIFAR100-LT with IF = $1$, $10$, $50$, and $100$.
In these experiments, across settings with different numbers of classes and IFs, the mean of the final $s^{\text{xx}}_{\mathrm{pos}}$ consistently stays close to 1, while the mean of $s^{\text{xx}}_{\mathrm{neg}}$ remains close to the corresponding theoretical extreme values, i.e., $-1/9$ on CIFAR10-LT and $-1/99$ on CIFAR100-LT. Meanwhile, the standard deviations of these two kinds of quantities are relatively small.
These results indicate that the re-balanced Softmax-based and Sigmoid-based loss functions using sample inverse frequencies effectively suppress the impact of class imbalance, and drive the learned features toward within-class collapse and uniformly separated class centers, as predicted by Theorem \ref{theorem_NC_imbalanced_CL}.
Furthermore, for the Sigmoid-based loss, the learned biases also remain close to their theoretical values, approximately $1.7778$ on CIFAR10-LT and $1.9798$ on CIFAR100-LT, supporting the theoretical analysis.

Compared with the Softmax-based loss, the Sigmoid-based loss exhibits relatively larger standard deviations in $s^{\text{xx}}_{\mathrm{neg}}$, particularly on CIFAR100-LT, although the means of $s^{\text{xx}}_{\mathrm{neg}}$ remain close to the theoretical values.
As the negative terms in the rebalanced Sigmoid-based loss are scaled by $\frac1{K-1}$, when the number of classes, $K$, is large, their contribution to contrastive learning is smaller and between-class separation may not fully converge within a finite training time. We recommend adding a weighting parameter before the negative terms when using the Sigmoid-based loss in practice.

\section{Conclusion}

This paper reformulates neural collapse (NC) in imbalanced datasets and open-set scenarios.
Unlike the prior works that use globally averaged losses, we analyze the minimizers of class-wise (CW) averaged losses re-balanced by sample frequency of each class. When the CW averaged losses reach their minima, the sample features and the classifier vectors form a standard NC, on both balanced and imbalanced datasets.
{\color{black} Furthermore, we study in an open-set scenario that includes empty classes, the geometric structure of features and classifier vectors under extreme conditions.}
In the open-set scenario, when the CW averaged losses achieve their minima, a standard NC still formed on the known classes that contain samples. For the unknown classes that contain no samples, the classifier vectors collapse into a single vector, which is orthogonal to the classifier vectors of the known classes. Finally, for contrastive learning on imbalanced datasets, the losses re-balanced by the inverse sample frequency can also achieve standard NC.

\noindent\textit{Data Availability.} The original CIFAR10 and CIFAR100 datasets used in this paper are publicly available, see \url{https://cave.cs.toronto.edu/kriz/cifar.html} for details.

\acks{This work was supported by the National Natural Science Foundation of China (Grant Nos. 62576217 and 62576216), Guangdong Provincial Key Laboratory (Grant No. 2023B1212060076), and Natural Science Foundation of Ningxia Province (Grant No. 2025AAC020002). This work was also supported by the Intelligent Computing Center of Shenzhen University.}

\clearpage
\section*{Appendix: Supplementary Material}
\setcounter{section}{0}
\setcounter{equation}{0}
\setcounter{figure}{0}
\setcounter{table}{0}
\setcounter{theorem}{0}
\renewcommand{\thesection}{S.\arabic{section}}
\renewcommand{\theequation}{S.\arabic{equation}}
\renewcommand{\thefigure}{S.\arabic{figure}}
\renewcommand{\thetable}{S.\arabic{table}}
\renewcommand{\thetheorem}{S.\arabic{theorem}}

This supplementary material provides additional experimental
settings and results, as well as complete proofs of the
theoretical results presented in the main paper.

\section{Experimental Settings and More Results} \label{experimental_results_app}

We summarize the experimental settings for supervised classification and supervised contrastive training, and provide additional results for ResNet18 and ViT-Tiny on CIFAR10 and CIFAR100.

\subsection{Experimental Settings}
We conduct experiments using ResNet18 and ViT-Tiny on CIFAR10 and CIFAR100 \citep{krizhevsky2009learning} to verify the theorems.
Table \ref{tab_supp_supervised_settings} summarizes the detailed training settings for classification, open-set scenarios, and supervised contrastive learning.
All models are trained from scratch. The feature dimensions are $d=512$ for ResNet18 and $d=192$ for ViT-Tiny. No additional projection head or terminal ReLU activation is used. All input images are channel-wise normalized, without weight decay, feature regularization, warmup, label smoothing, or other data augmentation. For optimization, SGD is with a momentum of $0.9$, and AdamW is with $(\beta_1,\beta_2)=(0.9,0.999)$.

\begin{table}[!htb]
	\centering
	\setlength{\tabcolsep}{3pt}
	\begin{tabular}{c|l||cc||cc||cc}
		\hline
		\multicolumn{2}{c||}{} & \multicolumn{2}{c||}{\shortstack{Classification\\Training}} & \multicolumn{2}{c||}{Open-set Scenario} & \multicolumn{2}{c}{\shortstack{Supervised Contrastive\\Learning}} \\
		\cline{3-8}
		\multicolumn{2}{c||}{} & ResNet18 & ViT-Tiny & ResNet18 & ViT-Tiny & ResNet18 & ViT-Tiny \\
		\hline
		\multirow{9}{*}{\rotatebox{90}{Hyper-parameter}} & epochs & 800 & 800 & 1000 & 1000 & 5000 & 5000 \\
		& optimizer & SGD & AdamW & SGD & AdamW & SGD & AdamW \\
		& batch size & 128 & 128 & 128 & 128 & 8192 & 8192 \\
		& learning rate & $0.01$ & $0.0001$ & $0.01$ & $0.0001$ & $0.5$ & $0.0002$ \\
		& \shortstack[l]{LR schedule} & step & step & step & step & step & cosine \\
		& scale factor & $\gamma=8$ & $\gamma=8$ & $\gamma=8$ & $\gamma=8$ & $\gamma=4$ & $\gamma=4$ \\
		& weight decay $\lambda$ & \ding{55} & \ding{55} & \ding{55} & \ding{55} & \ding{55} & \ding{55} \\
		& warmup epochs & 0 & 0 & 0 & 0 & 0 & 0 \\
		& Normalization & \ding{51} & \ding{51} & \ding{51} & \ding{51} & \ding{51} & \ding{51} \\
		\hline
	\end{tabular}
	\caption{Detailed experimental settings.}
	\label{tab_supp_supervised_settings}
\end{table}

\begin{table}[!htb]
	\centering
	\begin{tabular}{c|ccc|ccc}
		\hline
		\multirow{2}{*}{Dataset} & \multicolumn{3}{c|}{$n_{\max}$ for classification} & \multicolumn{3}{c}{$n_{\max}$ for contrastive learning} \\
		\cmidrule(lr){2-4}\cmidrule(lr){5-7}
		& IF = 10 & IF = 50 & IF = 100 & IF = 10 & IF = 50 & IF = 100 \\
		\hline
		CIFAR10-LT  & 5000 & 5000 & 5000& 2005 & 2927 & 3302 \\
		CIFAR100-LT & 500  & 500  & 500& 210  & 325  & 378  \\
		\hline
	\end{tabular}
	\caption{The number of samples, $n_{\max}$, of the largest class in CIFAR10-LT and CIFAR100-LT for classification and contrastive learning.}
	\label{tab_supcon_data_setting}
\end{table}

We achieve CIFAR10-LT and CIFAR100-LT \citep{cao2019learning} by sampling from the original CIFAR10 and CIFAR100, such that the number of samples in class $k$ is
\begin{equation}
    n_k = \left\lfloor n_{\max} * \mathrm{IF}^{-(k-1)/(K-1)}\right\rfloor, \quad\mathrm{IF}\in\{1,10,50,100\},\quad k\in[K],\quad K\in\{10, 100\}.
\end{equation}
For classification experiments, $n_{\max}=5000$ on CIFAR10 and $n_{\max}=500$ on CIFAR100, and $\mathrm{IF}=1$ corresponds to the original balanced datasets.
For supervised contrastive learning, we construct long-tailed datasets using the same sampling rules but set $n_{\max}$ to a smaller value to obtain smaller long-tailed datasets, which accelerates the convergence of the models in contrastive learning on the imbalanced datasets.

\subsection{More Results on Open-set Scenario}
Table \ref{table_results_on_datasets_w_empty_vit} reports the results of ViT-Tiny trained on CIFAR10 and CIFAR100 in open-set scenarios. In total, ViT-Tiny exhibits the same behaviors predicted by Theorems \ref{theorem_NC_open_set_BCE_main} and \ref{theorem_NC_open_set_CE_main}, demonstrating that the theoretical conclusions are not specific to the CNN architecture.

For the known classes, $k\in[K_1]$, their sample features exhibit strong within-class collapse, as indicated by $s^{\mathrm{xx}}_{\mathrm{pos}}$ being close to $1$ in most configurations. Meanwhile, their classifier vectors approximately form an ETF, with $s^{\mathrm{ww}}_{\mathrm{neg}}$ closely matching the theoretical value $-1/(K_1-1)$. For example, on CIFAR100 with $K_1=30$, the mean and standard deviation
of $s^{\mathrm{ww}}_{\mathrm{neg}}$ are $-0.0345\pm0.0010$ for the model trained by CE, and they are $-0.0345\pm0.0005$ for the model trained by BCE, both closely matching the theoretical value $-1/29\approx -0.0345$.

The classifier vectors, $\{\bm w_j: K_1<j\leq K\}$, for the empty unknown classes also exhibit the predicted collapse behavior. $s^{\mathrm{ww}}_{\mathrm{emp}}$ is consistently close to $1$, indicating that the classifier vectors of the empty classes become nearly identical. Meanwhile, $s^{\mathrm{ww}}_{\mathrm{emp2}}$ remains close to $0$, showing that these collapsed empty-class classifier vectors are approximately orthogonal to the classifier vectors of the known classes.
When $\|\tilde{\bm x}\|$ is close to zero, the similarities between the empty-class classifier vectors $\{\bm w_j: K_1<j\leq K\}$ and $\tilde{\bm x}$ are correspondingly small. For example, for the model trained by CE on CIFAR100 with $K_1=50$, $\|\tilde{\bm x}\|=0.0013$ and $s^{\mathrm{wx}}_{\mathrm{emp}}=-0.0442\pm0.0003$, which does not conform to the conclusion of the theorem. We believe this stems from the fact that when the magnitude of $\tilde{\bm x}$ is too small, its direction is prone to being affected by interference.
In contrast, when the magnitude of $\tilde{\bm x}$ is large enough, the empty-class classifier vectors become strongly aligned with the direction predicted by the theory. For the model trained by BCE on CIFAR100 with $K_1=30$, $\|\tilde{\bm x}\|=0.0808$ and $s^{\mathrm{wx}}_{\mathrm{emp}}=-1.000\pm0.0000$.

In total, the experimental results of ViT-Tiny consistently reproduce the similar geometric structures observed in the experiments of ResNet18, including within-class collapse, ETF-like geometry among known-class classifiers, collapse of empty-class classifiers, and their near orthogonality to the known-class classifiers.

\begin{table*}[!htb]
    \centering
    \begin{tabular}{cc|ccccc}
\hline
\multirow{2}{*}{$\mathcal D$} & \multirow{2}{*}{$K_1$} & \multicolumn{5}{c}{ViT-Tiny trained using CE} \\
\cline{3-7}
 &  & $s^{\text{xx}}_{\text{pos}}$ & $s^{\text{xx}}_{\text{neg}}$ & $s^{\text{wx}}_{\text{pos}}$ & $s^{\text{wx}}_{\text{neg}}$ & $s^{\text{ww}}_{\text{neg}}$ \\
\hline
\multirow{4}{*}{\rotatebox{90}{CIFAR10}} & 9 & $1.000\pm.0000$ & $-.1250\pm.0001$ & $1.000\pm.0000$ & $-.1250\pm.0019$ & $-.1250\pm.0029$ \\
 & 7 & $.9909\pm.0023$ & $-.1646\pm.0207$ & $.9948\pm.0014$ & $-.1658\pm.0170$ & $-.1666\pm.0117$ \\
 & 5 & $.9996\pm.0000$ & $-.2499\pm.0004$ & $.9998\pm.0000$ & $-.2499\pm.0020$ & $-.2500\pm.0031$ \\
 & 3 & $.9979\pm.0002$ & $-.4977\pm.0020$ & $.9984\pm.0001$ & $-.4992\pm.0029$ & $-.4998\pm.0046$ \\
\cline{1-7}
\multirow{4}{*}{\rotatebox{90}{CIFAR100}} & 90 & $.9881\pm.0060$ & $-.0107\pm.0067$ & $.9935\pm.0032$ & $-.0112\pm.0067$ & $-.0112\pm.0060$ \\
 & 70 & $.9972\pm.0005$ & $-.0145\pm.0045$ & $.9986\pm.0002$ & $-.0145\pm.0045$ & $-.0145\pm.0044$ \\
 & 50 & $.9997\pm.0003$ & $-.0204\pm.0010$ & $.9998\pm.0001$ & $-.0204\pm.0012$ & $-.0204\pm.0014$ \\
 & 30 & $1.0000\pm.0000$ & $-.0345\pm.0002$ & $1.000\pm.0000$ & $-.0345\pm.0007$ & $-.0345\pm.0010$ \\
\hline\hline
\multirow{2}{*}{$\mathcal D$} & \multirow{2}{*}{$K_1$} & \multicolumn{5}{c}{ViT-Tiny trained using BCE} \\
\cline{3-7}
 &  & $s^{\text{xx}}_{\text{pos}}$ & $s^{\text{xx}}_{\text{neg}}$ & $s^{\text{wx}}_{\text{pos}}$ & $s^{\text{wx}}_{\text{neg}}$ & $s^{\text{ww}}_{\text{neg}}$ \\
\hline
\multirow{4}{*}{\rotatebox{90}{CIFAR10}} & 9 & $.9998\pm.0000$ & $-.1250\pm.0002$ & $.9999\pm.0000$ & $-.1250\pm.0008$ & $-.1250\pm.0012$ \\
 & 7 & $.9951\pm.0020$ & $-.1657\pm.0037$ & $.9974\pm.0011$ & $-.1663\pm.0039$ & $-.1666\pm.0023$ \\
 & 5 & $1.000\pm.0000$ & $-.2500\pm.0003$ & $1.000\pm.0000$ & $-.2500\pm.0017$ & $-.2500\pm.0028$ \\
 & 3 & $.9999\pm.0000$ & $-.4992\pm.0003$ & $.9997\pm.0000$ & $-.4999\pm.0003$ & $-.5000\pm.0005$ \\
\cline{1-7}
\multirow{4}{*}{\rotatebox{90}{CIFAR100}} & 90 & $.9925\pm.0053$ & $-.0110\pm.0035$ & $.9961\pm.0027$ & $-.0112\pm.0034$ & $-.0112\pm.0028$ \\
 & 70 & $.9997\pm.0002$ & $-.0145\pm.0005$ & $.9998\pm.0001$ & $-.0145\pm.0007$ & $-.0145\pm.0008$ \\
 & 50 & $.9998\pm.0002$ & $-.0204\pm.0003$ & $.9999\pm.0001$ & $-.0204\pm.0008$ & $-.0204\pm.0011$ \\
 & 30 & $.9999\pm.0000$ & $-.0277\pm.0001$ & $.9966\pm.0001$ & $-.0345\pm.0004$ & $-.0345\pm.0005$ \\
\hline
\end{tabular}
\caption{Statistical results (mean $\pm$ std) for different metrics of ViT-Tiny trained by CE and BCE on CIFAR10 and CIFAR100 with $K-K_1$ empty classes. The CE and BCE are class-wise averaged using Eqs. (\ref{equ-minf_object_os_bce_main}) and (\ref{equ-minf_object-ce_os_main}).}
\label{table_results_on_datasets_w_empty_vit}
\end{table*}

\begin{table*}[!htb]
\centering
\begin{tabular}{cc|cccc}
\hline
\multirow{2}{*}{$\mathcal D$} & \multirow{2}{*}{$K_1$} & \multicolumn{4}{c}{ViT-Tiny trained using CE} \\
\cline{3-6}
 &  & $s^{\text{wx}}_{\text{emp}}$ & $s^{\text{ww}}_{\text{emp}}$ & $s^{\text{ww}}_{\text{emp2}}$ & $\|\tilde{\bm x}\|$ \\
\hline
\multirow{4}{*}{\rotatebox{90}{CIFAR10}} & 9 & $.1177\pm.0000$ & --- & $.0001\pm.0015$ & $0.0002$ \\
 & 7 & $-.7712\pm.0043$ & $.9997\pm.0002$ & $-.0004\pm.0072$ & $0.0217$ \\
 & 5 & $-.9965\pm.0000$ & $1.000\pm.0000$ & $.0000\pm.0009$ & $0.0057$ \\
 & 3 & $-.9113\pm.0016$ & $.9999\pm.0001$ & $.0000\pm.0053$ & $0.0292$ \\
\cline{1-6}
\multirow{4}{*}{\rotatebox{90}{CIFAR100}} & 90 & $-.0011\pm.0000$ & $.9997\pm.0004$ & $.0000\pm.0048$ & $0.0211$ \\
 & 70 & $-.0150\pm.0016$ & $.9999\pm.0002$ & $.0000\pm.0030$ & $0.0012$ \\
 & 50 & $-.0442\pm.0003$ & $.9999\pm.0001$ & $.0000\pm.0007$ & $0.0013$ \\
 & 30 & $-.9929\pm.0000$ & $1.000\pm.0000$ & $.0000\pm.0005$ & $0.0039$ \\
\hline\hline
\multirow{2}{*}{$\mathcal D$} & \multirow{2}{*}{$K_1$} & \multicolumn{4}{c}{ViT-Tiny trained using BCE} \\
\cline{3-6}
 &  & $s^{\text{wx}}_{\text{emp}}$ & $s^{\text{ww}}_{\text{emp}}$ & $s^{\text{ww}}_{\text{emp2}}$ & $\|\tilde{\bm x}\|$ \\
\hline
\multirow{4}{*}{\rotatebox{90}{CIFAR10}} & 9 & $-.0370\pm.0000$ & --- & $.0001\pm.0010$ & $0.0004$ \\
 & 7 & $-.7554\pm.0001$ & $1.000\pm.0000$ & $.0071\pm.0024$ & $0.0107$ \\
 & 5 & $.0551\pm.0001$ & $1.000\pm.0000$ & $-.0001\pm.0002$ & $0.0003$ \\
 & 3 & $-.9998\pm.0000$ & $1.000\pm.0000$ & $.0018\pm.0038$ & $0.0220$ \\
\cline{1-6}
\multirow{4}{*}{\rotatebox{90}{CIFAR100}} & 90 & $-.0032\pm.0000$ & $1.000\pm.0000$ & $.0001\pm.0015$ & $0.0104$ \\
 & 70 & $-.0120\pm.0001$ & $1.000\pm.0000$ & $.0000\pm.0003$ & $0.0007$ \\
 & 50 & $-.5992\pm.0000$ & $1.000\pm.0000$ & $.0000\pm.0004$ & $0.0005$ \\
 & 30 & $-1.000\pm.0000$ & $1.000\pm.0000$ & $.0011\pm.0012$ & $0.0808$ \\
\hline
\end{tabular}

\noindent\textbf{Table~\ref{table_results_on_datasets_w_empty_vit} (continued)} \par
\end{table*}

\subsection{Statistical Results for Models Trained on Imbalanced Datasets}
\begin{figure*}[!htb]
    \centering
    \subfigure[BCE]{\includegraphics[width=0.49\columnwidth]{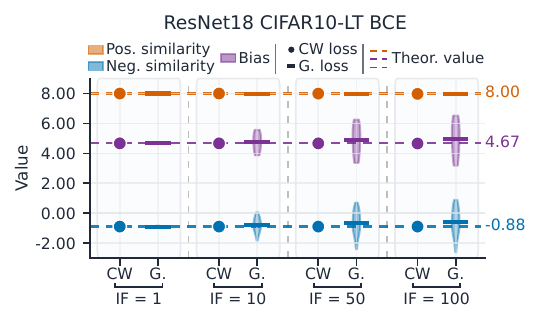}}
    \subfigure[CE]{\includegraphics[width=0.49\columnwidth]{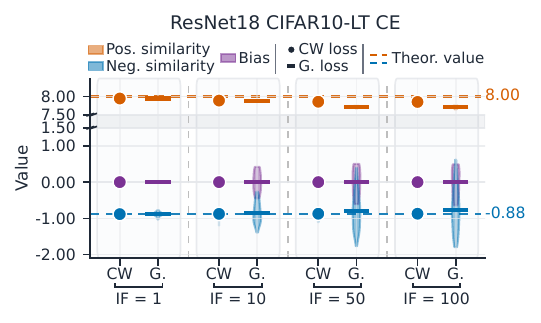}}
    \caption{Distributions of positive and negative scaled similarities and biases for ResNet18 trained on CIFAR10-LT. ``CW'' and ``G.'' denote class-wise and globally averaged losses, respectively.}
    \label{fig_supp_distribution_ResNet18_CIFAR10}
\end{figure*}

\begin{figure*}[!htb]
    \centering
    \subfigure[BCE]{\includegraphics[width=0.49\columnwidth]{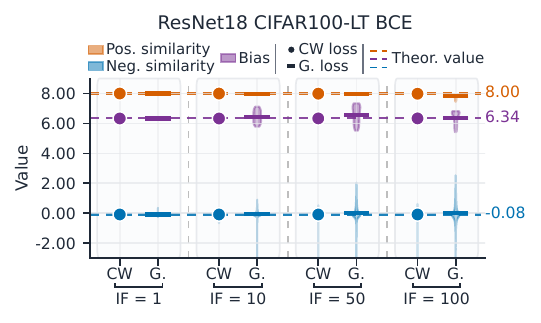}}
    \subfigure[CE]{\includegraphics[width=0.49\columnwidth]{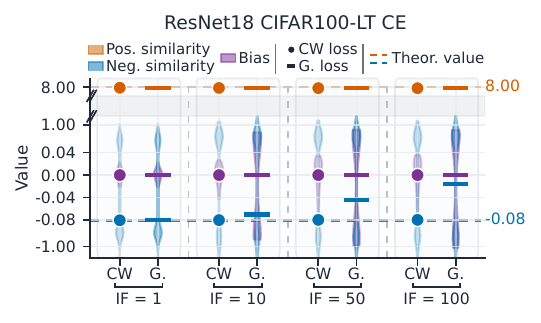}}
    \caption{Distributions of positive and negative scaled similarities and biases for ResNet18 trained on CIFAR100-LT. ``CW'' and ``G.'' denote class-wise and globally averaged losses, respectively.}
    \label{fig_supp_distribution_ResNet18_CIFAR100}
\end{figure*}

\begin{figure*}[!htb]
    \centering
    \subfigure[BCE]{\includegraphics[width=0.49\columnwidth]{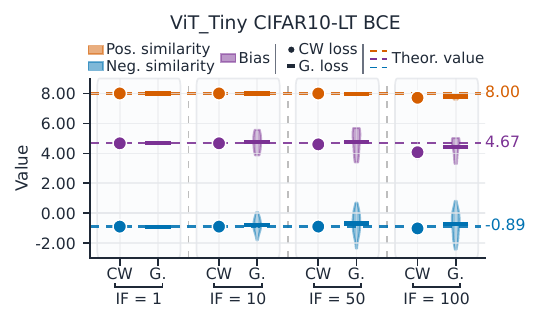}}
    \subfigure[CE]{\includegraphics[width=0.49\columnwidth]{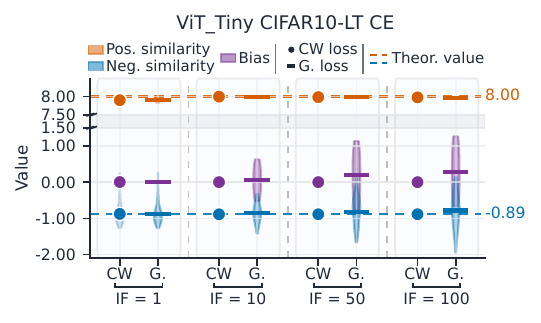}}
    \caption{Distributions of positive and negative scaled similarities and biases for ViT-Tiny trained on CIFAR10-LT. ``CW'' and ``G.'' denote class-wise and globally averaged losses, respectively.}
    \label{fig_supp_distribution_ViTTiny_CIFAR10}
\end{figure*}

\begin{figure*}[!htb]
    \centering
    \subfigure[BCE]{\includegraphics[width=0.49\columnwidth]{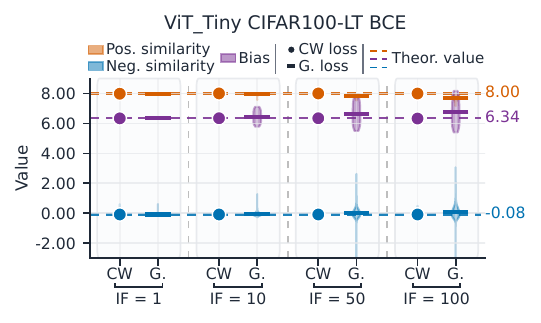}}
    \subfigure[CE]{\includegraphics[width=0.49\columnwidth]{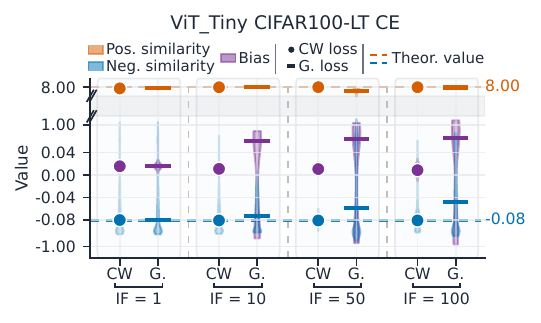}}
    \caption{Distributions of positive and negative scaled similarities and biases for ViT-Tiny trained on CIFAR100-LT. ``CW'' and ``G.'' denote class-wise and globally averaged losses, respectively.}
\label{fig_supp_distribution_ViTTiny_CIFAR100}
\end{figure*}

Figs. \ref{fig_supp_distribution_ResNet18_CIFAR10}--\ref{fig_supp_distribution_ViTTiny_CIFAR100} visually present statistical results for ResNet18 and ViT-Tiny trained on different imbalanced datasets. Consistent with the conclusions of the theorem, for models trained with CW averaged losses, their negative scaled similarity $\gamma\bm w_k^\top\bar{\bm x}_j$ is close to the NC value $-\gamma/(K-1)$ in most cases, especially for the models trained with BCE, where this remains true even as the IF increases. In contrast, for models trained with G. averaged losses, the negative feature similarity becomes increasingly more dispersed as the IF increases. For the models trained by different losses, their positive similarity remains close to $\gamma$ in most settings. For the BCE loss, the learned bias also tends toward its theoretical value.


\subsection{Similarity Matrices for Models Trained on Imbalanced Datasets}
\begin{figure*}[!htb]
	\centering
	\subfigure[CW averaged BCE]{
		\includegraphics*[scale=0.36, viewport=5 0 355 275]{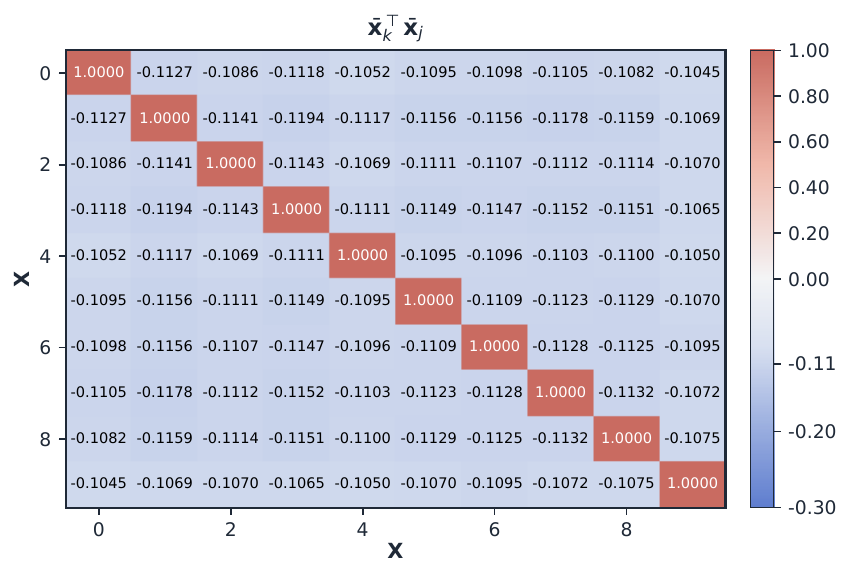}
		\includegraphics*[scale=0.36, viewport=5 0 355 275]{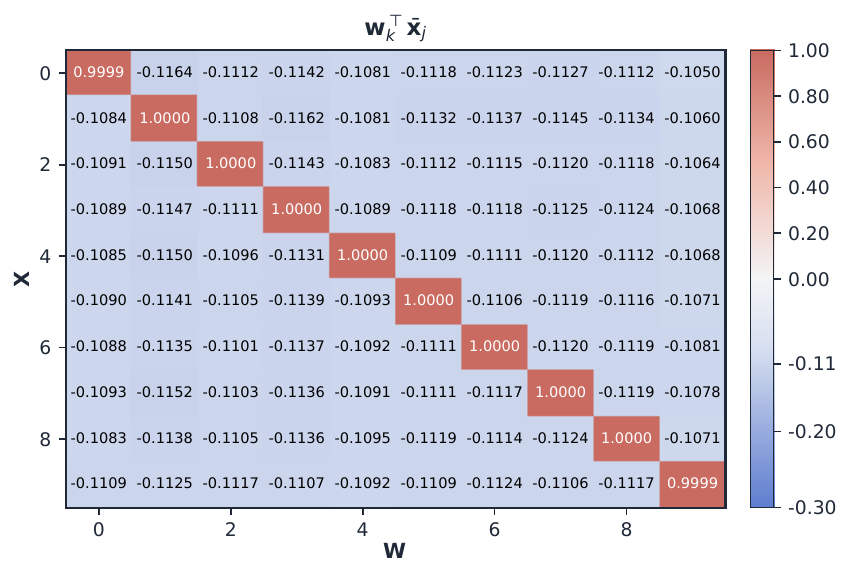}
		\includegraphics*[scale=0.36, viewport=5 0 440 275]{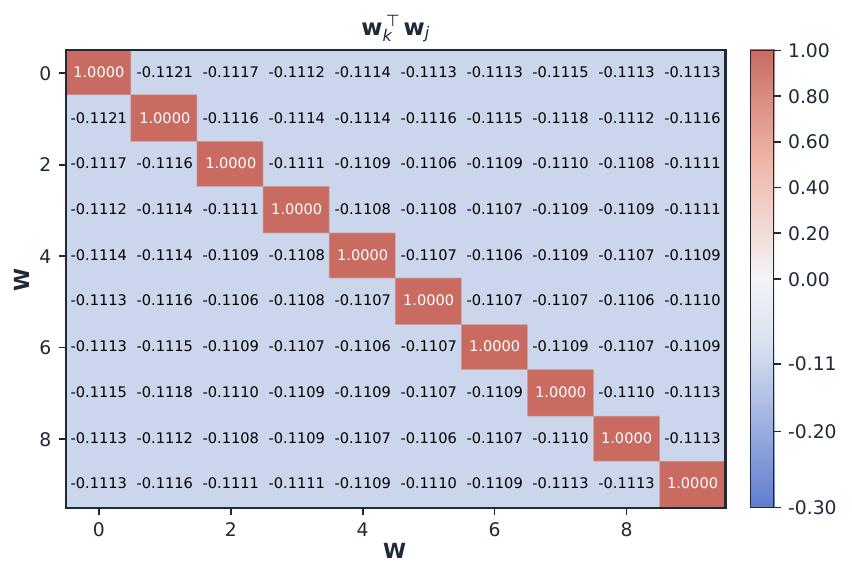}
	}
	\subfigure[G. averaged BCE]{
		\includegraphics*[scale=0.36, viewport=5 0 355 275]{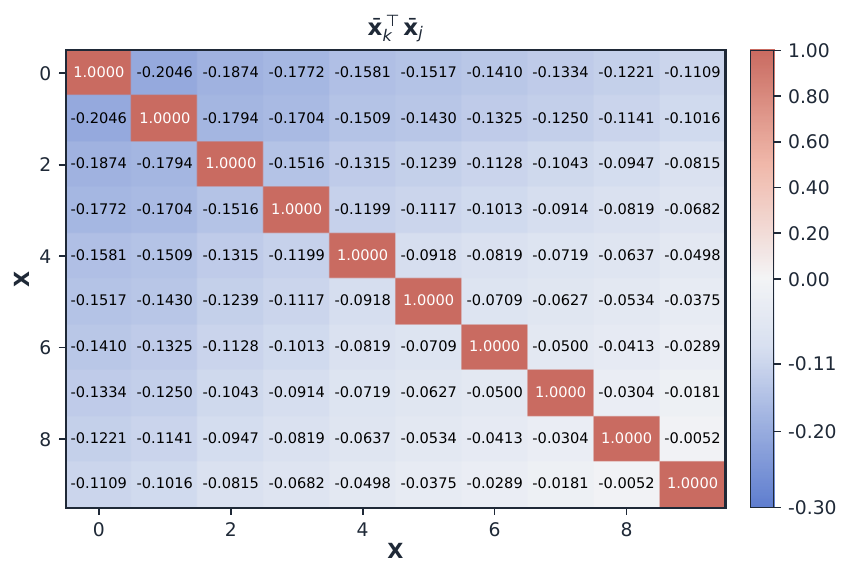}
		\includegraphics*[scale=0.36, viewport=5 0 355 275]{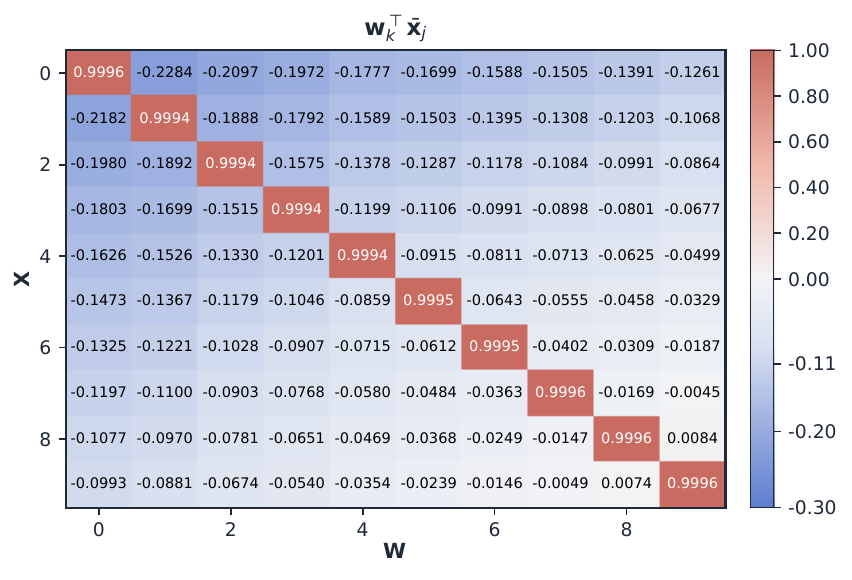}
		\includegraphics*[scale=0.36, viewport=5 0 440 275]{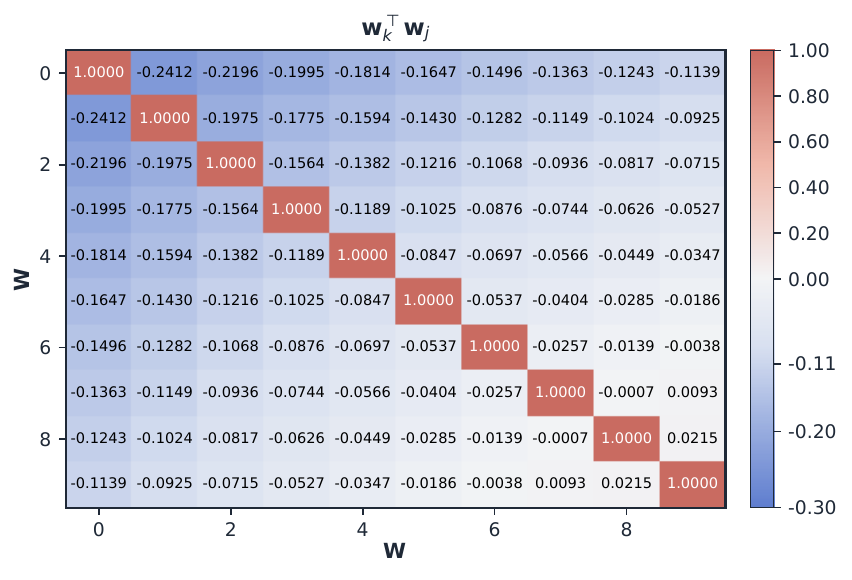}
	}
	\caption{Similarity matrices for class centers $\{\bar{\bm x}_k\}_{k=1}^K$ and classifier vectors $\{\bm w_k\}_{k=1}^K$ of ResNet18 trained with BCE on CIFAR10-LT with IF$=10$.}
	\label{fig_supp_matrix_ResNet18_cifar10_bce_10}
\end{figure*}

\begin{figure*}[!htb]
	\centering
	\subfigure[CW averaged BCE]{
		\includegraphics*[scale=0.36, viewport=5 0 355 275]{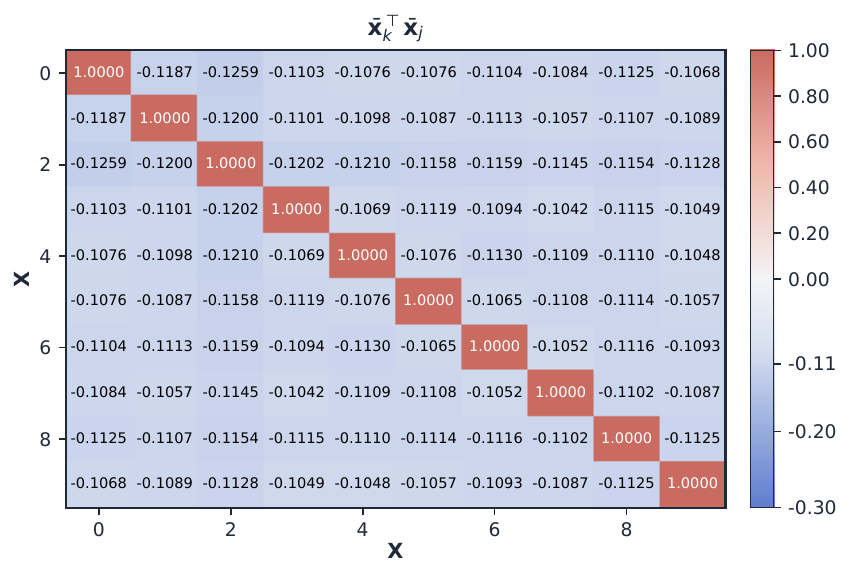}
		\includegraphics*[scale=0.36, viewport=5 0 355 275]{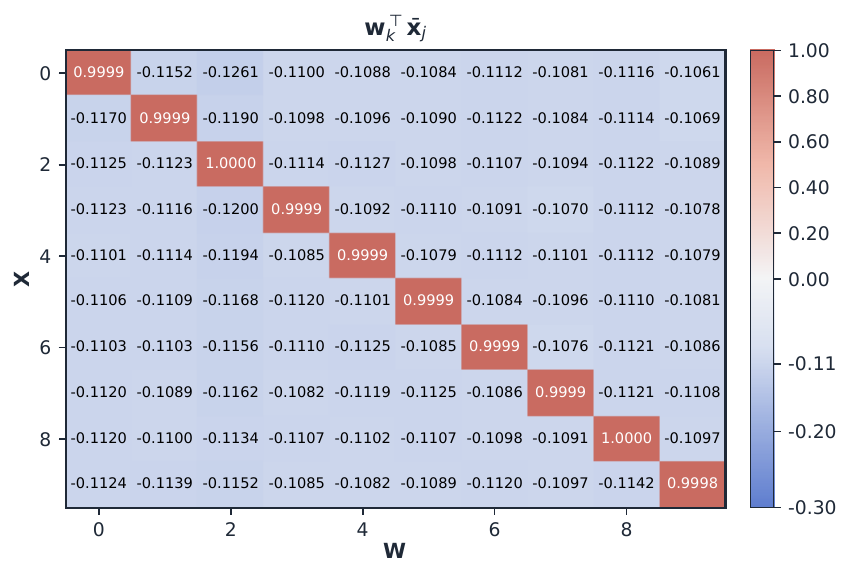}
		\includegraphics*[scale=0.36, viewport=5 0 440 275]{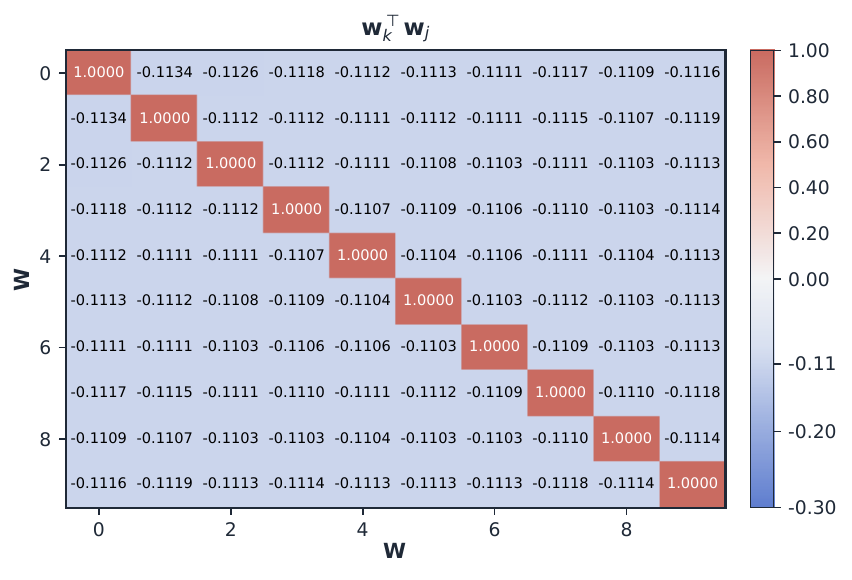}
	}
	\subfigure[G. averaged BCE]{
		\includegraphics*[scale=0.36, viewport=5 0 355 275]{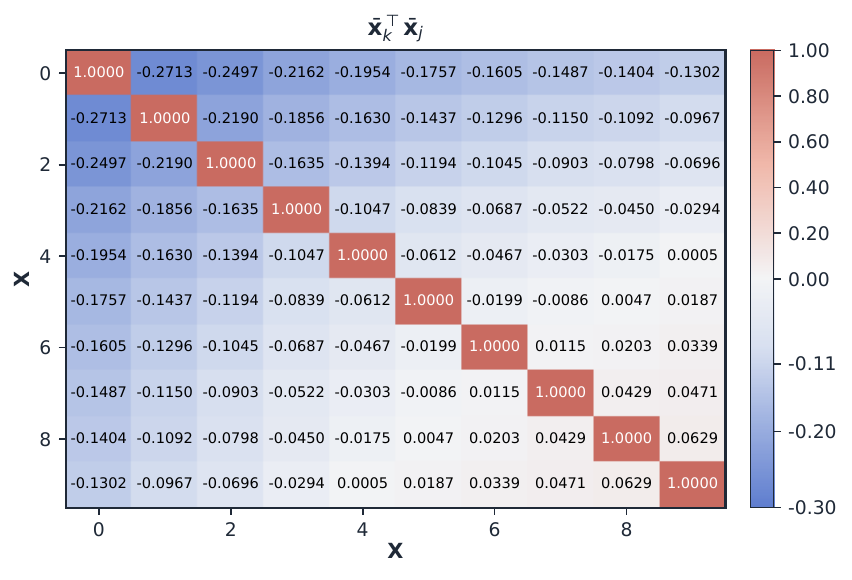}
		\includegraphics*[scale=0.36, viewport=5 0 355 275]{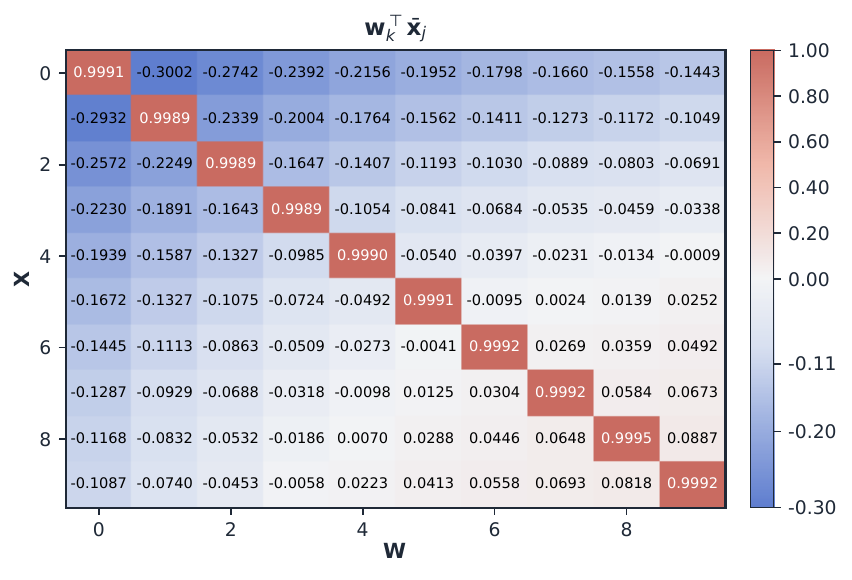}
		\includegraphics*[scale=0.36, viewport=5 0 440 275]{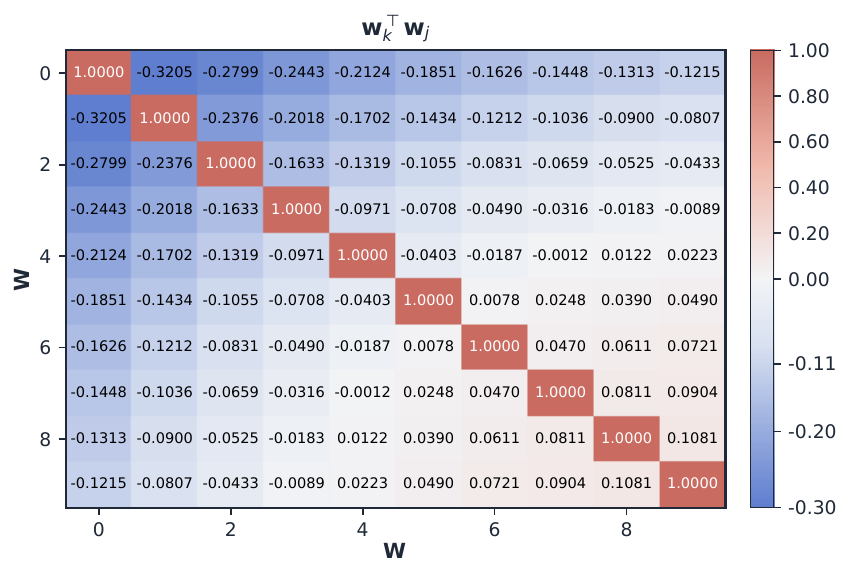}
	}
	\caption{Similarity matrices for class centers $\{\bar{\bm x}_k\}_{k=1}^K$ and classifier vectors $\{\bm w_k\}_{k=1}^K$ of ResNet18 trained with BCE on CIFAR10-LT with IF$=50$.}
	\label{fig_supp_matrix_ResNet18_cifar10_bce_50}
\end{figure*}

\begin{figure*}[!htb]
	\centering
	\subfigure[CW averaged BCE]{
		\includegraphics*[scale=0.36, viewport=5 0 355 275]{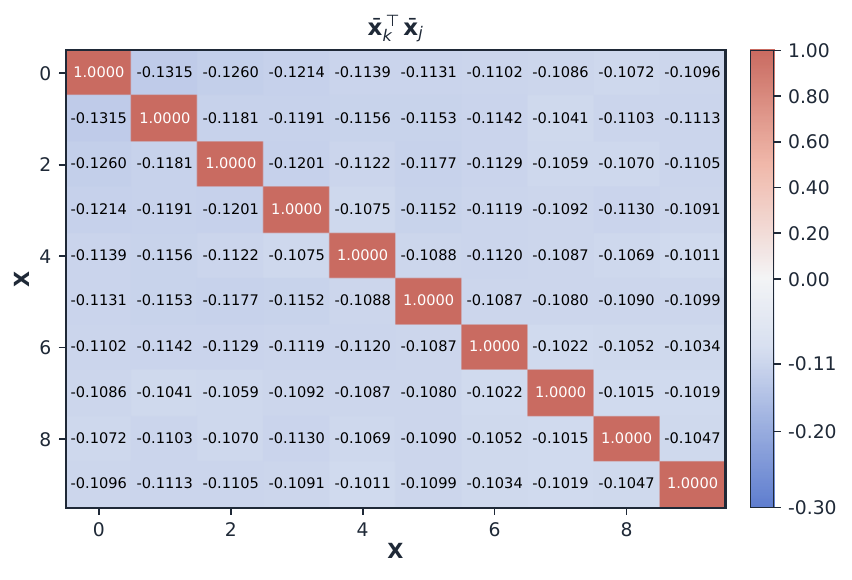}
		\includegraphics*[scale=0.36, viewport=5 0 355 275]{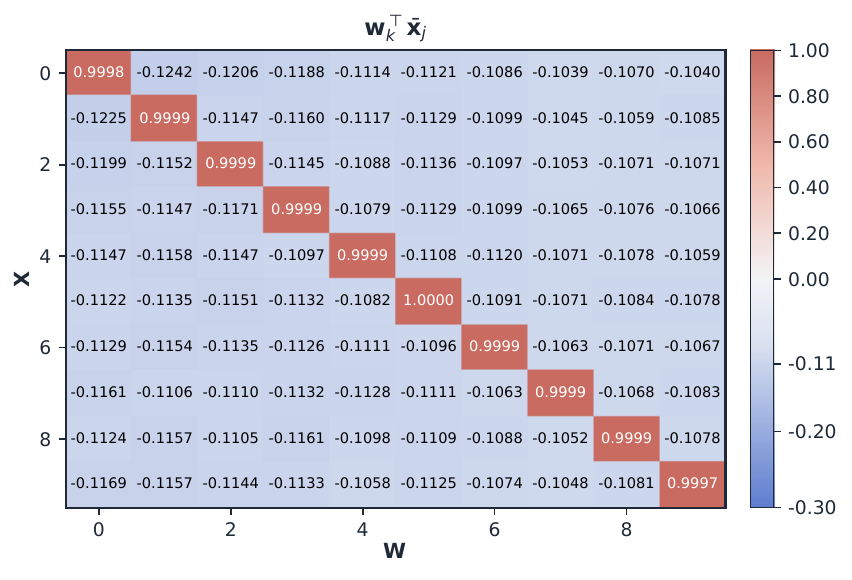}
		\includegraphics*[scale=0.36, viewport=5 0 440 275]{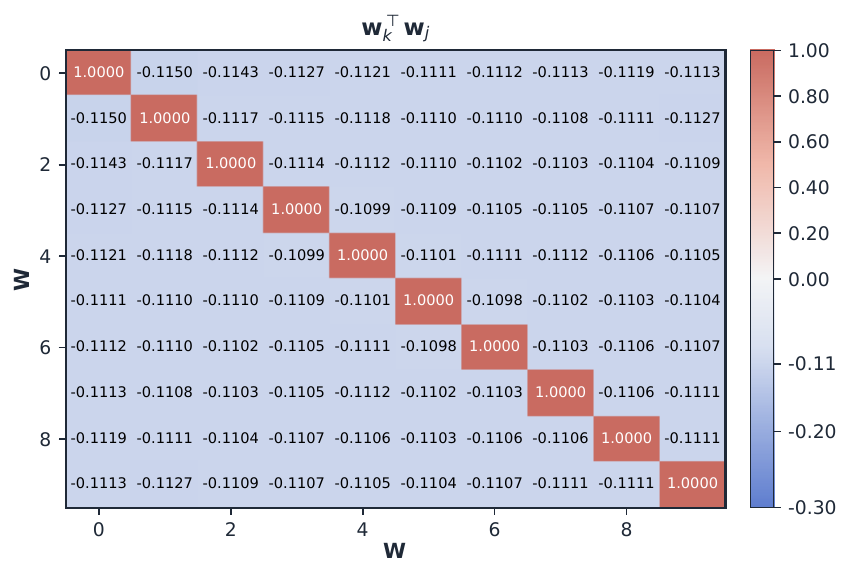}
	}
	\subfigure[G. averaged BCE]{
		\includegraphics*[scale=0.36, viewport=5 0 355 275]{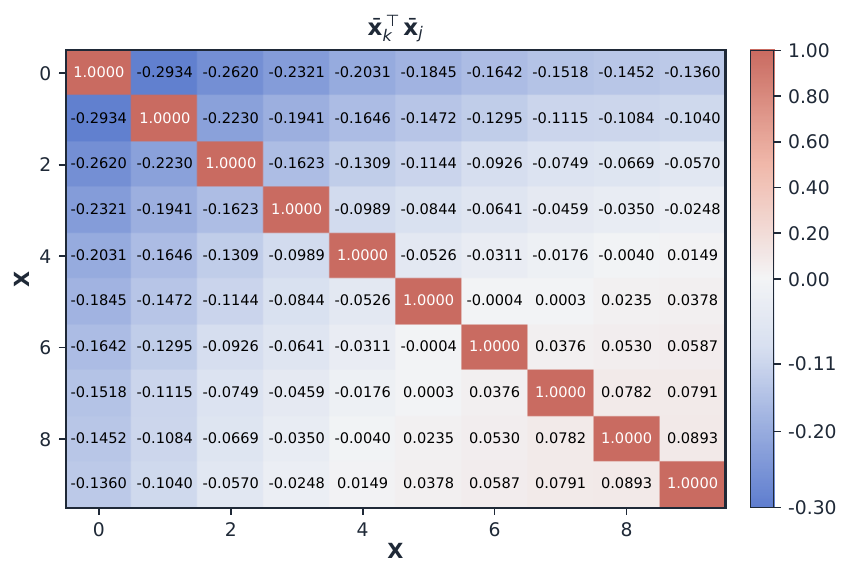}
		\includegraphics*[scale=0.36, viewport=5 0 355 275]{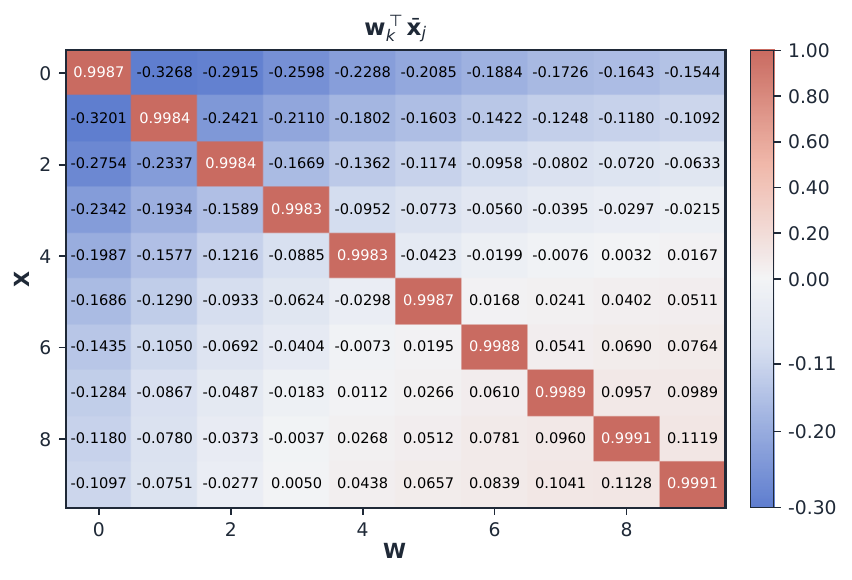}
		\includegraphics*[scale=0.36, viewport=5 0 440 275]{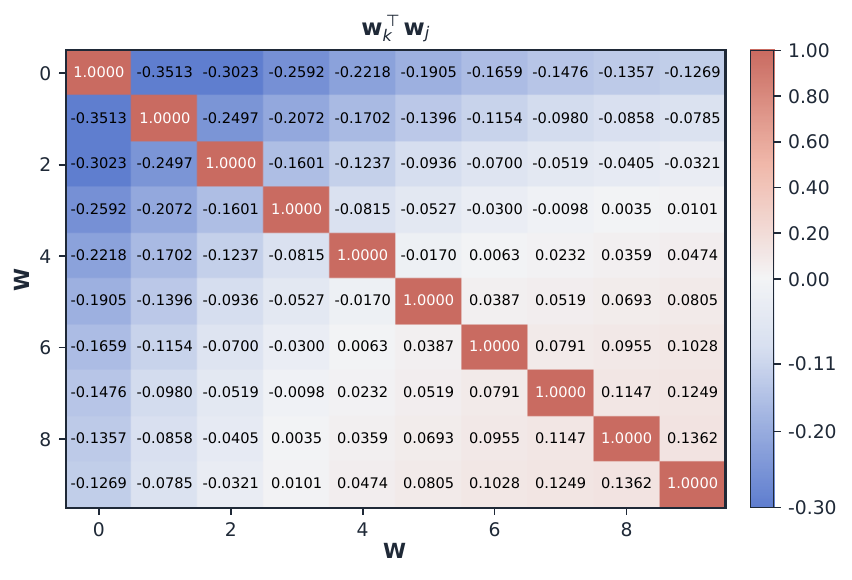}
	}
	\caption{Similarity matrices for class centers $\{\bar{\bm x}_k\}_{k=1}^K$ and classifier vectors $\{\bm w_k\}_{k=1}^K$ of ResNet18 trained with BCE on CIFAR10-LT with IF$=100$.}
	\label{fig_supp_matrix_ResNet18_cifar10_bce_100}
\end{figure*}

\begin{figure*}[!htb]
	\centering
	\subfigure[CW averaged CE]{
		\includegraphics*[scale=0.36, viewport=5 0 355 275]{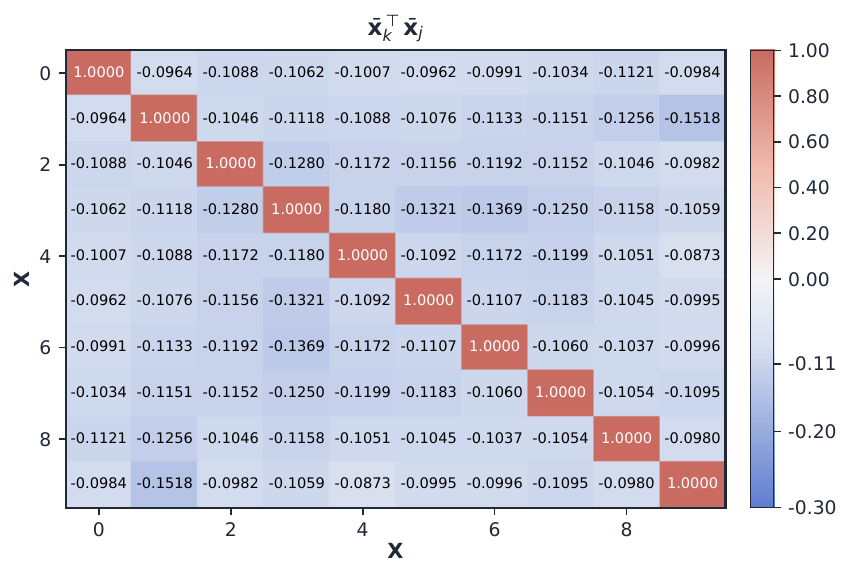}
		\includegraphics*[scale=0.36, viewport=5 0 355 275]{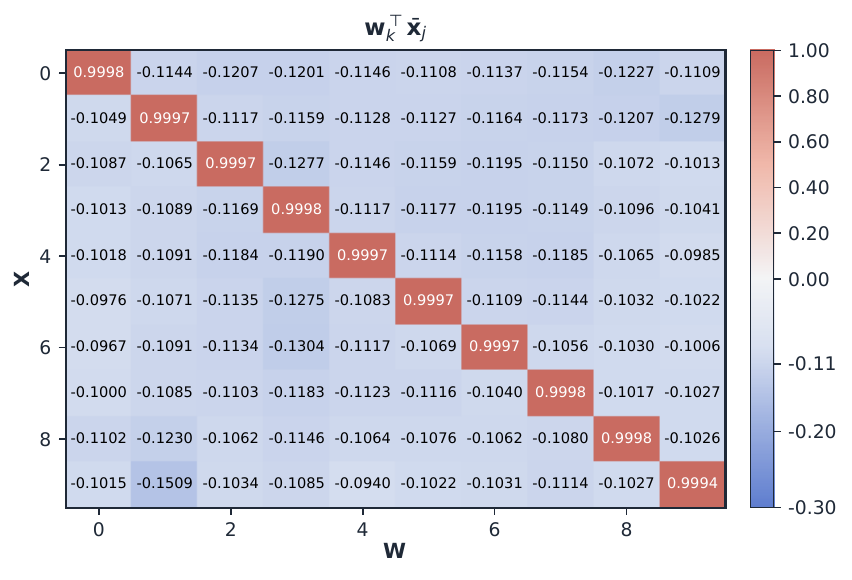}
		\includegraphics*[scale=0.36, viewport=5 0 440 275]{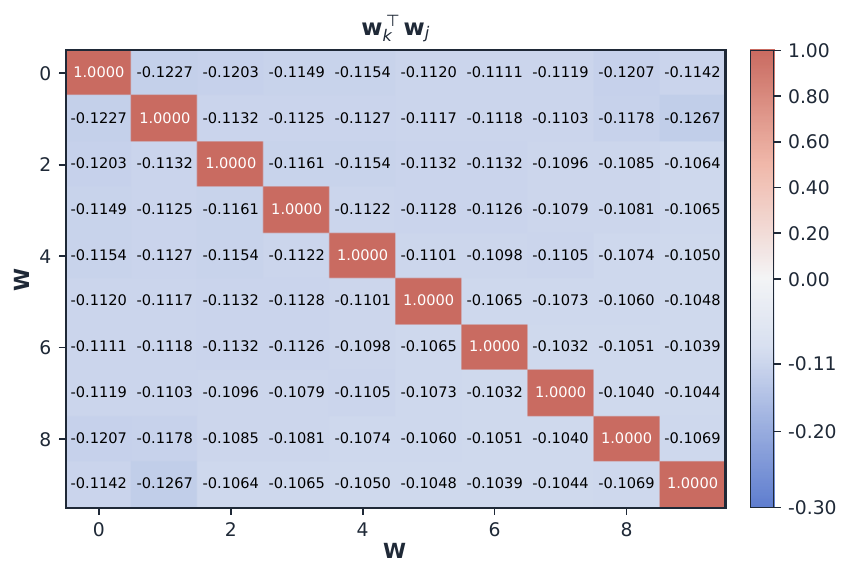}
	}
	\subfigure[G. averaged CE]{
		\includegraphics*[scale=0.36, viewport=5 0 355 275]{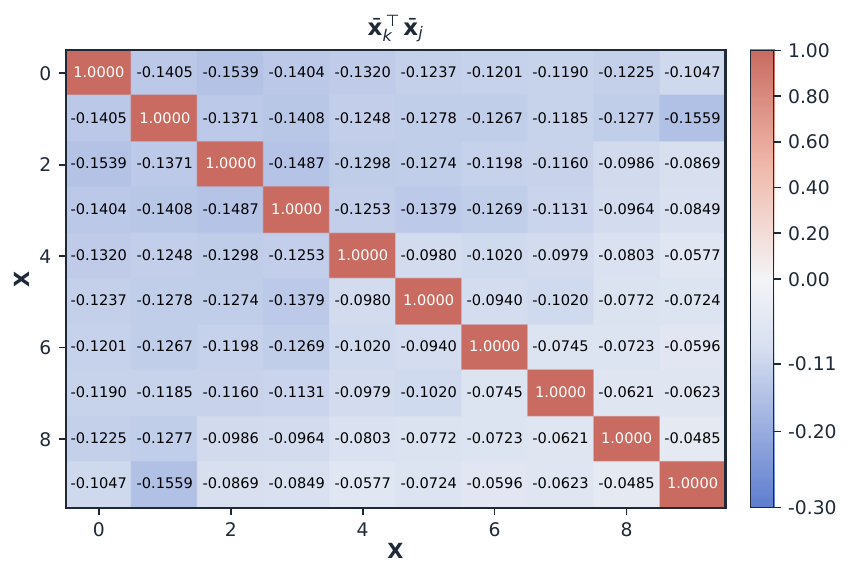}
		\includegraphics*[scale=0.36, viewport=5 0 355 275]{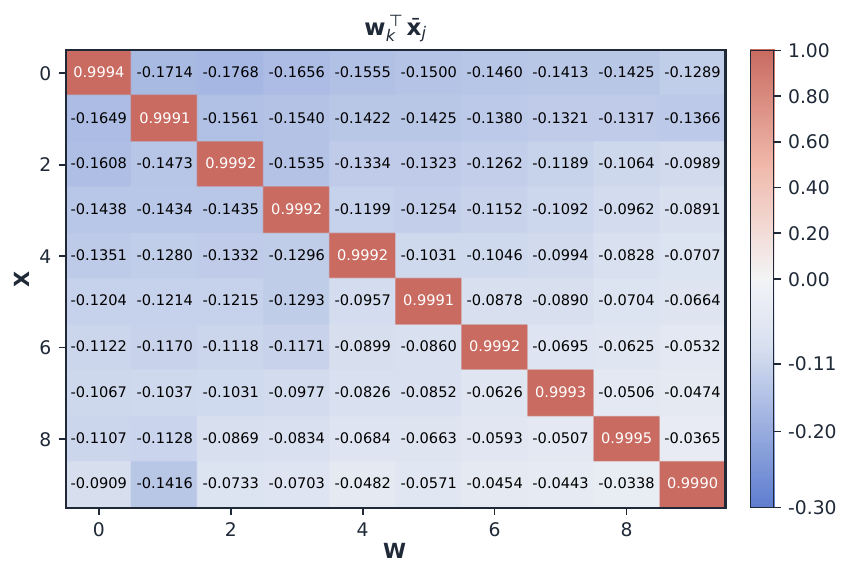}
		\includegraphics*[scale=0.36, viewport=5 0 440 275]{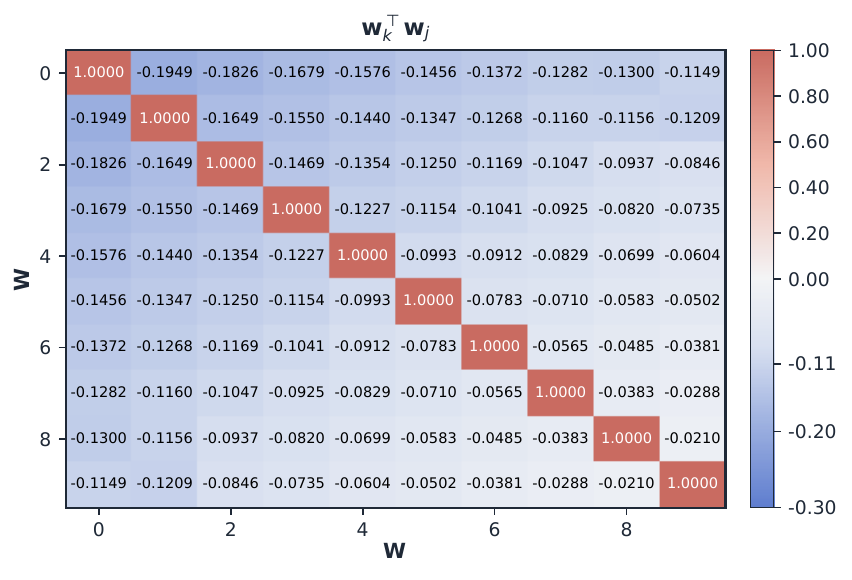}
	}
	\caption{Similarity matrices for class centers $\{\bar{\bm x}_k\}_{k=1}^K$ and classifier vectors $\{\bm w_k\}_{k=1}^K$ of ResNet18 trained with CE on CIFAR10-LT with IF$=10$.}
	\label{fig_supp_matrix_ResNet18_cifar10_ce_10}
\end{figure*}

\begin{figure*}[!htb]
	\centering
	\subfigure[CW averaged CE]{
		\includegraphics*[scale=0.36, viewport=5 0 355 275]{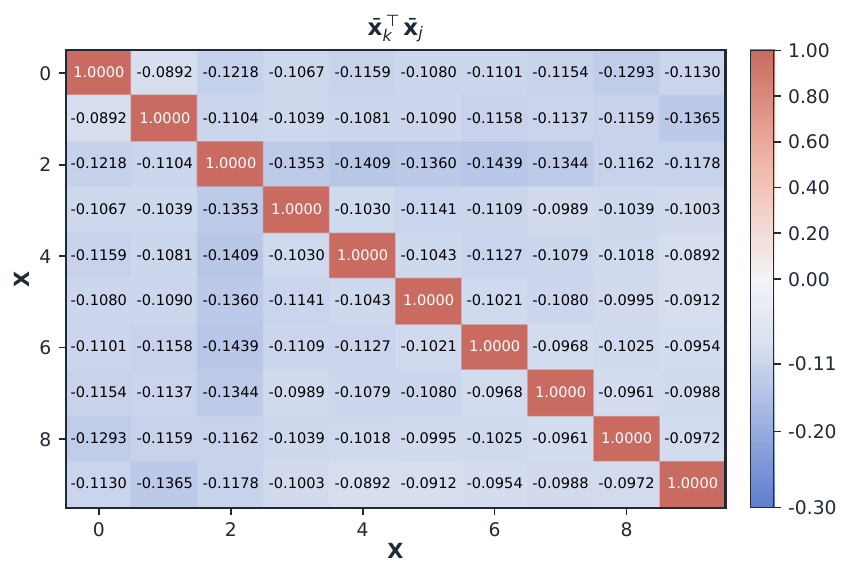}
		\includegraphics*[scale=0.36, viewport=5 0 355 275]{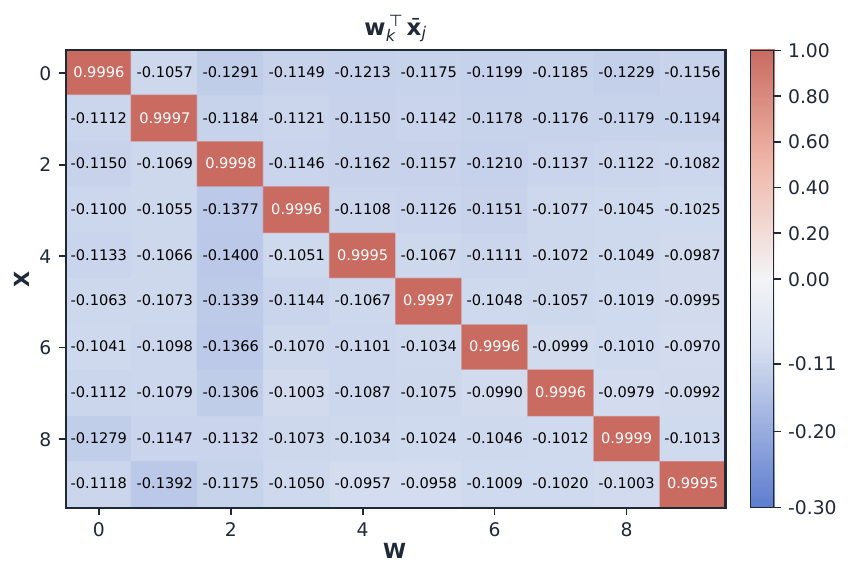}
		\includegraphics*[scale=0.36, viewport=5 0 440 275]{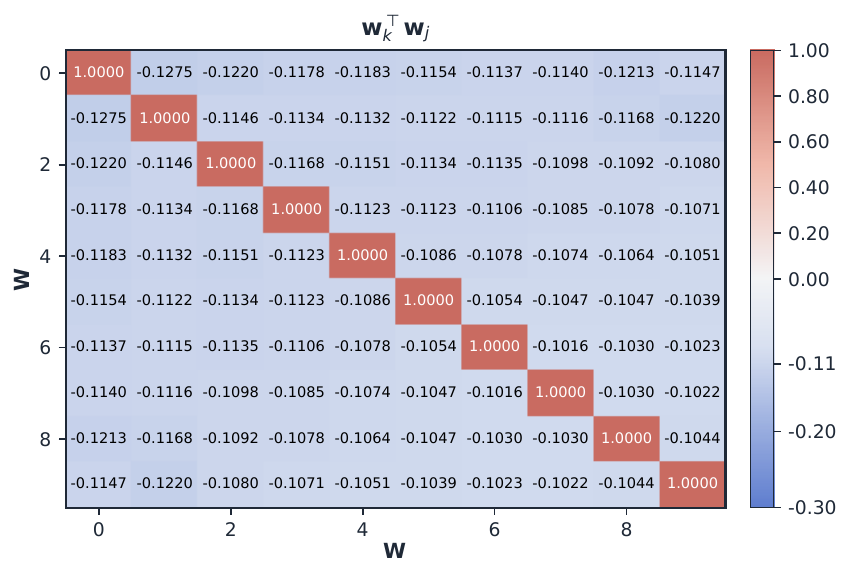}
	}
	\subfigure[G. averaged CE]{
		\includegraphics*[scale=0.36, viewport=5 0 355 275]{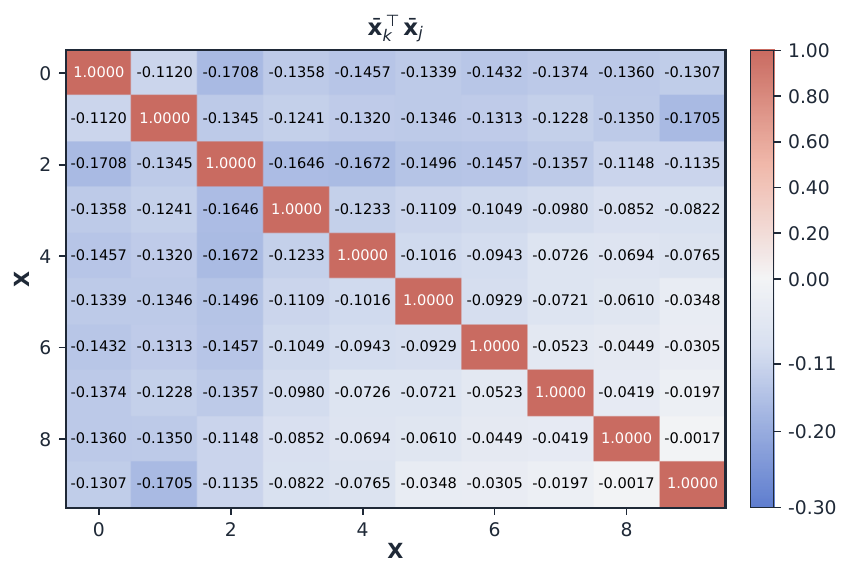}
		\includegraphics*[scale=0.36, viewport=5 0 355 275]{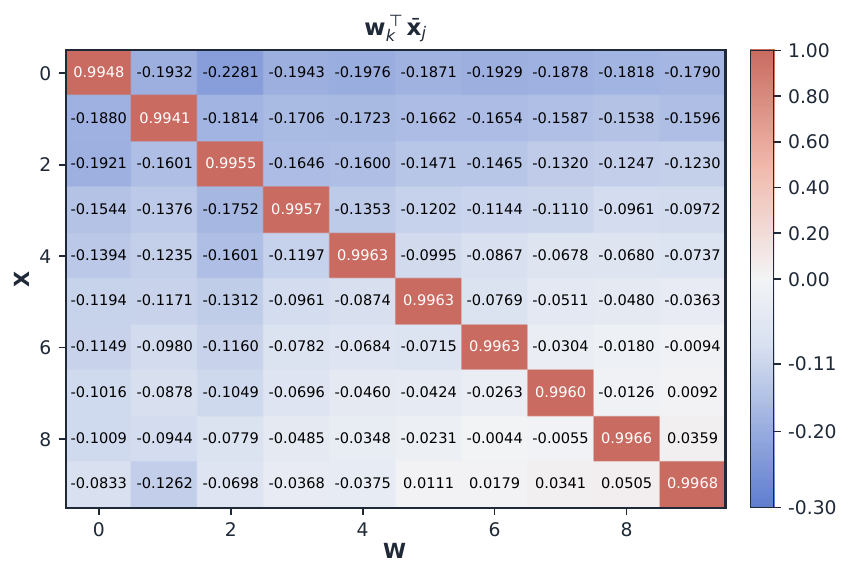}
		\includegraphics*[scale=0.36, viewport=5 0 440 275]{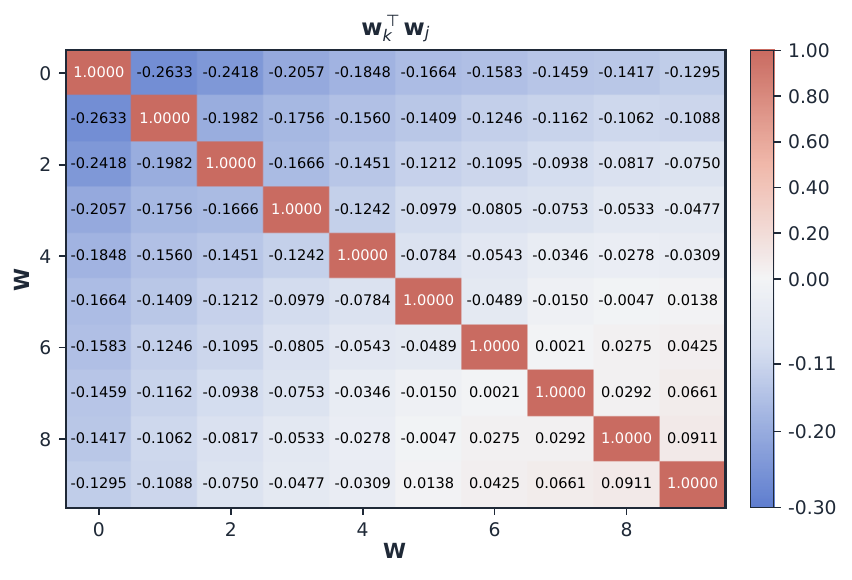}
	}
	\caption{Similarity matrices for class centers $\{\bar{\bm x}_k\}_{k=1}^K$ and classifier vectors $\{\bm w_k\}_{k=1}^K$ of ResNet18 trained with CE on CIFAR10-LT with IF$=50$.}
	\label{fig_supp_matrix_ResNet18_cifar10_ce_50}
\end{figure*}

\begin{figure*}[!htb]
	\centering
	\subfigure[CW averaged CE]{
		\includegraphics*[scale=0.36, viewport=5 0 355 275]{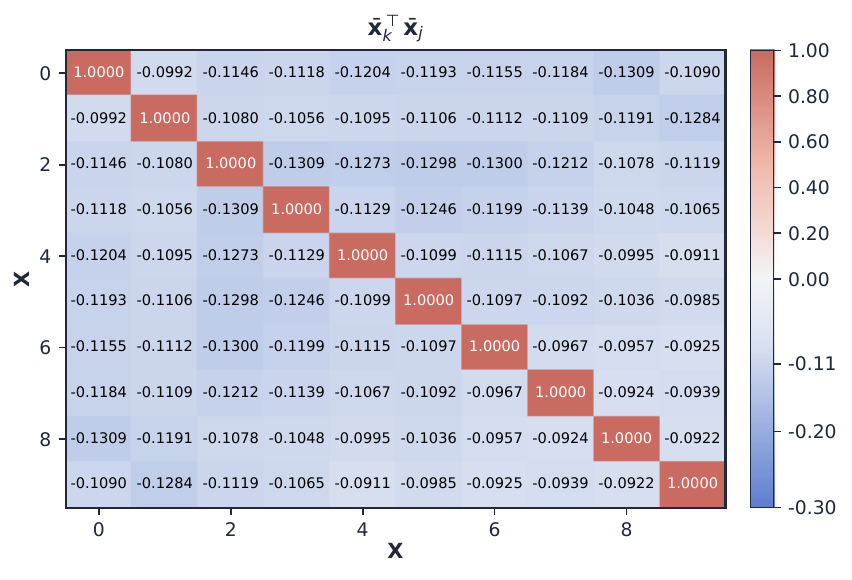}
		\includegraphics*[scale=0.36, viewport=5 0 355 275]{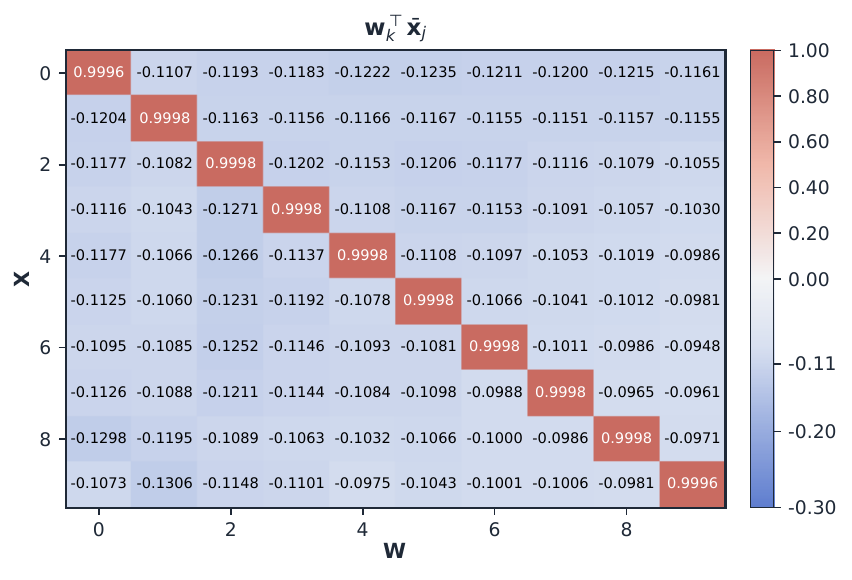}
		\includegraphics*[scale=0.36, viewport=5 0 440 275]{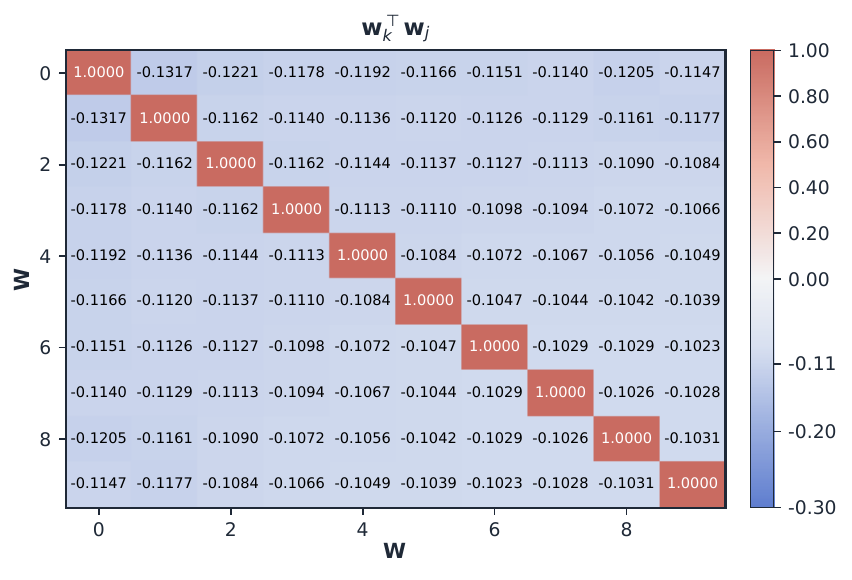}
	}
	\subfigure[G. averaged CE]{
		\includegraphics*[scale=0.36, viewport=5 0 355 275]{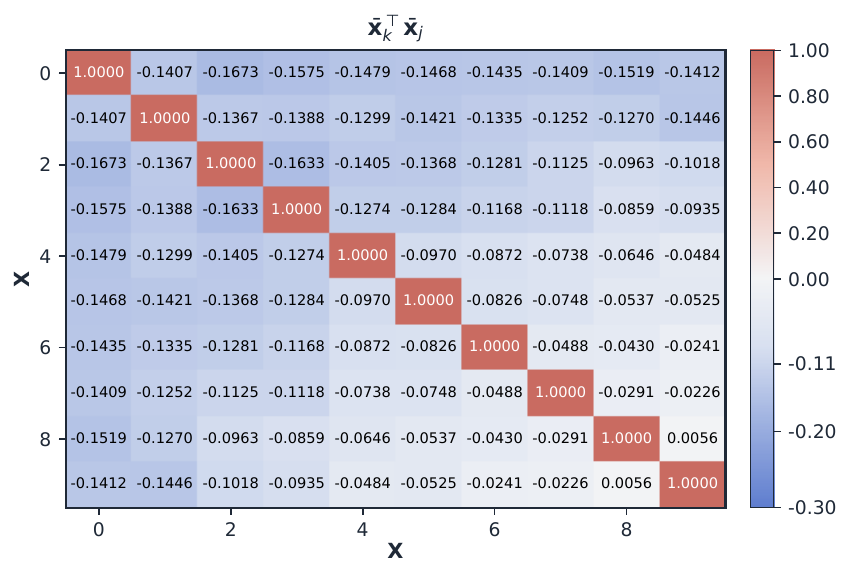}
		\includegraphics*[scale=0.36, viewport=5 0 355 275]{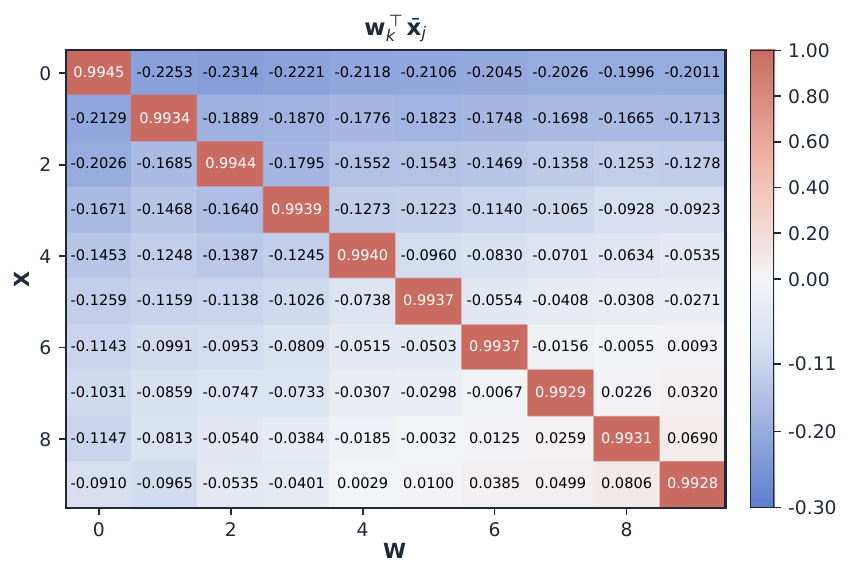}
		\includegraphics*[scale=0.36, viewport=5 0 440 275]{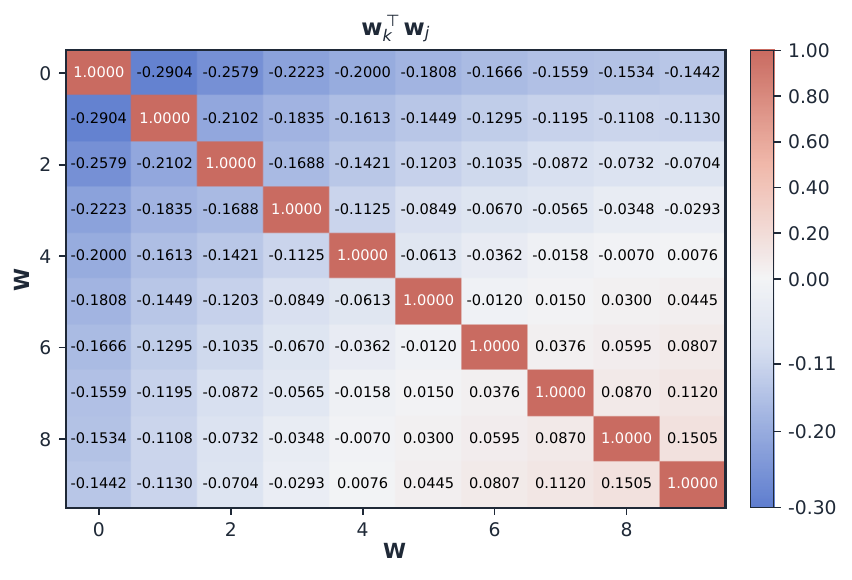}
	}
	\caption{Similarity matrices for class centers $\{\bar{\bm x}_k\}_{k=1}^K$ and classifier vectors $\{\bm w_k\}_{k=1}^K$ of ResNet18 trained with CE on CIFAR10-LT with IF$=100$.}
	\label{fig_supp_matrix_ResNet18_cifar10_ce_100}
\end{figure*}

\begin{figure*}[!htb]
	\centering
	\subfigure[CW averaged BCE]{
		\includegraphics*[scale=0.36, viewport=5 0 355 275]{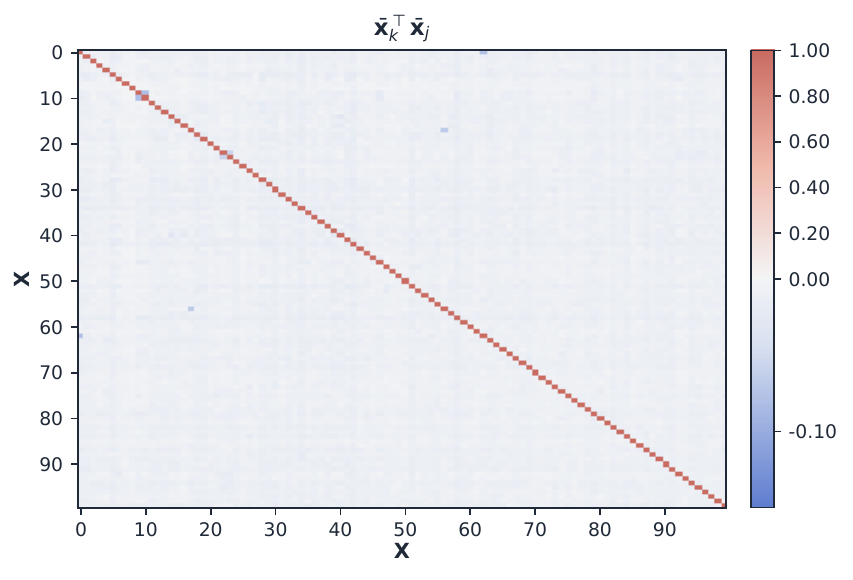}
		\includegraphics*[scale=0.36, viewport=5 0 355 275]{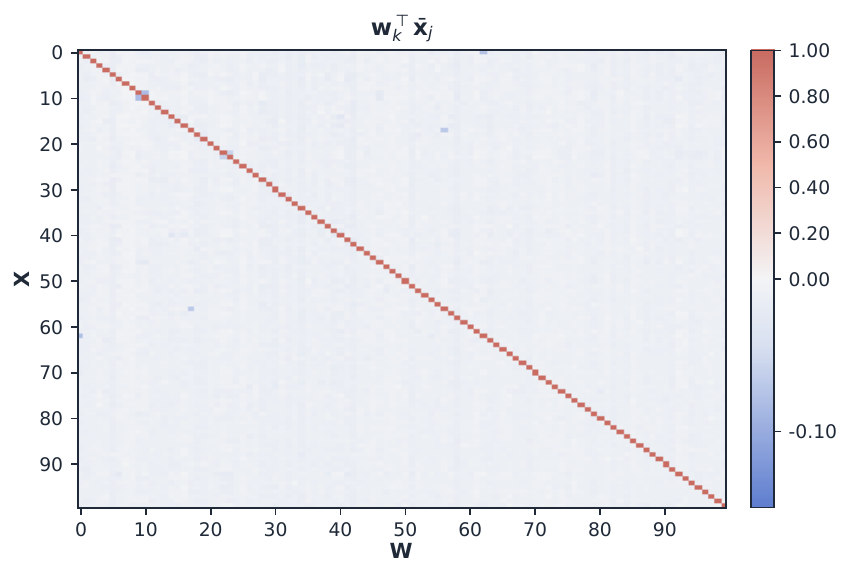}
		\includegraphics*[scale=0.36, viewport=5 0 440 275]{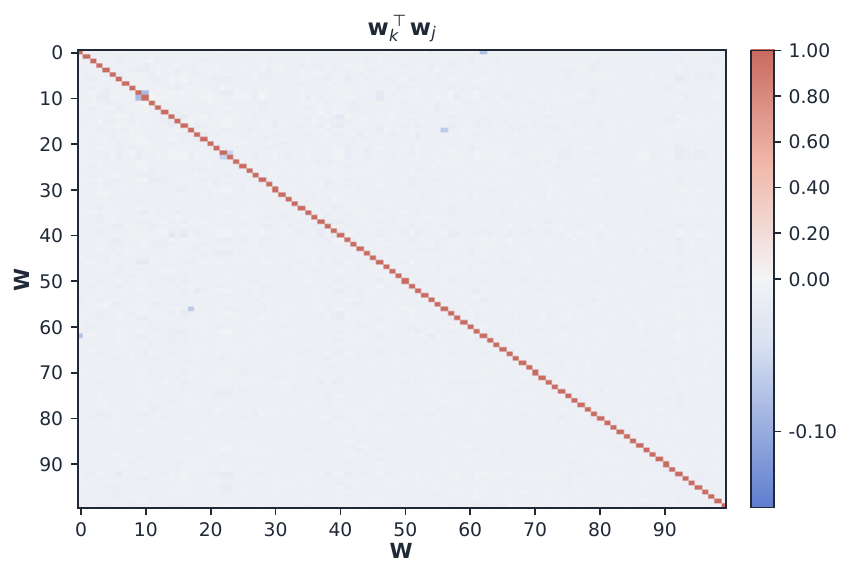}
	}
	\subfigure[G. averaged BCE]{
		\includegraphics*[scale=0.36, viewport=5 0 355 275]{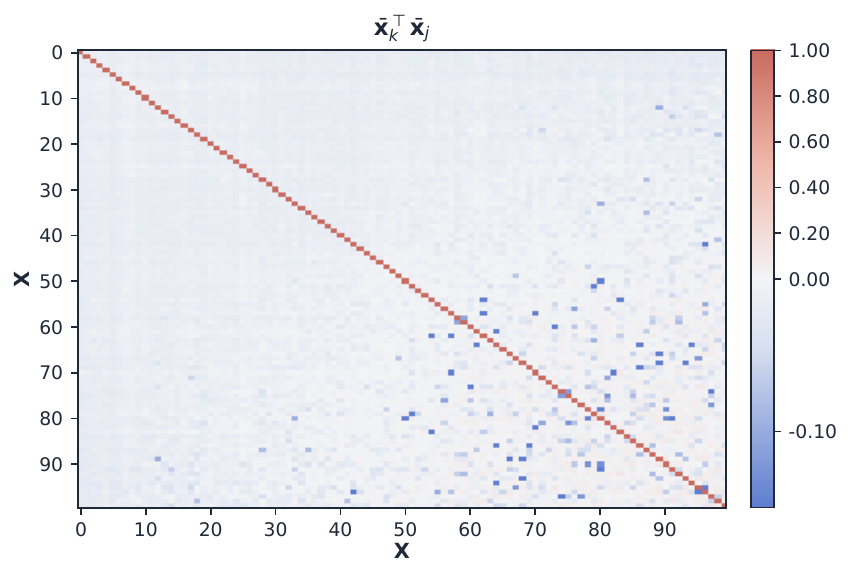}
		\includegraphics*[scale=0.36, viewport=5 0 355 275]{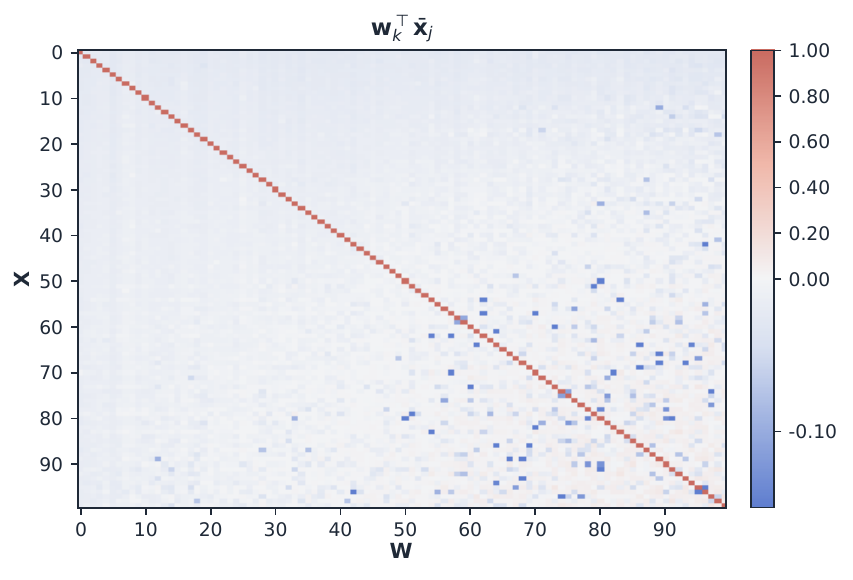}
		\includegraphics*[scale=0.36, viewport=5 0 440 275]{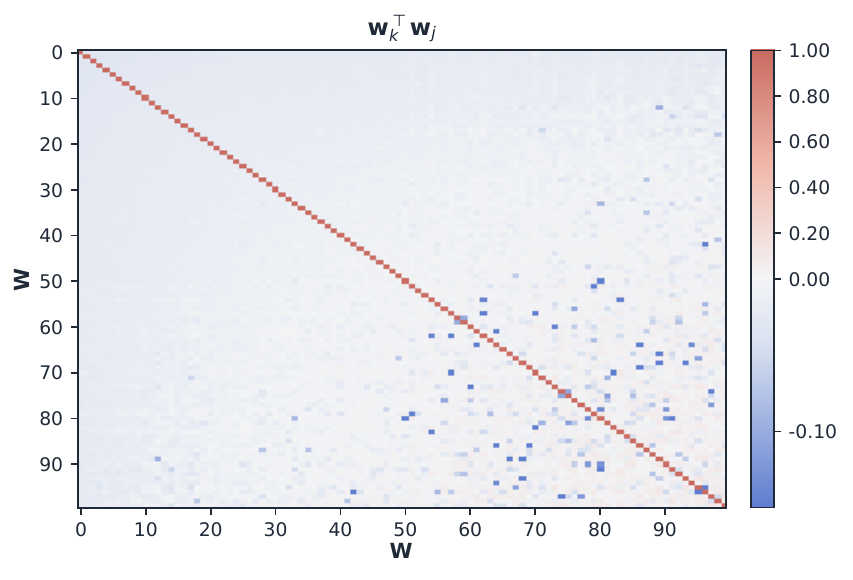}
	}
	\caption{Similarity matrices for class centers $\{\bar{\bm x}_k\}_{k=1}^K$ and classifier vectors $\{\bm w_k\}_{k=1}^K$ of ResNet18 trained with BCE on CIFAR100-LT with IF$=10$.}
	\label{fig_supp_matrix_ResNet18_cifar100_bce_10}
\end{figure*}

\begin{figure*}[!htb]
	\centering
	\subfigure[CW averaged BCE]{
		\includegraphics*[scale=0.36, viewport=5 0 355 275]{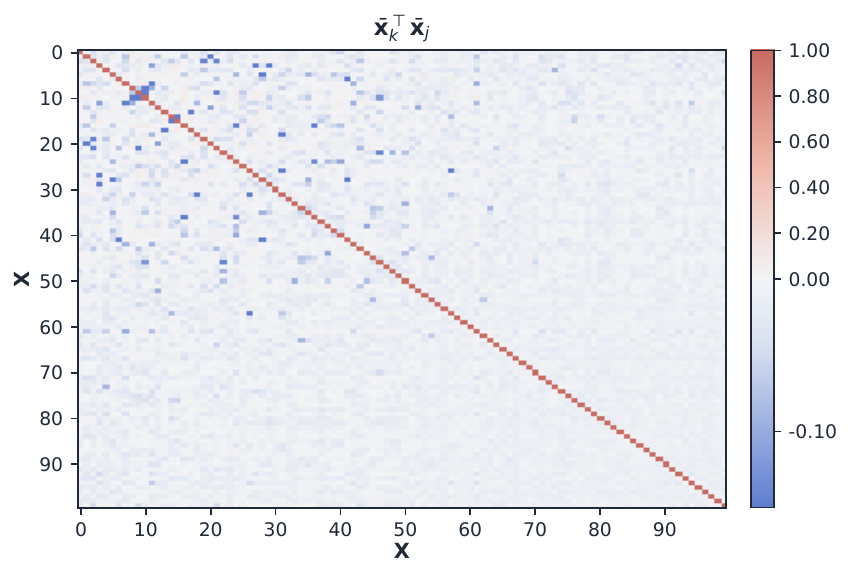}
		\includegraphics*[scale=0.36, viewport=5 0 355 275]{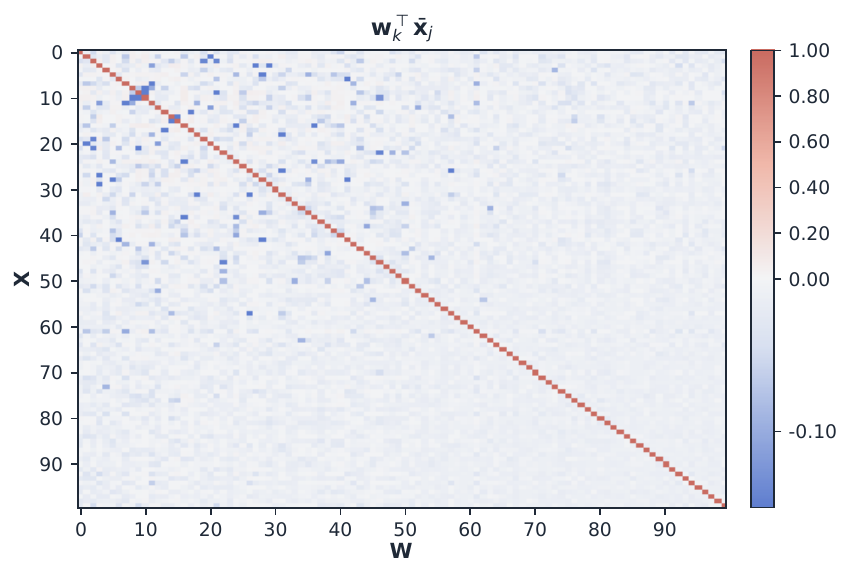}
		\includegraphics*[scale=0.36, viewport=5 0 440 275]{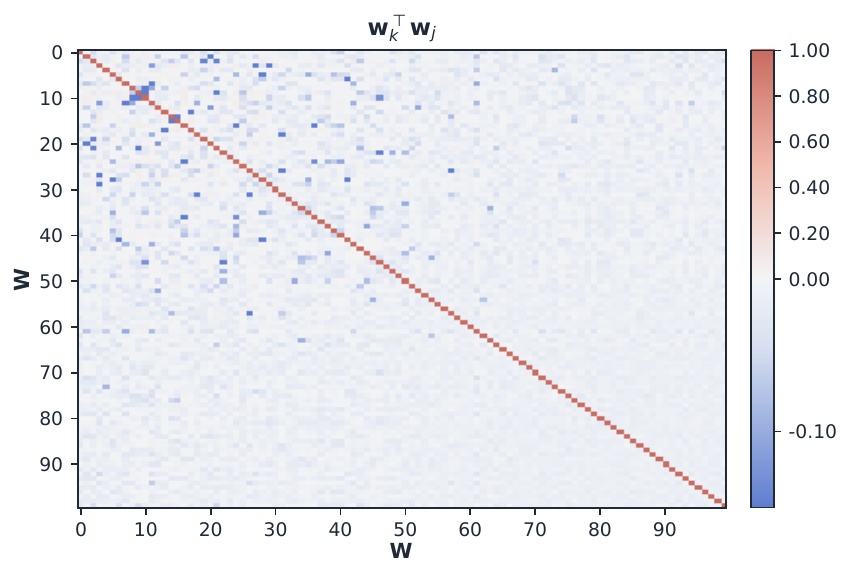}
	}
	\subfigure[G. averaged BCE]{
		\includegraphics*[scale=0.36, viewport=5 0 355 275]{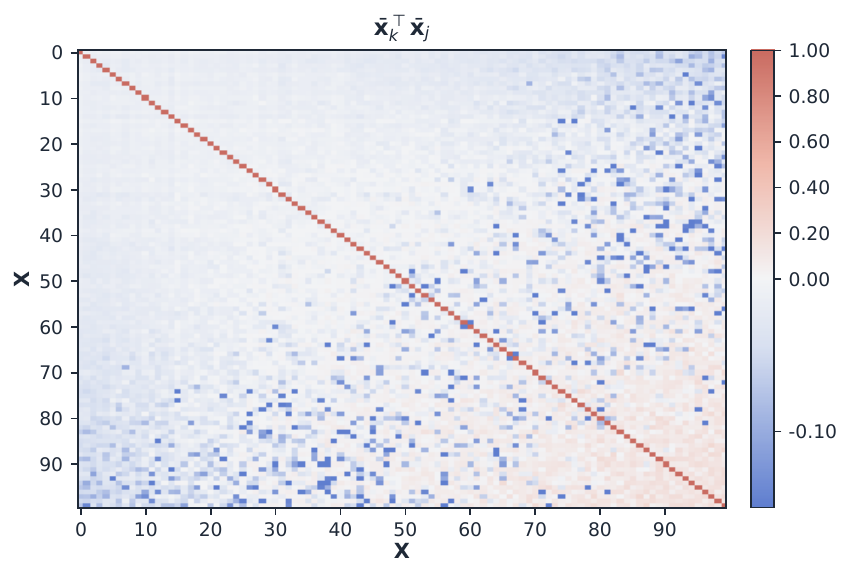}
		\includegraphics*[scale=0.36, viewport=5 0 355 275]{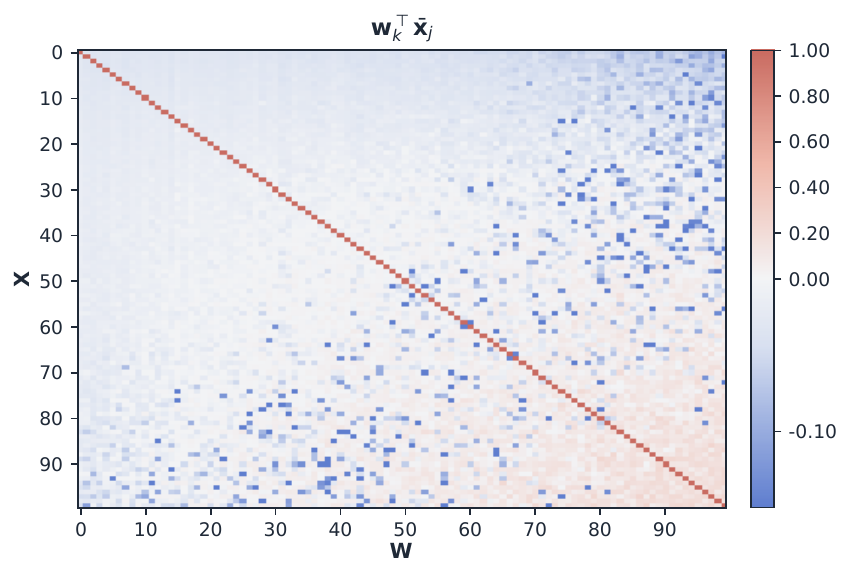}
		\includegraphics*[scale=0.36, viewport=5 0 440 275]{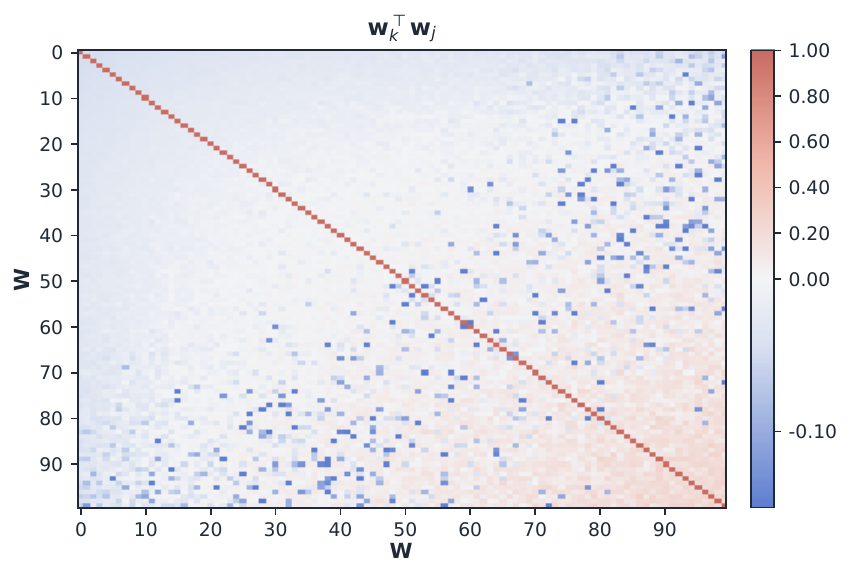}
	}
	\caption{Similarity matrices for class centers $\{\bar{\bm x}_k\}_{k=1}^K$ and classifier vectors $\{\bm w_k\}_{k=1}^K$ of ResNet18 trained with BCE on CIFAR100-LT with IF$=50$.}
	\label{fig_supp_matrix_ResNet18_cifar100_bce_50}
\end{figure*}

\begin{figure*}[!htb]
	\centering
	\subfigure[CW averaged BCE]{
		\includegraphics*[scale=0.36, viewport=5 0 355 275]{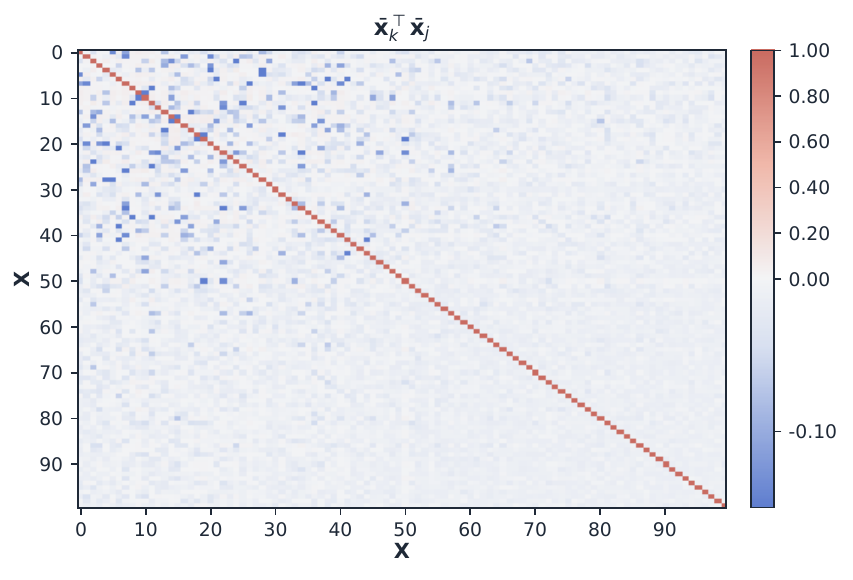}
		\includegraphics*[scale=0.36, viewport=5 0 355 275]{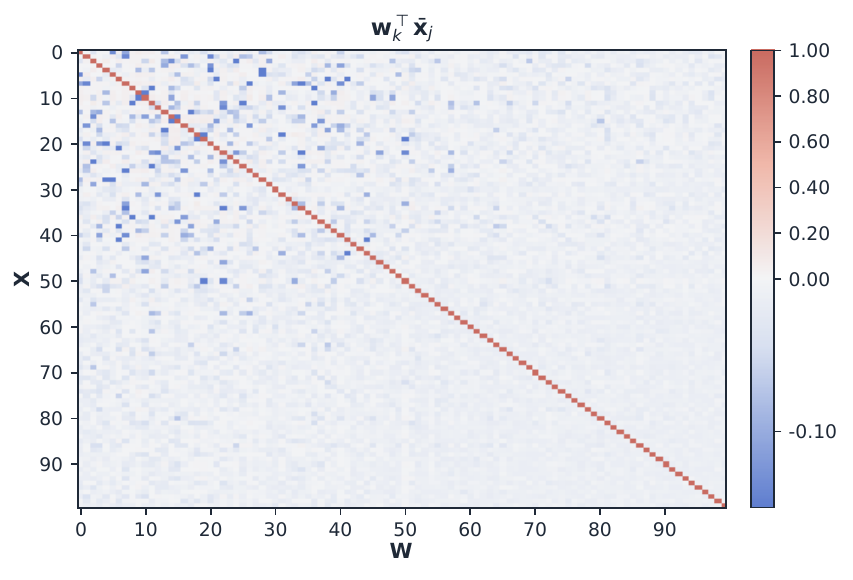}
		\includegraphics*[scale=0.36, viewport=5 0 440 275]{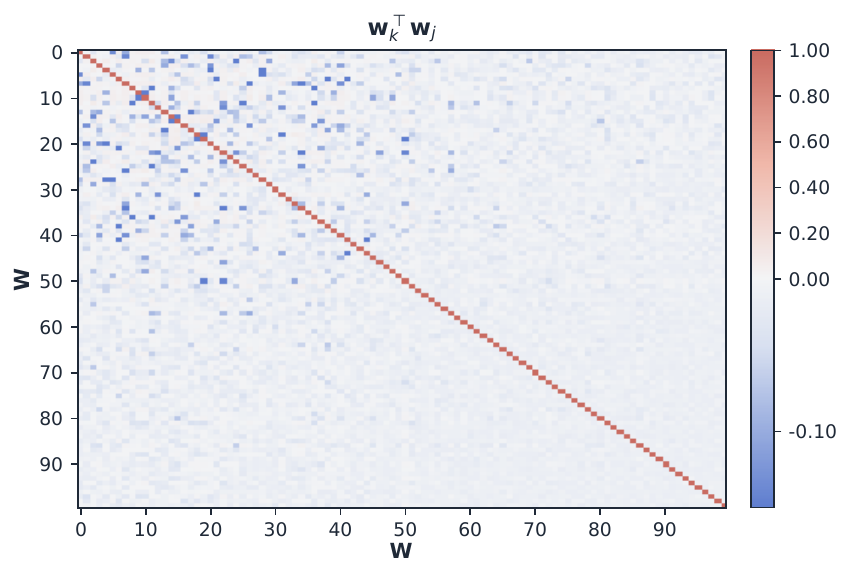}
	}
	\subfigure[G. averaged BCE]{
		\includegraphics*[scale=0.36, viewport=5 0 355 275]{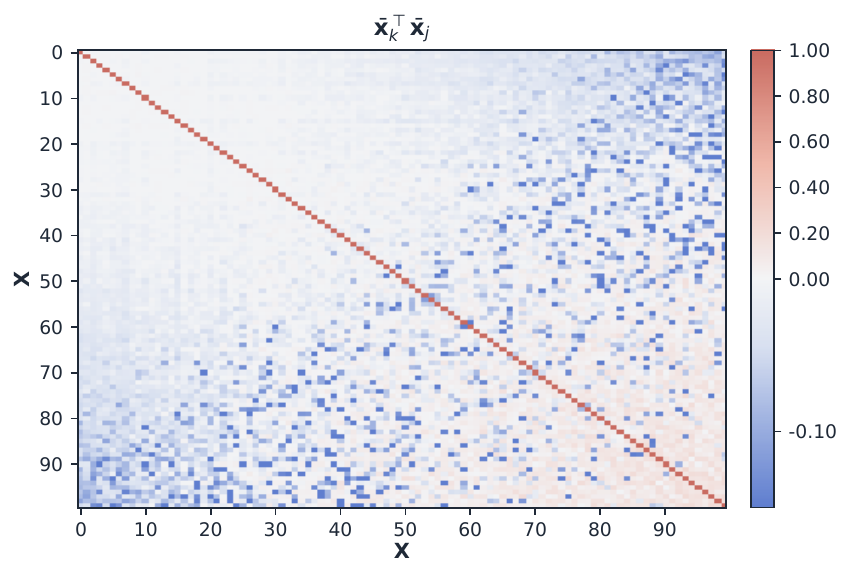}
		\includegraphics*[scale=0.36, viewport=5 0 355 275]{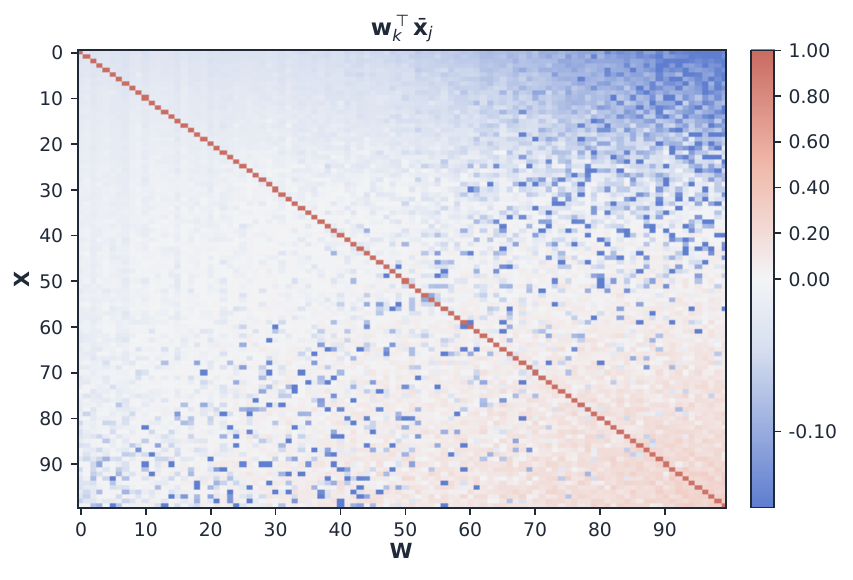}
		\includegraphics*[scale=0.36, viewport=5 0 440 275]{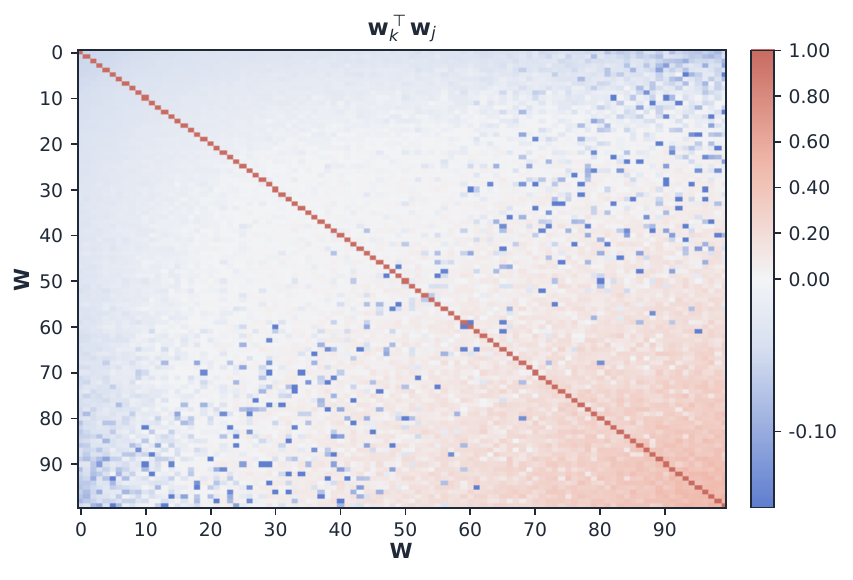}
	}
	\caption{Similarity matrices for class centers $\{\bar{\bm x}_k\}_{k=1}^K$ and classifier vectors $\{\bm w_k\}_{k=1}^K$ of ResNet18 trained with BCE on CIFAR100-LT with IF$=100$.}
	\label{fig_supp_matrix_ResNet18_cifar100_bce_100}
\end{figure*}

Figs. \ref{fig_supp_matrix_ResNet18_cifar10_bce_10}-\ref{fig_supp_matrix_ResNet18_cifar100_bce_100} present the similarity matrices of feature centers and classifier vectors for ResNet18 trained with CW averaged losses and G. averaged losses.
For models trained with CW averaged losses, the off-diagonal values in the similarity matrices are predominantly close to the NC value $-1/(K-1)$, consistent with the conclusion of the theorem. In contrast, for models trained with G. averaged losses, the between-class separation becomes obviously diverse, manifested as a gradual increase in the off-diagonal values of the matrix from the upper-left to the lower-right corner, exhibiting a clear correlation with the sample frequencies from the head classes to the tail classes. These results demonstrate that on imbalanced datasets, the CW. averaged losses maintain balanced between-class separability, whereas the G. averaged losses do not.



\clearpage

\section{Proofs}\label{theorem_proof_app}
We provide the proof of the theorems in our main paper, along with the necessary lemmas.

\subsection{Basics}
We first formally present the definition of canonical $K$-simplex ETF in $d$-dimension space, and two lemmas required by the proofs.

\begin{definition}  \label{simplex_ETF_definition}
    (K-simplex ETF) A K-simplex ETF is a collection of vectors in $\mathbb{R}^d$ specified by the columns of matrix
    \begin{equation}
        \bm{M}=\sqrt{\frac{K}{K-1}}\bm{P}\left(\bm{I}_{K}-\frac{1}{K}\bm{1}_{K}\bm{1}_{K}^{\top}\right),
    \end{equation}
    where $\bm{I}_{K} \in \mathbb{R}^{K\times K}$ is the identity matrix and $\bm{1}_{K} \in \mathbb{R}^K$ is the ones vector, and $\bm{P} \in \mathbb{R}^{d\times K} (d\geq K)$ is a partial-orthogonal matrix such that $\bm{P}^{\top}\bm{P}=\bm{I}_K$.

    Note that the matrix $\bm{M}$ satisfies:
    \begin{equation}
        \bm{M}^{\top}\bm{M}=\frac{K}{K-1}\left(\bm{I}_{K}-\frac{1}{K}\mathbf{1}_{K}\mathbf{1}_{K}^{\top}\right).
    \end{equation}
\end{definition}

\begin{lemma}   \label{young_inequality_definition}
    (Young's Inequality) Let $p, q$ be positive numbers satisfying $\frac{1}{p}+\frac{1}{q}=1$. Then for any $a, b \in \mathbb{R}$, we have
    \begin{equation}
        |ab|\leq\frac{|a|^p}{p}+\frac{|b|^q}{q},
    \end{equation}
    where the equality holds if and only if $|a|^p=|b|^q$. The case for $p=q=2$ is just the AM-GM inequality, $|ab|\leq\frac{1}{2}(a^2+b^2)$, and the equality holds if and only if $|a|=|b|$.
\end{lemma}

\begin{lemma}   \label{jensen_inequality_definition}
    (Jensen's Inequality) Let $f$ be a convex function on an interval $I$. Then
    \begin{equation}
        f(\lambda_1x_1+\cdots+\lambda_nx_n)\leq\lambda_1f(x_1)+\cdots+\lambda_nf(x_n),
    \end{equation}
    where
    \begin{equation}
        x_1,x_2,\cdots,x_n\in I,\quad \lambda_1,\lambda_2,\cdots,\lambda_n\in[0,1], \quad \sum_{i=1}^n \lambda_i = 1.
    \end{equation}
    When $f$ is strictly convex, let
    \begin{equation}
        I_1=\{k:\lambda_k\in(0,1]\}, ~\text{and}~ I_2=\{k:\lambda_k=0\}.
    \end{equation}
    The equality holds if and only if all $x_{k}, k \in I_1$ are equal. In practice, if the function $f$ likes $\exp(x)$, $\log(x)$, or $\log(1+\e^x)$, Jensen's Inequality can be written as
    \begin{align}
        \exp\left({\frac{\sum_{i=1}^{n}x_{i}}{n}}\right)&\leq{\frac{\sum_{i=1}^{n}\exp(x_{i})}{n}},\\
        \log\left({\frac{\sum_{i=1}^{n}x_{i}}{n}}\right)&\geq{\frac{\sum_{i=1}^{n}\log(x_{i})}{n}}, \\ 
        \log\left(1+\exp\left(\frac{\sum_{i=1}^{n}x_{i}}{n}\right)\right) &\leq \frac{1}{n}\sum_{i=1}^{n}\log\big(1+\exp(x_{i})\big).
    \end{align}
\end{lemma}

\subsection{Proof of Theorem \ref{theorem_NC_imbalanced_SC}}\label{proof:collapse_on_imbalanced_data}
Theorem 1 consists of two parts, namely, the class-wise averaged CE and BCE losses can lead to neural collapse (NC) on imbalanced datasets. We prove them in turn.

\begin{theorem}\label{them:BCE collapse_supp}
    (\textbf{BCE})
    Suppose that each class in dataset $\mathcal D$ contains at least one sample, and that the feature dimension is greater than the number of classes, i.e., $n_k = |\mathcal D_k| > 0$ for $\forall k \in [K]$ and $d \geq K-1$.
Then, for given scale factor $\gamma$ and class number $K$, any global minimizer of
    \begin{equation} \label{equ-minf_object}
        f_{\mathrm{bce}}(\bm W,\bm{b},\mathcal F)\triangleq \frac{1}{K}\sum_{k=1}^K\frac1{n_k}\sum_{\bm X^{(k)}\in \mathcal D_k} L_{\mathrm{bce}}(\bm X^{(k)})
    \end{equation}
    satisfies
    \begin{align}
        &\bm x^{(k)} = \bm w_k,\quad\forall\bm x^{(k)} \in \mathcal F_k,~\forall k\in[K], \label{eq_x=w_nc_imb_sc}\\
        &\langle\bm w_k, \bm w_j\rangle = - \frac1{K-1},\quad \forall k\neq j\in[K], \label{eq_w=w_etf_imb_sc}\\
        &b_k = b_j = b^\star, \quad\forall k,j\in[K], \label{eq_b=b*_imb_sc}
    \end{align}
    where $b^\star$ is the unique real solution of
    \begin{align}
        \e^{\frac{\gamma}{K-1}} \e^{2 b} - (K-2)\e^{b} - (K-1)\e^{\gamma} = 0.
    \end{align}
\end{theorem}
\begin{proof}

    According to Lemma \ref{lem:lemmaA.1}, for any $c_1, c_2 > 0$ and any $\bm X^{(k)} \in \mathcal D_k$ with $\forall k\in[K]$,
    \begin{align}
        {L}_{\mathrm{bce}}(\bm X^{(k)}) \geq \frac{1}{1+c_{1}}\Big(-\gamma \bm{w}_{k}^{\top}\bm x^{(k)}+b_{k}\Big) + \frac{1}{1+c_{2}}\sum_{j=1 \atop j\neq k}^{K}\Big(\gamma \bm{w}_{j}^{\top}\bm x^{(k)}-b_{j}\Big)+C, \nonumber
    \end{align}
    where
    \begin{equation}
        C=\log(1+c_1)-\frac{c_1}{1+c_1}\log(c_1) + (K-1)\Big(\log(1+c_2)-\frac{c_2}{1+c_2}\log(c_2)\Big), \nonumber
    \end{equation}
    and the inequality becomes an equality if and only if
    \begin{align}
        &c_1=\exp\Big(\gamma \bm{w}_k^\top\bm x^{(k)}-b_k\Big),  \quad c_2=\exp\Big(b_j-\gamma \bm{w}_j^\top\bm x^{(k)}\Big),\quad \forall j\neq k\in[K]. \nonumber
    \end{align}

    We take the same $c_1$ and $c_2$ for all $\bm X^{(k)}$ with $\forall k\in[K]$, then
    \begin{align}
        ~&f_{\mathrm{bce}}(\bm{W}, \bm b, \mathcal F) = \frac{1}{K}\sum_{k=1}^K \frac 1{n_k} \sum_{\bm X^{(k)}\in \mathcal D_k} {L}_\mathrm{bce}\big(\bm X^{(k)}\big) \nonumber\\
        ~
        \geq~&\frac{1}{K}\sum_{k=1}^K \frac1{n_k} \sum_{\bm x^{(k)}\in \mathcal F_k} \bigg[\frac{1}{1+c_1}\Big(-\gamma \bm{w}_k^\top \bm x^{(k)} + b_k\Big) + \frac{1}{1+c_2} \sum_{j=1\atop j\neq k}^K \Big(\gamma \bm{w}_j^\top \bm x^{(k)} - b_j\Big)\bigg] + C\\
        ~
        =~&\frac{1}{K}\frac{1}{1+c_2}\sum_{k=1}^K\frac1{n_k}\sum_{\bm x^{(k)}\in \mathcal F_k} \sum_{j=1 \atop j\neq k}^K \Big(\gamma \bm{w}_j^\top \bm x^{(k)} - b_j\Big) \nonumber\\
        &\quad - \frac1{K}\frac1{1+c_1}\sum_{k=1}^K\frac1{n_k}\sum_{\bm x^{(k)}\in \mathcal F_k} \Big(\gamma \bm{w}_k^\top \bm x^{(k)} - b_k\Big)  + C \nonumber\\
        ~
        =~&\frac1{K}\frac1{1+c_2}\sum_{k=1}^K\frac1{n_k}\sum_{\bm x^{(k)}\in \mathcal F_k}\sum_{j=1}^K\left(\gamma \bm{w}_j^\top \bm x^{(k)}-b_j - \gamma \bm{w}_k^\top \bm x^{(k)} + b_k\right) + C \nonumber \\
        &~+\frac1{K}\frac1{1+c_2}\sum_{k=1}^K \frac1{n_k} \sum_{\bm x^{(k)}\in \mathcal F_k} \sum_{j=1 \atop j\neq k}^K \left(\gamma \bm{w}_k^\top \bm x^{(k)} - b_k\right) \nonumber\\
        &\quad - \frac1{K}\frac1{1+c_1}\sum_{k=1}^K\frac1{n_k} \sum_{\bm x^{(k)}\in \mathcal F_k} \big(\gamma \bm{w}_k^\top \bm x^{(k)} - b_k\big)\nonumber\\
        ~
        =~&\frac1{K}\frac1{1+c_2}\Bigg[\sum_{k=1}^K \frac1{n_k} \sum_{\bm x^{(k)} \in \mathcal F_k} \sum_{j=1}^K\Big(\gamma \bm{w}_j^\top \bm x^{(k)} - \gamma \bm{w}_k^\top \bm x^{(k)} \Big) - \underbrace{\sum_{k=1}^K \frac1{n_k} \sum_{\bm x^{(k)} \in \mathcal F_k} \sum_{j=1}^K (b_j-b_k)}_0\Bigg] \nonumber \\
        &~+\frac1{K}\bigg(\frac{K-1}{1+c_2}-\frac1{1+c_1}\bigg)\sum_{k=1}^K\frac1{n_k}\sum_{\bm x^{(k)}\in\mathcal F_k} \gamma \bm{w}_k^\top \bm x^{(k)} \nonumber\\
        &\quad - \frac1{K}\bigg(\frac{K-1}{1+c_2}-\frac1{1+c_1}\bigg)\sum_{k=1}^{K}b_k  + C\nonumber\\
        ~
        =~&\frac1{1+c_2}\sum_{k=1}^K\gamma \bm{w}_k^\top\big(\tilde{\bm x} - \bar{\bm x}_k\big) - \frac1{K} \bigg(\frac{K-1}{1+c_2} -\frac1{1+c_1}\bigg) \sum_{k=1}^{K}b_k \nonumber\\
        &\quad + \frac1{K}\bigg(\frac{K-1}{1+c_2}-\frac1{1+c_1}\bigg) \sum_{k=1}^K\gamma \bm{w}_k^\top\bar{\bm{x}}_k + C \nonumber\\
        ~
        \triangleq~&\hat{f}_{\text{bce}}(\bm{W},\bm b,\mathcal F),
    \end{align}
    where $\tilde{\bm x} = \frac{1}{K}\sum_{j=1}^{K}\bar{\bm{x}}_j$ and $\bar{\bm{x}}_j = \frac1{n_j}\sum_{\bm x^{(j)}\in\mathcal F_j} \bm x^{(j)},~\forall j\in[K]$.

    For the function $\hat{f}_{\text{bce}}(\bm{W},\bm b,\mathcal F)$, its global minima are located at its stationary points in terms of $b_k$ for $\forall k\in[K]$, then
    \begin{equation}
        0 = \frac{\partial \hat{f}_{\mathrm{bce}}}{\partial b_k} = -\frac1{K}\bigg(\frac{K-1}{1+c_2}-\frac1{1+c_1}\bigg). \label{b_condition}
    \end{equation}
    Therefore,
    \begin{align}
        \hat{f}_{\mathrm{bce}}(\bm{W},\bm b,\mathcal F) \geq \frac{\gamma}{1+c_2}\sum_{k=1}^K \bm{w}_k^\top\big(\tilde{\bm{x}} - \bar{\bm x}_k\big) + C. \nonumber
    \end{align}
    Furthermore, according to the AM-GM inequality (see Lemma \ref{young_inequality_definition}), we have
    \begin{equation}
        \bm{u}^\top\bm{v}\geq - \Big(\frac{c}{2}\|\bm{u}\|_2^2+\frac{1}{2c}\|\bm{v}\|_2^2\Big), \quad\forall\bm{u}, \bm{v}\in\mathbb{R}^d, ~\forall c > 0, \nonumber
    \end{equation}
    which becomes an equality when $\bm{v} = -c\bm{u}$. Then,
    \begin{align}
        & f_{\mathrm{bce}}(\bm{W},\bm b,\mathcal F) \geq \hat{f}_{\mathrm{bce}}(\bm{W},\bm b,\mathcal F) \geq \frac{\gamma}{1+c_2}\sum_{k=1}^K \bm{w}_k^\top\big(\tilde{\bm{x}} - \bar{\bm{x}}_k\big) + C \nonumber \\
        \geq& -\frac{\gamma}{1+c_2}\bigg(\frac{c_3}2\sum_{k=1}^K\|\bm{w}_k\|_2^2+\frac1{2c_3}\sum_{k=1}^K\|\tilde{\bm{x}}-\bar{\bm{x}}_k\|_2^2\bigg) + C \label{c3_ineq} \\
        =& -\frac{\gamma}{1+c_2}\bigg[\frac{c_3}2\sum_{k=1}^K\|\bm{w}_k\|_2^2 + \frac1{2c_3}\bigg(\sum_{k=1}^K\|\bar{\bm{x}}_k\|_2^2 -K\|\tilde{\bm{x}}\|_2^2\bigg)\bigg] + C \nonumber\\
        \geq& -\frac{\gamma}{1+c_2}\bigg(\frac{c_3}{2} K +\frac{1}{2c_3}K\bigg) +\frac{\gamma}{1+c_2}\frac{1}{2c_3}K\|\tilde{\bm{x}}\|_2^2 + C\label{h_mean=h}\\
        \geq& -\frac{\gamma}{1+c_2}\bigg(\frac{c_3}{2} K +\frac{1}{2c_3}K\bigg) + C, \label{h_mean=0}
    \end{align}
    where inequality (\ref{c3_ineq}) becomes an equality if and only if
    \begin{equation}
        c_{3} \bm{w}_{k} = \bar{\bm{x}}_k - \tilde{\bm{x}}, \quad \forall k\in[K], ~\forall c_3 > 0, \label{equ-c3_hold}
    \end{equation}
    inequality (\ref{h_mean=h}) becomes an equality if and only if
    \begin{align}
        & 1 = \|\bar{\bm x}_k\| = \bigg\|\frac1{n_k}\sum_{\bm x^{(k)}\in\mathcal F_k} \bm x^{(k)}\bigg\| \leq \frac1{n_k}\sum_{\bm x^{(k)}\in\mathcal F_k} \big\|\bm x^{(k)}\big\| = 1 \nonumber \\
        &\quad\Leftrightarrow\quad \bm x^{(k)} = \bar{\bm{x}}_k,~\forall \bm x^{(k)} \in\mathcal F_k,\quad \forall k\in[K], \label{equ-h_mean=h}
    \end{align}
    and, the inequality (\ref{h_mean=0}) becomes an equality if and only if
    \begin{equation}
        \tilde{\bm{x}} = \bm 0, \label{equ-hmean=0}
    \end{equation}
    implying
    \begin{equation}
		\begin{aligned}
		 & c_3 \bm{w}_k = \bar{\bm{x}}_k \quad {\Longrightarrow} \quad \|c_3 \bm{w}_k\|_2^2 = \|\bar{\bm{x}}_k\|_2^2 \quad \overset{\ref{equ-h_mean=h}}{\Longrightarrow}\quad c_3^2=1 \nonumber \\
		 &\quad \Longrightarrow\quad c_3=1~~\text{and}~~\bm w_k = \bm x^{(k)},\quad \forall \bm x^{(k)} \in \mathcal F_k,~\forall k\in[K].
		\end{aligned} \label{equ-c3_solution}
\end{equation}
    Combining Eqs. (\ref{h_mean=0}) and (\ref{equ-c3_solution}), we get
    \begin{align}
        f_{\mathrm{bce}}(\bm{W}, \bm b, \mathcal F) \geq& -\frac{\gamma}{1+c_2}\bigg(\frac{c_3}{2} K +\frac{1}{2c_3}K\bigg) + C = -\frac{\gamma K}{1+c_2} + C, \label{f_bce_low}
    \end{align}
    which becomes an equality if and only if \textit{Eqs.} (\ref{equ-c1_1}), (\ref{equ-c2_1}), (\ref{equ-c3_hold}), (\ref{equ-h_mean=h}), and (\ref{equ-hmean=0}) hold.

    When \textit{Eqs.} (\ref{equ-c3_hold}), (\ref{equ-h_mean=h}), and (\ref{equ-hmean=0}) hold, one can deduce that
    \begin{align}
        c_1 &=\exp\Big(\gamma \bm{w}_k^\top\bm{x}^{(k)}-b_k\Big) = \exp\Big(\gamma\| \bm{w}_k\|_2^2-b_k\Big), \quad\forall k \in [K], \label{equ-c1_formula_simple} \\
        c_2 &=\exp\Big(b_j-\gamma \bm{w}_j^\top\bm{x}^{(k)}\Big) = \exp\Big(b_j-\gamma \bm{w}_k^\top\bm{w}_j\Big), \quad \forall j\neq k \in [K]. \label{equ-c2_formula_simple}
    \end{align}
    Since the $c_1, c_2 > 0$ are the same for different $j \neq k \in [K]$, we have
    \begin{align}
        &\gamma\|\bm{w}_k\|_2^2-b_k=\gamma\|\bm{w}_j\|_2^2-b_j, \quad \forall k,j\in[K], \label{equ-c1_same_for_all} \\
        &\gamma \bm{w}_k^\top\bm{w}_j-b_j=\gamma \bm{w}_k^\top\bm{w}_\ell-b_\ell, \quad \forall j\neq\ell\in[K], ~\forall k\in[K]. \label{equ-c2_same_for_all}
    \end{align}
    In the normalized feature space, \textit{Eq.} (\ref{equ-c1_same_for_all}) implies
    \begin{align}
        \gamma\|\bm{w}_k\|_2^2-b_k &=\gamma\|\bm{w}_j\|_2^2-b_j \quad\Rightarrow\quad \gamma-b_k=\gamma-b_j \quad\Leftrightarrow\quad b_k=b_j, \quad \forall k, j \in [K]. \label{equ-bk=bj}
    \end{align}
    Therefore, at the minimum points of $f_{\mathrm{bce}}$, one can write $\bm{b}=b^\star\mathbf{1}$ for some $b^\star \in \mathbb R$.
    Then, from Eq. (\ref{equ-c2_same_for_all}),
    \begin{align}
         \gamma \bm{w}_k^\top\bm{w}_j=\gamma \bm{w}_k^\top\bm{w}_\ell \quad\Rightarrow\quad \bm{w}_k^\top\bm{w}_j=\bm{w}_k^\top\bm{w}_\ell, \quad\forall j\neq\ell\in[K], \quad \forall k\in[K],
    \end{align}
    and from \textit{Eqs.} (\ref{equ-c3_hold}), (\ref{equ-h_mean=h}), and (\ref{equ-hmean=0}) hold, one deduces that $\sum_{k=1}^K\bm w_k = \bm 0$ implying
    \begin{align}
        & 1 = \|\bm{w}_{k}\|_{2}^{2}=-\sum_{j=1 \atop j \neq k}^{K}\bm{w}_{k}^\top\bm{w}_{j} = -\left(K-1\right)\bm{w}_{k}^\top\bm{w}_{j} \nonumber \\
        &\Rightarrow \quad \bm{w}_{k}^\top\bm{w}_{j} = -\frac{1}{K-1}, \quad \forall j\neq k\in[K]. \label{equ-wkwj}
    \end{align}
    Plugging \textit{Eqs.} (\ref{equ-bk=bj}), (\ref{equ-wkwj}) into \textit{Eqs.} (\ref{equ-c1_formula_simple}), (\ref{equ-c2_formula_simple}),
    \begin{align}
        c_1 &=\exp(\gamma-b^\star)\quad\text{and}\quad c_2 =\exp\bigg(b^\star+\frac{\gamma}{K-1}\bigg). \label{equ-c1c2_formula_bce}
    \end{align}

    When $f_\mathrm{bce}$ defined in \textit{Eq.} (\ref{equ-minf_object}) achieves its minimum value, it theoretically satisfies
    \begin{align}
        0 = \frac{\partial f_{\mathrm{bce}}}{\partial b_{k}} = ~& \frac{1}{K}\sum_{k=1}^K\frac1{n_k}\sum_{\bm x^{(k)}\in\mathcal F_k}\bigg(1-\frac1{1+\exp\big(-\gamma \bm w_k^\top \bm x^{(k)} + b_k \big)}\bigg) \nonumber\\
        ~& -\frac1K\sum_{j=1}^K\frac1{n_j}\sum_{\bm x^{(j)}\in\mathcal F_j} \sum_{k=1\atop k\neq j}^K\bigg(1-\frac1{1+\exp\big(\gamma\bm w_k^\top\bm x^{(j)} - b_k \big)}\bigg),\quad\forall k\in[K],
    \end{align}
    thus, combining \textit{Eqs.} (\ref{equ-h_mean=h}) and (\ref{equ-wkwj}),
    \begin{align}
        &0 = 1-\frac1{1+\exp\big(-\gamma + b_k \big)} - (K-1)\bigg(1-\frac1{1+\exp\big(-\gamma\frac1{K-1} - b_k \big)}\bigg)\nonumber\\
        \Leftrightarrow\qquad&\frac1{1+\exp\big(\gamma - b_k \big)} = \frac{K-1}{1+\exp\big(\frac{\gamma}{K-1} + b_k \big)}\nonumber\\
        \Leftrightarrow\qquad&\e^{\frac{\gamma}{K-1}} \e^{2b_k} - (K-2)\e^{b_k} - (K-1)\e^{\gamma} = 0, \quad \forall k\in[K], \label{eq_b*_bce}
    \end{align}
    which satisfies Eq. (\ref{b_condition}).

    In total,
    \begin{align}
        f_{\mathrm{bce}}(\bm W,\bm{b},\mathcal F)
        \geq~& -\frac{\gamma K}{1+c_2} -\frac{\gamma K}{1+c_2} + \log(1+c_1) + C \nonumber \\
        ~& -\frac{c_1}{1+c_1}\log(c_1) + (K-1)\bigg(\log(1+c_2) -\frac{c_2}{1+c_2}\log(c_2)\bigg) \nonumber\\
        \overset{\ref{equ-c1c2_formula_bce},\ref{b_condition}}{=}~& \log\big(1+\e^{b^{\star}-\gamma}\big) + (K  -1)\log\Big(1+\e^{-b^{\star}-\frac{\gamma}{K-1}}\Big),
    \end{align}
    and $f_{\mathrm{bce}}$ achieves this constant lower boundary if and only if \textit{Eqs.} (\ref{equ-c1_1}), (\ref{equ-c2_1}), (\ref{equ-c3_hold}), (\ref{equ-h_mean=h}), (\ref{equ-hmean=0}), and (\ref{eq_b*_bce}) hold.
    It is easy deduce \textit{Eqs.} (\ref{eq_x=w_nc_imb_sc}), (\ref{eq_w=w_etf_imb_sc}), and (\ref{eq_b=b*_imb_sc}) from these conditions, and vice versa, which completes the proof.
\end{proof}

\begin{lemma} \label{lem:lemmaA.1}
    For any $\bm X^{(k)}\in\mathcal D_k, \forall k\in[K]$, with any $c_1, c_2 > 0$ and $\gamma > 0$, the BCE loss $L_{\mathrm{bce}}(\bm X^{(k)})$ defined in Eq. (\ref{equ-bce-single-sample}) is lower bounded,
    \begin{align} \label{equ-Lbce_lowerBound}
        {L}_{\mathrm{bce}}(\bm X^{(k)}) \geq \frac{1}{1+c_{1}}\Big(-\gamma \bm{w}_{k}^{\top}\bm x^{(k)}+b_{k}\Big) + \frac{1}{1+c_{2}}\sum_{j=1 \atop j\neq k}^{K}\Big(\gamma \bm{w}_{j}^{\top}\bm x^{(k)}-b_{j}\Big)+C,
    \end{align}
    where
    \begin{equation} \label{equ-C_formula_2}
        C = \log(1+c_1)-\frac{c_1}{1+c_1}\log(c_1) + (K-1)\Big(\log(1+c_2)-\frac{c_2}{1+c_2}\log(c_2)\Big).
    \end{equation}
    The inequality becomes an equality if and only if
    \begin{align}
        &c_1=\exp\Big(\gamma \bm{w}_k^\top\bm x^{(k)}-b_k\Big), \label{equ-c1_1}\\
        & c_2=\exp\Big(b_j-\gamma \bm{w}_j^\top\bm x^{(k)}\Big),\quad \forall j\neq k\in[K]. \label{equ-c2_1}
    \end{align}
\end{lemma}

\begin{proof}
    By the concavities of $\log(x)$, and according to Lemma \ref{jensen_inequality_definition}, the BCE loss $L_{\mathrm{bce}}(\bm X^{(k)})$ for any $\bm X^{(k)}\in\mathcal D$ can be lower bounded by
    \begin{align}
        {L}_{\mathrm{bce}}(\bm X^{(k)})
        =~ &\log\Big(1+\exp\big(-\gamma \bm{w}_k^\top\bm x^{(k)}+b_k\big)\Big) + \sum_{j=1 \atop j\neq k}^K\log\Big(1  +\exp\big(\gamma \bm{w}_j^\top\bm x^{(k)}-b_j\big)\Big) \nonumber \\
        =~ &\log\bigg(\frac{c_1}{1+c_1}\frac{1+c_1}{c_1}+\frac{1+c_1}{1+c_1}\exp\big(-\gamma \bm{w}_k^\top\bm x^{(k)}+b_k\big)\bigg) \nonumber \\
        &~ + \sum_{j=1 \atop j\neq k}^K \log\bigg(\frac{c_2}{1+c_2}\frac{1+c_2}{c_2} +\frac{1+c_2}{1+c_2}\exp\big(\gamma \bm{w}_j^\top\bm x^{(k)}-b_j\big)\bigg) \nonumber\\
        \geq~ & \frac{c_1}{1+c_1}\log\bigg(\frac{1+c_1}{c_1}\bigg) + \frac{1}{1+c_1}\log\Big[(1+c_1)\exp\big(-\gamma \bm{w}_k^\top \bm x^{(k)} +b_k\big)\Big] \nonumber \\
        &~ + (K-1)\frac{c_2}{1+c_2}\log\bigg(\frac{1+c_2}{c_2}\bigg) \nonumber \\
        &~ +\frac{1}{1+c_2}\sum_{j=1 \atop j\neq k}^K\log\Big[(1  +c_2)\exp\big(\gamma \bm{w}_j^\top\bm x^{(k)}-b_j\big)\Big] \nonumber \\
        =~ &\frac{1}{1+c_1}\bigg(-\gamma \bm{w}_k^\top\bm x^{(k)}+b_k\bigg) + \frac1{1+c_2}\sum_{j=1 \atop j \neq k}^K\bigg(\gamma \bm{w}_j^\top \bm x^{(k)} - b_j\bigg) \nonumber \\
        &~ +\underbrace{\log(1+c_1)-\frac{c_1}{1+c_1}\log(c_1) + (K-1)\Big(\log(1+c_2) - \frac{c_2}{1+c_2}\log(c_2)\Big)}_{C}. 
        \label{use_C}
    \end{align}
    The inequality becomes an equality if and only if
    \begin{align}
        \frac{1+c_1}{c_1} &=\left(1+c_1\right)\exp\left(-\gamma \bm{w}_k^\top \bm x^{(k)} + b_k\right) ~~\text{or}~~ c_1=0 ~~\text{or}~~ c_1=+\infty, ~\text{and} \nonumber\\
        \frac{1+c_2}{c_2} &=\left(1+c_2\right)\exp\Big(\gamma \bm{w}_j^\top \bm x^{(k)}-b_j\Big) ~~\text{or}~~ c_2=0 ~~\text{or}~~ c_2=+\infty,\quad \forall j\neq k \in[K].\nonumber
    \end{align}
    These equalities are trivial when \(c_1 = 0\) or \(c_1 = +\infty\) and \(c_2 = 0\) or \(c_2 = +\infty\).
    Therefore, if and only if
    \begin{align}
        c_1 &=\exp\left(\gamma \bm{w}_k^\top\bm x^{(k)}-b_k\right), \quad\text{and}\nonumber\\
        c_2 &=\exp\left(b_j-\gamma \bm{w}_j^\top\bm x^{(k)}\right), \quad\forall j\neq k\in [K],\nonumber
    \end{align}
    Eq. (\ref{equ-Lbce_lowerBound}) becomes an equality, which completes the proof.
\end{proof}

\clearpage
\begin{theorem} \label{them:lemmaB.1}(\textbf{CE})
    Suppose that each class in dataset $\mathcal D$ contains at least one sample, and that the feature dimension is greater than the number of classes, i.e., $n_k = |\mathcal D_k| > 0$ for $\forall k \in [K]$ and $d \geq K-1$.
Then, at any minimum point of loss
\begin{align}
    f_{\mathrm{ce}}(\bm W, \bm b, \mathcal F) \triangleq
    \frac1K\sum_{k=1}^K\frac1{n_k}\sum_{\bm X^{(k)}}L_{\mathrm{ce}}(\bm X^{(k)}) \label{equ-minf_object-ce}
\end{align}
satisfies
    \begin{align}
        &\bm x^{(k)} = \bm w_k,\quad\forall\bm x^{(k)} \in \mathcal F_k,~\forall k\in[K], \label{eq_x=w_nc_imb_sc_ce}\\
        &\langle\bm w_k, \bm w_j\rangle = - \frac1{K-1},\quad \forall k\neq j\in[K], \label{eq_w=w_etf_imb_sc_ce}\\
        &b_k = b_j = b^\star,\quad\forall k,j\in[K], \label{eq_b=b*_imb_sc_ce}
    \end{align}
while $b^\star\in\mathbb R$ could be arbitrary number.
\end{theorem}
\begin{proof}
    According to Lemma \ref{lem:lemmaB.1}, for any $c_1 > 0$ and any $\bm{X}^{(k)}\in\mathcal D_k$ with $\forall k\in[K]$,
    \begin{align}
        {L}_{\mathrm{ce}}(\bm X^{(k)}) \geq \frac{\sum_{j=1}^K\big(\gamma\bm{w}_j^{\top}\bm{x}^{(k)}-b_j\big) - K(\gamma\bm{w}_k^{\top}\bm{x}^{(k)}-b_k)}{\big(1+c_1\big)(K-1)}+C,\nonumber
    \end{align}
    where
    \begin{align}
        C =\frac{c_1}{1+c_1}\log\bigg(\frac{1+c_1}{c_1}\bigg) + \frac{1}{1+c_1}\log\Big[\big(1+c_1\big)(K-1)\Big], \nonumber
    \end{align}
    and inequality becomes an equality if and only if
    \begin{align}
        &c_1=\Bigg[(K-1)\exp\Bigg(\frac{\sum_{j=1}^K \big(\gamma\bm{w}_j^{\top}\bm{x}^{(k)}-b_j\big) -K\big(\gamma\bm{w}_k^{\top}\bm{x}^{(k)}-b_k\big)}{K-1}\Bigg)\Bigg]^{-1} \qquad{\text{and}} \nonumber\\
        &\gamma\bm{w}_j^{\top}\bm{x}^{(k)}-b_j = \gamma\bm{w}_\ell^{\top}\bm{x}^{(k)}-b_\ell,
        \quad \forall j, \ell \neq k \in [K].\nonumber
    \end{align}

    We take the same $c_1$ for all $\bm{X}^{(k)}\in\mathcal D$, then one can deduce the following lower bound for the function $f_{\mathrm{ce}}$
    \begin{align}
        & (1+c_1)(K-1)\big(f_{\mathrm{ce}}(\bm{W},\bm{b}, \mathcal F) - C \big) \nonumber \\
        = &~(1+c_1)(K  -1)\bigg[\frac{1}{K}\sum_{k=1}^K\frac1{n_k}\sum_{\bm X^{(k)}\in\mathcal D_k}{L}_\mathrm{ce}(\bm X^{(k)}) - C \bigg] \nonumber \\
        \geq &~\frac1{K}\sum_{k=1}^K\frac1{n_k}\sum_{\bm x^{(k)}\in\mathcal F_k}\bigg[\sum_{j=1}^K\big(\gamma \bm{w}_j^\top \bm{x}^{(k)}-b_j\big) -K\big(\gamma \bm{w}_k^\top\bm{x}^{(k)}-b_k\big)\bigg] \nonumber\\
        = &~\frac{1}{K}\sum_{k=1}^K\frac1{n_k}\sum_{\bm x^{(k)}\in\mathcal F_k}\sum_{j=1}^K \gamma \bm w_j^\top \bm x^{(k)} - \sum_{k=1}^K\frac1{n_k}\sum_{\bm x^{(k)}\in\mathcal F_k} \gamma\bm w_k^\top \bm x^{(k)}
        \underbrace{-\frac1K\sum_{k=1}^{K}\sum_{j=1}^{K}b_{j} +\sum_{k=1}^K b_{k}}_{0} \nonumber\\
        = &~\frac{1}{K}\sum_{k=1}^K\sum_{j=1}^K \gamma \bm w_j^\top \bar{\bm x}_k - \sum_{k=1}^K \gamma\bm w_k^\top \bar{\bm x}_k = \gamma \sum_{k=1}^K\bm w_k^\top \big(\tilde{\bm x} - \bar{\bm x}_k\big), \label{equ-fce_medal_lower_bound}
    \end{align}
    where $\tilde{\bm{x}} = \frac{1}{K}\sum_{j=1}^{K}\bar{\bm{x}}_{j}$ and $\bar{\bm{x}}_{j} = \frac1{n_j}\sum_{\bm x^{(j)}\in\mathcal F_j} \bm x^{(j)},~\forall j\in[K]$.

    According to the AM-GM inequality in Lemma \ref{young_inequality_definition},
    \begin{equation}
        \bm{u}^\top\bm{v}\leq\frac{c}{2}\|\bm{u}\|_2^2+\frac{1}{2c}\|\bm{v}\|_2^2, ~\forall\bm{u},\bm{v}\in\mathbb{R}^d, ~\forall c > 0, \nonumber
    \end{equation}
    which becomes an equality when $\bm{v}=c\bm{u}$. Then, based on \textit{Eq.} (\ref{equ-fce_medal_lower_bound}),
    \begin{align}
(1+c_1)(K-1)\big[f_{\mathrm{ce}}(\bm{W},\bm{b},\mathcal F)-C\big]
        \geq &~\gamma \sum_{k=1}^K\bm w_k \big(\tilde{\bm x} - \bar{\bm x}_k\big) \nonumber\\
        \geq &~-\gamma \frac{c_2}2\sum_{k=1}^K\|\bm{w}_k\|_2^2 -\frac\gamma{2c_2}\sum_{k=1}^K\|\tilde{\bm{x}}-\bar{\bm{x}}_k\|_2^2 \nonumber\\
        =&~-K\gamma \frac{c_2}2 - \frac\gamma{2c_2}\bigg(\sum_{k=1}^K\|\bar{\bm x}_k\|^2 - K\|\tilde{\bm x}\|^2\bigg), \nonumber
    \end{align}
    where the second inequality becomes an equality if and only if
    \begin{align}
        c_2 \bm{w}_{k} =  - \tilde{\bm{x}} + \bar{\bm{x}}_{k}, \quad\forall k\in[K], ~\forall c_2 > 0. \label{equ-fce_c3_hold}
    \end{align}
    Therefore,
    \begin{align}
        f_{\mathrm{ce}}(\bm{W}, \bm{b}, \mathcal F)
        \geq &~-\frac{c_2\gamma K}{2(1+c_1)(K-1)} - \frac{\gamma}{2c_2(1+c_1)(K-1)}\bigg(\sum_{k=1}^K\|\bar{\bm x}_k\|^2 - K\|\tilde{\bm x}\|^2\bigg) + C \nonumber\\
        ~
        \geq &~-\frac{c_2\gamma K}{2(1+c_1)(K-1)} - \frac{\gamma}{2c_2(1+c_1)(K-1)} \sum_{k=1}^K \|\bar{\bm x}_k\|^2 + C \nonumber\\
        ~
        = &~-\frac{c_2\gamma K}{2(1+c_1)(K-1)} - \frac{\gamma}{2c_2(1+c_1)(K-1)} \sum_{k=1}^K \bigg\|\frac1{n_k}\sum_{\bm x^{(k)}\in \mathcal F_k}\bm x^{(k)}\bigg\|^2 + C \nonumber\\
        ~
        \geq &~-\frac{c_2\gamma K}{2(1+c_1)(K-1)} - \frac{\gamma}{2c_2(1+c_1)(K-1)} \sum_{k=1}^K\frac1{n_k}\sum_{\bm x^{(k)}\in \mathcal F_k}\big\|\bm x^{(k)}\big\|^2 + C \nonumber\\
        ~
        = &~-\frac{c_2\gamma K}{2(1+c_1)(K-1)} - \frac{K\gamma}{2c_2(1+c_1)(K-1)} + C,
        \label{equ-gce_final_equation}
    \end{align}
    where the second inequality becomes an equality if and only if
    \begin{equation}
        \tilde{\bm{x}}=\frac{1}{K}\sum_{j=1}^{K}\bar{\bm{x}}_{j}=\bm{0}, \label{equ-fce_h_mean=0}
    \end{equation}
    and the third inequality becomes an equality if and only if
    \begin{align}
        \bm x^{(k)} = \bm x_*^{(k)}, \quad \forall \bm x^{(k)}, \bm x_*^{(k)} \in \mathcal F_k,~\forall k\in [K]. \label{equ_x=x*}
    \end{align}
    From \textit{Eqs.} (\ref{equ-fce_c3_hold}), (\ref{equ-fce_h_mean=0}), and (\ref{equ_x=x*}), we get
    \begin{equation}
\begin{aligned}
 & c_2 \bm{w}_k = \bar{\bm{x}}_k \quad\Rightarrow\quad \|c_2 \bm{w}_k\|_2^2 = \|\bar{\bm{x}}_k\|_2^2 \quad\Rightarrow\quad (c_2)^2=1 \quad\Rightarrow\quad c_2=1, ~~\text{and}~~ \nonumber \\
 & \bm w_k = \bm x^{(k)},~\forall \bm x^{(k)}\in\mathcal F_k,~\forall k\in[K].
\end{aligned} \label{equ-fce_c3_solution}
\end{equation}
    Plugging \textit{Eq.} (\ref{equ-fce_c3_solution}) into (\ref{equ-gce_final_equation}), we have
    \begin{align}
        f_{\mathrm{ce}}(\bm W,\bm{b},\mathcal F) \geq -\frac{\gamma K}{2(1+c_1)(K-1)} - \frac{K\gamma}{2(1+c_1)(K-1)} + C =-\frac{\gamma K}{(1+c_1)(K-1)}+C. \nonumber
    \end{align}

    Furthermore, \textit{Eqs.} (\ref{equ-fce_c3_hold}), (\ref{equ-fce_h_mean=0}), and (\ref{equ_x=x*}) imply
    \begin{equation}
        \sum_{k=1}^{K}\bm{w}_{k}=\bm{0}.\label{equ-fce_sumw=sumh=0}
    \end{equation}
    Then,
    \begin{align}
        \sum_{j=1}^{K}\big(\gamma \bm{w}_j^\top\bm{x}^{(k)}-b_j\big) &=\gamma\bigg(\sum_{j=1}^{K}\bm{w}_j^\top\bigg)\bm{x}^{(k)}-\sum_{j=1}^{K}b_j =0-\sum_{j=1}^{K}b_j=-K\bar{b}, \label{equ-sum_zji}
    \end{align}
    where $\bar{b}=\frac{1}{K}\sum_{j=1}^{K}b_j$, and
    \begin{equation}\label{equ-k_zki}
        K\big(\gamma \bm{w}_k^\top\bm{x}^{(k)}-b_k\big) = K\big(\gamma\|\bm{w}_k\|_2^2-b_k\big) = K \gamma - K b_k.
    \end{equation}
    Combining \textit{Eqs.} (\ref{equ-jensen_hold_condition_a}), (\ref{equ-jensen_hold_condition_b}), (\ref{equ-sum_zji}), and (\ref{equ-k_zki}), we have
    \begin{align}
        c_1 &=\Bigg[(K-1)\exp\bigg(\frac{\sum_{j=1}^K(\gamma \bm{w}_j^\top\bm{x}^{(k)}-b_j)-K(\gamma \bm{w}_k^\top\bm{x}^{(k)}-b_k)}{K-1}\bigg)\Bigg]^{-1} \nonumber \\
        &=\bigg[(K-1)\exp\bigg(\frac{-K\bar{b}-\big(K\gamma-Kb_k\big)}{K-1}\bigg)\bigg]^{-1} =\bigg[(K-1)\exp\bigg(\frac{K}{K-1}\big(b_k-\bar{b}-\gamma\big)\bigg)\bigg]^{-1}. \label{equ-fce_c1_medal_solution}
    \end{align}

    Since the $c_1 > 0$ can be arbitrary number, we choose the same $c_1$ for any $k,j\in[K]$.
    Then, we have
    \begin{align}
        b_k-\bar{b}-\gamma &=b_j-\bar{b}-\gamma \quad\Rightarrow\quad b_k=b_j, ~\forall j \neq k \in [K]. \label{equ-fce_bj=bk}
    \end{align}
    Therefore, we can write $\bm{b}=b^\star\mathbf{1}_K$.
    Plugging the results in \textit{Eq.} (\ref{equ-fce_bj=bk}) into (\ref{equ-fce_c1_medal_solution}), we have
    \begin{align}
       c_1 &=\bigg[(K-1)\exp\bigg(\frac{K}{K-1}\big(b_k-\bar{b}-\gamma\big)\bigg)\bigg]^{-1} =\bigg[(K-1)\exp\bigg(-\frac{\gamma K}{K-1}\bigg)\bigg]^{-1}. \nonumber
    \end{align}

    Moreover, plugging \textit{Eqs.} (\ref{equ-fce_c3_solution}) and (\ref{equ-fce_bj=bk}) into \textit{Eq.} (\ref{equ-k_zki}),
    \begin{align}
        \gamma\bm{w}_j^{\top}\bm{x}^{(k)}-b_j=\gamma\bm{w}_\ell^{\top}\bm{x}^{(k)}-b_\ell &\quad \Rightarrow\quad \bm{w}_j^\top\bm{x}^{(k)}=\bm{w}_\ell^\top\bm{x}^{(k)} \quad\Rightarrow\quad \bm{w}_j^\top\bm{w}_k=\bm{w}_\ell^\top\bm{w}_k. \nonumber
    \end{align}
    Combining with \textit{Eq.} (\ref{equ-fce_sumw=sumh=0}),
    \begin{align}
        1 = \|\bm{w}_{k}\|_{2}^{2}&=-\sum_{j=1 \atop j \neq k}^{K}\bm{w}_{k}^{\top}\bm{w}_{j}=-\left(K-1\right)\bm{w}_{k}^{\top}\bm{w}_{j} ~\Rightarrow~ \bm{w}_{k}^{\top}\bm{w}_{j}=-\frac{1}{K-1}, ~\forall j\neq k\in[K]. 
        \label{equ-fce_wkwj_solution}
    \end{align}

    In total,
    \begin{align}
        f_{\mathrm{ce}}(\bm W,\bm{b},\mathcal F) \geq-\frac{\gamma K}{(1+c_1)(K-1}+C =\log\Big(1+(K-1)\mathrm{e}^{-\frac{\gamma K}{K-1}}\Big), \nonumber
    \end{align}
    where this constant lower boundary is achieved if and only if \textit{Eqs.} (\ref{equ-jensen_hold_condition_a}), (\ref{equ-jensen_hold_condition_b}), (\ref{equ-fce_c3_hold}), (\ref{equ-fce_h_mean=0}), and (\ref{equ_x=x*}) hold.
    One can deduce \textit{Eqs.} (\ref{eq_x=w_nc_imb_sc_ce}), (\ref{eq_w=w_etf_imb_sc_ce}), and (\ref{eq_b=b*_imb_sc_ce}) from these conditions, and vice versa, which completes the proof.
\end{proof}

\begin{lemma} \label{lem:lemmaB.1}
    For any $\bm X^{(k)} \in \mathcal D$ with any $c_1 > 0$ and $\gamma > 0$, the CE loss $L_{\mathrm{ce}}(\bm X^{(k)})$ defined in Eq. (\ref{equ-ce-single-sample}) is lower bounded by
    \begin{equation} \label{equ-Lce_lowerBound}
        {L}_{\mathrm{ce}}(\bm X^{(k)}) \geq \frac{\sum_{j=1}^K\big(\gamma\bm{w}_j^{\top}\bm{x}^{(k)}-b_j\big) - K(\gamma\bm{w}_k^{\top}\bm{x}^{(k)}-b_k)}{\big(1+c_1\big)(K-1)}+C,
    \end{equation}
    where
    \begin{equation} \label{equ-fce_c2_formula_2}
        C =\frac{c_1}{1+c_1}\log\bigg(\frac{1+c_1}{c_1}\bigg) + \frac{1}{1+c_1}\log\Big[\big(1+c_1\big)(K-1)\Big].
    \end{equation}
    The inequality becomes an equality if and only if
    \begin{equation} \label{equ-jensen_hold_condition_a}
        c_1=\Bigg[(K-1)\exp\Bigg(\frac{\sum_{j=1}^K \big(\gamma\bm{w}_j^{\top}\bm{x}^{(k)}-b_j\big) -K\big(\gamma\bm{w}_k^{\top}\bm{x}^{(k)}-b_k\big)}{K-1}\Bigg)\Bigg]^{-1},
    \end{equation}
    and
    \begin{equation} \label{equ-jensen_hold_condition_b}
        \gamma\bm{w}_j^{\top}\bm{x}^{(k)}-b_j = \gamma\bm{w}_\ell^{\top}\bm{x}^{(k)}-b_\ell,
        \quad \forall j, \ell \neq k \in [K].
    \end{equation}
\end{lemma}

\begin{proof}
    According to Jensen's inequality in Lemma \ref{jensen_inequality_definition}, the CE loss $L_{\mathrm{ce}}(\bm X^{(k)})$ can be lower bounded,
    \begin{align}
        {L}_{\mathrm{ce}}(\bm X^{(k)}) 
        = &\log\Bigg(1 + \sum_{j=1 \atop j\neq k}^K \frac{\exp\big(\gamma\bm{w}_j^{\top}\bm{x}^{(k)}-b_j\big)}{ \exp\big(\gamma\bm{w}_k^{\top}\bm{x}^{(k)}-b_k\big)}\Bigg) \nonumber \\
        = &\log\Bigg(1+(K -1)\sum_{j=1\atop j\neq k}^K \frac{\exp\big(\gamma\bm{w}_j^{\top}\bm{x}^{(k)}-b_j\big)}{ (K-1)\exp\big(\gamma\bm{w}_k^{\top}\bm{x}^{(k)}-b_k\big)}\Bigg) \nonumber\\
        ~
        \geq &\log\Bigg(1+(K-1)\exp\bigg(\sum_{j=1\atop j\neq k}^K \frac{\gamma\bm{w}_j^{\top}\bm{x}^{(k)}-b_j  - \gamma\bm{w}_k^{\top}\bm{x}^{(k)}+b_k}{K-1}\bigg)\Bigg) \nonumber\\
        ~
        = &\log\Bigg(1+(K-1)\exp\bigg(\frac{\sum_{j=1}^K (\gamma\bm{w}_j^{\top}\bm{x}^{(k)}-b_j) - K(\gamma\bm{w}_k^{\top}\bm{x}^{(k)}-b_k)}{K-1}\bigg)\Bigg) \label{equ-Lce_Jensen},
    \end{align}
    which achieves the equality if and only if $\gamma\bm{w}_j^{\top}\bm{x}^{(k)}-b_j = \gamma\bm{w}_\ell^{\top}\bm{x}^{(k)}-b_\ell$ for all $j,\ell \neq k \in [K]$.

    Second, by the concavity of the function $\log(1+\mathrm{e}^x)$, for any $c_1>0$, we get
    \begin{align}
        & {L}_{\mathrm{ce}}(\bm X^{(k)}) \nonumber \\
        \geq &~\log\Bigg(1+(K-1)\exp\bigg(\frac{\sum_{j=1}^K(\gamma\bm{w}_j^{\top}\bm{x}^{(k)}-b_j)-K(\gamma\bm{w}_k^{\top}\bm{x}^{(k)}-b_k)}{K-1}\bigg)\Bigg) \nonumber\\
        = &~\log\Bigg(\frac{c_1}{1+c_1}\frac{1+c_1}{c_1} +\frac{1+c_1}{1+c_1}(K-1)\exp\bigg(\frac{\sum_{j=1}^K(\gamma\bm{w}_j^{\top}\bm{x}^{(k)}-b_j)-K(\gamma\bm{w}_k^{\top}\bm{x}^{(k)}-b_k)}{K-1}\bigg)\Bigg) \nonumber \\
        \geq &~\frac{c_1}{1+c_1}\log\bigg(\frac{1+c_1}{c_1}\bigg) \nonumber \\
        &\quad +\frac{1}{1+c_1}\log\Bigg(\big(1+c_1\big) (K-1)\exp\bigg(\frac{\sum_{j=1}^K(\gamma\bm{w}_j^{\top}\bm{x}^{(k)}-b_j)-K(\gamma\bm{w}_k^{\top}\bm{x}^{(k)}-b_k)}{K-1}\bigg)\Bigg) \nonumber\\
        = &~\frac1{1+c_1}\frac{\sum_{j=1}^K(\gamma\bm{w}_j^{\top}\bm{x}^{(k)}-b_j)-K(\gamma\bm{w}_k^{\top}\bm{x}^{(k)}-b_k)}{K-1} \nonumber \\
        &\quad +\underbrace{\frac{c_1}{1+c_1}\log\bigg(\frac{1+c_1}{c_1}\bigg) + \frac{1}{1+c_1}\log\big[(1+c_1)(K-1)\big]}_{C}. \nonumber
    \end{align}
    The last inequality achieves its equality if and only if
    \begin{align}
        \frac{1+c_1}{c_1} =& (1+c_1)(K  -1)\exp\bigg(\frac{\sum_{j=1}^K(\gamma\bm{w}_j^{\top}\bm{x}^{(k)}-b_j)-K(\gamma\bm{w}_k^{\top}\bm{x}^{(k)}-b_k)}{K-1}\bigg) \nonumber \\
        &~\text{or}\quad c_1 =0 \quad\text{or} \quad c_1=+\infty. \nonumber
    \end{align}
    Actually, when $c_1 = 0$ or $c_1 = +\infty$, the equality is trivial. Therefore, we have
    \begin{equation}
        c_1=\Bigg[(K-1)\exp\bigg(\frac{\sum_{j=1}^K(\gamma\bm{w}_j^{\top}\bm{x}^{(k)}-b_j) -K(\gamma\bm{w}_k^{\top}\bm{x}^{(k)}-b_k)}{K-1}\bigg)\Bigg]^{-1}, \nonumber
    \end{equation}
    as desired.
\end{proof}

\clearpage
\subsection{Theorems in Open-set Scenario}\label{proof:collapse_on_open_set}
We here present the proof Theorems \ref{theorem_NC_open_set_BCE_main} and \ref{theorem_NC_open_set_CE_main}.

\begin{theorem}\label{them:BCE collapse_supp_open_set}
    (\textbf{BCE})
    Suppose that the first $K_1$ classes in dataset $\mathcal D = \bigcup_{k=1}^K \mathcal D_k$ contains at least one sample, $1 < K_1 < K$, while the other classes are empty, and that the feature dimension is greater than the number of classes, i.e., $n_k = |\mathcal D_k| > 0$ for $\forall k \in [K_1]$, $n_k = |\mathcal D_k| = 0$ for $ K_1 < k \leq K$, and $d \geq K_1$.
    Denote $\tilde{\bm x} = \frac1{K_1}\sum_{k=1}^{K_1}\bar{\bm x}_k$ with $\bar{\bm x}_k = \frac1{n_k}\sum_{\bm x\in\mathcal F_k}\bm x$.
    On the dataset $\mathcal D$, define BCE loss
    \begin{equation}
        f_{\mathrm{bce}\text{-}\mathrm{os}}(\bm W,\bm{b},\mathcal F)\triangleq \frac{1}{K_1}\sum_{k=1}^{K_1}\frac1{n_k}\sum_{\bm X^{(k)}\in \mathcal D_k} L_{\mathrm{bce}}(\bm X^{(k)}),\label{equ-minf_object_os}
    \end{equation}
    where
    \begin{align}
        L_{\mathrm{bce}}(\bm X^{(k)}) = \log\big(1+\e^{-\gamma \bm w_k^\top \bm x^{(k)} + b_k}\big) + \sum_{j=1,j\neq k}^K \log\big(1+\e^{\gamma \bm w_j^\top \bm x^{(k)} - b_j}\big), \nonumber
    \end{align}
    with $\bm x^{(k)} = \mathcal M(\bm X^{(k)})$ for $\forall \bm X^{(k)} \in \mathcal D_k \subset \mathcal D$.

    Then, the BCE loss $f_{\mathrm{bce}\text{-}\mathrm{os}}(\bm W,\bm{b},\mathcal F)$ satisfies
    \begin{align}
        f_{\mathrm{bce}\text{-}\mathrm{os}}(\bm W,\bm{b},\mathcal F) \geq& \log\big(1 + \e^{b^\star - \gamma\sqrt{1 - \|\tilde{\bm x}\|^2}}\big) + (K_1-1) \log\big(1 + \e^{- \frac{\gamma}{K_1-1}\sqrt{1 - \|\tilde{\bm x}\|^2} - b^\star}\big) \nonumber \\
        &~ +(K - K_1)\log\big(1 + \e^{- \gamma \|\tilde{\bm x}\| - b^\dag}\big),
        \label{eq_lower_bound_bce_os}
    \end{align}
    where $b^\star$ is the unique real solution of
    \begin{align}
         \e^{\frac{\gamma}{K_1-1}\sqrt{1-\|\tilde{\bm x}\|^2}}\e^{2b} - (K_1-2)\e^b - (K_1-1)\e^{\gamma \sqrt{1-\|\tilde{\bm x}\|^2}} =0,\nonumber
    \end{align}
    and $b^\dag \in \mathbb R$. When the BCE loss $f_{\mathrm{bce}\text{-}\mathrm{os}}(\bm W,\bm{b},\mathcal F)$ reaching its lower bound in Eq. (\ref{eq_lower_bound_bce_os}), it satisfies
    \begin{itemize}
      \item (collapse)
          \begin{align}
            \bm x^{(k)} = \bar{\bm x}_k = \frac1{n_k}\sum_{\bm x\in\mathcal F_k} \bm x,\quad\forall \bm x^{(k)} \in\mathcal F_k,~\forall k\in[K_1]; \nonumber
          \end{align}
      \item (equiangular)
        \begin{align}
            &\bm w_k^\top \bm w_j = -\frac{1}{K_1-1},\quad\forall j\neq k\in[K_1], \nonumber\\
            &\bar{\bm x}_k^\top \bar{\bm x}_j = -\frac{1}{K_1-1} + \frac{K_1}{K_1-1}\|\tilde{\bm x}\|^2,\quad\forall j\neq k\in[K_1]; \nonumber
        \end{align}
      \item (isogonal)
        \begin{align}
            &\bm w_k^\top \bar{\bm x}_k = \sqrt{1-\|\tilde{\bm x}\|^2},\quad\forall k\in[K_1], \nonumber\\
            &\bm w_j^\top \bar{\bm x}_k = -\frac1{K_1-1}\sqrt{1-\|\tilde{\bm x}\|^2},\quad\forall j\neq k\in[K_1], \nonumber\\
            &\bm w_k^\top \tilde{\bm x} = 0, \quad\forall k\in[K_1]; \nonumber
        \end{align}
      \item (minority collapse) for $\forall j = K_1+1, K_1+2, \cdots, K$, if $\|\tilde{\bm x}\| \neq 0$,
        \begin{align}
            &\bm w_j = -\frac{\tilde{\bm x}}{\|\tilde{\bm x}\|}, \nonumber
        \end{align}
        and for $\forall k\in[K_1],~\forall j = K_1+1, K_1+2, \cdots, K$,
        \begin{align}
            &\bm w_j^\top \bar{\bm x}_k = -\|\tilde{\bm x}\|, \nonumber\\
            &\bm w_j^\top \bm w_k = 0, \nonumber
        \end{align}
    \end{itemize}
    and
    \begin{align}
        &b_k = b^\star,\quad \forall k\in[K_1], \nonumber\\
        &b_j = b^\dagger,\quad\forall j = K_1+1, K_1+2, \cdots, K.\nonumber
    \end{align}
    As the BCE loss $f_{\mathrm{bce}\text{-}\mathrm{os}}(\bm W,\bm{b},\mathcal F)$ approaching its minimum value, $b^\dag \rightarrow +\infty$ and $\|\tilde{\bm x}\| \rightarrow 0$.
\end{theorem}
\begin{proof}

    According to Lemma \ref{lem:lemmaA.1_open_set}, for any $c_1, c_2, c_3 > 0$ and any $\bm X^{(k)} \in \mathcal D_k$ with $\forall k\in[K_1]$,
    \begin{align}
        {L}_{\mathrm{bce}}(\bm X^{(k)}) \geq&~ \frac{1}{1+c_{1}}\big(-\gamma \bm{w}_{k}^{\top}\bm x^{(k)}+b_{k}\big) + \frac{1}{1+c_{2}}\sum_{j=1 \atop j\neq k}^{K_1}\big(\gamma \bm{w}_{j}^{\top}\bm x^{(k)}-b_{j}\big)  \nonumber \\
        &\quad + \frac{1}{1+c_{3}}\sum_{j=K_1 + 1}^{K} \big(\gamma \bm{w}_{j}^{\top}\bm x^{(k)} - b_j\big) + C, \nonumber
    \end{align}
    where
    \begin{equation}
		\begin{aligned}
		 C=& \log(1+c_1)-\frac{c_1}{1+c_1}\log(c_1) + (K_1-1)\bigg(\log(1+c_2)-\frac{c_2}{1+c_2}\log(c_2)\bigg) \nonumber \\
		 &~ +(K-K_1)\bigg(\log(1+c_3) -\frac{c_3}{1+c_3}\log(c_3)\bigg), \nonumber
		\end{aligned} 
\end{equation}
    and the inequality becomes an equality if and only if
    \begin{align}
        &c_1=\exp\big(\gamma \bm{w}_k^\top\bm x^{(k)}-b_k\big),\quad 1\leq k\leq K_1, \nonumber\\
        & c_2=\exp\big(b_j-\gamma \bm{w}_j^\top\bm x^{(k)}\big),\quad 1 \leq j\neq k \leq K_1, \nonumber\\
        & c_3=\exp\big(b_j- \gamma \bm{w}_j^\top\bm x^{(k)}\big),\quad 1 \leq k\leq K_1,~K_1 < j\leq K. \nonumber
    \end{align}

    We take the same $c_1, c_2$, and $c_3$ for all $\bm X^{(k)}$, $\forall k\in[K_1]$, then
    \begin{align}
        &f_{\mathrm{bce}\text{-}\mathrm{os}}(\bm{W}, \bm b, \mathcal F) = \frac{1}{K_1}\sum_{k=1}^{K_1} \frac 1{n_k} \sum_{\bm X^{(k)}\in \mathcal D_k} {L}_\mathrm{bce}\big(\bm X^{(k)}\big)  \nonumber\\
        \geq~&\frac{1}{K_1}\sum_{k=1}^{K_1} \frac1{n_k} \sum_{\bm x^{(k)}\in \mathcal F_k} \bigg[\frac{1}{1+c_1}\Big(-\gamma \bm{w}_k^\top \bm x^{(k)} + b_k\Big) + \frac{1}{1+c_2} \sum_{j=1\atop j\neq k}^{K_1} \Big(\gamma \bm{w}_j^\top \bm x^{(k)} - b_j\Big)  \nonumber \\
        &\quad + \frac{1}{1+c_3} \sum_{j=K_1+1}^K \big(\gamma \bm{w}_j^\top \bm x^{(k)} - b_j\big)\bigg] + C \nonumber\\
        =~&\frac{1}{K_1}\frac{1}{1+c_2}\sum_{k=1}^{K_1}\frac1{n_k}\sum_{\bm x^{(k)}\in \mathcal F_k} \sum_{j=1 \atop j\neq k}^{K_1} \Big(\gamma \bm{w}_j^\top \bm x^{(k)} - b_j\Big) \nonumber\\
        &\quad + \frac{1}{K_1}\frac{1}{1+c_3}\sum_{k=1}^{K_1}\frac1{n_k}\sum_{\bm x^{(k)}\in \mathcal F_k} \sum_{j=K_1+1}^{K} \big(\gamma \bm{w}_j^\top \bm x^{(k)}-b_j\big) \nonumber \\
        &\quad -  \frac1{K_1}\frac1{1+c_1}\sum_{k=1}^{K_1}\frac1{n_k}\sum_{\bm x^{(k)}\in \mathcal F_k} \Big(\gamma \bm{w}_k^\top \bm x^{(k)} - b_k\Big)  + C  \nonumber \\
        =~&\frac1{K_1}\frac1{1+c_2}\sum_{k=1}^{K_1}\frac1{n_k}\sum_{\bm x^{(k)}\in \mathcal F_k}\sum_{j=1}^{K_1}\Big(\gamma \bm{w}_j^\top \bm x^{(k)}-b_j - \gamma \bm{w}_k^\top \bm x^{(k)} + b_k\Big)
         \nonumber \\
        &\quad +\frac1{K_1}\frac1{1+c_2}\sum_{k=1}^{K_1} \frac1{n_k} \sum_{\bm x^{(k)}\in \mathcal F_k} \sum_{j=1 \atop j\neq k}^{K_1} \Big(\gamma \bm{w}_k^\top \bm x^{(k)} - b_k\Big) \nonumber \\
        &\quad + \frac{1}{K_1}\frac{1}{1+c_3}\sum_{k=1}^{K_1}\frac1{n_k}\sum_{\bm x^{(k)}\in \mathcal F_k} \sum_{j=K_1+1}^{K} \big(\gamma \bm{w}_j^\top \bm x^{(k)} - b_j\big) \nonumber \\
        &\quad - \frac1{K_1}\frac1{1+c_1}\sum_{k=1}^{K_1}\frac1{n_k} \sum_{\bm x^{(k)}\in \mathcal F_k} \Big(\gamma \bm{w}_k^\top \bm x^{(k)} - b_k\Big) + C \nonumber \\
        =~&\frac1{K_1}\frac1{1+c_2}\bigg[\sum_{k=1}^{K_1} \frac1{n_k} \sum_{\bm x^{(k)} \in \mathcal F_k} \sum_{j=1}^{K_1}\Big(\gamma \bm{w}_j^\top \bm x^{(k)} - \gamma \bm{w}_k^\top \bm x^{(k)} \Big) - \underbrace{\sum_{k=1}^{K_1} \frac1{n_k} \sum_{\bm x^{(k)} \in \mathcal F_k} \sum_{j=1}^{K_1} (b_j-b_k)}_0\bigg] \nonumber \\
        &\quad+\frac1{K_1}\bigg(\frac{K_1-1}{1+c_2}-\frac1{1+c_1}\bigg)\sum_{k=1}^{K_1}\frac1{n_k}\sum_{\bm x^{(k)}\in\mathcal F_k} \gamma \bm{w}_k^\top \bm x^{(k)} \nonumber \\
        &\quad - \frac1{K_1}\bigg(\frac{K_1-1}{1+c_2}-\frac1{1+c_1}\bigg)\sum_{k=1}^{K_1}b_k + \frac{1}{K_1}\frac{1}{1+c_3}\sum_{k=1}^{K_1}\frac1{n_k}\sum_{\bm x^{(k)}\in \mathcal F_k} \sum_{j=K_1+1}^{K} \gamma \bm{w}_j^\top \bm x^{(k)} \nonumber \\
        &\quad - \frac{1}{K_1}\frac{1}{1+c_3}\sum_{k=1}^{K_1}\frac1{n_k}\sum_{\bm x^{(k)}\in \mathcal F_k} \sum_{j=K_1+1}^{K} b_j + C \nonumber \\
        =~&\frac1{1+c_2}\sum_{k=1}^{K_1}\gamma \bm{w}_k^\top\big(\tilde{\bm x} - \bar{\bm x}_k\big)
        + \frac1{K_1} \bigg(\frac{K_1-1}{1+c_2} - \frac1{1+c_1}\bigg) \sum_{k=1}^{K_1}\big(\gamma \bm{w}_k^\top\bar{\bm{x}}_k - b_k\big)
         \nonumber\\
        &\quad + \frac{1}{1+c_3} \sum_{j=K_1+1}^{K}\big(\gamma \bm{w}_j^\top \tilde{\bm x} - b_j\big) + C \nonumber \\
        \triangleq~&\hat{f}^{(1)}_{\text{bce-os}}(\bm{W},\bm b,\mathcal F),\nonumber
    \end{align}
    where $\tilde{\bm x} = \frac{1}{K_1}\sum_{j=1}^{K_1}\bar{\bm{x}}_j$ and $\bar{\bm{x}}_j = \frac1{n_j}\sum_{\bm x^{(j)}\in\mathcal F_j} \bm x^{(j)},~\forall j\in[K_1]$.

    For the function $\hat{f}^{(1)}_{\text{bce-os}}(\bm{W},\bm b,\mathcal F)$, its global minima are located at its stationary points in terms of $b_k$ for $\forall k\in[K_1]$,
    \begin{equation}
        0 = \frac{\partial \hat{f}_{\text{bce-os}}}{\partial b_k} = -\frac1{K_1}\bigg(\frac{K_1-1}{1+c_2}-\frac1{1+c_1}\bigg). \label{b_condition_bce_os}
    \end{equation}
    Therefore,
    \begin{align}
        \hat{f}^{(1)}_{\text{bce-os}}(\bm{W},\bm b,\mathcal F) \geq \frac1{1+c_2}\sum_{k=1}^{K_1}\gamma \bm{w}_k^\top\big(\tilde{\bm x} - \bar{\bm x}_k\big) + \frac{1}{1+c_3} \sum_{j=K_1+1}^{K} \gamma \bm{w}_j^\top \tilde{\bm x} - \frac{1}{1+c_3} \sum_{j=K_1+1}^{K} b_j + C. \nonumber
    \end{align}
    Furthermore, according to the AM-GM inequality (see Lemma \ref{young_inequality_definition}),
    \begin{equation}
        \bm{u}^\top\bm{v}\geq - \Big(\frac{c}{2}\|\bm{u}\|_2^2+\frac{1}{2c}\|\bm{v}\|_2^2\Big), \quad\forall\bm{u}, \bm{v}\in\mathbb{R}^d, ~\forall c > 0, \nonumber
    \end{equation}
    which becomes an equality when $\bm{v} = -c\bm{u}$. Then,
    \begin{align}
        & f_{\text{bce-os}}(\bm{W},\bm b,\mathcal F) \geq ~\hat{f}^{(1)}_{\text{bce-os}}(\bm{W},\bm b,\mathcal F) \nonumber \\
        \geq& \frac1{1+c_2}\sum_{k=1}^{K_1}\gamma \bm{w}_k^\top\big(\tilde{\bm x} - \bar{\bm x}_k\big) + \frac{1}{1+c_3} \sum_{j=K_1+1}^{K} \gamma \bm{w}_j^\top \tilde{\bm x} - \frac{1}{1+c_3} \sum_{j=K_1+1}^{K} b_j + C \nonumber \\
        \geq&  -\frac{\gamma}{1+c_2}\bigg(\frac{c_4}2\sum_{k=1}^{K_1}\|\bm{w}_k\|_2^2 + \frac1{2c_4}\sum_{k=1}^{K_1}\|\tilde{\bm{x}} - \bar{\bm{x}}_k\|_2^2\bigg)
        + \frac{1}{1+c_3} \sum_{j=K_1+1}^{K} \gamma \bm{w}_j^\top \tilde{\bm x}  \nonumber\\
        &\quad - \frac{1}{1+c_3} \sum_{j=K_1+1}^{K} b_j + C \label{c4_ineq_os} \\
        \geq& -\frac{\gamma}{1+c_2}\bigg(\frac{c_4}2\sum_{k=1}^{K_1}\|\bm{w}_k\|_2^2 + \frac1{2c_4}\sum_{k=1}^{K_1}\|\tilde{\bm{x}} - \bar{\bm{x}}_k\|_2^2\bigg)
        - \frac{\gamma}{1+c_3} \sum_{j=K_1+1}^{K} \|\bm{w}_j\| \|\tilde{\bm x}\|  \nonumber\\
        &\quad - \frac{1}{1+c_3} \sum_{j=K_1+1}^{K} b_j + C \label{c5_ineq_os} \\
        =& -\frac{\gamma}{1+c_2}\bigg[\frac{c_4}2\sum_{k=1}^K\|\bm{w}_k\|_2^2 + \frac1{2c_4}\bigg(\sum_{k=1}^{K_1}\|\bar{\bm{x}}_k\|_2^2  -K_1\|\tilde{\bm{x}}\|_2^2\bigg)\bigg] \nonumber \\
        &\quad -\gamma\frac{K-K_1}{1+c_3} \|\tilde{\bm x}\| - \frac{1}{1+c_3} \sum_{j=K_1+1}^{K} b_j + C \nonumber \\
        \geq& -\frac{\gamma}{1+c_2}\bigg(\frac{c_4}{2} K_1 +\frac{1}{2c_4}K_1\bigg) +\frac{\gamma}{1+c_2}\frac{1}{2c_4}K_1\|\tilde{\bm{x}}\|_2^2 \nonumber\\
        &\quad - \gamma\frac{K-K_1}{1+c_3} \|\tilde{\bm x}\| - \frac{1}{1+c_3} \sum_{j=K_1+1}^{K} b_j + C \label{h_mean=h_os} \\
        =& -\frac{\gamma K_1}{1+c_2}\bigg(\frac{c_4}{2} +\frac{1}{2c_4}\bigg) + \frac{\gamma K_1}{1+c_2}\frac{1}{2c_4} \|\tilde{\bm x}\|^2 - \gamma\frac{K-K_1}{1+c_3} \|\tilde{\bm x}\| - \frac{1}{1+c_3} \sum_{j=K_1+1}^{K} b_j + C, \nonumber
    \end{align}
    where inequality (\ref{c4_ineq_os}) becomes an equality if and only if
    \begin{equation}
        c_{4} \bm{w}_{k} = \bar{\bm{x}}_k - \tilde{\bm{x}}, \quad \forall k\in[K], ~\forall c_4 > 0, \label{equ-c4_hold_os}
    \end{equation}
    inequality (\ref{c5_ineq_os}) becomes an equality if and only if
    \begin{align}
        c_5 \bm w_j = - \tilde{\bm x},\quad K_1 + 1\leq j\leq K,~\forall c_5 >0, \label{equ-c5_hold_os}
    \end{align}
    and inequality (\ref{h_mean=h_os}) becomes an equality if and only if
    \begin{align}
        &1 = \|\bar{\bm x}_k\| = \bigg\|\frac1{n_k}\sum_{\bm x^{(k)}\in\mathcal F_k} \bm x^{(k)}\bigg\| \leq \frac1{n_k}\sum_{\bm x^{(k)}\in\mathcal F_k} \big\|\bm x^{(k)}\big\| = 1 \nonumber \\
        &\Leftrightarrow \bm x^{(k)}  = \bar{\bm x}_k,~\forall \bm x^{(k)} \in\mathcal F_k,\quad \forall k\in[K_1]. \label{equ-h_mean=h_os}
    \end{align}
    With Eq. (\ref{equ-c5_hold_os}), $\|\tilde{\bm x}\| = c_5$, then
    \begin{align}
    	& f_{\text{bce-os}}(\bm{W},\bm b,\mathcal F) \nonumber \\
        \geq~&-\frac{\gamma K_1}{1+c_2}\bigg(\frac{c_4}{2} +\frac{1}{2c_4}\bigg) + \frac{\gamma K_1}{1+c_2}\frac{1}{2c_4} \|\tilde{\bm x}\|^2-\frac{\gamma(K-K_1)}{1+c_3} \|\tilde{\bm x}\| - \frac{1}{1+c_3} \sum_{j=K_1+1}^{K} b_j + C\nonumber\\
        =~&-\frac{\gamma K_1}{1+c_2}\bigg(\frac{c_4}{2} +\frac{1}{2c_4}\bigg) + \frac{\gamma K_1}{1+c_2}\frac{c_5^2}{2c_4} - \frac{\gamma(K-K_1)}{1+c_3} c_5 - \frac{1}{1+c_3} \sum_{j=K_1+1}^{K} b_j + C. \label{eq_bce_lower_boundary_os}
    \end{align}

    Finally, Lemma \ref{lem:lemmaA.2_open_set} completes the proof.
\end{proof}

\begin{lemma} \label{lem:lemmaA.2_open_set}
    Suppose that the first $K_1$ classes in dataset $\mathcal D$ contains at least one sample, $K_1 < K$, while the other classes are empty, and that the feature dimension is greater than the number of classes, i.e., $n_k = |\mathcal D_k| > 0$ for $\forall k \in [K_1]$, $n_k = |\mathcal D_k| = 0$ for $ K_1 < k \leq K$, and $d \geq K_1$.
    Then, for the BCE loss $f_{\mathrm{bce}\text{-}\mathrm{os}}(\bm W,\bm{b},\mathcal F)$ defined in Eq. (\ref{equ-minf_object_os}),
    \begin{align}
        f_{\mathrm{bce}}(\bm W,\bm{b},\mathcal F)\geq&~ \log\bigg(1+\e^{b^\star - \gamma \sqrt{1-\|\tilde{\bm x}\|^2}}\bigg) + (K_1-1)\log\bigg(1 +\e^{-\frac{\gamma }{K_1-1}\sqrt{1-\|\tilde{\bm x}\|^2}-b^\star}\bigg) \nonumber \\
        &\quad +(K-K_1)\log\bigg(1+\e^{-\gamma \|\tilde{\bm x}\|-b^\dagger}\bigg),
    \end{align}
    where $\tilde{\bm x} = \frac1{K_1}\sum_{k=1}^{K_1}\bar{\bm x}_k$ with $\bar{\bm x}_k = \frac1{n_k}\sum_{\bm x^{(k)}\in\mathcal F_k} \bm x^{(k)}$, $b^\dagger\in\mathbb R$, and $b^\star$ is the unique real solution of
    \begin{align}
         \e^{\frac{\gamma \sqrt{1-\|\tilde{\bm x}\|^2}}{K_1-1}}\e^{2b} - (K_1-2)\e^b - (K_1-1)\e^{\gamma \sqrt{1-\|\tilde{\bm x}\|^2}} =0. \label{eq_b*_of_bce_os}
    \end{align}

    When $f_{\mathrm{bce}}(\bm W,\bm{b},\mathcal F)$ approaching its minimum value, $b^\dag \rightarrow +\infty$ and $\|\tilde{\bm x}\| \rightarrow 0$.
\end{lemma}
\begin{proof}
    According to Theorem \ref{them:BCE collapse_supp_open_set}, the BCE loss $f_{\text{bce-os}}(\bm W,\bm{b},\mathcal F)$ satisfies
    \begin{align}
    	f_{\text{bce-os}}(\bm{W},\bm b,\mathcal F)
        \geq~&-\frac{\gamma K_1}{1+c_2}\bigg(\frac{c_4}{2} +\frac{1}{2c_4}\bigg) + \frac{\gamma K_1}{1+c_2}\frac{c_5^2}{2c_4} - \frac{\gamma (K-K_1)}{1+c_3} {c_5}  \nonumber\\
  &\quad - \frac{1}{1+c_3} \sum_{j=K_1+1}^{K} b_j + C, \label{eq_bce_lower_boundary_os_lem}
    \end{align}
    and the loss achieves this constant lower bound if and only if Eqs. (\ref{equ-c1_1_os}), (\ref{equ-c2_1_os}), (\ref{equ-c3_1_os}), (\ref{b_condition_bce_os}), (\ref{equ-c4_hold_os}), (\ref{equ-c5_hold_os}), and (\ref{equ-h_mean=h_os}) hold.

    From Eq. (\ref{equ-c4_hold_os}), for $\forall c_4 > 0$,
    \begin{align}
        c_4 \sum_{k=1}^{K_1} \bm w_k = \sum_{k=1}^{K_1} (\bar{\bm x}_k - \tilde{\bm x}) = \sum_{k=1}^{K_1} \bar{\bm x}_k - K_1 \tilde{\bm x} = \sum_{k=1}^{K_1} \bar{\bm x}_k - \frac{K_1}{K_1} \sum_{k=1}^{K_1} \bar{\bm x}_k = \bm 0 \Leftrightarrow \sum_{k=1}^{K_1} \bm w_k = \bm 0.
        \label{eq_sum_of_w_k}
    \end{align}
    Then, from Eq. (\ref{equ-c4_hold_os}),
    \begin{align}
        c_4 \bm w_k = \bar{\bm x}_k - \tilde{\bm x} \quad & \Rightarrow\quad c_4 = c_4 \bm w_k^\top\bm w_k = \bm w_k^\top\bar{\bm x}_k - \bm w_k^\top\tilde{\bm x}  \nonumber \\
        & \Rightarrow\quad K_1c_4 = \sum_{k=1}^{K_1} \bm w_k^\top\bar{\bm x}_k - \bigg(\sum_{k=1}^{K_1} \bm w_k\bigg)^\top\tilde{\bm x} \nonumber \\
        &\overset{\ref{eq_sum_of_w_k}}{\Rightarrow}\quad \sum_{k=1}^{K_1} \bm w_k^\top\bar{\bm x}_k = c_4 K_1, \label{eq_sum_wx_bce_os} \\
        c_4 \bm w_k = \bar{\bm x}_k - \tilde{\bm x} \quad &\Rightarrow\quad c_4 \bm w_k^\top\bar{\bm x}_k \overset{\ref{equ-h_mean=h_os}}{=} 1 - \tilde{\bm x}^\top\bar{\bm x}_k \nonumber \\
        &\overset{\ref{eq_sum_wx_bce_os}}{\Rightarrow}\quad c_4^2K_1 = c_4 \sum_{k=1}^{K_1} \bm w_k^\top\bar{\bm x}_k = K_1 - \sum_{k=1}^{K_1} \tilde{\bm x}^\top\bar{\bm x}_k = K_1 - K_1 \|\tilde{\bm x}\|^2 \nonumber \\
        &\Rightarrow\quad c_4^2 = 1 - \|\tilde{\bm x}\|^2, \label{eq_value_of_c4_bce_os} \\
        c_4 \bm w_k = \bar{\bm x}_k - \tilde{\bm x} \quad &\Rightarrow\quad \tilde{\bm x} = \bar{\bm x}_k - c_4 \bm w_k \nonumber \\
        &\overset{\ref{equ-h_mean=h_os},\ref{eq_value_of_c4_bce_os}}{\Rightarrow}\quad
        1-c_4^2 = \|\tilde{\bm x}\|^2 = \|\bar{\bm x}_k\|^2 + c_4^2\|\bm w_k\|^2 - 2c_4 \bm w_k^\top\bar{\bm x}_k = 1+c_4^2 - 2c_4\bm w_k^\top\bar{\bm x}_k \nonumber \\
        &\Rightarrow\quad \bm w_k^\top\bar{\bm x}_k = c_4 = \sqrt{1 - \|\tilde{\bm x}\|^2},\quad\forall k\in[K_1], \label{eq_wxk_c_4_bce_os} \\
        c_4 \bm w_k = \bar{\bm x}_k - \tilde{\bm x} \quad &\Rightarrow\quad c_4 = c_4 \bm w_k^\top\bm w_k = \bm w_k^\top\bar{\bm x}_k - \bm w_k^\top\tilde{\bm x} \nonumber\\
        &\overset{\ref{eq_wxk_c_4_bce_os}}{\Rightarrow}\quad \bm w_k^\top\tilde{\bm x} = 0,\quad\forall k\in[K_1]. \label{eq_wxc_bce_os}
    \end{align}
    Meanwhile,
    \begin{align}
        &c_4 \bm w_k = \bar{\bm x}_k - \tilde{\bm x} \quad \Rightarrow\quad c_4 \bm w_k^\top \bm w_j = \bm w_j^\top \bar{\bm x}_k - \bm w_j^\top \tilde{\bm x} \nonumber \\
         \Rightarrow\quad &c_4 \gamma \bm w_k^\top \bm w_j = \gamma\bm w_j^\top \bar{\bm x}_k - b_j - \gamma\bm w_j^\top \tilde{\bm x} + b_j =\gamma\bm w_j^\top \bar{\bm x}_k - b_j - \frac1{K_1}\sum_{k=1}^{K_1}\big(\gamma \bm w_j^\top \bar{\bm x}_k-b_j\big) \nonumber \\
         &\qquad\quad
        \overset{\ref{equ-c1_1_os},\ref{equ-c2_1_os},\ref{equ-h_mean=h_os}}{=} -c'_2 - \frac{c'_1 - (K_1-1)c'_2}{K_1} = - \frac{c'_1 + c'_2}{K_1}\nonumber\\
        ~
        \Rightarrow\quad & \bm w_k^\top \bm w_j = - \frac{c'_1 + c'_2}{c_4\gamma K_1},\quad \forall j\neq k\in [K_1], \label{eq_wk_wj_os_1}
    \end{align}
    where $c'_1 = \log(c_1), c'_2 = \log(c_2)$.
    From Eq. (\ref{eq_sum_of_w_k})
    \begin{align}
        \sum_{j=1}^{K_1} \bm w_j = \bm 0 \quad\Rightarrow\quad &\bm w_k^\top\sum_{j=1}^{K_1} \bm w_j = 0 \quad\Rightarrow\quad 1 = -\sum_{j=1\atop j\neq k}^{K_1} \bm w_k^\top\bm w_j \overset{\ref{eq_wk_wj_os_1}}{=} (K_1-1) \frac{c'_1 + c'_2}{c_4\gamma K_1} \quad \nonumber \\
        \quad\Rightarrow\quad& \frac{c'_1 + c'_2}{c_4\gamma K_1} = \frac1{K_1 - 1} \quad\Rightarrow\quad \bm w_k^\top \bm w_j = -\frac1{K_1-1}, ~\forall j\neq k\in[K_1]. \label{eq_wk_wj_bce_os}
    \end{align}

    Then,
    \begin{align}
         &2 -2\bar{\bm x}_k^\top\bar{\bm x}_j = \|\bar{\bm x}_k - \bar{\bm x}_j\|^2 \overset{\ref{equ-c4_hold_os}}{=} \|c_4 \bm w_k - c_4 \bm w_j\|^2 \nonumber \\
         &\qquad\qquad~~ = c_4^2\big(\|\bm w_k\|^2 + \|\bm w_j\|^2 -2\bm w_k^\top \bm w_j\big) \overset{\ref{eq_wk_wj_bce_os}}{=} 2 c_4^2 \frac{K_1}{K_1-1} \nonumber \\
         \Rightarrow& \quad\bar{\bm x}_k^\top\bar{\bm x}_j = 1 - \frac{K_1}{K_1-1}c_4^2 = 1 - \frac{K_1}{K_1-1}\big(1 - \|\tilde{\bm x}\|^2\big) \nonumber \\
         &\qquad\qquad~~ =-\frac{1}{K_1-1} +\frac{K_1}{K_1-1}\|\tilde{\bm x}\|^2,\quad \forall j\neq k\in[K_1], 
         \label{eq_xk_xj_bce_os}
    \end{align}
    and
    \begin{align}
         & c_4\bm w_k^\top\bar{\bm x}_j \overset{\ref{equ-c4_hold_os}}{=} \bar{\bm x}_k^\top\bar{\bm x}_j - \tilde{\bm x}^\top \bar{\bm x}_j = \bar{\bm x}_k^\top\bar{\bm x}_j - \frac{1}{K_1}\sum_{\ell=1}^{K_1}\bar{\bm x}_\ell^\top \bar{\bm x}_j \nonumber \\
         &\qquad\quad~ \overset{\ref{eq_xk_xj_bce_os}}{=} -\frac1{K_1} + \bigg(1 - \frac{K_1-1}{K_1}\bigg)\bigg(1 - \frac{K_1}{K_1-1}c_4^2\bigg) = -\frac{1}{K_1-1}c_4^2\nonumber\\
          \Rightarrow\quad& \bm w_k^\top\bar{\bm x}_j = -\frac{1}{K_1-1}c_4 = -\frac{\sqrt{1 - \|\tilde{\bm x}\|^2}}{K_1-1},\quad\forall j\neq k\in[K_1]. \label{eq_wk_xj_bce_os}
    \end{align}

    Now, we try to express the lower bound of $f_{\mathrm{bce}}$ using $c_4$; to this end, it is necessary to derive the relationship between $c_4$ with $c_1, c_2, c_3$, and $c_5$. In first, from Eq. (\ref{b_condition_bce_os}),
    \begin{align}
        0 = \frac{K_1-1}{1+c_2} - \frac{1}{1+c_1}
        ~ \overset{\ref{equ-c1_1_os},\ref{equ-c2_1_os},\ref{equ-h_mean=h_os}} \Rightarrow~ &(K_1-2)+(K_1  -1)\e^{\gamma \bm w_k^\top\bar{\bm x}_k -b_k} = \e^{b_k-\gamma\bm w_k^\top\bar{\bm x}_j}
        ~\overset{\ref{eq_wxk_c_4_bce_os},\ref{eq_wk_xj_bce_os}} \nonumber \\
        \Rightarrow\quad~ &(K_1-2) + (K_1-1)\e^{\gamma c_4 - b_k} = \e^{b_k + \frac{\gamma}{K_1-1}c_4} \nonumber,
    \end{align}
    then,
    \begin{align}
        b_k = b^\star,\quad \forall k\in[K_1],
    \end{align}
    where $b^\star$ is the unique real solution of
    \begin{align}
         \e^{\frac{\gamma c_4}{K_1-1}}\e^{2b} - (K_1-2)\e^b - (K_1-1)\e^{\gamma c_4} =0, \nonumber
    \end{align}
    i.e., the solution of Eq. (\ref{eq_b*_of_bce_os}).
    Then,
    \begin{align}
        c_1 &~= \exp\big(\gamma\bm w_k^\top\bar{\bm x}_k - b_k\big) = \exp\big(\gamma c_4 - b^\star\big),\label{eq_value_of_c1_bce_os}\\
        c_2 &~= \exp\big(b_k - \gamma\bm w_k^\top\bar{\bm x}_j\big) = \exp\big(b^\star + \frac{\gamma c_4}{K_1-1} \big).\label{eq_value_of_c2_bce_os}
    \end{align}
    From Eq. (\ref{equ-c5_hold_os}),
    \begin{align}
        c_5 \bm w_j = -\tilde{\bm x}\quad\Rightarrow\quad& c_5^2 \|\bm w_j\|^2 = \|\tilde{\bm x}\|^2 \quad\Rightarrow\quad c_5^2 \overset{\ref{eq_value_of_c4_bce_os}}{=} 1-c_4^2, \label{eq_value_of_c5_bce_os} \\
        c_5 \bm w_j = -\tilde{\bm x}\quad\Rightarrow\quad& c_5 \bm w_j^\top \bar{\bm x}_k = -\tilde{\bm x}^\top\bar{\bm x}_k = -\frac1{K_1}\sum_{\ell = 1}^{K_1}\bar{\bm x}_{\ell}^\top\bar{\bm x}_k \nonumber \\
        &\qquad\qquad \overset{\ref{eq_xk_xj_bce_os}}= -\frac1{K_1}\bigg[1 + (K_1-1)\bigg(1-\frac{K_1}{K_1-1}c_4^2\bigg)\bigg] \nonumber \\
        &\qquad\qquad = -\frac1{K_1}\big(K_1 -K_1 c_4^2\big) \nonumber \\
        \Rightarrow\quad& \bm w_j^\top \bar{\bm x}_k = -\frac{1-c_4^2}{c_5} \overset{\ref{eq_value_of_c5_bce_os}}= -\sqrt{1-c_4^2},\quad K_1<j\leq K,~\forall k\in[K_1] \\
        \Rightarrow\quad& c_3 = \exp\big(b_j-\gamma\bm w_j^\top\bar{\bm x}_k\big) = \exp\Big(b_j+\gamma\sqrt{1-c_4^2}\Big),\quad K_1<j \nonumber \\
        &\quad \leq K,~\forall k\in[K_1] \label{eq_value_of_c3_bce_os} \\
        \Rightarrow\quad& b_j = b_\ell = b^\dag,\quad\forall K_1< j\neq \ell \leq K. \label{eq_value_of_bj_bce_os}
    \end{align}

    Therefore,
    \begin{align}
    	&f_{\text{bce-os}}(\bm{W},\bm b,\mathcal F) \nonumber \\
        \geq~&-\frac{\gamma K_1}{1+c_2}\bigg(\frac{c_4}{2} +\frac{1}{2c_4}\bigg) + \frac{\gamma K_1}{1+c_2}\frac{c_5^2}{2c_4} - \frac{\gamma (K-K_1)}{1+c_3} c_5 - \frac{1}{1+c_3} \sum_{j=K_1+1}^{K} b_j + C \nonumber \\
        =~& -\frac{\gamma K_1}{1+c_2}\bigg(\frac{c_4}{2} +\frac{1}{2c_4}\bigg) + \frac{\gamma K_1}{1+c_2}\frac{1-c_4^2}{2c_4} - \frac{\gamma (K-K_1)}{1+c_3} c_5 -\frac{1}{1+c_3} \sum_{j=K_1+1}^{K} b_j + C \nonumber \\
        =~& -\frac{\gamma K_1}{1+c_2}c_4 - \frac{\gamma (K-K_1)}{1+c_3} c_5 - \frac{1}{1+c_3} \sum_{j=K_1+1}^{K} b_j + C \nonumber\\
        =~& \log\bigg(1+\frac1{c_1}\bigg) + (K_1-1)\log\bigg(1+\frac1{c_2}\bigg) + (K-K_1)\log\bigg(1+\frac1{c_3}\bigg) \nonumber \\
        &\quad +(\gamma c_4 - b^\star)\underbrace{\bigg(\frac1{1+c_1} - \frac{K_1-1}{1+c_2}\bigg)}_0 - \frac{1}{1+c_3} \sum_{j=K_1+1}^{K} b_j \nonumber \\
        =~& \log\big(1+\e^{b^\star - \gamma c_4}\big) + (K_1-1)\log\bigg(1+\e^{-\frac{\gamma c_4}{K_1-1}-b^\star}\bigg) \nonumber \\
        &\quad + (K-K_1)\log\bigg(1+\e^{-\gamma\sqrt{1-c_4^2}-b^\dag}\bigg) \nonumber \\
        \triangleq~&\hat{f}^{(2)}_{\text{bce-os}}\big(\|\tilde{\bm x}\|, b^\star, b^\dag\big). \nonumber
    \end{align}
    For the function $\hat{f}^{(2)}_{\text{bce-os}}\big(\|\tilde{\bm x}\|, b^\star, b^\dag\big)$,
    \begin{align}
         \frac{\partial \hat{f}^{(2)}_{\text{bce-os}}\big(\|\tilde{\bm x}\|, b^\star, b^\dag\big)}{\partial b^\dag}
          = -\frac{\e^{-\gamma\|\tilde{\bm x}\| - b^\dag }}{1+\e^{-\gamma\|\tilde{\bm x}\| - b^\dag }} < 0\quad \forall b^\dag\in\mathbb R,\nonumber
    \end{align}
    then, when $\hat{f}^{(2)}_{\text{bce-os}}\big(\|\tilde{\bm x}\|, b^\star, b^\dag\big)$ approaching its minimum value,
    \begin{align}
        b^\dag \rightarrow +\infty,
    \end{align}
    and
    \begin{align}
        & f_{\text{bce-os}}(\bm{W},\bm b,\mathcal F) \geq \hat{f}^{(2)}_{\text{bce-os}}\big(\|\tilde{\bm x}\|, b^\star, b^\dag\big) \nonumber \\
        \geq&~ \log\big(1+\e^{b^\star - \gamma \sqrt{1-\|\tilde{\bm x}\|^2}}\big) + (K_1-1)\log\bigg(1 +\e^{-\frac{\gamma \sqrt{1-\|\tilde{\bm x}\|^2}}{K_1-1}-b^\star}\bigg)  \nonumber \\
        \triangleq&~ \hat{f}^{(3)}_{\text{bce-os}}\big(\|\tilde{\bm x}\|, b^\star\big).
    \end{align}

    For $\hat{f}^{(3)}_{\text{bce-os}}\big(\|\tilde{\bm x}\|, b^\star\big)$ on $\mathbb D_1 = \big\{(\|\tilde{\bm x}\|, b^\star): \|\tilde{\bm x}\| \in [0,1], b^\star ~\text{satisfies}~ Eq.~ (\ref{eq_b*_of_bce_os})\big\}$, as $\mathbb D_1 \subset \mathbb D_2 = \big\{(\|\tilde{\bm x}\|, b^\star): \|\tilde{\bm x}\| \in [0,1], b^\star \in \mathbb R \big\}$,
    \begin{align}
        \min_{\mathbb D_1} \hat{f}^{(3)}_{\text{bce-os}}\big(\|\tilde{\bm x}\|, b^\star\big) \geq \min_{\mathbb D_2} \hat{f}^{(3)}_{\text{bce-os}}\big(\|\tilde{\bm x}\|, b^\star\big). \nonumber
    \end{align}
    For $\hat{f}^{(3)}_{\text{bce-os}}\big(\|\tilde{\bm x}\|, b^\star\big)$ on $\mathbb D_2$, it is easily to deduce that $\frac{\partial \hat{f}^{(3)}_{\text{bce-os}}\big(\|\tilde{\bm x}\|, b^\star\big)}{\partial \|\tilde{\bm x}\|} > 0$; therefore on $\mathbb D_2$, when $\hat{f}^{(3)}_{\text{bce-os}}\big(\|\tilde{\bm x}\|, b^\star\big)$ reaches its minimum value, $\|\tilde{\bm x}\| = 0$.
    In terms of $b^\star$ on $\mathbb D_2$, when $\hat{f}^{(3)}_{\text{bce-os}}\big(\|\tilde{\bm x}\|, b^\star\big)$ reaches its minimum value, $\frac{\partial \hat{f}^{(3)}_{\text{bce-os}}\big(\|\tilde{\bm x}\|, b^\star\big)}{\partial b^\star} = 0$, which aligns with Eq. (\ref{eq_b*_of_bce_os}), indicating that the minimum points of $\hat{f}^{(3)}_{\text{bce-os}}\big(\|\tilde{\bm x}\|, b^\star\big)$ are contained in $\mathbb D_1$ and when $\hat{f}^{(3)}_{\text{bce-os}}\big(\|\tilde{\bm x}\|, b^\star\big)$ reaching its minimum value on $\mathbb D_1$, $\|\tilde{\bm x}\| = 0$.

    In total, when $f_{\text{bce-os}}(\bm{W},\bm b,\mathcal F)$ approaching its minimum value, $b^\dag \rightarrow +\infty$ and $\|\tilde{\bm x}\| \rightarrow 0$, which completes the proof.
\end{proof}

\begin{lemma} \label{lem:lemmaA.1_open_set}
    For any $\bm X^{(k)}\in\mathcal D$ with any $c_1, c_2 > 0$ and $\gamma > 0$, the BCE loss $L_{\mathrm{bce}}(\bm X^{(k)})$ defined in Eq. (\ref{equ-bce-single-sample}) is lower bounded by
    \begin{align}
        {L}_{\mathrm{bce}}(\bm X^{(k)}) \geq& \frac{1}{1+c_{1}}\big(-\gamma \bm{w}_{k}^{\top}\bm x^{(k)}+b_{k}\big) + \frac{1}{1+c_{2}}\sum_{j=1 \atop j\neq k}^{K_1}\big(\gamma \bm{w}_{j}^{\top}\bm x^{(k)}-b_{j}\big)  \nonumber \\
        &\quad + \frac{1}{1+c_{3}}\sum_{j=K_1 + 1}^{K} \big(\gamma \bm{w}_{j}^{\top}\bm x^{(k)} - b_j\big) + C,
        \label{equ-Lbce_lowerBound_os}
    \end{align}
    where
    \begin{equation}
		\begin{aligned}
		 C& =\log(1+c_1)-\frac{c_1}{1+c_1}\log(c_1) + (K_1 - 1)\bigg(\log(1+c_2)-\frac{c_2}{1+c_2}\log(c_2)\bigg) \nonumber \\
		 &\quad +(K-K_1)\bigg(\log(1+c_3)-\frac{c_3}{1+c_3}\log(c_3)\bigg).
		\end{aligned} 
		\label{equ-C_formula_2_os}
\end{equation}
    The inequality becomes an equality if and only if
    \begin{align}
        & c_1=\exp\big(\gamma \bm{w}_k^\top\bm x^{(k)}-b_k\big), \quad 1\leq k\leq K_1\label{equ-c1_1_os}\\
        & c_2=\exp\big(b_j - \gamma \bm{w}_j^\top\bm x^{(k)}\big), \quad 1 \leq j\neq k\leq K_1. \label{equ-c2_1_os}\\
        & c_3=\exp\big(b_j - \gamma \bm{w}_j^\top\bm x^{(k)}\big),\quad 1 \leq k \leq K_1,~K_1 < j \leq K. \label{equ-c3_1_os}
    \end{align}
\end{lemma}

\begin{proof}
    By the concavities of $\log(x)$, {\color{black}and according} to Lemma \ref{jensen_inequality_definition}, the BCE loss $L_{\mathrm{bce}}(\bm X^{(k)})$ for any $\bm X^{(k)}\in\mathcal D$ can be lower bounded by
    \begin{align}
        {L}_{\mathrm{bce}}(\bm X^{(k)})
        =&~ \log\Big(1+\exp\big(-\gamma \bm{w}_k^\top\bm x^{(k)} +b_k\big)\Big) + \sum_{j=1 \atop j\neq k}^{K_1}\log\Big(1+\exp\big(\gamma \bm{w}_j^\top\bm x^{(k)}-b_j\big)\Big) \nonumber \\
        &\quad +\sum_{j= K_1 + 1}^K \log\Big(1+\exp\big(\gamma \bm{w}_j^\top\bm x^{(k)}-b_j\big)\Big) \nonumber\\
        =&~ \log\bigg(\frac{c_1}{1+c_1}\frac{1+c_1}{c_1}+\frac{1+c_1}{1+c_1}\exp\big(-\gamma \bm{w}_k^\top\bm x^{(k)}+b_k\big)\bigg) \nonumber \\
        &\quad + \sum_{j=1 \atop j\neq k}^{K_1} \log\bigg(\frac{c_2}{1+c_2}\frac{1+c_2}{c_2}+\frac{1+c_2}{1+c_2}\exp\big(\gamma \bm{w}_j^\top\bm x^{(k)}-b_j\big)\bigg) \nonumber \\
        &\quad + \sum_{j=K_1 + 1}^K \log\bigg(\frac{c_3}{1+c_3}\frac{1+c_3}{c_3}+\frac{1+c_3}{1+c_3}\exp\big(\gamma \bm{w}_j^\top\bm x^{(k)}-b_j\big)\bigg) \nonumber \\
        \geq&~ \frac{c_1}{1+c_1}\log\bigg(\frac{1+c_1}{c_1}\bigg) + \frac{1}{1+c_1}\log\Big[(1+c_1)\exp\big(-\gamma \bm{w}_k^\top \bm x^{(k)} +b_k\big)\Big] \nonumber \\
        &\quad+(K_1-1)\frac{c_2}{1+c_2}\log\bigg(\frac{1+c_2}{c_2}\bigg) +(K-K_1)\frac{c_3}{1+c_3}\log\bigg(\frac{1+c_3}{c_3}\bigg) \nonumber \\
        &\quad + \frac{1}{1+c_2}\sum_{j=1 \atop j\neq k}^{K_1}\log\Big[(1+c_2)\exp\big(\gamma \bm{w}_j^\top\bm x^{(k)}-b_j\big)\Big] \nonumber \\
        &\quad + \frac{1}{1+c_3}\sum_{j=K_1 + 1}^{K}\log\Big[(1 +c_3)\exp\big(\gamma \bm{w}_j^\top\bm x^{(k)}-b_j\big)\Big] \nonumber \\
        =&~ \frac{1}{1+c_1}\Big(-\gamma \bm{w}_k^\top\bm x^{(k)}+b_k\Big) + \frac1{1+c_2}\sum_{j=1 \atop j \neq k}^{K_1} \Big(\gamma \bm{w}_j^\top \bm x^{(k)}- b_j\Big) \nonumber \\
        &\quad +\frac1{1+c_3}\sum_{j=K_1 + 1}^{K} \big(\gamma \bm{w}_j^\top \bm x^{(k)}-b_j\big) + C. \label{use_C_os}
    \end{align}
    where
    \begin{align}
    	C =& 
    	\log(1+c_1)-\frac{c_1}{1+c_1}\log(c_1)
    	+(K_1-1)\bigg(\log(1+c_2)-\frac{c_2}{1+c_2}\log(c_2)\bigg) \nonumber \\
    	&\quad +(K-K_1)\bigg(\log(1+c_3)-\frac{c_3}{1+c_3}\log(c_3)\bigg). \nonumber
    \end{align}
    The inequality becomes an equality if and only if
    \begin{align}
        \frac{1+c_1}{c_1} &=(1+c_1)\exp\big(-\gamma \bm{w}_k^\top \bm x^{(k)} + b_k\big) ~\text{or}~ c_1=0 ~\text{or}~ c_1=+\infty, ~~ 1\leq k \leq K_1,  \nonumber\\
        \frac{1+c_2}{c_2} &=(1+c_2)\exp\big(\gamma \bm{w}_j^\top \bm x^{(k)}-b_j\big) ~\text{or}~ c_2=0 ~\text{or}~ c_2=+\infty,~~ 1\leq j\neq k \leq K_1,  \nonumber\\
        \frac{1+c_3}{c_3} &=(1+c_3)\exp\big(\gamma \bm{w}_j^\top \bm x^{(k)}-b_j\big) ~\text{or}~ c_3=0 ~\text{or}~ c_3=+\infty, ~~ 1\leq k \leq K_1, ~ K_1 + 1\leq j \leq K. \nonumber
    \end{align}
    These equalities are trivial when \(c_1 = 0\), \(c_1 = +\infty\), \(c_2 = 0\), \(c_2 = +\infty\), \(c_3 = 0\), or \(c_3 = +\infty\).
    Therefore, if and only if
    \begin{align}
        c_1 &=\exp\big(\gamma \bm{w}_k^\top\bm x^{(k)}-b_k\big), \quad 1\leq k \leq K_1, \nonumber\\
        c_2 &=\exp\big(b_j-\gamma \bm{w}_j^\top\bm x^{(k)}\big), \quad 1\leq j\neq k \leq K_1, \nonumber\\
        c_3 &=\exp\big(b_j-\gamma \bm{w}_j^\top\bm x^{(k)}\big), \quad 1\leq k \leq K_1,~ K_1 < j \leq K, \nonumber
    \end{align}
    Eq. (\ref{equ-Lbce_lowerBound_os}) becomes an equality, which completes the proof.
\end{proof}

\clearpage

\begin{theorem} \label{them:lemmaB.1_OS}(\textbf{CE})
    Suppose that the first $K_1$ classes in dataset $\mathcal D$ contains at least one sample, $K_1 < K$, while the other classes are empty, and that the feature dimension is greater than the number of classes, i.e., $n_k = |\mathcal D_k| > 0$ for $\forall k \in [K_1]$, $n_k = |\mathcal D_k| = 0$ for $ K_1 < k \leq K$, and $d \geq K_1$.
    Denote $\tilde{\bm x} = \frac1{K_1}\sum_{k=1}^{K_1}\bar{\bm x}_k$ with $\bar{\bm x}_k = \frac1{n_k}\sum_{\bm x\in\mathcal F_k}\bm x$.
    On the dataset $\mathcal D$, define
    \begin{align}
        f_{\mathrm{ce}\text{-}\mathrm{os}}(\bm W, \bm b, \mathcal F) \triangleq
        \frac1{K_1}\sum_{k=1}^{K_1}\frac1{n_k}\sum_{\bm X^{(k)}\in\mathcal D_k}L_{\mathrm{ce}}(\bm X^{(k)}), \label{equ-minf_object-ce_os}
    \end{align}
    where
    \begin{align}
        L_{\mathrm{ce}}(\bm X^{(k)}) = -\log\Big(\frac{\exp\big(\gamma \bm w_k^\top \bm x^{(k)} - b_k\big)}{\sum_{j=1}^K \exp\big(\gamma \bm w_j^\top \bm x^{(k)} - b_j\big)}\Big),
    \end{align}
    with $\bm x^{(k)} = \mathcal M(\bm X^{(k)})$ for $\bm X^{(k)} \in \mathcal D_k \subset \mathcal D$.

    Then, the CE loss $f_{\mathrm{ce}\text{-}\mathrm{os}}(\bm W, \bm b, \mathcal F)$ satisfies
    \begin{align}
        f_{\mathrm{ce}\text{-}\mathrm{os}}(\bm W, \bm b, \mathcal F) \geq \log(1+c_1+c_2), \label{eq_lower_bound_ce_os}
    \end{align}
    where
    \begin{align}
        c_1 &~= (K_1-1)\exp\bigg(\frac{\gamma}{\sqrt{g(q^\star)}}\frac{K_1}{K_1-1}\big(\|\tilde{\bm x}\|^2 - 1\big)\bigg)\nonumber\\
        c_2 &~= K_1\frac{1-q^\star}{q^\star}\exp\bigg(\frac{\gamma}{\sqrt{g(q^\star)}}\frac{K_1}{K_1-1}\big(\|\tilde{\bm x}\|^2 - 1\big)\bigg)\nonumber\\
        c_3 &~= \frac{1}{q^\star}\sqrt{g(q^\star)}\exp\bigg(\frac{\gamma}{\sqrt{g(q^\star)}}\frac{K_1}{K_1-1}\big(\|\tilde{\bm x}\|^2 - 1\big)\bigg), \nonumber
    \end{align}
    and $g(q) = 1-2 q \|\tilde{\bm x}\|^2 + q^2 \|\tilde{\bm x}\|^2$ with a certain $q^\star\in(0,1)$.
    When the CE loss $f_{\mathrm{ce}\text{-}\mathrm{os}}(\bm W, \bm b, \mathcal F)$ reaching its lower bound in Eq. (\ref{eq_lower_bound_ce_os}), it satisfies
    \begin{itemize}
      \item (collapse)
        \begin{align}
            &\bm x^{(k)} = \bar{\bm x}_k = \frac1{n_k}\sum_{\bm x\in\mathcal F_k} \bm x,\quad\forall \bm x^{(k)}, \bm x^{(k)}_*\in\mathcal F_k,~\forall k\in[K_1],\nonumber
        \end{align}
      \item (equiangular)
        \begin{align}
            &\bar{\bm x}_k^\top\bar{\bm x}_j = 1 - \frac{K_1}{K_1-1}\big(1 - \|\tilde{\bm x}\|^2\big) = -\frac1{K_1-1}+\frac{K_1}{K_1-1}\|\tilde{\bm x}\|^2,\quad \forall j\neq k\in[K_1],\nonumber\\
            &\bm w_k^\top \bm w_j =
        \frac{1}{K_1}\frac{1}{K_1-1}\bigg(\frac{c_2}{c_3}\bigg)^2\|\tilde{\bm x}\|^2 - \frac{1}{K_1-1},\quad \forall j\neq k\in[K_1],\nonumber
        \end{align}
      \item (isogonal)
        \begin{align}
            &\bm w_k^\top \bar{\bm x}_k =  \bigg(\frac{c_1}{K_1-1}+\frac{c_2}{K_1}\bigg)\frac{1}{c_3} - \frac{c_1}{K_1-1}\frac1{c_3}\|\tilde{\bm x}\|^2,\quad\forall k\in[K_1], \nonumber \\
            &\bm w_k^\top \bar{\bm x}_j = \frac{1}{c_3}\bigg(\frac{1}{K_1-1}\frac{c_1}{K_1-1} + \frac{K_1}{K_1-1}\frac{c_2}{K_1}\bigg)\|\tilde{\bm x}\|^2 \nonumber \\
            &\qquad\qquad -\frac{1}{c_3}\frac{1}{K_1-1}  \bigg(\frac{c_1}{K_1-1} + \frac{c_2}{K_1}\bigg),\quad\forall j\neq k\in[K_1], \nonumber \\
            &\bm w_k^\top \tilde{\bm x} = \frac{c_2}{c_3}\frac{\|\tilde{\bm x}\|^2}{K_1},\quad\forall k\in[K_1], \nonumber\\
        \end{align}
      \item (minority collapse) for $\forall j = K_1+1, K_1+2, \cdots, K$, if $\|\tilde{\bm x}\|\neq 0$,
        \begin{align}
            &\bm w_j = -\frac{\tilde{\bm x}}{\|\tilde{\bm x}\|},\quad ,\nonumber
        \end{align}
        and for $\forall k\in[K_1],~\forall j = K_1+1, K_1+2, \cdots, K$,
        \begin{align}
            &\bm w_j^\top \bar{\bm x}_k = -\|\tilde{\bm x}\|,\nonumber\\
            &\bm w_j^\top \bm w_k = -\frac{c_2}{c_3}\frac{\|\tilde{\bm x}\|}{K_1},\nonumber
        \end{align}
    \end{itemize}
    and $\exists b^\star, b^\dag$, such that
    \begin{align}
            &b_k = b^\star \in \mathbb R,\quad \forall k\in[K_1],\nonumber\\
            &b_j = b^\dagger \in \mathbb R,\quad K_1 < j \leq K.\nonumber
    \end{align}

    As the CE loss $f_{\mathrm{ce}\text{-}\mathrm{os}}(\bm W, \bm b, \mathcal F)$ approaching its minimum value, $b^\dag - b^\star \rightarrow+\infty$, $q^\star\rightarrow1$, and $\|\tilde{\bm x}\|\rightarrow0$.
\end{theorem}

\begin{proof}
    According to Lemma \ref{lem:lemmaB.1}, for any $c_1, c_2 > 0$ and any $\bm{X}^{(k)}\in\mathcal D_k$ with $\forall k\in[K]$,
    \begin{align}
        {L}_{\mathrm{ce}}(\bm X^{(k)}) \geq&~ \frac{c_1}{1+c_1+c_2} \sum_{j=1\atop j\neq k}^{K_1} \frac{\gamma\bm{w}_j^{\top}\bm{x}^{(k)}-b_j - \gamma\bm{w}_k^{\top}\bm{x}^{(k)}+b_k}{K_1-1} \nonumber \\
        &\quad + \frac{c_2}{1+c_1+c_2}\sum_{j=K_1+1}^{K} \frac{\gamma\bm{w}_j^{\top}\bm{x}^{(k)}-b_j - \gamma\bm{w}_k^{\top}\bm{x}^{(k)}+b_k}{K-K_1} + C,  \nonumber
    \end{align}
    where
    \begin{equation}
		\begin{aligned}
			& C = \frac1{1+c_1+c_2}\log\big(1+c_1+c_2\big) + \frac{c_1}{1+c_1+c_2}\log\bigg(\frac{1+c_1+c_2}{c_1}(K_1-1)\bigg) \\
			&\qquad + \frac{c_2}{1+c_1+c_2}\log\bigg(\frac{1+c_1+c_2}{c_2}(K-K_1)\bigg), \nonumber
\end{aligned} 
\end{equation}
    and the inequality becomes an equality if and only if Eqs. \ref{equ-Lce_neg_score_K1_os}, \ref{equ-Lce_neg_score_K_os}, \ref{equ-Lce_c1_os}, and \ref{equ-Lce_c2_os} hold.

    We take the same $c_1, c_2$ for all $\bm{X}^{(k)}\in\mathcal D$, then one can deduce the following lower bound for the function $f_{\text{ce-os}}$
    \begin{align}
        &f_{\text{ce-os}}(\bm{W},\bm{b}, \mathcal F) = \frac1{K_1}\sum_{k=1}^{K_1}\frac1{n_k}\sum_{\bm X^{(k)}\in\mathcal D_k}L_{\mathrm{ce}}(\bm X^{(k)}) \nonumber \\
        \geq&~ \frac1{K_1}\sum_{k=1}^{K_1}\frac1{n_k}\sum_{\bm x^{(k)}\in\mathcal F_k} \Bigg[\frac{c_1}{1+c_1+c_2} \sum_{j=1\atop j\neq k}^{K_1} \frac{\gamma\bm{w}_j^{\top}\bm{x}^{(k)}-b_j  - \gamma\bm{w}_k^{\top}\bm{x}^{(k)}+b_k}{K_1-1} \nonumber \\
        &\quad + \frac{c_2}{1+c_1+c_2}\sum_{j=K_1+1}^{K} \frac{\gamma\bm{w}_j^{\top}\bm{x}^{(k)}-b_j - \gamma\bm{w}_k^{\top}\bm{x}^{(k)}+b_k}{K-K_1}\Bigg] + C \nonumber\\
        =&~ \frac{c_1}{1+c_1+c_2} \frac1{K_1-1} \frac1{K_1} \sum_{k=1}^{K_1}\frac1{n_k}\sum_{\bm x^{(k)}\in\mathcal F_k} \sum_{j=1}^{K_1} \big(\gamma\bm{w}_j^{\top}\bm{x}^{(k)}-b_j- \gamma\bm{w}_k^{\top}\bm{x}^{(k)}+b_k\big) \nonumber \\
        &\quad +\frac{c_2}{1+c_1+c_2} \frac1{K - K_1} \frac1{K_1} \sum_{k=1}^{K_1}\frac1{n_k}\sum_{\bm x^{(k)}\in\mathcal F_k} \sum_{j=K_1+1}^{K} \big(\gamma\bm{w}_j^{\top}\bm{x}^{(k)} - b_j - \gamma\bm{w}_k^{\top}\bm{x}^{(k)}+b_k\big) + C \nonumber \\
         = &~\frac{c_1}{1+c_1+c_2} \frac{\gamma}{K_1-1}  \Bigg(\sum_{j=1}^{K_1} \bm{w}_j^{\top}\tilde{\bm{x}} - \sum_{k=1}^{K_1} \bm w_k^\top \bar{\bm x}_k\Bigg) + \frac{c_2}{1+c_1+c_2} \frac{\gamma}{K - K_1} \sum_{j=K_1+1}^{K} \bm w_j^\top\tilde{\bm x} \nonumber \\
         &\quad +\frac{c_1}{1+c_1+c_2} \frac1{K_1-1} \frac1{K_1}\Bigg(K_1\sum_{k=1}^{K_1}b_k - K_1 \sum_{j=1}^{K_1}b_j\Bigg) + C - \frac{c_2}{1+c_1+c_2} \frac{\gamma}{ K_1} \sum_{k=1}^{K_1} \bm w_k^\top\bar{\bm x}_k \nonumber \\
         &\quad + \frac{c_2}{1+c_1+c_2} \frac1{K - K_1} \frac1{K_1}\Bigg[(K  -K_1)\sum_{k=1}^{K_1}b_k - K_1\sum_{j=K_1+1}^K b_j\Bigg] \nonumber \\
         =&~\frac{\gamma}{1+c_1+c_2} \sum_{k=1}^{K_1}\bm w_k^\top\bigg[\frac{c_1}{K_1-1}\tilde{\bm x} - \bigg(\frac{c_1}{K_1-1}+\frac{c_2}{K_1}\bigg)\bar{\bm x}_k\bigg] + C \nonumber\\
         &\quad +\frac{c_2}{1+c_1+c_2} \frac{\gamma}{K - K_1} \sum_{j=K_1+1}^{K} \bm w_j^\top\tilde{\bm x} + \frac{c_2}{1+c_1+c_2} \Bigg(\frac{1}{K_1}\sum_{k=1}^{K_1}b_k - \frac1{K-K_1}\sum_{j=K_1+1}^K b_j\Bigg), \label{equ-fce_medal_lower_bound_os}
    \end{align}
    where $\tilde{\bm{x}} = \frac{1}{K}\sum_{j=1}^{K}\bar{\bm{x}}_{j}$ and $\bar{\bm{x}}_{j} = \frac1{n_j}\sum_{\bm x^{(j)}\in\mathcal F_j} \bm x^{(j)},~\forall j\in[K_1]$.

    According to the AM-GM inequality (see Lemma \ref{young_inequality_definition}), we have
    \begin{equation}
        \bm{u}^\top\bm{v}\geq - \Big(\frac{c}{2}\|\bm{u}\|_2^2+\frac{1}{2c}\|\bm{v}\|_2^2\Big), \quad\forall\bm{u}, \bm{v}\in\mathbb{R}^d, ~\forall c > 0, \nonumber
    \end{equation}
    which becomes an equality if and only if $\bm{v} = -c\bm{u}$,
    and
    \begin{align}
        \bm{u}^\top\bm{v}\geq - \|\bm u\| \|\bm v\| = -c\|\bm v\|^2, \quad\forall\bm{u}, \bm{v}\in\mathbb{R}^d, ~\forall c > 0, \nonumber
    \end{align}
    which becomes an equality if and only if $\bm{u} = -c\bm{v}$.
    Then,
    \begin{align}
    	f_{\text{ce-os}}(\bm{W},\bm{b}, \mathcal F)
        \geq &~ \frac{\gamma}{1+c_1+c_2} \sum_{k=1}^{K_1}\bm w_k^\top\bigg[\frac{c_1}{K_1-1}\tilde{\bm x} - \bigg(\frac{c_1}{K_1-1}+\frac{c_2}{K_1}\bigg)\bar{\bm x}_k\bigg] \nonumber \\
        &\quad +\frac{c_2}{1+c_1+c_2} \frac{\gamma}{K - K_1} \sum_{j=K_1+1}^{K} \bm w_j^\top\tilde{\bm x} \nonumber \\
        &\quad + \frac{c_2}{1+c_1+c_2} \Bigg(\frac{1}{K_1}\sum_{k=1}^{K_1}b_k - \frac1{K-K_1}\sum_{j=K_1+1}^K b_j\Bigg)+C \nonumber \\
        \geq&~-\frac{\gamma}{1+c_1+c_2} \sum_{k=1}^{K_1}\frac{1}{c_3}\bigg\|\frac{c_1}{K_1-1}\tilde{\bm x} - \bigg(\frac{c_1}{K_1-1} + \frac{c_2}{K_1}\bigg)\bar{\bm x}_k\bigg\|^2 \nonumber \\
        &\quad - \frac{c_2}{1+c_1+c_2} \frac{\gamma}{K - K_1} \sum_{j=K_1+1}^{K} \frac{1}{c_4}\|\tilde{\bm x}\|^2 \nonumber \\
        &\quad + \frac{c_2}{1+c_1+c_2} \Bigg(\frac{1}{K_1}\sum_{k=1}^{K_1}b_k - \frac1{K-K_1}\sum_{j=K_1+1}^K b_j\Bigg)+C \nonumber \\
        =&~-\frac{\gamma}{1+c_1+c_2} \sum_{k=1}^{K_1}\frac{1}{c_3}\Bigg[\bigg(\frac{c_1}{K_1-1}\bigg)^2\|\tilde{\bm x}\|^2 - 2\bigg(\frac{c_1}{K_1-1}\bigg)\bigg(\frac{c_1}{K_1-1}+\frac{c_2}{K_1}\bigg)\bar{\bm x}_k^\top \tilde{\bm x} \nonumber \\
        &\quad + \bigg(\frac{c_1}{K_1-1} +\frac{c_2}{K_1}\bigg)^2\|\bar{\bm x}_k\|^2\Bigg] - \frac{c_2}{1+c_1+c_2} \frac{\gamma}{K - K_1} \sum_{j=K_1+1}^{K} \frac{1}{c_4}\|\tilde{\bm x}\|^2 \nonumber \\
        &\quad + \frac{c_2}{1+c_1+c_2} \Bigg(\frac{1}{K_1}\sum_{k=1}^{K_1}b_k - \frac1{K-K_1}\sum_{j=K_1+1}^K b_j\Bigg)+C \nonumber \\
        \geq&~-\frac{\gamma}{1+c_1+c_2} \frac{1}{c_3}\Bigg[\bigg(\frac{c_1}{K_1-1}\bigg)^2 K_1 \|\tilde{\bm x}\|^2 - 2\bigg(\frac{c_1}{K_1-1}\bigg)\bigg(\frac{c_1}{K_1-1}+\frac{c_2}{K_1}\bigg) K_1 \|\tilde{\bm x}\|^2 \nonumber \\
        &\quad + \bigg(\frac{c_1}{K_1-1}+\frac{c_2}{K_1}\bigg)^2 K_1\Bigg] - \frac{c_2 \gamma}{1+c_1+c_2} \frac{1}{c_4}\|\tilde{\bm x}\|^2 \nonumber\\
        &\quad + \frac{c_2}{1+c_1+c_2} \Bigg(\frac{1}{K_1}\sum_{k=1}^{K_1}b_k - \frac1{K-K_1}\sum_{j=K_1+1}^K b_j\Bigg)+C, \nonumber
    \end{align}
    where the second inequality becomes an equality if and only if
    \begin{align}
        -c_3 \bm w_k &~= \frac{c_1}{K_1-1}\tilde{\bm x} - \bigg(\frac{c_1}{K_1-1}+\frac{c_2}{K_1}\bigg)\bar{\bm x}_k,\quad\forall c_3 >0,~\forall k\in[K_1], \label{eq_CE_wk_xk_c3}\\
        -c_4\bm w_j &~= \tilde{\bm x},\quad \forall c_4>0,~K_1 < j\leq K, \label{eq_CE_wj_xk_c4}
    \end{align}
    and the third inequality becomes an equality if and only if
    \begin{align}
        &1 = \|\bar{\bm x}_k\| = \bigg\|\frac1{n_k}\sum_{\bm x^{(k)}\in\mathcal F_k}\bm x^{(k)}\bigg\| \leq \frac1{n_k}\sum_{\bm x^{(k)}\in\mathcal F_k}\|\bm x^{(k)}\| = 1 = \|\bm x^{(k)}\| \nonumber\\
        &\quad \Leftrightarrow\quad \bm x^{(k)} = \bar{\bm x}_k,\quad\forall \bm x^{(k)}\in\mathcal F_k,~\forall k\in[K_1].\label{eq_ce_os_xk=xk}
    \end{align}

    As $\|\bm w_j\| = 1$, from Eq. (\ref{eq_CE_wj_xk_c4}), one can deduce
    \begin{align}
        c_4 = \|\tilde{\bm x}\|. \label{eq_c4_ce_os}
    \end{align}
    Then,
    \begin{align}
    	&~f_{\text{ce-os}}(\bm{W},\bm{b}, \mathcal F) \nonumber \\
        \geq&~ -\frac{\gamma}{1+c_1+c_2} \frac{1}{c_3}\Bigg[\bigg(\frac{c_1}{K_1-1}\bigg)^2 K_1 \|\tilde{\bm x}\|^2 - 2\bigg(\frac{c_1}{K_1-1}\bigg)\bigg(\frac{c_1}{K_1-1} +\frac{c_2}{K_1}\bigg) K_1 \|\tilde{\bm x}\|^2 \nonumber \\
        &\quad + \bigg(\frac{c_1}{K_1-1}+\frac{c_2}{K_1}\bigg)^2 K_1\Bigg] - \frac{c_2 \gamma}{1+c_1+c_2} \|\tilde{\bm x}\| \nonumber \\
        &\quad + \frac{c_2}{1+c_1+c_2} \Bigg(\frac{1}{K_1}\sum_{k=1}^{K_1}b_k - \frac1{K-K_1}\sum_{j=K_1+1}^K b_j\Bigg)+C \nonumber \\
        =&~\frac{\gamma}{1+c_1+c_2} \Bigg\{\bigg[\frac{K_1}{c_3}\bigg(\frac{c_1}{K_1-1}\bigg)^2 + \frac{2K_1}{c_3}\frac{c_1}{K_1-1}\frac{c_2}{K_1}\bigg] \|\tilde{\bm x}\|^2 - c_2 \|\tilde{\bm x}\|\Bigg\} + C \nonumber \\
        &\quad -\frac{\gamma}{1+c_1+c_2} \frac{K_1}{c_3}\bigg(\frac{c_1}{K_1-1} + \frac{c_2}{K_1}\bigg)^2 + \frac{c_2}{1+c_1+c_2} \Bigg(\frac{1}{K_1}\sum_{k=1}^{K_1}b_k - \frac1{K-K_1}\sum_{j=K_1+1}^K b_j\Bigg). \nonumber
    \end{align}
    In summary, $f_{\text{ce-os}}(\bm{W},\bm{b}, \mathcal F)$ achieves the lower bound
    if and only if Eqs. (\ref{equ-Lce_neg_score_K1_os}, \ref{equ-Lce_neg_score_K_os}, \ref{equ-Lce_c1_os}, \ref{equ-Lce_c2_os}) and Eqs. (\ref{eq_CE_wk_xk_c3}, \ref{eq_CE_wj_xk_c4}, \ref{eq_ce_os_xk=xk}) hold.

    Finally, Lemma \ref{lem:lemmaB.2_open_set} completes the proof.
\end{proof}

\begin{lemma} \label{lem:lemmaB.2_open_set}
    Suppose that the first $K_1$ classes in dataset $\mathcal D$ contains at least one sample, $K_1 < K$, while the other classes are empty, and that the feature dimension is greater than the number of classes, i.e., $n_k = |\mathcal D_k| > 0$ for $\forall k \in [K_1]$, $n_k = |\mathcal D_k| = 0$ for $ K_1 < k \leq K$, and $d \geq K_1$.
    Then, for the CE loss $f_{\mathrm{ce}\text{-}\mathrm{os}}(\bm W,\bm{b},\mathcal F)$ defined in Eq. (\ref{equ-minf_object-ce_os}),
    \begin{align}
        f_{\mathrm{ce}}(\bm W,\bm{b},\mathcal F) \geq
        &~\log\big(1+c_1+c_2\big),
    \end{align}
    where
    \begin{align}
        c_1 &~= (K_1-1)\exp\bigg(\frac{\gamma}{\sqrt{g(q^\star)}}\frac{K_1}{K_1-1}\big(\|\tilde{\bm x}\|^2 - 1\big)\bigg)\nonumber\\
        c_2 &~= K_1\frac{1-q^\star}{q^\star}\exp\bigg(\frac{\gamma}{\sqrt{g(q^\star)}}\frac{K_1}{K_1-1}\big(\|\tilde{\bm x}\|^2 - 1\big)\bigg)\nonumber\\
        c_3 &~= \frac{1}{q^\star}\sqrt{g(q^\star)}\exp\bigg(\frac{\gamma}{\sqrt{g(q^\star)}}\frac{K_1}{K_1-1}\big(\|\tilde{\bm x}\|^2 - 1\big)\bigg), \nonumber
    \end{align}
    and $g(q) = 1-2 q \|\tilde{\bm x}\|^2 + q^2 \|\tilde{\bm x}\|^2$ with a certain $q^\star\in(0,1)$.

    When $f_{\mathrm{ce}\text{-}\mathrm{os}}(\bm W,\bm{b},\mathcal F)$ approaching its minimum value, $q^\star\rightarrow1$ and $\|\tilde{\bm x}\|\rightarrow0$.
\end{lemma}

\begin{proof}
    According to Theorem \ref{them:lemmaB.1_OS}, the CE loss $f_{\mathrm{ce}}(\bm W,\bm{b},\mathcal F)$ satisfies
    \begin{align}
    	f_{\text{ce-os}}(\bm{W},\bm b,\mathcal F) \geq& -\frac{\gamma}{1+c_1+c_2}\frac{c_2}{2}\|\tilde{\bm x}\|  -\frac{\gamma}{1+c_1+c_2} \frac{K_1}{c_3}\bigg(\frac{c_1}{K_1-1} + \frac{c_2}{K_1}\bigg)^2 \nonumber \\
    	&~ + \frac{c_2}{1+c_1+c_2} \Bigg(\frac{1}{K_1}\sum_{k=1}^{K_1}b_k - \frac1{K-K_1}\sum_{j=K_1+1}^K b_j\Bigg) + C, \label{eq_ce_lower_boundary_os_lem}
    \end{align}
    and the loss achieves this lower boundary if and only if Eqs. (\ref{equ-Lce_neg_score_K1_os}, \ref{equ-Lce_neg_score_K_os}, \ref{equ-Lce_c1_os}, \ref{equ-Lce_c2_os}) and Eqs. (\ref{eq_CE_wk_xk_c3}, \ref{eq_CE_wj_xk_c4}, \ref{eq_ce_os_xk=xk}) hold.

    From Eq. (\ref{eq_CE_wk_xk_c3}),
    \begin{align}
        &~-c_3 \bm w_k= \frac{c_1}{K_1-1}\tilde{\bm x} - \bigg(\frac{c_1}{K_1-1}+\frac{c_2}{K_1}\bigg)\bar{\bm x}_k \nonumber \\
        \Rightarrow\quad&~ -c_3 \sum_{k=1}^{K_1}\bm w_k = \frac{c_1}{K_1-1}K_1\tilde{\bm x} - \bigg(\frac{c_1}{K_1-1}+\frac{c_2}{K_1}\bigg)K_1\tilde{\bm x} \nonumber \\
        \Rightarrow\quad&~ -c_3 \sum_{k=1}^{K_1}\bm w_k = -\frac{c_2}{K_1} K_1 \tilde{\bm x}\quad\Rightarrow\quad \sum_{k=1}^{K_1}\bm w_k = \frac{c_2}{c_3} \tilde{\bm x}, \label{eq_sum_of_wk_ce_os} \\
        &~-c_3 \bm w_k= \frac{c_1}{K_1-1}\tilde{\bm x} - \bigg(\frac{c_1}{K_1-1}+\frac{c_2}{K_1}\bigg)\bar{\bm x}_k \nonumber \\
        \Rightarrow\quad&~ -c_3 = -c_3 \bm w_k^\top\bm w_k = \frac{c_1}{K_1-1}\bm w_k^\top\tilde{\bm x} - \bigg(\frac{c_1}{K_1-1}+\frac{c_2}{K_1}\bigg)\bm w_k^\top \bar{\bm x}_k \nonumber \\
        \Rightarrow\quad
        &~-K_1c_3 = \frac{c_1}{K_1-1}\bigg(\sum_{k=1}^{K_1}\bm w_k\bigg)^\top\tilde{\bm x} - \bigg(\frac{c_1}{K_1-1} +\frac{c_2}{K_1}\bigg)\sum_{k=1}^{K_1}\bm w_k^\top \bar{\bm x}_k \nonumber \\
        \Rightarrow\quad&~ \sum_{k=1}^{K_1}\bm w_k^\top \bar{\bm x}_k \overset{\ref{eq_sum_of_wk_ce_os}}{=}\bigg(\frac{c_1}{K_1-1} +\frac{c_2}{K_1}\bigg)^{-1} \Bigg(K_1c_3 + \frac{c_1}{K_1-1}\frac{c_2}{c_3} \|\tilde{\bm x}\|^2\Bigg), \label{eq_sum_of_wk_xk_ce_os}\\
        ~
        &~-c_3 \bm w_k= \frac{c_1}{K_1-1}\tilde{\bm x} - \bigg(\frac{c_1}{K_1-1}+\frac{c_2}{K_1}\bigg)\bar{\bm x}_k \nonumber \\
        \Rightarrow\quad&~ -c_3 \bm w_k^\top\bar{\bm x}_k = \frac{c_1}{K_1-1}\tilde{\bm x}^\top\bar{\bm x}_k - \bigg(\frac{c_1}{K_1-1}+\frac{c_2}{K_1}\bigg)\bar{\bm x}_k^\top\bar{\bm x}_k \nonumber \\
        \Rightarrow\quad&~ \sum_{k=1}^{K_1}\bm w_k^\top\bar{\bm x}_k = \bigg(\frac{c_1}{K_1-1}+\frac{c_2}{K_1}\bigg)\frac{K_1}{c_3} - \frac{c_1}{K_1-1}\frac{K_1}{c_3}\|\tilde{\bm x}\|^2\nonumber\\
        \Rightarrow\quad
        &~ c_3^2\overset{\ref{eq_sum_of_wk_xk_ce_os}}{=}\bigg(\frac{c_1}{K_1-1}+\frac{c_2}{K_1}\bigg)^2 - \frac{c_1}{K_1-1}\bigg(\frac{c_1}{K_1-1}+2\frac{c_2}{K_1}\bigg)\|\tilde{\bm x}\|^2, \label{eq_c3_c2_c1_ce_os_2} \\
        &~-c_3 \bm w_k= \frac{c_1}{K_1-1}\tilde{\bm x} - \bigg(\frac{c_1}{K_1-1}+\frac{c_2}{K_1}\bigg)\bar{\bm x}_k \nonumber \\
        \Rightarrow\quad&~
        \bigg(\frac{c_1}{K_1-1}\bigg)^2\|\tilde{\bm x}\|^2 = \bigg\|\bigg(\frac{c_1}{K_1-1}+\frac{c_2}{K_1}\bigg)\bar{\bm x}_k-c_3 \bm w_k\bigg\|^2 \nonumber \\
        \Rightarrow\quad
        &~2\bigg(\frac{c_1}{K_1-1}+\frac{c_2}{K_1}\bigg)c_3 \bm w_k^\top \bar{\bm x}_k = \bigg(\frac{c_1}{K_1-1}+\frac{c_2}{K_1}\bigg)^2 + c_3^2 - \bigg(\frac{c_1}{K_1-1}\bigg)^2\|\tilde{\bm x}\|^2 \nonumber \\
        &~\qquad \overset{\ref{eq_c3_c2_c1_ce_os_2}}{=} 2\bigg(\frac{c_1}{K_1-1}+\frac{c_2}{K_1}\bigg)^2 -  2\bigg(\frac{c_1}{K_1-1}+\frac{c_2}{K_1}\bigg)\frac{c_1}{K_1-1}\|\tilde{\bm x}\|^2\nonumber\\
        \Rightarrow\quad&~\bm w_k^\top \bar{\bm x}_k = \bigg(\frac{c_1}{K_1-1}+\frac{c_2}{K_1}\bigg)\frac{1}{c_3} - \frac{c_1}{K_1-1}\frac1{c_3}\|\tilde{\bm x}\|^2, \quad\forall k\in[K_1]. \label{eq_wk_xk_ce_os}
    \end{align}

    Furthermore,
    \begin{align}
        &~-c_3 \bm w_k= \frac{c_1}{K_1-1}\tilde{\bm x} - \bigg(\frac{c_1}{K_1-1}+\frac{c_2}{K_1}\bigg)\bar{\bm x}_k \nonumber \\
        \Rightarrow\quad&~ c_3\big(\bm w_k - \bm w_j\big) = \bigg(\frac{c_1}{K_1-1}+\frac{c_2}{K_1}\bigg)\big(\bar{\bm x}_k - \bar{\bm x}_j\big) \nonumber \\
        \Rightarrow\quad
        &~c_3\big(\bm w_k^\top\bm w_j - 1\big) = \bigg(\frac{c_1}{K_1-1}+\frac{c_2}{K_1}\bigg)\big(\bm w_j^\top\bar{\bm x}_k - \bm w_j^\top\bar{\bm x}_j\big) \nonumber \\
        &\qquad \text{and}~~c_3\big(1 - \bm w_k^\top\bm w_j\big) = \bigg(\frac{c_1}{K_1-1}+\frac{c_2}{K_1}\bigg)\big(\bm w_k^\top\bar{\bm x}_k - \bm w_k^\top\bar{\bm x}_j\big) \nonumber \\
        \Rightarrow\quad
        &~2\gamma c_3\big(\bm w_k^\top\bm w_j - 1\big) = \bigg(\frac{c_1}{K_1-1}+\frac{c_2}{K_1}\bigg)\big(\gamma\bm w_j^\top\bar{\bm x}_k -b_j - \gamma\bm w_k^\top\bar{\bm x}_k +b_k + \nonumber \\
        &\qquad \gamma\bm w_k^\top\bar{\bm x}_j -b_k - \gamma\bm w_j^\top\bar{\bm x}_j + b_j\big) \nonumber \\
        \Rightarrow\quad
        &~\bm w_k^\top\bm w_j \overset{\ref{equ-Lce_c1_os},\ref{eq_ce_os_xk=xk}}{=} \frac{1}{\gamma c_3}\bigg(\frac{c_1}{K_1-1}+\frac{c_2}{K_1}\bigg)\log\bigg(\frac{c_1}{K_1-1}\bigg),\quad \forall j\neq k\in[K_1], \label{eq_wk_wj_ce_os_1}\\
        \Rightarrow\quad
        &~\bm w_k^\top\bm w_j \overset{\ref{eq_sum_of_wk_ce_os}}{=}
        \frac{1}{K_1}\frac{1}{K_1-1}\bigg(\frac{c_2}{c_3}\bigg)^2\|\tilde{\bm x}\|^2 - \frac{1}{K_1-1},\quad \forall j\neq k\in[K_1], \label{eq_wk_wj_ce_os_2} \\
        &~-c_3 \bm w_k= \frac{c_1}{K_1-1}\tilde{\bm x} - \bigg(\frac{c_1}{K_1-1}+\frac{c_2}{K_1}\bigg)\bar{\bm x}_k \nonumber \\
        \Rightarrow\quad&~ \bigg(\frac{c_1}{K_1-1}+\frac{c_2}{K_1}\bigg)\big(\bar{\bm x}_k - \bar{\bm x}_j\big) = c_3\big(\bm w_k - \bm w_j\big) \nonumber \\
        \Rightarrow\quad
        &~\bigg(\frac{c_1}{K_1-1}+\frac{c_2}{K_1}\bigg)^2\big(2-2\bar{\bm x}_k^\top\bar{\bm x}_j\big) = c_3^2\big(2-2\bm w_k^\top\bm w_j\big) \nonumber \\
        \Rightarrow\quad&~
        \bar{\bm x}_k^\top\bar{\bm x}_j \overset{\ref{eq_wk_wj_ce_os_2}}{=} 1 - \bigg(\frac{c_1}{K_1-1}+\frac{c_2}{K_1}\bigg)^{-2}c_3^2\frac{K_1}{K_1-1} + \frac{1}{K_1}\frac{1}{K_1-1}c_2^2\|\tilde{\bm x}\|^2\nonumber\\
        \Rightarrow\quad
        &~\bar{\bm x}_k^\top\bar{\bm x}_j = \frac{K_1}{K_1-1}\|\tilde{\bm x}\|^2 - \frac{1}{K_1-1},\quad\forall j\neq k\in[K_1], \label{eq_xk_xj_ce_os} \\
        &~-c_3 \bm w_k = \frac{c_1}{K_1-1}\tilde{\bm x} - \bigg(\frac{c_1}{K_1-1}+\frac{c_2}{K_1}\bigg)\bar{\bm x}_k \nonumber \\
        \Rightarrow\quad&~ -c_3 \bm w_k^\top\bar{\bm x}_j = \frac{c_1}{K_1-1}\bigg(\frac{1}{K_1}\sum_{k=1}^{K_1}\bar{\bm x}_k\bigg)^\top\bar{\bm x}_j - \bigg(\frac{c_1}{K_1-1}+\frac{c_2}{K_1}\bigg)\bar{\bm x}_k^\top\bar{\bm x}_j \nonumber \\
        \Rightarrow\quad
        &~ \bm w_k^\top\bar{\bm x}_j = \frac{1}{c_3}\bigg(\frac{1}{K_1-1}\frac{c_1}{K_1-1} + \frac{K_1}{K_1-1}\frac{c_2}{K_1}\bigg)\|\tilde{\bm x}\|^2 \nonumber \\
        &\quad -\frac{1}{c_3}\frac{1}{K_1-1}  \bigg(\frac{c_1}{K_1-1} + \frac{c_2}{K_1}\bigg),\quad\forall j\neq k\in[K_1]. \label{eq_wk_xj_ce_os}
    \end{align}

    According to Eqs. (\ref{eq_CE_wj_xk_c4},\ref{eq_c4_ce_os}),
    \begin{align}
        \bm w_j^\top \bar{\bm x}_k = -\|\tilde{\bm x}\|, \quad\forall k\in[K_1],~ K_1< j \leq K; \label{eq_wj_xk_ce_os}
    \end{align}
    according to Eqs. (\ref{equ-Lce_neg_score_K1_os}, \ref{eq_wk_xj_ce_os}), we denote
    \begin{align}
        b^\star = b_k = b_j,\quad\forall j\neq k\in[K_1]; \label{eq_bk=bj_ce_os}
    \end{align}
    according to Eqs. (\ref{equ-Lce_neg_score_K_os}, \ref{eq_wj_xk_ce_os}), we denote
    \begin{align}
        b^\dag = b_j = b_\ell, \quad \forall K_1< j,~\ell \leq K; \label{eq_bj=bl_ce_os}
    \end{align}
    according to Eqs. (\ref{eq_wk_xj_ce_os}, \ref{eq_xk_xj_ce_os}),
    \begin{align}
        \log\bigg(\frac{c_1}{K_1-1}\bigg) = \frac{\gamma}{c_3}\frac{K_1}{K_1-1}\bigg(\frac{c_1}{K_1-1} + \frac{c_2}{K_1}\bigg)\big(\|\tilde{\bm x}\|^2 - 1\big); \label{eq_c1_of_x_c3_c2_ce_os}
    \end{align}
    according to Eqs. (\ref{eq_wj_xk_ce_os},\ref{eq_wk_xk_ce_os}),
    \begin{align}
        \log\bigg(\frac{c_2}{K-K_1}\bigg) = -\gamma\|\tilde{\bm x}\| - \bigg(\frac{c_1}{K_1-1}+\frac{c_2}{K_1}\bigg)\frac{\gamma}{c_3} + \frac{c_1}{K_1-1}\frac{\gamma}{c_3}\|\tilde{\bm x}\|^2 + b^\star - b^\dag. \label{eq_c2_of_x_c3_c1_ce_os}
    \end{align}

    In total, when the CE loss $f_{\text{ce-os}}(\bm W, \bm b, \mathcal F)$ reaching its lower bound in Eq. (\ref{eq_ce_lower_boundary_os_lem}), $c_1,c_2,c_3$ satisfy Eqs. (\ref{eq_c3_c2_c1_ce_os_2}, \ref{eq_c1_of_x_c3_c2_ce_os}, \ref{eq_c2_of_x_c3_c1_ce_os}).
    According to Lemma \ref{lem:lemmaB.3_os}, for any given $\gamma > 0$, $K>K_1>1$, $b^\star, b^\dag\in\mathbb R$, and $\|\tilde{\bm x}\|\in(0,1)$, there exists a unique set of $c_1,c_2,c_3$. Then, they can be expressed as functions of $x = \|\tilde{\bm x}\|, \delta_b=b^\star-b^\dag$ and $\gamma, K, K_1$.
    Therefore,
    \begin{align}
    	&~f_{\text{ce-os}}(\bm{W},\bm{b}, \mathcal F) \nonumber \\
        \geq &~\frac{\gamma}{1+c_1+c_2} \Bigg\{\bigg[\frac{K_1}{c_3}\bigg(\frac{c_1}{K_1-1}\bigg)^2 + \frac{2K_1}{c_3}\frac{c_1}{K_1-1}\frac{c_2}{K_1}\bigg] \|\tilde{\bm x}\|^2 - c_2 \|\tilde{\bm x}\|\Bigg\} + C \nonumber \\
        &\quad -\frac{\gamma}{1+c_1+c_2} \frac{K_1}{c_3}\bigg(\frac{c_1}{K_1-1} + \frac{c_2}{K_1}\bigg)^2 + \frac{c_2}{1+c_1+c_2} \Bigg(\frac{1}{K_1}\sum_{k=1}^{K_1}b_k - \frac1{K-K_1}\sum_{j=K_1+1}^K b_j\Bigg) \nonumber\\
        =&~\frac{\gamma}{1+c_1+c_2} \Bigg\{\bigg[\frac{K_1}{c_3}\bigg(\frac{c_1}{K_1-1}\bigg)^2 + \frac{2K_1}{c_3}\frac{c_1}{K_1-1}\frac{c_2}{K_1}\bigg] \|\tilde{\bm x}\|^2 - c_2 \|\tilde{\bm x}\| \Bigg\}   \nonumber\\
        &\quad -\frac{\gamma}{1+c_1+c_2} \frac{K_1}{c_3}\bigg(\frac{c_1}{K_1-1} + \frac{c_2}{K_1}\bigg)^2 + \frac{c_2}{1+c_1+c_2} \big(b^\star - b^\dag\big) \nonumber \\
        &\quad +\frac1{1+c_1+c_2}\log\big(1+c_1+c_2\big) + \frac{c_1}{1+c_1+c_2}\log\bigg(\frac{1+c_1+c_2}{c_1}(K_1-1)\bigg) \nonumber\\
        &\quad + \frac{c_2}{1+c_1+c_2}\log\bigg(\frac{1+c_1+c_2}{c_2}(K-K_1)\bigg) \nonumber \\
        =&~\frac{\gamma}{1+c_1+c_2} \Bigg\{\bigg[\frac{K_1}{c_3}\bigg(\frac{c_1}{K_1-1}\bigg)^2 + \frac{2K_1}{c_3}\frac{c_1}{K_1-1}\frac{c_2}{K_1}\bigg] \|\tilde{\bm x}\|^2 - c_2 \|\tilde{\bm x}\| \Bigg\}   \nonumber \\
        &\quad -\frac{\gamma}{1+c_1+c_2} \frac{K_1}{c_3}\bigg(\frac{c_1}{K_1-1} + \frac{c_2}{K_1}\bigg)^2 + \frac{c_2}{1+c_1+c_2} \big(b^\star - b^\dag\big)\nonumber\\
        &\quad +\log\big(1+c_1+c_2\big) - \frac{c_1}{1+c_1+c_2}\log\bigg(\frac{c_1}{K_1-1}\bigg) - \frac{c_2}{1+c_1+c_2}\log\bigg(\frac{c_2}{K-K_1}\bigg) \nonumber \\
        =&~\log\big(1+c_1+c_2\big) = \log\Big(1+c_1\big(x,\delta_b;\gamma, K, K_1\big)+c_2\big(x,\delta_b;\gamma, K, K_1\big)\Big) \nonumber \\
        \triangleq&~\hat{f}^{(1)}_{\text{ce-os}}(x,\delta_b;\gamma,K,K_1). \nonumber
    \end{align}

    According to Lemma \ref{lem:lemmaB.4_os}, $\frac{\partial\hat{f}^{(1)}_{\text{ce-os}}(x,\delta_b;\gamma,K,K_1)}{\partial \delta_b} > 0$, therefore, as $\hat{f}^{(1)}_{\text{ce-os}}(x,\delta_b;\gamma,K,K_1)$ approaches its minimum,
    \begin{align}
        \delta_b = b^\star - b^\dag \rightarrow -\infty.\nonumber\
    \end{align}
    According to Eq. (\ref{eq_value_of_q_ce_os}), $q^\star \rightarrow 1$, as $\delta_b \rightarrow -\infty$, and then according to Eq. (\ref{eq_value_c1_ce_os},\ref{eq_value_c2_ce_os},\ref{eq_value_c3_ce_os}),
    \begin{align}
        c_1 &~\rightarrow (K_1 - 1) \exp\Big(-\gamma\frac{K_1}{K_1-1}\sqrt{1-\|\tilde{\bm x}\|^2}\Big), \nonumber\\
        c_2 &~\rightarrow 0,\nonumber\\
        c_3 &~\rightarrow \sqrt{1-\|\tilde{\bm x}\|^2} \exp\Big(-\gamma\frac{K_1}{K_1-1}\sqrt{1-\|\tilde{\bm x}\|^2}\Big). \nonumber
    \end{align}
    Therefore,
    \begin{align}
        f_{\text{ce-os}}(\bm{W},\bm{b}, \mathcal F) &~\geq \hat{f}^{(1)}_{\text{ce-os}}(x,\delta_b;\gamma,K,K_1) = \log\big(1+c_1+c_2\big) \nonumber\\
        &~\geq \log\big(1+c_1\big) = \log\Big(1+(K_1 - 1) \exp\Big(-\gamma\frac{K_1}{K_1-1}\sqrt{1-\|\tilde{\bm x}\|^2}\Big)\Big) \nonumber\\
        \triangleq&~\hat{f}^{(2)}_{\text{ce-os}}(\|\tilde{\bm x}\|;\gamma,K_1).\nonumber
    \end{align}
    One can easily deduce that $\frac{\partial\hat{f}^{(2)}_{\text{ce-os}}(\|\tilde{\bm x}\|;\gamma,K_1)}{\partial\|\tilde{\bm x}\|} > 0$, then when $\hat{f}^{(2)}_{\text{ce-os}}(\|\tilde{\bm x}\|;\gamma,K_1)$ approaching its minimum value, $\|\tilde{\bm x}\| \rightarrow 0$.
\end{proof}

\begin{lemma} \label{lem:lemmaB.4_os}
    Suppose $\gamma > 0$, $K>K_1>1$, $\delta_b\in\mathbb R$, $x\in(0,1)$, and
    \begin{align}
        &~c_3^2 = \bigg(\frac{c_1}{K_1-1}+\frac{c_2}{K_1}\bigg)^2 - \Bigg[\bigg(\frac{c_1}{K_1-1}+\frac{c_2}{K_1}\bigg)\frac{c_1}{K_1-1} + \frac{c_1}{K_1-1}\frac{c_2}{K_1}\Bigg] x^2 \nonumber \\
        &~\log\bigg(\frac{c_1}{K_1-1}\bigg) = \frac{\gamma}{c_3}\frac{K_1}{K_1-1}\bigg(\frac{c_1}{K_1-1} + \frac{c_2}{K_1}\bigg)\big(x^2 - 1\big) \nonumber\\
        &~\log\bigg(\frac{c_2}{K-K_1}\bigg) = -\gamma x - \bigg(\frac{c_1}{K_1-1}+\frac{c_2}{K_1}\bigg)\frac{\gamma}{c_3} + \frac{c_1}{K_1-1}\frac{\gamma}{c_3} x^2 + \delta_b.\nonumber
    \end{align}
    Then, for the function $\hat{f}^{(1)}_{\text{ce-os}}(x,\delta_b;\gamma,K,K_1) = \log(1+c_1+c_2)$,
    \begin{align}
        \frac{\partial\hat{f}^{(1)}_{\text{ce-os}}(x,\delta_b;\gamma,K,K_1)}{\partial \delta_b} = \frac{1}{1+c_1+c_2}\bigg(\frac{\partial c_1}{\partial \delta_b} + \frac{\partial c_2}{\partial \delta_b}\bigg) > 0. \nonumber
    \end{align}
\end{lemma}
\begin{proof}
    Denote $u = \frac{c_1}{K_1-1}, v = \frac{c_2}{K_1}$ and $s = u+v, q= \frac{u}{u+v}\in(0,1)$.
    According to Lemma \ref{lem:lemmaB.3_os},
    \begin{align}
        &~c_3 = s\sqrt{g(q,x)},\nonumber\\
        &~c_1 = (K_1-1)\exp\big(-\phi(q,x)\big) > 0,\nonumber\\
        &~c_2 = K_1 \frac{1-q}{q} \exp\big(-\phi(q,x)\big) > 0,\nonumber
    \end{align}
    where
    \begin{align}
        &~ g(q,x) = 1 - 2x^2 q + x^2q^2, \nonumber\\
        &~ \phi(q,x) = \gamma \frac{K_1}{K_1-1} \frac{1-x^2}{\sqrt{g(q,x)}}.\nonumber
    \end{align}
    Let
    \begin{align}
    	\delta_b &~= \log\bigg(\frac{c_2}{K-K_1}\bigg) + \gamma x + \bigg(\frac{c_1}{K_1-1}+\frac{c_2}{K_1}\bigg)\frac{\gamma}{c_3} - \frac{c_1}{K_1-1}\frac{\gamma}{c_3} x^2 \nonumber \\
    	&~ = \log\bigg(\frac{c_2}{K_1}\bigg) + \log\bigg(\frac{K_1}{K-K_1}\bigg) + \gamma x + s \frac{\gamma}{s\sqrt{g(q,x)}} - u \frac{\gamma}{s\sqrt{g(q,x)}} x^2 \nonumber \\
        &~=\log\bigg(\frac{1-q}{q}\bigg) + \log\bigg(\frac{K_1}{K-K_1}\bigg) + \gamma x - \gamma T(q,x) \nonumber \\
        &~\triangleq B(q,x), \nonumber
    \end{align}
    where
    \begin{align}
        T(q,x) = \frac{1}{\sqrt{g(q,x)}}\bigg[\frac{K_1}{K_1-1}(1-x^2) -1 + q x^2\bigg]. \nonumber
    \end{align}
    According to Lemma \ref{lem:lemmaB.3_os}, given $\gamma, K, K_1$, $q$ is a function in terms of $x$ and $\delta_b$. Denote it as $q(x, \delta_b)$. Then,
    \begin{align}
        \frac{\partial B(q,x)}{\partial q} \frac{\partial q(x, \delta_b)}{\partial\delta_b} = 1. \nonumber
    \end{align}

    One can deduce that
    \begin{align}
        \frac{\partial B(q,x)}{\partial q} = -\frac{1}{1-q}-\frac{1}{q} -\gamma \frac{\partial T(q,x)}{\partial q} = -\frac{1}{1-q}-\frac{1}{q} -\gamma \frac{x^2(1-x^2)(K_1-q)}{(K_1-1)\big(g(q,x)^{3/2}\big)} <0; \nonumber
    \end{align}
    therefore, $\frac{\partial q(x, \delta_b)}{\partial\delta_b} < 0$.
    As
    \begin{align}
        \frac{\partial\phi(q,x)}{\partial q} = \frac{1}{\big(g(q,x)\big)^{3/2}}\bigg(\gamma\frac{K_1}{K_1-1}(1-x^2)x^2(1-q)\bigg) > 0, \nonumber
    \end{align}
    and
    \begin{align}
        c_1 = (K_1-1)\e^{-\phi(q,x)} &\quad\Rightarrow\quad \log(c_1) = \log(K_1-1) - \phi(q,x) \quad \nonumber\\
        &\quad \Rightarrow\quad \frac{\partial c_1}{\partial\delta_b} \frac{1}{c_1} = -\frac{\partial\phi(q,x)}{\partial q}\frac{\partial q}{\partial\delta_b} \quad \nonumber\\
        &\quad \Rightarrow\quad
         \frac{\partial c_1}{\partial\delta_b} = - c_1 \frac{\partial\phi(q,x)}{\partial q}\frac{\partial q}{\partial\delta_b} > 0, \nonumber \\
        c_2 = K_1 \frac{1-q}{q} \e^{-\phi(q,x)}) &\quad\Rightarrow\quad \frac{\partial c_2}{\partial\delta_b} \frac{1}{c_2} = \bigg(-\frac{1}{1-q} -\frac{1}{q} - \frac{\partial\phi(q,x)}{\partial q} \bigg)\frac{\partial q}{\partial\delta_b} \nonumber \\
        &\quad\Rightarrow\quad \frac{\partial c_2}{\partial\delta_b} = -c_2\bigg(\frac{1}{1-q} + \frac{1}{q} + \frac{\partial\phi(q,x)}{\partial q} \bigg)\frac{\partial q}{\partial\delta_b} > 0, \nonumber
    \end{align}
    one can deduce that
    \begin{align}
        \frac{\partial\hat{f}^{(1)}_{\text{ce-os}}(x,\delta_b;\gamma,K,K_1)}{\partial \delta_b} = \frac{1}{1+c_1+c_2}\bigg(\frac{\partial c_1}{\partial \delta_b} + \frac{\partial c_2}{\partial \delta_b}\bigg) > 0, \nonumber
    \end{align}
    which completes the proof.
\end{proof}

\begin{lemma} \label{lem:lemmaB.3_os}
    Given $\gamma > 0$, $K>K_1>1$, $\delta_b\in\mathbb R$, and $x\in(0,1)$, Eqs. (\ref{eq_c3_c2_c1_ce_os_2}, \ref{eq_c1_of_x_c3_c2_ce_os}, \ref{eq_c2_of_x_c3_c1_ce_os}), i.e.,
    \begin{align}
        &~c_3^2 = \bigg(\frac{c_1}{K_1-1}+\frac{c_2}{K_1}\bigg)^2 - \Bigg[\bigg(\frac{c_1}{K_1-1}+\frac{c_2}{K_1}\bigg)\frac{c_1}{K_1-1} + \frac{c_1}{K_1-1}\frac{c_2}{K_1}\Bigg] x^2 \nonumber \\
        &~\log\bigg(\frac{c_1}{K_1-1}\bigg) = \frac{\gamma}{c_3}\frac{K_1}{K_1-1}\bigg(\frac{c_1}{K_1-1} + \frac{c_2}{K_1}\bigg)\big(x^2 - 1\big) \nonumber\\
        &~\log\bigg(\frac{c_2}{K-K_1}\bigg) = -\gamma x - \bigg(\frac{c_1}{K_1-1}+\frac{c_2}{K_1}\bigg)\frac{\gamma}{c_3} + \frac{c_1}{K_1-1}\frac{\gamma}{c_3} x^2 + \delta_b,\nonumber
    \end{align}
    determine a unique set of values for $c_1, c_2, c_3$.
\end{lemma}

\begin{proof}
    Denote
    \begin{align}
        u = \frac{c_1}{K_1-1}, \quad v = \frac{c_2}{K_1},  \nonumber
    \end{align}
    then $u>0, v>0, \delta_b\in\mathbb R, x\in(0,1)$ and the set of equations evolves into
    \begin{align}
        &~c_3^2 = \big(u + v\big)^2 - \big(u^2 + 2uv\big) x^2, \label{eq_c3_u_v_ce_os} \\
        &~\log\big(u\big) = \frac{\gamma}{c_3}\frac{K_1}{K_1-1}\big(u+v\big)\big(x^2 - 1\big), \label{eq_u_c3_v_ce_os}\\
        &~\log\bigg(\frac{K_1}{K-K_1}v\bigg) = -\gamma x - \frac{\gamma}{c_3} \big(u + v\big) + \frac{\gamma}{c_3} u x^2 + \delta_b.\label{eq_v_c3_u_ce_os}
    \end{align}

    Let
    \begin{align}
        q = \frac{u}{u+v}\in(0,1),\quad s = u+v \in \mathbb R, \nonumber
    \end{align}
    then
    \begin{align}
        u = qs,\quad v = (1-q)s = \frac{1-q}{q}u,\nonumber
    \end{align}
    and for $c_3$, according to Eq. (\ref{eq_c3_u_v_ce_os}),
    \begin{align}
        c_3^2 = s^2 \big(1 - 2x^2 q + x^2q^2\big). \nonumber
    \end{align}
    Denote
    \begin{align}
        g(q) = 1 - 2x^2 q + x^2q^2, \nonumber
    \end{align}
    then as $x\in(0,1), q\in(0,1)$, one can deduce that $g(q) > 0$. Therefore,
    \begin{align}
        c_3 = s\sqrt{g(q)}. \nonumber
    \end{align}

    According to Eq. (\ref{eq_u_c3_v_ce_os}),
    \begin{align}
    	&\log\big(u\big) = \frac{\gamma}{s\sqrt{g(q)}}\frac{K_1}{K_1-1}s\big(x^2 - 1\big) = \frac{\gamma}{\sqrt{g(q)}}\frac{K_1}{K_1-1}\big(x^2 - 1\big) \nonumber \\
    	&\quad\Rightarrow\quad u = \exp\bigg(\frac{\gamma}{\sqrt{g(q)}}\frac{K_1}{K_1-1}\big(x^2 - 1\big)\bigg). \nonumber
    \end{align}
    According to Eq. (\ref{eq_v_c3_u_ce_os})
    \begin{align}
        \log\bigg(\frac{K_1}{K-K_1}v\bigg) &= \delta_b -\gamma x - \frac{\gamma}{c_3} \big(s - u x^2\big) \nonumber \\
        &= \delta_b -\gamma x - \frac{\gamma}{s\sqrt{g(q)}} \big(s - qs x^2\big) \nonumber \\
        &= \delta_b -\gamma x - \frac{\gamma}{\sqrt{g(q)}} \big(1 - q x^2\big). \nonumber
    \end{align}
    As $v = (1-q)s = \frac{1-q}{q}u$,
    \begin{align}
        &~\log\bigg(\frac{K_1}{K-K_1}\frac{1-q}{q}u\bigg) = \delta_b -\gamma x - \frac{\gamma}{\sqrt{g(q)}} \big(1 - q x^2\big)\nonumber\\
        \Rightarrow\quad
        &~\underbrace{\log\frac{1-q}{q}}_{\alpha(q)} =  \underbrace{\delta_b - \gamma x - \frac{\gamma}{\sqrt{g(q)}}\bigg(\frac{K_1}{K_1-1}\big(x^2 - 1\big) + 1 - q x^2 \bigg) - \log\bigg(\frac{K_1}{K-K_1}\bigg)}_{\beta(q)}. \label{eq_value_of_q_ce_os}
    \end{align}

    When $q\rightarrow 0^+$,
    \begin{align}
        \alpha(q) \rightarrow +\infty,\quad \beta(q)\rightarrow \delta_b - \gamma x - \gamma\bigg(\frac{K_1}{K_1-1}\big(x^2 - 1\big) + 1 \bigg) - \log\bigg(\frac{K_1}{K-K_1}\bigg), \nonumber
    \end{align}
    and when $q\rightarrow 1^-$,
    \begin{align}
        \alpha(q) \rightarrow -\infty, \quad  \beta(q)\rightarrow \delta_b - \gamma x + \gamma\frac{\sqrt{1-x^2}}{K_1-1} - \log\bigg(\frac{K_1}{K-K_1}\bigg), \nonumber
    \end{align}
    which indicate that $\alpha(q)$ and $\beta(q)$ must intersect. Meanwhile
    \begin{align}
        \frac{\mathrm d \alpha(q)}{\mathrm d q} &= -\frac{1}{1-q}\frac{1}{q} <0,\quad\forall q\in(0,1),\nonumber\\
        \frac{\mathrm d \beta(q)}{\mathrm d q} &= -\gamma \frac{-x^2 \sqrt{g(q)}-\bigg(\frac{K_1}{K_1-1}\big(x^2 - 1\big) + 1 - q x^2 \bigg) \mathrm d\big(\sqrt{g(q)}\big)\big/ \mathrm dq}{g(q)} \nonumber \\
        & = \frac{\gamma x^2 (1-x^2)(K_1 - q)}{(K_1-1)\big(g(q)\big)^{3/2}} > 0,\quad \forall q\in(0,1). \nonumber
    \end{align}
    Therefore, there exists only one $q^\star\in(0,1)$ such that $\alpha(q^\star) = \beta(q^\star)$, and
    \begin{align}
    	u = \exp\bigg(\frac{\gamma}{\sqrt{g(q^\star)}}\frac{K_1}{K_1-1}\big(x^2 - 1\big)\bigg) ~\Rightarrow~ &c_1 = (K_1  -1)\exp\bigg(\frac{\gamma}{\sqrt{g(q^\star)}}\frac{K_1}{K_1-1}\big(x^2 - 1\big)\bigg), \label{eq_value_c1_ce_os} \\
        v = \frac{1-q^\star}{q^\star} u ~\Rightarrow~ &c_2 = K_1\frac{1-q^\star}{q^\star}\exp\bigg(\frac{\gamma}{\sqrt{g(q^\star)}}\frac{K_1}{K_1-1}\big(x^2 - 1\big)\bigg), \label{eq_value_c2_ce_os} \\
        &~c_3 = \bigg(\frac{c_1}{K_1-1} + \frac{c_2}{K_1}\bigg)\sqrt{g(q^\star)} \nonumber\\
        &\quad = \frac{1}{q^\star}\sqrt{g(q^\star)}\exp\bigg(\frac{\gamma}{\sqrt{g(q^\star)}}\frac{K_1}{K_1-1}\big(x^2 - 1\big)\bigg), \label{eq_value_c3_ce_os}
    \end{align}
    which completes the proof.
\end{proof}

\begin{lemma} \label{lem:lemmaB.1_os}
    For any $\bm X^{(k)} \in \mathcal D$ with any $c_1, c_2 > 0$ and $\gamma > 0$, the CE loss $L_{\mathrm{ce}}(\bm X^{(k)})$ defined in Eq. (\ref{equ-ce-single-sample}) is lower bounded by
    \begin{align}
        {L}_{\mathrm{ce}}\big(\bm X^{(k)}\big) \geq&~ \frac{c_1}{1+c_1+c_2} \sum_{j=1\atop j\neq k}^{K_1} \frac{\gamma\bm{w}_j^{\top}\bm{x}^{(k)}-b_j  - \gamma\bm{w}_k^{\top}\bm{x}^{(k)}+b_k}{K_1-1} \nonumber \\
        &\quad + \frac{c_2}{1+c_1+c_2}\sum_{j=K_1+1}^{K} \frac{\gamma\bm{w}_j^{\top}\bm{x}^{(k)}-b_j - \gamma\bm{w}_k^{\top}\bm{x}^{(k)}+b_k}{K-K_1} + C,  \label{equ-Lce_lowerBound_os}
    \end{align}
    where
    \begin{equation}
		\begin{aligned}
			&C = \frac1{1+c_1+c_2}\log\big(1+c_1+c_2\big) + \frac{c_1}{1+c_1+c_2}\log\bigg(\frac{1+c_1+c_2}{c_1}(K_1-1)\bigg) \\
			&\qquad + \frac{c_2}{1+c_1+c_2}\log\bigg(\frac{1+c_1+c_2}{c_2}(K-K_1)\bigg).
		\end{aligned} \label{equ-Lce_C_os}
\end{equation}
    The inequality becomes an equality if and only if
    \begin{align}
        &\gamma\bm{w}_j^{\top}\bm{x}^{(k)}-b_j = \gamma\bm{w}_\ell^{\top}\bm{x}^{(k)}-b_\ell, \quad \forall j, \ell \neq k \in [K_1], \label{equ-Lce_neg_score_K1_os}\\
        &\gamma\bm{w}_j^{\top}\bm{x}^{(k)}-b_j = \gamma\bm{w}_\ell^{\top}\bm{x}^{(k)}-b_\ell,\quad \forall k \in [K_1],~K_1< j,\ell\leq K, \label{equ-Lce_neg_score_K_os},
    \end{align}
    and
    \begin{align}
        &~ c_1=  (K_1-1)\exp\bigg(\sum_{j=1\atop j\neq k}^{K_1}\frac{\gamma\bm{w}_j^{\top}\bm{x}^{(k)}-b_j -\gamma\bm{w}_k^{\top}\bm{x}^{(k)}+b_k}{K-1}\bigg), \label{equ-Lce_c1_os}\\
        &~ c_2= (K-K_1)\exp\bigg(\sum_{j=K_1+1}^{K} \frac{\gamma\bm{w}_j^{\top}\bm{x}^{(k)}-b_j - \gamma\bm{w}_k^{\top}\bm{x}^{(k)}+b_k}{K-K_1}\bigg). \label{equ-Lce_c2_os}
    \end{align}
\end{lemma}

\begin{proof}
    According to Jensen's inequality in Lemma \ref{jensen_inequality_definition}, the CE loss $L_{\mathrm{ce}}(\bm X^{(k)})$ can be lower bounded,
    \begin{align}
        &~{L}_{\mathrm{ce}}(\bm X^{(k)})
        =\log\Bigg(1 + \sum_{j=1 \atop j\neq k}^K \frac{\exp\big(\gamma\bm{w}_j^{\top}\bm{x}^{(k)}-b_j\big)}{ \exp\big(\gamma\bm{w}_k^{\top}\bm{x}^{(k)}-b_k\big)}\Bigg) \nonumber \\
        = &~\log\Bigg(1+(K_1-1)\sum_{j=1\atop j\neq k}^{K_1} \frac{\exp\big(\gamma\bm{w}_j^{\top}\bm{x}^{(k)}-b_j\big)}{ (K_1-1)\exp\big(\gamma\bm{w}_k^{\top}\bm{x}^{(k)}-b_k\big)} \nonumber \\
        &\quad +(K-K_1)\sum_{j=K_1+1}^{K} \frac{\exp\big(\gamma\bm{w}_j^{\top}\bm{x}^{(k)}-b_j\big)}{(K-K_1)\exp\big(\gamma\bm{w}_k^{\top}\bm{x}^{(k)}-b_k\big)}\Bigg) \nonumber \\
        \geq &~\log\Bigg(1+(K_1-1)\exp\bigg(\sum_{j=1\atop j\neq k}^{K_1} \frac{\gamma\bm{w}_j^{\top}\bm{x}^{(k)}-b_j  - \gamma\bm{w}_k^{\top}\bm{x}^{(k)}+b_k}{K_1-1}\bigg) \nonumber \\
        &\quad +(K-K_1)\sum_{j=K_1+1}^{K} \frac{\exp\big(\gamma\bm{w}_j^{\top}\bm{x}^{(k)}-b_j\big)}{(K-K_1)\exp\big(\gamma\bm{w}_k^{\top}\bm{x}^{(k)}-b_k\big)}\Bigg) \nonumber \\
        \geq &~\log\Bigg(1+(K_1-1)\exp\bigg(\sum_{j=1\atop j\neq k}^{K_1} \frac{\gamma\bm{w}_j^{\top}\bm{x}^{(k)}-b_j  - \gamma\bm{w}_k^{\top}\bm{x}^{(k)}+b_k}{K_1-1}\bigg) \nonumber \\
        &\quad +(K-K_1)\exp\bigg(\sum_{j=K_1+1}^{K} \frac{\gamma\bm{w}_j^{\top}\bm{x}^{(k)}-b_j - \gamma\bm{w}_k^{\top}\bm{x}^{(k)}+b_k}{K-K_1}\bigg)\Bigg), \nonumber
    \end{align}
    where the first inequality becomes an equality if and only if
    \begin{align}
        \gamma\bm{w}_j^{\top}\bm{x}^{(k)}-b_j = \gamma\bm{w}_\ell^{\top}\bm{x}^{(k)}-b_\ell,\quad \forall j,\ell \neq k \in [K_1], \nonumber
    \end{align}
    and the second inequality becomes an equality if and only if
    \begin{align}
        \gamma\bm{w}_j^{\top}\bm{x}^{(k)}-b_j = \gamma\bm{w}_\ell^{\top}\bm{x}^{(k)}-b_\ell,\quad \forall k \in [K_1],~K_1< j,\ell\leq K.\nonumber
    \end{align}

    Then, by the concavity of the function $\log(1+\mathrm{e}^x)$, for any $c_1, c_2>0$, we get
    \begin{align}
        &{L}_{\mathrm{ce}}(\bm X^{(k)}) \nonumber \\
        \geq &~\log\Bigg[1+(K_1-1)\exp\bigg(\sum_{j=1\atop j\neq k}^{K_1} \frac{\gamma\bm{w}_j^{\top}\bm{x}^{(k)}-b_j  - \gamma\bm{w}_k^{\top}\bm{x}^{(k)}+b_k}{K_1-1}\bigg) \nonumber \\
        &\quad +(K-K_1)\exp\bigg(\sum_{j=K_1+1}^{K} \frac{\gamma\bm{w}_j^{\top}\bm{x}^{(k)}-b_j - \gamma\bm{w}_k^{\top}\bm{x}^{(k)}+b_k}{K-K_1}\bigg)\Bigg] \nonumber \\
        = &~\log\Bigg[\frac1{1+c_1+c_2}\frac{1+c_1+c_2}1 \nonumber \\
        &\quad +\frac{c_1}{1+c_1+c_2}\frac{1+c_1+c_2}{c_1}(K_1-1)\exp\bigg(\sum_{j=1\atop j\neq k}^{K_1} \frac{\gamma\bm{w}_j^{\top}\bm{x}^{(k)}-b_j  - \gamma\bm{w}_k^{\top}\bm{x}^{(k)}+b_k}{K_1-1}\bigg) \nonumber \\
        &\quad +\frac{c_2}{1+c_1+c_2}\frac{1+c_1+c_2}{c_2}(K-K_1)\exp\bigg(\sum_{j=K_1+1}^{K} \frac{\gamma\bm{w}_j^{\top}\bm{x}^{(k)}-b_j - \gamma\bm{w}_k^{\top}\bm{x}^{(k)}+b_k}{K-K_1}\bigg)\Bigg] \nonumber \\
        \geq &~ \frac1{1+c_1+c_2}\log\bigg(\frac{1+c_1+c_2}1\bigg) \nonumber \\
        &\quad + \frac{c_1}{1+c_1+c_2}\log\Bigg[\frac{1+c_1+c_2}{c_1}(K_1-1)\exp\bigg(\sum_{j=1\atop j\neq k}^{K_1} \frac{\gamma\bm{w}_j^{\top}\bm{x}^{(k)}-b_j  - \gamma\bm{w}_k^{\top}\bm{x}^{(k)}+b_k}{K_1-1}\bigg)\Bigg]\nonumber \\
        &\quad + \frac{c_2}{1+c_1+c_2}\log\Bigg[\frac{1+c_1+c_2}{c_2}(K-K_1)\exp\bigg(\sum_{j=K_1+1}^{K} \frac{\gamma\bm{w}_j^{\top}\bm{x}^{(k)}-b_j - \gamma\bm{w}_k^{\top}\bm{x}^{(k)}+b_k}{K-K_1}\bigg)\Bigg]\nonumber \\
        =&~ \frac{c_1}{1+c_1+c_2} \sum_{j=1\atop j\neq k}^{K_1} \frac{\gamma\bm{w}_j^{\top}\bm{x}^{(k)}-b_j  - \gamma\bm{w}_k^{\top}\bm{x}^{(k)}+b_k}{K_1-1} \nonumber \\
        &\quad + \frac{c_2}{1+c_1+c_2}\sum_{j=K_1+1}^{K} \frac{\gamma\bm{w}_j^{\top}\bm{x}^{(k)}-b_j - \gamma\bm{w}_k^{\top}\bm{x}^{(k)}+b_k}{K-K_1} + C, \nonumber
    \end{align}
    where
    \begin{align}
    	C =& \frac1{1+c_1+c_2}\log\big(1+c_1+c_2\big) + \frac{c_1}{1+c_1+c_2}\log\bigg(\frac{1+c_1+c_2}{c_1}(K_1-1)\bigg) \nonumber \\
    	&\quad + \frac{c_2}{1+c_1+c_2}\log\bigg(\frac{1+c_1+c_2}{c_2}(K-K_1)\bigg).
    \end{align}
    The second inequality achieves its equality if and only if
    \begin{align}
        \frac{1+c_1+c_2}1
        =~& \frac{1+c_1+c_2}{c_1}(K_1-1)\exp\bigg(\sum_{j=1\atop j\neq k}^{K_1} \frac{\gamma\bm{w}_j^{\top}\bm{x}^{(k)}-b_j  - \gamma\bm{w}_k^{\top}\bm{x}^{(k)}+b_k}{K_1-1}\bigg) \nonumber\\
        =~& \frac{1+c_1+c_2}{c_2}(K-K_1)\exp\bigg(\sum_{j=K_1+1}^{K} \frac{\gamma\bm{w}_j^{\top}\bm{x}^{(k)}-b_j - \gamma\bm{w}_k^{\top}\bm{x}^{(k)}+b_k}{K-K_1}\bigg),\nonumber
    \end{align}
    which is equivalent to
    \begin{align}
        &~ c_1=  (K_1-1)\exp\bigg(\sum_{j=1\atop j\neq k}^{K_1}\frac{\gamma\bm{w}_j^{\top}\bm{x}^{(k)}-b_j -\gamma\bm{w}_k^{\top}\bm{x}^{(k)}+b_k}{K-1}\bigg) \quad\text{and}\nonumber\\
        &~ c_2= (K-K_1)\exp\bigg(\sum_{j=K_1+1}^{K} \frac{\gamma\bm{w}_j^{\top}\bm{x}^{(k)}-b_j - \gamma\bm{w}_k^{\top}\bm{x}^{(k)}+b_k}{K-K_1}\bigg),\nonumber
    \end{align}
    as desired.
\end{proof}

\clearpage
\allowdisplaybreaks[2]
\subsection{Theorems in Contrastive Learning}
\begin{theorem}\label{theorem_NC_imbalanced_CL_sigmoid_supp}
Suppose that each class in dataset $\mathcal D$ contains at least two samples, and that the feature dimension is greater than the number of classes, i.e., $n_k = |\mathcal D_k| > 0$ for $\forall k \in [K]$ and $d \geq K-1$.
On the dataset $\mathcal D$, define losses
\begin{align}
    f_{\mathrm{sigm}}(\mathcal F) = \frac{1}{K}\sum_{k=1}^K\frac{1}{n_k}\sum_{\bm X^{(k)}\in\mathcal D_k} L_{\mathrm{sigm}}(\bm X^{(k)}),
    \label{eq_losses_CL_dataset_supp}
\end{align}
where
\begin{align}
    L_{\mathrm{sigm}}(\bm X^{(k)}) = \frac{1}{n_k}\sum_{\bm x_\text{p} \in \mathcal F_k}\log\big(1 + \e^{- \gamma \bm x_\text{p}^\top \bm x^{(k)} + b_k}\big) + \lambda\sum_{j=1, j\neq k}^K\frac{1}{n_j}\sum\limits_{\bm x_\text{n}\in\mathcal F_j}\log\big(1+\e^{\gamma \bm x_\text{n}^\top \bm x^{(k)} - b_j}\big), \label{eq_sigmoid_CL_single_sample_supp}
\end{align}
where $\lambda > 0$.
Then, at its minimum point,
    \begin{align}
        &\bm x^{(k)} = \frac{1}{n_k}\sum_{\bm x\in\mathcal F_k} \bm x = \bar{\bm x}_k ,\quad\forall\bm x^{(k)} \in \mathcal F_k,~\forall k\in[K],\label{eq_x_center_sigmoid_supp}\\
        &\langle\bar{\bm x}_k,\bar{\bm x}_j\rangle = -\frac1{K-1},\quad\forall k\neq j\in[K],\label{eq_x_centers_sigmoid_supp}\\
        &b_k = b^\star,\quad\forall k\in[K],
    \end{align}
where $b^\star$ is the unique solution of
    \begin{align}
        \e^{\frac{\gamma}{K-1}}\e^{2b} - \big[\lambda(K-1)-1\big]\e^{b} - \lambda(K-1)\e^{\gamma} = 0. \nonumber
    \end{align}
\end{theorem}

\begin{proof}
    The Sigmoid-based contrastive loss can be divided into two parts,
    \begin{align}
    	f_{\mathrm{sigm}}(\mathcal F) =&~ \frac{1}{K}\sum_{k=1}^K\frac{1}{n_k}\sum_{\bm X^{(k)}\in\mathcal D_k} L_{\mathrm{sigm}}(\bm X^{(k)}) \nonumber \\
        = &~\underbrace{\frac{1}{K}\sum_{k=1}^K\frac{1}{n_k}\sum_{\bm x^{(k)}\in\mathcal F_k} \frac{1}{n_k}\sum_{\bm x_\text{p} \in \mathcal F_k}\log\big(1+\e^{- \gamma \bm x_\text{p}^\top \bm x^{(k)} + b_k}\big)}_{f_{\mathrm{sigm}}^{\text{(intra)}}} \nonumber \\
        &\quad +\lambda \cdot \underbrace{\frac{1}{K}\sum_{k=1}^K\frac{1}{n_k}\sum_{\bm x^{(k)}\in\mathcal F_k} \sum_{j=1, j\neq k}^K\frac{1}{n_j}\sum\limits_{\bm x_\text{n}\in\mathcal F_j}\log\big(1+\e^{\gamma \bm x_\text{n}^\top \bm x^{(k)} - b_j}\big)}_{f_{\text{sigm}}^{\text{(inter)}}}. \nonumber
    \end{align}
    According to the convexity of $\kappa(x) = \log(1+\e^{-x})$,
    \begin{align}
        f_{\mathrm{sigm}}^{\text{(intra)}} &~= \frac{1}{K}\sum_{k=1}^K\frac{1}{n_k}\sum_{\bm x^{(k)}\in\mathcal F_k} \frac{1}{n_k}\sum_{\bm x_\text{p} \in \mathcal F_k}\log\big(1  + \e^{- \gamma \bm x_\text{p}^\top \bm x^{(k)} + b_k}\big)
         \nonumber \\
        &~ = \frac{1}{K}\sum_{k=1}^K\frac{1}{n_k}\sum_{\bm x^{(k)}\in\mathcal F_k} \frac{1}{n_k}\sum_{\bm x_\text{p} \in \mathcal F_k}\kappa\big(\gamma \bm x_\text{p}^\top \bm x^{(k)} - b_k\big) \nonumber \\
        &~\geq \frac{1}{K}\sum_{k=1}^K\kappa\bigg(\frac{1}{n_k^2}\sum_{\bm x^{(k)}\in\mathcal F_k} \sum_{\bm x_\text{p} \in \mathcal F_k}\big(\gamma \bm x_\text{p}^\top \bm x^{(k)} - b_k\big)\bigg) \nonumber \\
        &~\geq \kappa\bigg(\frac{1}{K}\sum_{k=1}^K\frac{1}{n_k^2}\sum_{\bm x^{(k)}\in\mathcal F_k} \sum_{\bm x_\text{p} \in \mathcal F_k}\big(\gamma \bm x_\text{p}^\top \bm x^{(k)} - b_k\big)\bigg) \nonumber \\
        &~= \kappa\bigg(\frac{1}{K}\sum_{k=1}^K\frac{1}{n_k^2}\sum_{\bm x^{(k)}\in\mathcal F_k} \sum_{\bm x_\text{p} \in \mathcal F_k}\gamma \bm x_\text{p}^\top \bm x^{(k)} - \frac{1}{K}\sum_{k=1}^K b_k\bigg),
    \end{align}
    where the first inequality becomes an equality if and only if
    \begin{align}
        \gamma \bm x_\text{p}^\top \bm x^{(k)} - b_k =  \gamma \hat{\bm x}_\text{p}^\top \hat{\bm x}^{(k)} - b_k, \quad\text{for}\quad \forall  \bm x_\text{p}, \hat{\bm x}_\text{p}, \bm x^{(k)}, \hat{\bm x}^{(k)} \in \mathcal F_k,~\forall k\in[K], \label{eq_intra_compactness_sigmoid_1}
    \end{align}
    and the second inequality becomes an equality if and only if
    \begin{align}
        &~\frac{1}{n_k^2}\sum_{\bm x^{(k)}\in\mathcal F_k} \sum_{\bm x_\text{p} \in \mathcal F_k}\big(\gamma \bm x_\text{p}^\top \bm x^{(k)} - b_k\big) = \frac{1}{n_\ell^2}\sum_{\bm x^{(\ell)}\in\mathcal F_\ell} \sum_{\bm x_\text{p} \in \mathcal F_\ell}\big(\gamma \bm x_\text{p}^\top \bm x^{(\ell)} - b_\ell\big)\nonumber\\
        \overset{[\ref{eq_intra_compactness_sigmoid_1}]}{\Longleftrightarrow} \quad
        &~ \gamma \bm x_\text{p}^\top \bm x^{(k)} - b_k = \gamma \hat{\bm x}_\text{p}^\top \bm x^{(\ell)} - b_\ell,\quad\forall \bm x_\text{p}, \bm x^{(k)} \in \mathcal F_k,~\forall \hat{\bm x}_\text{p}, \bm x^{(\ell)} \in \mathcal F_\ell,~\forall k,\ell\in[K]. \label{eq_intra_compactness_sigmoid_2}
    \end{align}

    Similarly, for the second term, according to the convexity of $\kappa(x) = \log(1+\e^{-x})$,
    \begin{align}
        f_{\text{sigm}}^{\text{(inter)}} &~= \frac{1}{K}\sum_{k=1}^K\frac{1}{n_k}\sum_{\bm x^{(k)}\in\mathcal F_k} \sum_{j=1, j\neq k}^K\frac{1}{n_j}\sum\limits_{\bm x_\text{n}\in\mathcal F_j}\log\big(1  +\e^{\gamma \bm x_\text{n}^\top \bm x^{(k)} - b_j}\big)  \nonumber\\
        &~ = \frac{1}{K}\sum_{k=1}^K\sum_{j=1, j\neq k}^K\frac{1}{n_k n_j}\sum_{\bm x^{(k)}\in\mathcal F_k}\sum\limits_{\bm x_\text{n}\in\mathcal F_j}\kappa\big(b_j - \gamma \bm x_\text{n}^\top \bm x^{(k)} \big) \nonumber \\
        &~\geq \frac{1}{K}\sum_{k=1}^K\sum_{j=1, j\neq k}^K \kappa\bigg(\frac{1}{n_k n_j}\sum_{\bm x^{(k)}\in\mathcal F_k}\sum_{\bm x_\text{n}\in\mathcal F_j} \big(b_j - \gamma \bm x_\text{n}^\top \bm x^{(k)} \big) \bigg) \nonumber \\
        &~\geq (K-1) \kappa\bigg(\frac{1}{K(K-1)}\sum_{k=1}^K\sum_{j=1, j\neq k}^K\frac{1}{n_k n_j}\sum_{\bm x^{(k)}\in\mathcal F_k}\sum_{\bm x_\text{n}\in\mathcal F_j} \big(b_j - \gamma \bm x_\text{n}^\top \bm x^{(k)} \big)\bigg) \nonumber \\
        &~= (K-1)\kappa\bigg(\frac{1}{K} \sum_{k=1}b_k - \frac{\gamma}{K(K-1)}\sum_{k=1}^K\sum_{j=1, j\neq k}^K\frac{1}{n_k n_j}\sum_{\bm x^{(k)}\in\mathcal F_k}\sum_{\bm x_\text{n}\in\mathcal F_j} \bm x_\text{n}^\top \bm x^{(k)} \bigg) \nonumber \\
        &~= (K-1)\kappa\bigg(\frac{1}{K} \sum_{k=1}b_k - \frac{\gamma}{K(K-1)}\sum_{k=1}^K\sum_{j=1}^K\frac{1}{n_k n_j}\sum_{\bm x^{(k)}\in\mathcal F_k}\sum_{\bm x_\text{n}\in\mathcal F_j} \bm x_\text{n}^\top \bm x^{(k)}  \nonumber \\
        &\qquad + \frac{\gamma}{K(K-1)}\sum_{k=1}^K\frac{1}{n_k n_k}\sum_{\bm x^{(k)}\in\mathcal F_k}\sum_{\bm x_\text{p}\in\mathcal F_k} \bm x_\text{p}^\top \bm x^{(k)}\bigg) \nonumber \\
        &~\geq (K-1) \kappa\bigg(\frac{1}{K} \sum_{k=1}b_k - \frac{\gamma}{K(K-1)}\sum_{k=1}^K\sum_{j=1}^K\frac{1}{n_k n_j}\sum_{\bm x^{(k)}\in\mathcal F_k}\sum_{\bm x_\text{n}\in\mathcal F_j} \bm x_\text{n}^\top \bm x^{(k)} \nonumber \\
        &\qquad +\frac{\gamma}{K(K-1)}\sum_{k=1}^K 1\bigg) \nonumber \\
        &~= (K-1) \kappa\Bigg(\frac{1}{K} \sum_{k=1}b_k - \frac{\gamma}{K(K-1)}\bigg(\sum_{k=1}^K\frac{1}{n_k}\sum_{\bm x^{(k)}\in\mathcal F_k} \bm x^{(k)} \bigg)^\top \bigg(\sum_{j=1}^K\frac{1}{n_j}\sum_{\bm x_\text{n}\in\mathcal F_j} \bm x_\text{n}\bigg)  \nonumber \\
        &\qquad + \frac{\gamma}{K-1}\Bigg) \nonumber \\
        &~= (K-1) \kappa\bigg(\frac{1}{K} \sum_{k=1}b_k - \frac{\gamma K}{K-1}\|\tilde{\bm x}\|^2 + \frac{\gamma}{K-1}\bigg) \nonumber\\
        &~\geq (K-1) \kappa\bigg(\frac{1}{K} \sum_{k=1}b_k + \frac{\gamma}{K-1}\bigg),
    \end{align}
    where the first inequality is derived from the convexity of $\kappa(x)$ and it becomes an equality if and only if
    \begin{align}
        b_j - \gamma \bm x_\text{n}^\top \bm x^{(k)} = b_j - \gamma \hat{\bm x}_\text{n}^\top \hat{\bm x}^{(k)},\quad \forall \bm x^{(k)}, \hat{\bm x}^{(k)} \in\mathcal F_k,~\forall \bm x_\text{n}, \hat{\bm x}_\text{n} \in \mathcal F_j,~\forall k\neq j\in[K], \label{eq_inter_separability_sigmoid_1}
    \end{align}
    the second inequality is derived from the convexity of $\kappa(x)$ and it becomes an equality if and only if
    \begin{align}
        &\frac{1}{n_k n_j}\sum_{\bm x^{(k)}\in\mathcal F_k}\sum_{\bm x_\text{n}\in\mathcal F_j} \big(b_j - \gamma \bm x_\text{n}^\top \bm x^{(k)} \big) =
        \frac{1}{n_{k'} n_{j'}}\sum_{\bm x^{(k')}\in\mathcal F_{k'}}\sum_{\bm x_\text{n}\in\mathcal F_{j'}} \big(b_{j'} - \gamma \bm x_\text{n}^\top \bm x^{(k')} \big)\nonumber\\
        \overset{\ref{eq_inter_separability_sigmoid_1}}\Longleftrightarrow\quad
        &b_j - \gamma \bm x_\text{n}^\top \bm x^{(k)} = b_{j'} - \gamma \hat{\bm x}_\text{n}^\top \hat{\bm x}^{(k')}, \nonumber \\
        &\qquad \forall \bm x^{(k)}\in\mathcal F_k, ~\forall \bm x_\text{n} \in \mathcal F_j, ~\forall \hat{\bm x}^{(k')}\in\mathcal F_{k'}, ~\forall \hat{\bm x}_\text{n} \in \mathcal F_{j'},~\forall k\neq j, k'\neq j'\in[K], \label{eq_inter_separability_sigmoid_2}
    \end{align}
    the third inequality is derived from the monotonic decreasing of $\kappa(x)$ and it becomes an equality if and only if
    \begin{align}
        \bm x^{(k)} = \bm x_{\text{p}},\quad \text{for}~\forall \bm x^{(k)}, \bm x_{\text{p}} \in\mathcal F_k,~\forall k\in[K], \label{eq_inter_separability_sigmoid_3}
    \end{align}
    and the fourth inequality is derived from the monotonic decreasing of $\kappa(x)$ and it becomes an equality if and only if
    \begin{align}
        \tilde{\bm x} = \frac{1}{K}\sum_{k=1}^K\bar{\bm x}_k = \frac{1}{K}\sum_{k=1}^K \frac{1}{n_k}\sum_{\bm x^{(k)}\in\mathcal F_k} = \bm 0. \label{eq_inter_separability_sigmoid_4}
    \end{align}

    Combining Eqs. (\ref{eq_intra_compactness_sigmoid_2},\ref{eq_inter_separability_sigmoid_3}), one can deduce that, $\exists b\in\mathbb R$,
    \begin{align}
        b_k = b,\quad \forall k\in[K].
    \end{align}
    Based on the above analysis, if and only if Eqs. (\ref{eq_intra_compactness_sigmoid_1}, \ref{eq_intra_compactness_sigmoid_2}, \ref{eq_inter_separability_sigmoid_1}, \ref{eq_inter_separability_sigmoid_2}, \ref{eq_inter_separability_sigmoid_3}, \ref{eq_inter_separability_sigmoid_4}),
    \begin{align}
        &~f_{\mathrm{sigm}}(\mathcal F) = f_{\text{sigm}}^{\text{(intra)}} + \lambda f_{\text{sigm}}^{\text{(inter)}} \nonumber \\
        \geq &~\kappa\bigg(\frac{1}{K}\sum_{k=1}^K\frac{1}{n_k^2}\sum_{\bm x^{(k)}\in\mathcal F_k} \sum_{\bm x_\text{p} \in \mathcal F_k}\gamma \bm x_\text{p}^\top \bm x^{(k)} - \frac{1}{K}\sum_{k=1}^K b_k\bigg) + \lambda (K-1) \kappa\bigg(\frac{1}{K} \sum_{k=1}b_k + \frac{\gamma}{K-1}\bigg) \nonumber \\
        = &~\kappa\bigg(\frac{1}{K}\sum_{k=1}^K \gamma - \frac{1}{K}\sum_{k=1}^K b_k\bigg) + \lambda (K-1) \kappa\bigg(\frac{1}{K} \sum_{k=1}b_k + \frac{\gamma}{K-1}\bigg) \nonumber \\
        =&~ \kappa\big(\gamma - b\big) + \lambda (K-1) \kappa\bigg(b + \frac{\gamma}{K-1}\bigg)\nonumber\\
        =&~\log\bigg(1+\exp\big(b-\gamma\big)\bigg) + \lambda (K-1)\log\bigg(1+\exp\big(-b-\frac{\gamma}{K-1}\big)\bigg) \triangleq \hat{f}_{\text{sigm}}(b).
    \end{align}
    At the minimum points of $\hat{f}_{\text{sigm}}(b)$,
    \begin{align}
        &0 = \frac{\partial\hat{f}_{\text{sigm}}(b)}{\partial b} = \frac{\exp\big(b-\gamma\big)}{1+\exp\big(b-\gamma\big)} - \lambda(K  -1)\frac{\exp\big(-b-\frac{\gamma}{K-1}\big)}{1+\exp\big(-b-\frac{\gamma}{K-1}\big)} \nonumber \\
        &\qquad\qquad\qquad = \frac{1}{1+\exp\big(\gamma-b\big)} - \frac{\lambda(K-1)}{1+\exp\big(b+\frac{\gamma}{K-1}\big)}\nonumber\\
        \Longleftrightarrow &\quad \e^{\frac{\gamma}{K-1}}\e^{2b} - \big[\lambda(K-1)-1\big]\e^{b} - \lambda(K-1)\e^{\gamma} = 0.
    \end{align}

    From Eq. (\ref{eq_inter_separability_sigmoid_3}), one can deduce that
    \begin{align}
        \bm x^{(k)} = \frac{1}{n_k}\sum_{\bm x\in\mathcal F_k} \bm x = \bar{\bm x}_k ,\quad\forall\bm x^{(k)} \in \mathcal F_k,~\forall k\in[K]. \nonumber
    \end{align}
    Similar to the proof of Theorem \ref{them:BCE collapse_supp}, from Eq. (\ref{eq_inter_separability_sigmoid_4}), one can deduce
    \begin{align}
        \langle\bar{\bm x}_k,\bar{\bm x}_j\rangle = -\frac1{K-1},\quad\forall k\neq j\in[K].\nonumber
    \end{align}
    It completes the proof.

    Finally, if set $b_k = 0$ for $\forall k\in[K]$ in (\ref{eq_sigmoid_CL_single_sample_supp}), Eqs. (\ref{eq_inter_separability_sigmoid_3}, \ref{eq_inter_separability_sigmoid_4}) can still be derived from the minimization of $f_{\text{sigm}}^{\text{(inter)}}$, and in this case it suffices for each class in $\mathcal D$ to contain at least one sample.
    Therefore, if remove the biases in the Sigmoid-based contrastive loss, the main conclusions of this theorem, i.e., Eqs. (\ref{eq_x_center_sigmoid_supp}) and (\ref{eq_x_centers_sigmoid_supp}) still hold when reaching its minimum value, provided that there are no empty classes in $\mathcal D$.
\end{proof}

\clearpage
\allowdisplaybreaks[2]
\begin{theorem}\label{theorem_NC_imbalanced_CL_softmax_supp}
Suppose that each class in dataset $\mathcal D$ contains at least two samples, and that the feature dimension is greater than the number of classes, i.e., $n_k = |\mathcal D_k| > 0$ for $\forall k \in [K]$ and $d > K$.
On the dataset $\mathcal D$, define losses
\begin{align}
    f_{\mathrm{soft}}(\mathcal F) = \frac{1}{K}\sum_{k=1}^K\frac{1}{n_k}\sum_{\bm X^{(k)}\in\mathcal D_k} L_{\mathrm{soft}}(\bm X^{(k)}),
    \label{eq_losses_CL_dataset_soft_supp}
\end{align}
where
\begin{align}
    L_{\mathrm{soft}}(\bm X^{(k)})
        =&-\frac{1}{n_k}\sum_{\bm x_\text{p} \in \mathcal F_k}\log\bigg(\frac{\e^{\gamma\bm x_\text{p}^\top \bm x^{(k)}}}{\sum_{j=1}^K\frac{1}{n_j}\sum_{\bm x\in\mathcal F_j}\e^{\gamma\bm x^\top \bm x^{(k)}}}\bigg). \label{eq_softmax_CL_single_sample_supp}
\end{align}
Then, at its minimum point,
    \begin{align}
        &\bm x^{(k)} = \frac{1}{n_k}\sum_{\bm x\in\mathcal F_k} \bm x = \bar{\bm x}_k ,\quad\forall\bm x^{(k)} \in \mathcal F_k,~\forall k\in[K],\label{eq_x_center_softmax_supp}\\
        &\langle\bar{\bm x}_k,\bar{\bm x}_j\rangle = -\frac1{K-1},\quad\forall k\neq j\in[K].\label{eq_x_centers_softmax_supp}
    \end{align}

\end{theorem}

\begin{proof}
    According to the convexities of $\e^x$ and $\log x$,
    \begin{align}
    	&~f_{\text{soft}}(\mathcal F) \nonumber \\
        =&~ -\frac{1}{K}\sum_{k=1}^K\frac{1}{n_k}\sum_{\bm x^{(k)}\in\mathcal F_k}\frac{1}{n_k}\sum_{\bm x_\text{p} \in \mathcal F_k}\log\bigg(\frac{\e^{\gamma\bm x_\text{p}^\top \bm x^{(k)}}}{\sum_{j=1}^K\frac{1}{n_j}\sum_{\bm x\in\mathcal F_j}\e^{\gamma\bm x^\top \bm x^{(k)}}}\bigg) \nonumber \\
        =&~\frac{1}{K}\sum_{k=1}^K\frac{1}{n_k}\sum_{\bm x^{(k)}\in\mathcal F_k}\frac{1}{n_k}\sum_{\bm x_\text{p} \in \mathcal F_k}\Bigg[\log\bigg({\sum_{j=1}^K\frac{1}{n_j}\sum_{\bm x\in\mathcal F_j}\e^{\gamma\bm x^\top \bm x^{(k)}}}\bigg) - \gamma\bm x_\text{p}^\top \bm x^{(k)}\Bigg] \nonumber \\
        =&~\frac{1}{K}\sum_{k=1}^K\frac{1}{n_k}\sum_{\bm x^{(k)}\in\mathcal F_k}\Bigg[\log\bigg({\sum_{j=1}^K\frac{1}{n_j}\sum_{\bm x\in\mathcal F_j}\e^{\gamma\bm x^\top \bm x^{(k)}}}\bigg) - \frac{1}{n_k}\sum_{\bm x_\text{p} \in \mathcal F_k} \gamma\bm x_\text{p}^\top \bm x^{(k)}\Bigg] \nonumber \\
        =&~\frac{1}{K}\sum_{k=1}^K\frac{1}{n_k}\sum_{\bm x^{(k)}\in\mathcal F_k}\Bigg[\log\bigg({\sum_{j=1}^K\frac{1}{n_j}\sum_{\bm x\in\mathcal F_j}\e^{\gamma\bm x^\top \bm x^{(k)}}}\bigg) - \log\exp\bigg(\frac{1}{n_k}\sum_{\bm x_\text{p} \in \mathcal F_k} \gamma\bm x_\text{p}^\top \bm x^{(k)}\bigg)\Bigg] \nonumber \\
        \geq &~ \frac{1}{K}\sum_{k=1}^K\frac{1}{n_k}\sum_{\bm x^{(k)}\in\mathcal F_k}\Bigg[\log\Bigg(\sum_{j=1}^K\exp\bigg(\frac{1}{n_j}\sum_{\bm x\in\mathcal F_j}\gamma\bm x^\top \bm x^{(k)}\bigg)\Bigg) - \log\exp\bigg(\frac{1}{n_k}\sum_{\bm x_\text{p} \in \mathcal F_k} \gamma\bm x_\text{p}^\top \bm x^{(k)}\bigg)\Bigg] \nonumber \\
        =&~ \frac{1}{K}\sum_{k=1}^K\frac{1}{n_k}\sum_{\bm x^{(k)}\in\mathcal F_k}\log\Bigg(1 + \sum_{j=1,j\neq k}^K\exp\bigg(\frac{1}{n_j}\sum_{\bm x\in\mathcal F_j}\gamma\bm x^\top \bm x^{(k)} - \frac{1}{n_k}\sum_{\bm x_\text{p} \in \mathcal F_k} \gamma\bm x_\text{p}^\top \bm x^{(k)}\bigg)\Bigg) \nonumber \\
        \geq&~ \frac{1}{K}\sum_{k=1}^K\frac{1}{n_k}\sum_{\bm x^{(k)}\in\mathcal F_k}\log\Bigg(1 + (K-1)\exp\bigg[\frac{1}{K-1}\sum_{j=1,j\neq k}^K\bigg(\frac{1}{n_j}\sum_{\bm x\in\mathcal F_j}\gamma\bm x^\top \bm x^{(k)} \nonumber \\
        &\quad - \frac{1}{n_k}\sum_{\bm x_\text{p} \in \mathcal F_k} \gamma\bm x_\text{p}^\top \bm x^{(k)}\bigg)\bigg]\Bigg) \nonumber \\
        =&~ \frac{1}{K}\sum_{k=1}^K\frac{1}{n_k}\sum_{\bm x^{(k)}\in\mathcal F_k}\log\Bigg\{\frac{c_1}{1+c_1}\frac{1+c_1}{c_1} \nonumber \\
        &\quad +\frac{1+c_1}{1+c_1} (K-1)\exp\bigg[\frac{1}{K-1}\sum_{j=1,j\neq k}^K\bigg(\frac{1}{n_j}\sum_{\bm x\in\mathcal F_j}\gamma\bm x^\top \bm x^{(k)} - \frac{1}{n_k}\sum_{\bm x_\text{p} \in \mathcal F_k} \gamma\bm x_\text{p}^\top \bm x^{(k)}\bigg)\bigg]\Bigg\} \nonumber \\
        \geq &~ \frac{1}{K}\sum_{k=1}^K\frac{1}{n_k}\sum_{\bm x^{(k)}\in\mathcal F_k}\Bigg\{\frac{c_1}{1+c_1}\log\bigg(\frac{1+c_1}{c_1}\bigg) \nonumber \\
        &\quad +\frac1{1+c_1}\log\Bigg((1+c_1)(K-1)\exp\bigg[\frac{1}{K-1}\sum_{j=1,j\neq k}^K\bigg(\frac{1}{n_j}\sum_{\bm x\in\mathcal F_j}\gamma\bm x^\top \bm x^{(k)} - \frac{1}{n_k}\sum_{\bm x_\text{p} \in \mathcal F_k} \gamma\bm x_\text{p}^\top \bm x^{(k)}\bigg)\bigg]\Bigg)\Bigg\} \nonumber \\
        = &~ \frac{1}{K}\sum_{k=1}^K\frac{1}{n_k}\sum_{\bm x^{(k)}\in\mathcal F_k} \frac{1}{K-1}\sum_{j=1,j\neq k}^K\bigg(\frac{1}{n_j}\sum_{\bm x\in\mathcal F_j}\gamma\bm x^\top \bm x^{(k)} - \frac{1}{n_k}\sum_{\bm x_\text{p} \in \mathcal F_k} \gamma\bm x_\text{p}^\top \bm x^{(k)}\bigg) \nonumber \\
        &\quad + \underbrace{\frac{c_1}{1+c_1}\log\bigg(\frac{1+c_1}{c_1}\bigg) + \frac1{1+c_1}\log\big[(1+c_1)(K-1)\big]}_C, \nonumber
    \end{align}
    where the first inequality is derived from the convexity of $\e^x$ and it becomes an equality if and only if
    \begin{align}
        \gamma \bm x^\top \bm x^{(k)} = \gamma \hat{\bm x}^\top \bm x^{(k)},\quad\forall \bm x, \hat{\bm x}\in \mathcal F_j,~\forall \bm x^{(k)}\in\mathcal F_k,~ \forall j,k\in[K], \label{eq_softmax_supp_1}
    \end{align}
    the second inequality is derived from the convexity of $\e^x$ and it becomes an equality if and only if
    \begin{align}
        &\frac{1}{n_j}\sum_{\bm x\in\mathcal F_j}\gamma\bm x^\top \bm x^{(k)} - \frac{1}{n_k}\sum_{\bm x_\text{p} \in \mathcal F_k} \gamma\bm x_\text{p}^\top \bm x^{(k)} = \frac{1}{n_{j'}}\sum_{\bm x\in\mathcal F_{j'}}\gamma\bm x^\top \bm x^{(k)} - \frac{1}{n_k}\sum_{\bm x_\text{p} \in \mathcal F_k} \gamma\bm x_\text{p}^\top \bm x^{(k)} \nonumber \\
        \overset{[\ref{eq_softmax_supp_1}]}{\Leftrightarrow}\quad&~ \gamma\bm x^\top \bm x^{(k)} - \gamma\bm x_\text{p}^\top \bm x^{(k)} = \gamma\hat{\bm x}^\top \hat{\bm x}^{(k)} - \gamma\hat{\bm x}_\text{p}^\top \hat{\bm x}^{(k)},\quad \forall \bm x\in\mathcal F_j, \hat{\bm x}\in\mathcal F_{j'}, \nonumber \\
        &\qquad ~\forall \bm x_\text{p}, \bm x^{(k)}, \hat{\bm x}_\text{p}, \hat{\bm x}^{(k)} \in \mathcal F_k, ~\forall k\in [K],~\forall j,j'\neq k\in [K], \label{eq_softmax_supp_2}
    \end{align}
    and the third inequality is derived from the convexity of $\log x$ and it becomes an equality if and only if
    \begin{align}
        &~\frac{1+c_1}{c_1} = (1+c_1)(K-1)\exp\bigg[\frac{1}{K-1}\sum_{j=1,j\neq k}^K\bigg(\frac{1}{n_j}\sum_{\bm x\in\mathcal F_j}\gamma\bm x^\top \bm x^{(k)} - \frac{1}{n_k}\sum_{\bm x_\text{p} \in \mathcal F_k} \gamma\bm x_\text{p}^\top \bm x^{(k)}\bigg)\bigg] \nonumber \\
        \Leftrightarrow\quad&~
        c_1 = \bigg\{(K  -1)\exp\bigg[\frac{1}{K-1}\sum_{j=1,j\neq k}^K\bigg(\frac{1}{n_j}\sum_{\bm x\in\mathcal F_j}\gamma\bm x^\top \bm x^{(k)} - \frac{1}{n_k}\sum_{\bm x_\text{p} \in \mathcal F_k} \gamma\bm x_\text{p}^\top \bm x^{(k)}\bigg)\bigg]\bigg\}^{-1}. \label{eq_softmax_supp_3}
    \end{align}

    Furthermore,
    \begin{align}
        f_{\text{soft}}(\mathcal F)
        \geq&~ \frac{1}{K}\sum_{k=1}^K\frac{1}{n_k}\sum_{\bm x^{(k)}\in\mathcal F_k} \frac{1}{K-1}\sum_{j=1,j\neq k}^K\bigg(\frac{1}{n_j}\sum_{\bm x\in\mathcal F_j}\gamma\bm x^\top \bm x^{(k)} - \frac{1}{n_k}\sum_{\bm x_\text{p} \in \mathcal F_k} \gamma\bm x_\text{p}^\top \bm x^{(k)}\bigg) + C \nonumber \\
        =&~ \frac{1}{K}\sum_{k=1}^K\frac{1}{n_k}\sum_{\bm x^{(k)}\in\mathcal F_k} \frac{1}{K-1}\bigg(\sum_{j=1}^K\frac{1}{n_j}\sum_{\bm x\in\mathcal F_j}\gamma\bm x^\top \bm x^{(k)} - K \frac{1}{n_k}\sum_{\bm x_\text{p} \in \mathcal F_k} \gamma\bm x_\text{p}^\top \bm x^{(k)}\bigg) + C \nonumber \\
        =&~ \frac{1}{K}\frac{1}{K-1} \sum_{k=1}^K\frac{1}{n_k}\sum_{\bm x^{(k)}\in\mathcal F_k} \sum_{j=1}^K\frac{1}{n_j}\sum_{\bm x\in\mathcal F_j}\gamma\bm x^\top \bm x^{(k)} \nonumber \\
        &\quad -\frac{1}{K-1}\sum_{k=1}^K\frac{1}{n_k}\sum_{\bm x^{(k)}\in\mathcal F_k}  \frac{1}{n_k}\sum_{\bm x_\text{p} \in \mathcal F_k} \gamma\bm x_\text{p}^\top \bm x^{(k)} + C \nonumber \\
        =&~ \frac{K}{K-1} \gamma\Bigg\|\frac{1}{K}\sum_{k=1}^K\frac{1}{n_k}\sum_{\bm x^{(k)}\in\mathcal F_k} \bm x^{(k)}\Bigg\|^2 \nonumber\\
        &\quad - \frac{1}{K-1}\sum_{k=1}^K\frac{1}{n_k}\sum_{\bm x^{(k)}\in\mathcal F_k}  \frac{1}{n_k}\sum_{\bm x_\text{p} \in \mathcal F_k} \gamma\bm x_\text{p}^\top \bm x^{(k)} + C \nonumber\\
        \geq&~ \frac{K}{K-1} \gamma\|\tilde{\bm x}\|^2 - \frac{K}{K-1}\gamma + C\nonumber\\
        \geq&~ - \frac{K}{K-1}\gamma + C,
    \end{align}
    where $\tilde{\bm x} = \frac{1}{K}\sum_{k=1}^K \bar{\bm x}_k = \frac{1}{K}\sum_{k=1}^K\frac{1}{n_k}\sum_{\bm x^{(k)}\in\mathcal F_k} \bm x^{(k)}$, the second inequality becomes an equality if and only if
    \begin{align}
        \bm x_{\text{p}} = \bm x^{(k)},\quad \forall \bm x, \bm x_{\text{p}} \in \mathcal F_k,~\forall k\in[K], \label{eq_softmax_supp_4}
    \end{align}
    and the third inequality becomes an equality if and only if
    \begin{align}
        \tilde{\bm x} = \frac{1}{K}\sum_{k=1}^K \bar{\bm x}_k = \bm 0. \label{eq_softmax_supp_5}
    \end{align}

    From Eq. (\ref{eq_softmax_supp_4}), one can deduce Eq. (\ref{eq_x_center_softmax_supp}), from Eq. (\ref{eq_softmax_supp_5}), one can deduce Eq. (\ref{eq_x_centers_softmax_supp}), and they are compatible with Eq. (\ref{eq_softmax_supp_1}) and (\ref{eq_softmax_supp_2}),
    which completes the proof.
\end{proof}

\bibliography{references}
\end{document}